\documentclass{article}
\usepackage{iclr2025_conference,times}

\usepackage{amsmath,amsfonts,bm}

\def\eqref#1{equation~\ref{#1}}

\def\1{\bm{1}}

\DeclareMathAlphabet{\mathsfit}{\encodingdefault}{\sfdefault}{m}{sl}
\SetMathAlphabet{\mathsfit}{bold}{\encodingdefault}{\sfdefault}{bx}{n}

\usepackage{hyperref}
\usepackage{url}

\addtocontents{toc}{\protect\setcounter{tocdepth}{-1}}

\title{\includegraphics[width=0.035\textwidth]{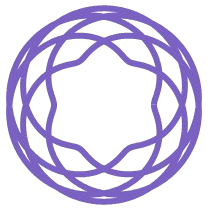}~\textbf{PhysBench}: Benchmarking and Enhancing \\ Vision-Language Models for \\Physical World Understanding}

\author{\vspace{-0.01in}
    \textbf{Wei Chow}$^{*1}$, 
    \textbf{Jiageng Mao}$^{*1}$, 
    \textbf{Boyi Li}$^{2}$, 
    \textbf{Daniel Seita}$^{1}$,
    \textbf{Vitor Guizilini}$^{3}$,
    \textbf{Yue Wang}$^{1}$\vspace{0.03in}\\
    $^1$University of Southern California, 
    $^2$UC Berkeley,
    $^3$Toyota Research Institute\vspace{0.1in}\\
}

\usepackage{tcolorbox}
\usepackage{xcolor}
\usepackage{array}
\usepackage{pifont}
\usepackage{booktabs}
\usepackage{graphicx}
\usepackage{multirow}
\usepackage{hyperref}
\usepackage{adjustbox}
\usepackage{caption}
\usepackage{subcaption}
\usepackage{arydshln}
\usepackage{tikz}
\usepackage{xspace}
\usepackage{algorithm}
\usepackage{algorithmic}
\usepackage{footmisc}

\definecolor{my_green}{RGB}{51,102,0}
\definecolor{my_yellow}{RGB}{255,165,0}
\definecolor{my_red}{RGB}{204, 0, 0}
\definecolor{light_pink}{RGB}{250,245,247} 
\definecolor{light_blue}{RGB}{240,245,255}
\definecolor{light_green}{RGB}{240,255,240}
\definecolor{light_yellow}{RGB}{255,255,240}
\definecolor{light_grey}{RGB}{240,240,240}

\newcommand{\red}[1]{\textcolor{red}{#1}}

\renewcommand{\checkmark}{\textcolor{my_green}{\ding{51}}} 
\newcommand{\crossmark}{\textcolor{my_red}{\ding{55}}} 
\usepackage{fontawesome}

\newcommand{\numobj}{679 }
\newcommand{\numhdr}{470 }

\definecolor{Aquamarine}{rgb}{0.0, 1, 0.8}
\definecolor{Fuchsia}{rgb}{1.0, 0.0, 1.0}
\definecolor{BurntOrange}{rgb}{0.8, 0.33, 0.0}

\newcommand{\blfootnote}[1]{%
  \begingroup
  \renewcommand\thefootnote{}%
  \footnote{#1}%
  \addtocounter{footnote}{-1}%
  \endgroup
}

\usepackage{tcolorbox}
\newtcolorbox{mycase}[1]{
  colback=#1,
  colframe=white,
  boxrule=1pt, 
  left=1pt,
  right=1pt,
  top=1pt,
  bottom=1pt,
}
\newtcolorbox{question}[1]{
  colback=#1, 
  colframe=#1, 
  boxrule=0pt,
  left=1pt,
  right=1pt,
  top=1pt,
  bottom=1pt,
  fonttitle=\bfseries\color{black}, 
  title=Question
}
\newtcolorbox{answer}[1]{
  colback=#1, 
  colframe=#1, 
  boxrule=0pt,
  left=1pt,
  right=1pt,
  top=1pt,
  bottom=1pt,
  fonttitle=\bfseries\color{black}, 
  title=Answer
}
\newcommand{\includeimage}[2]{
  \raisebox{-.5\height}{\includegraphics[height=#2\fontcharht\font`\B]{#1}}
}

\iclrfinalcopy 

\begin{document}

\maketitle

\vspace{-6mm}
\begin{abstract}
\vspace{-2mm}
Understanding the physical world is a fundamental challenge in embodied AI, critical for enabling agents to perform complex tasks and operate safely in real-world environments. While Vision-Language Models (VLMs) have shown great promise in reasoning and task planning for embodied agents, their ability to comprehend physical phenomena remains extremely limited.
To close this gap, we introduce PhysBench, a comprehensive benchmark designed to evaluate VLMs' physical world understanding capability across a diverse set of tasks. 
PhysBench contains 10,002 entries of interleaved video-image-text data, categorized into four major domains: physical object properties, physical object relationships, physical scene understanding, and physics-based dynamics, further divided into 19 subclasses and 8 distinct capability dimensions.
Our extensive experiments, conducted on 75 representative VLMs, reveal that while these models excel in common-sense reasoning, they struggle with understanding the physical world---likely due to the absence of physical knowledge in their training data and the lack of embedded physical priors.
To tackle the shortfall, we introduce PhysAgent, a novel framework that combines the generalization strengths of VLMs with the specialized expertise of vision models, significantly enhancing VLMs' physical understanding across a variety of tasks, including an 18.4\% improvement on GPT-4o.
Furthermore, our results demonstrate that enhancing VLMs' physical world understanding capabilities can help embodied agents such as MOKA.
We believe that PhysBench and PhysAgent offer valuable insights and contribute to bridging the gap between VLMs and physical world understanding.
\href{https://physbench.github.io/}{\color{blue}\textbf{Project Page is here}\xspace}\vspace{-0.1in}
\end{abstract}

\vspace{-5mm}
\section{Introduction}\label{sec: intro}
\vspace{-3mm}
Understanding the physical world is a fundamental challenge in embodied AI~\citep{gupta2021embodied, shen2021igibson}. 
Embodied agents are required to understand the physical properties of objects (e.g., mass, stiffness) to accurately interact with these objects. 
They also need to understand the relationships of physical objects to operate efficiently in cluttered environments, understand the structure of physical scenes for safe navigation and manipulation, and anticipate the outcomes of interactions and physics-based dynamics for better planning and preventing accidents.
These capabilities of intuitive physics~\citep{mccloskey1983intuitive, carey2000origin} are innate to humans and can also greatly benefit embodied agents, allowing them to perform complex tasks and operate safely in real-world scenarios~\citep{kill2020mental}.

Vision-language models (VLMs)~\citep{liu2024visual, achiam2023gpt, team2023gemini} have emerged as promising solutions for building embodied agents~\citep{liu2024moka, google2024pivot, huang2023voxposer}. Trained on large amounts of human knowledge, these models have developed strong capabilities in reasoning and task planning~\citep{yue2023mmmu, lu2024mathvista, kim24openvla, niu2024llarva, zhen20243dvla}.
However, relying solely on these capabilities is insufficient for developing generalist embodied agents. A series of studies have highlighted a gap in understanding the physical world, leading to operational errors~\citep{liu2024moka}, such as mishandling fragile objects~\citep{wang2023newton} or failing to recognize appropriate grasping affordances~\citep{Guo2024PhyGrasp}. \textbf{Since these agents operate in and interact with the real world, VLMs must possess a comprehensive understanding of the physical world—a critical yet underexplored domain}. This deficiency in physical world understanding limits the effective deployment of VLMs in embodied applications~\citep{liu2024moka, Guo2024PhyGrasp, pgvlm2024}.\blfootnote{$^*$Equal contribution.}

\begin{figure}[!t]
	\centering  
	\vspace{-7mm}
	\includegraphics[width=\linewidth]{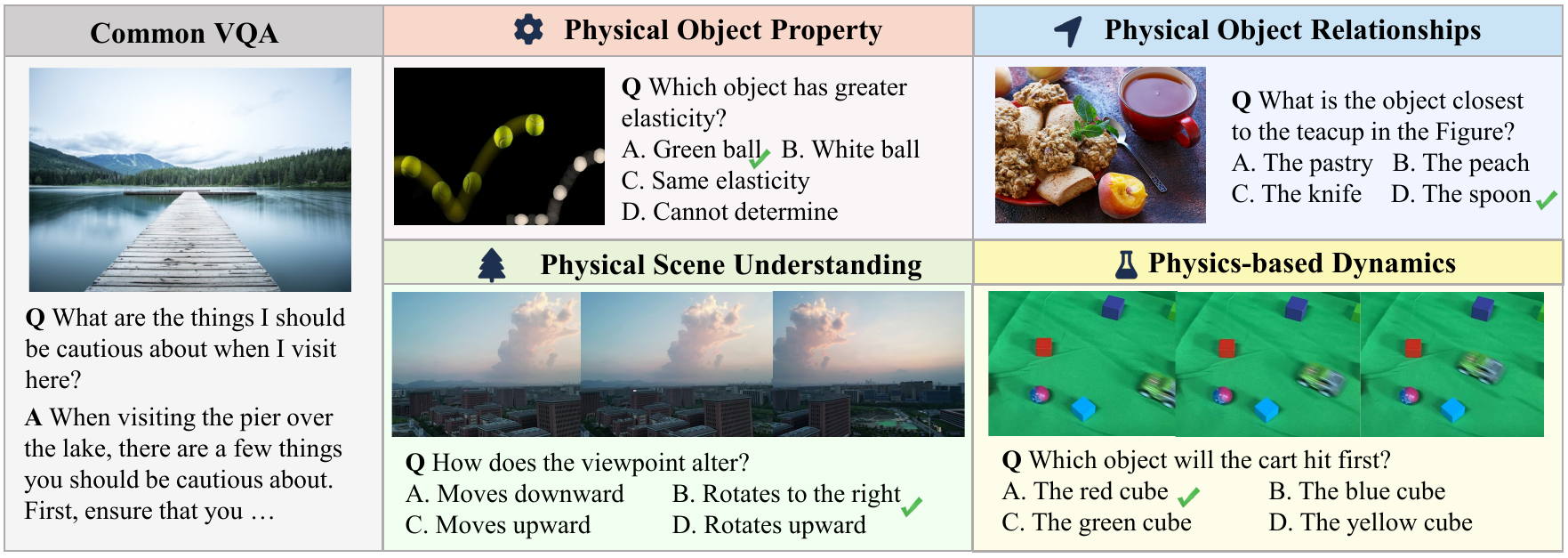}
	\vspace{-6mm}
        \caption{\textbf{Common VQA} tasks typically involve questions about visual content and general knowledge. \textbf{PhysBench} emphasizes understanding the physical world, encompassing 4 dimensions.}\label{fig:abstract}  
	\vspace{-4mm}
\end{figure}

To further investigate this issue, we pose \textbf{two fundamental questions}: \textit{(1) Do VLMs possess an understanding of the physical world, and if not, what factors contribute to this limitation?} \textit{(2) How can we enhance VLMs' physical world understanding capabilities and facilitate the effective deployment of embodied agents like MOKA}?

To answer the above questions and comprehensively assess the extent of the gap between VLMs and physical world understanding, we introduce PhysBench, a dataset comprising 10,002 interleaved video-image-text entries. 
Given the difficulty of acquiring such data, where expressing specific properties often requires multiple images, we undertook a five-step process, spending a total of 4,000 hours on annotation.
We systematically evaluate 75 representative VLM across four domains—physical object properties, physical object relationships, physical scene understanding, and physics-based dynamics—encompassing 19 sub-tasks, as shown in Figure~\ref{fig:abstract}.
Our extensive experiments reveal that (1) \textit{most current VLMs exhibit poor understanding of the physical world,} particularly in physical scene understanding and physics-based dynamics, with closed-source models significantly outperforming open-source ones; and (2) \textit{the training data for VLMs is likely a major factor contributing to their subpar performance}, as it often lacks the necessary physical knowledge. Notably, when VLMs were fine-tuned on our physically grounded data, their performance improved.

To further improve VLM's physical world understanding capabilities, we propose PhysAgent, a unified framework that incorporates vision foundation models and a physics knowledge memory. 
By analyzing the sources of errors for VLMs on PhysBench, we identified perceptual inaccuracies and insufficient knowledge as the primary causes of mistakes. To address these issues, we incorporated vision foundation models to enhance perceptual capabilities and assist VLMs in handling tasks they typically struggle with, such as depth estimation and numerical distance calculation. Additionally, we integrated a knowledge memory module to embed essential knowledge about the physical world, which can be selectively invoked by PhysAgent.
Unlike previous methods designed for physical reasoning~\citep{zheng2024contphy, physion++}, PhysAgent retains the strong generalization abilities of VLMs and their capacity to solve open-ended problems, without relying on manually predefined processing logic or being limited to specific tasks. Experimental results demonstrate that PhysAgent improves GPT-4o's zero-shot performance on PhysBench by 18.4\%.
Furthermore, we investigate how physical world understanding helps the deployment of embodied agents through extensive robotic manipulation experiments on MOKA~\citep{liu2024moka}. Specifically, we employ two approaches: fine-tuning the VLM with PhysBench and utilizing PhysAgent for zero-shot inference across five representative manipulation tasks.
The improvement in those tasks further validates that PhysBench and PhysAgent can facilitate the deployment of embodied agents like MOKA.

We hope this work offers valuable insights and contributes to bridging the gap between VLMs and physical world understanding, ultimately advancing embodied AI toward human-level capabilities. 
In summary, this paper has two technical contributions: (1) We present PhysBench, a large-scale benchmark for evaluating the performances of vision-language models in physical world understanding. We identify the key challenges through extensive studies and provide insights into why the existing VLMs have insufficient physical world understanding capabilities.
(2) We propose PhysAgent, a unified approach that improves VLMs' physical world understanding abilities. Through extensive experiments, we demonstrate that enhancing VLMs' comprehension of physical environments can significantly facilitate the deployment of embodied agents.
\clearpage
\section{Related Work}\label{sec: realate}
\textbf{Physical Comprehension Datasets}.
Early benchmarks~\citep{riochet2018intphys,rajani2020esprit} were developed primarily for vision-only models, while more recent efforts~\citep{yi2019clevrer,chen2022comphy,wang2024compositional} have predominantly focused on simple visual primitives, such as spheres, cubes, and rigid object collision events, often restricted to a limited set of simulated scenarios~\citep{zheng2024contphy, physion++}.
We summarize the key features of these various benchmarks and compare them against our benchmark in Table~\ref{table_benchmark_comparison_qa}.
However, existing VQA datasets assessing physical knowledge~\citep{lu2022learn, he2024olympiadbench} mainly focus on commonsense reasoning rather than physical world perception. Spatial VQA benchmarks~\citep{Chen_2024_CVPR, lyu2024mmscan, bonnen2024evaluating, wang2023embodiedscan} emphasize geometric relationships in 3D sence, which represent only a part of the physical.
In contrast, PhysBench is the first comprehensive dataset designed to evaluate models' understanding of the physical world, encompassing a wide variety of scenarios and tasks not covered by previous benchmarks.

\textbf{Physical Reasoning Models}.
Models for understanding the physical world generally fall into two categories. The first comprises physics-specialized models~\citep{guen2020disentangling, duan2022pip}, which are typically limited to predicting the next state and are not applicable to other tasks. The second includes physical oracle models~\citep{zheng2024contphy, physion++}, which are suitable for only a narrow range of tasks due to their reliance on predefined rules. These models often require training additional modules like R-CNN, and their probabilistic outputs restrict them to classification tasks, limiting their ability to handle open-ended questions. In contrast, PhysAgent offers greater flexibility and adaptability across a broader spectrum of problems without these limitations.

\textbf{Vision-Language Models}.
Vision-Language Models (VLMs) are large-scale models that integrate visual modalities with language understanding~\citep{Wu2023NExTGPTAM,Zhan2024AnyGPTUM,dai2024instructblip}. In recent years, there has been a surge of work leveraging VLMs as agents for embodied AI~\citep{liu2024moka, google2024pivot}. Although these approaches are generalizable, they face challenges due to weak physical world understanding capabilities~\citep{liu2024moka,Guo2024PhyGrasp}. 
By employing PhysBench and PhysAgent, these shortcomings can be mitigated, enhancing the physical world understanding capabilities of VLMs and enabling more reliable robotic control.
Additionally, spatial VLMs~\citep{bonnen2024evaluating} have identified that most VLMs lack 3D spatial reasoning capabilities due to insufficient data. However, since spatial reasoning represents only a subset of physical world understanding, our work aims to provide a more comprehensive evaluation and improvement of VLMs' physical world understanding abilities.
For additional related work, see Appendix~\ref{app_related_work}.

\begin{table}[h]
    \centering
    \small
    \vspace{-1mm}
    \caption{A comparison between PhysBench and other physical understanding question-answering benchmarks. PhysBench is a comprehensive dataset, covering a wide range of tasks related to physical world understanding.}
    \label{table_benchmark_comparison_qa}
    \vspace{-2mm}
    \resizebox{\textwidth}{!}{
\begin{tabular}{lcccccccccccrc}\toprule
           & Property & Attribute & Location & Motion & Temperature & Viewpoint & Light & Collision & Manipulation & Fluid & Interleaved & Size & More than cube \\\midrule
CLEVRER~\citep{yi2019clevrer}    & \checkmark& \crossmark& \crossmark & \crossmark & \crossmark  & \crossmark & \crossmark& \checkmark & \crossmark   & \crossmark& \crossmark   & 300,000 & \crossmark \\
Cater~\citep{girdhar2019cater}  & \checkmark& \crossmark& \crossmark & \crossmark & \crossmark  & \crossmark & \crossmark& \checkmark & \crossmark   & \crossmark& \crossmark   & 5,500 & \crossmark \\
CRIPP-VQA~\citep{patel2022cripp}  & \checkmark& \crossmark& \crossmark & \crossmark & \crossmark  & \crossmark & \crossmark& \checkmark & \crossmark   & \crossmark& \crossmark   & 5,000 & \crossmark \\
ComPhy~\citep{chen2022comphy}  & \checkmark& \crossmark& \crossmark & \crossmark & \crossmark  & \crossmark & \crossmark& \checkmark & \crossmark   & \crossmark& \crossmark   & 99,844 & \crossmark \\
EmbSpatial~\citep{du2024embspatial} & \crossmark & \crossmark& \checkmark& \crossmark & \crossmark  & \crossmark & \crossmark& \crossmark& \crossmark   & \crossmark& \crossmark   & 3,600 & \checkmark \\
Physion~\citep{physion}    & \checkmark& \crossmark& \crossmark & \crossmark & \crossmark  & \crossmark & \crossmark& \checkmark & \checkmark    & \crossmark& \crossmark   & 17,200 & \checkmark \\
Physion++~\citep{physion++}  & \checkmark& \checkmark & \crossmark & \crossmark & \crossmark  & \crossmark & \crossmark& \checkmark & \crossmark   & \crossmark& \crossmark   & 2,000 & \checkmark \\
ContPhy~\citep{zheng2024contphy}    & \checkmark& \crossmark& \crossmark & \crossmark & \crossmark  & \crossmark & \crossmark& \checkmark & \crossmark   & \checkmark& \crossmark   & 6,500 & \checkmark \\
SuperCLEVR~\citep{wang2024compositional} & \checkmark& \crossmark& \crossmark & \checkmark& \crossmark  & \crossmark & \crossmark& \checkmark & \crossmark   & \crossmark& \crossmark   & 1,200 & \crossmark \\
PhysBench  & \checkmark& \checkmark & \checkmark& \checkmark& \checkmark   & \checkmark & \checkmark& \checkmark & \checkmark    & \checkmark& \checkmark    & 10,002 & \checkmark \\\bottomrule              
\end{tabular}
    }
    \vspace{-4mm}
\end{table}

\section{PhysBench}\label{sec: datasets}
To assess VLMs' physical world understanding ability, we first define the concept of physical world understanding and introduce PhysBench in Section~\ref{sec3:1}. Next, we provide a detailed description of the data collection process in Section~\ref{sec3:2}.
Utilizing PhysBench, we conduct experiments to determine whether VLMs can effectively comprehend the physical world in Section~\ref{exp:main}. Finally, in Section~\ref{exp:task_analysis}, we discuss the potential reasons for poor performance.
\vspace{-2mm}
\subsection{Overview of PhysBench}\label{sec3:1}
\vspace{-2mm}
Understanding the physical world is essential yet fundamentally challenging for embodied AI, as systems must perceive, interpret, and predict the properties and dynamics of objects and environments. This involves comprehending object properties and relationships, interpreting environmental scenes, and anticipating interaction outcomes based on visual cues and core physical principles to ensure safe and effective operation.

\begin{figure}[t!] 
	\centering  
	\vspace{-8mm}
	\includegraphics[width=\linewidth]{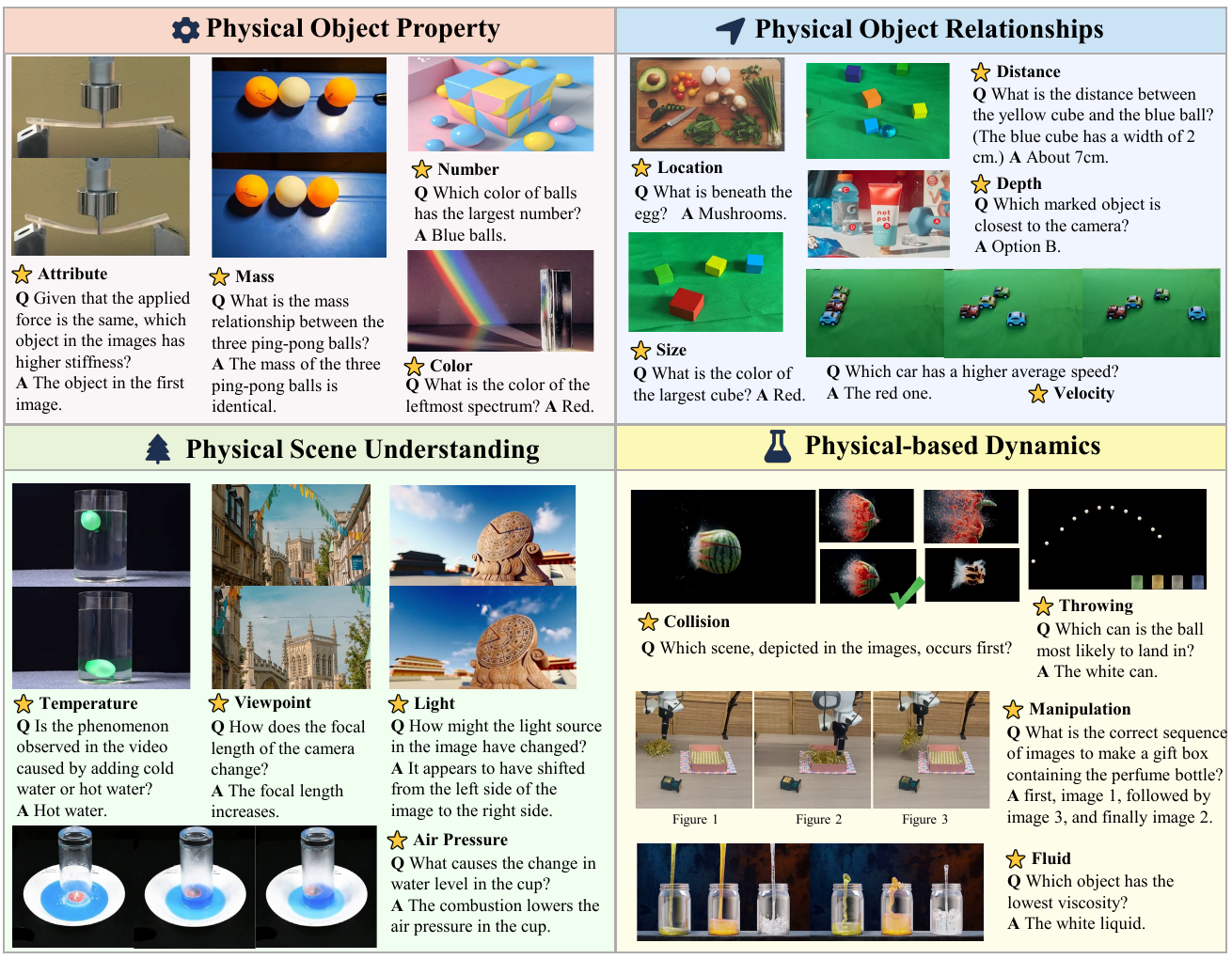}
	\vspace{-6mm}
        \caption{Sampled PhysBench examples from four major dimensions mentioned in Section~\ref{sec3:1}. 
        Due to space constraints, we present only the correct answers (as each question in our dataset is a four-option multiple-choice with one correct answer) and defer additional examples to Appendix~\ref{app:task_description}.} 
	\label{fig:data_cases_full}  
	\vspace{-6mm}
\end{figure}
\begin{figure}[!t]
\vspace{-2mm}
 \begin{minipage}{0.35\textwidth} 
     \centering
     \fontsize{6.2pt}{\baselineskip}\selectfont 
     \renewcommand\tabcolsep{0.9pt} 
     \renewcommand\arraystretch{0.7} 
             \scalebox{0.95}{
             \begin{tabular}{lc}
             \toprule
             \textbf{Statistic} & \textbf{Number} \\
             \midrule
              Total questions & 10,002 \\
              ~- only one image & 1,766 (18.6\%) \\
              ~- only one video & 2,749 (44.8\%) \\
              ~- interleave & 1,902 (20.1\%) \\
             \midrule
             Unique number of images & 10,058 \\
             Unique number of videos & 3,260 \\
             \midrule
             3D Assets & 678 \\
             \midrule
             Maximum question length & 48 \\
             Maximum choice length & 20 \\
             Average question length & 16.5 \\
             Average choice length & 4.4 \\
             \bottomrule
             \end{tabular}
     }
     \captionof{table}{Key statistics.}
     \label{tab:bench_statistics}
 \end{minipage} 
 \hfill
  \begin{minipage}{0.64\textwidth}
     \centering 
     \vspace{-1mm}
    \includegraphics[width=0.95\linewidth]{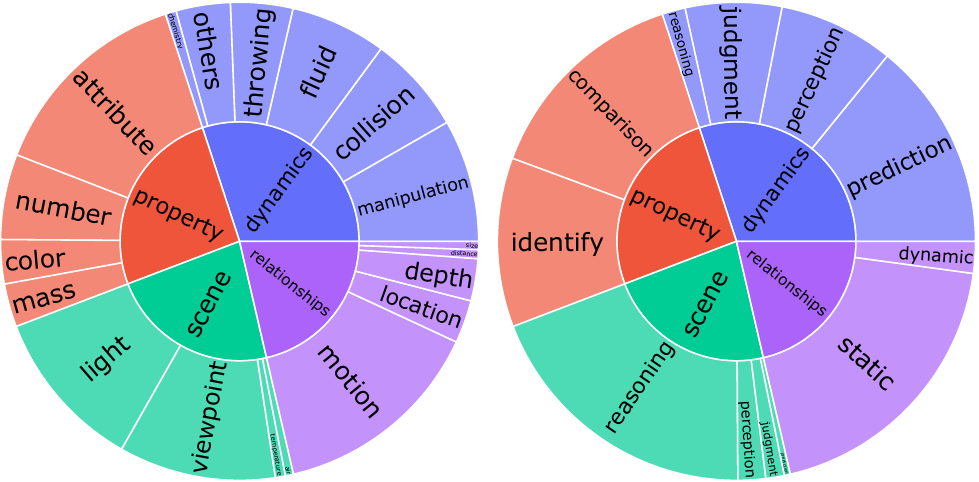}
    \vspace{-1mm}
     \caption{Subtype distribution and ability distribution}
     \label{fig:sub_dataset}
 \end{minipage}
 \vspace{-2mm}
\end{figure}

However, existing datasets often focus solely on image content and commonsense reasoning, neglecting the four fundamental aspects of the physical world mentioned above. To address this gap, we propose PhysBench, which comprehensively evaluates VLMs' perception of the physical world across four major task categories of the physical world:
\textit{(1) Physical Object Property}: Assessment of physical attributes of objects such as mass, size, density, tension, friction, bending stiffness, elasticity, and plasticity.
\textit{(2) Physical Object Relationships}: Evaluation of spatial relationships involving objects' relative or absolute positions and motions.
\textit{(3) Physical Scene Understanding}: Interpretation of environmental factors, including light sources, viewpoints, temperature, \textit{etc.}
\textit{(4) Physics-based Dynamics}: Understanding of physical events like collisions, throwing, fluid dynamics, explosions, and other phenomena. 
Each category corresponds to specific sub-task types and ability types, whose distributions are shown in Figures~\ref{fig:sub_dataset}. Detailed examples of specific tasks are illustrated in Figure~\ref{fig:data_cases_full}, with additional examples provided in Appendix~\ref{app:task_examples}. A comprehensive description of sub-task types and ability types is available in Appendix~\ref{app:task_description}.

PhysBench is structured as a multiple-choice questionnaire, presenting four options for each question, with only one correct answer.
The primary statistics of PhysBench are presented in Table~\ref{tab:bench_statistics} and detailed in Appendix~\ref{app_more_data_stat}.
Recognizing that different types of tasks possess unique characteristics, we utilize videos and multiple images to effectively convey features that are difficult to capture in a single image—such as elasticity, mass, density, and environmental factors like temperature, humidity, light source, and viewpoint. The dataset also includes objects with similar initial states but differing properties, leading to different future outcomes. This enriches the dataset and allows for a wider range of observable physical behaviors. Consequently, PhysBench draws its data from the internet, real-world captures, and simulations, making it a mixed-format benchmark that integrates text, images, and videos.
For convenience, PhysBench-test consists of 10,002 entries, organized into 19 subclasses, as the test set, and 200 entries as the validation set for parameter choosing. We also present 89,998 entries for further research. \textbf{The experimental results presented in this paper, unless otherwise specified, are based on the test set}. The performance of VLMs on PhysBench-val can be found in Appendix~\ref{exp_bench_val}. Benchmark release details can be found in Appendix~\ref{exp_bench_release}.

\subsection{Dataset Collection Process}\label{sec3:2}
To ensure data quality, all questions were manually annotated by graduate students in STEM fields and further refined through a rigorous review process after collecting and clipping the raw images or videos. To maintain consistency in annotations, we implemented multiple rounds of cleaning and validation throughout the following steps.
We have preserved intermediate outputs from the annotation process, such as depth and reflectance maps for simulator-generated data and human-annotated physical principles for many web-sourced videos. The process involves the following sequential steps: 
\textit{(a) Video Collection.} Videos and images are gathered from web searches, simulations, and real-world captures. The collection process uses predefined simulation rules, LLM-guided queries, and other strategies to find related images or videos (see Appendix~\ref{app:collection_process}). Human annotators further refine the data by clipping and annotating physical principles in the images or videos.
\textit{(b) Video Captioning.} Human-annotated raw videos are processed through automatic filtering, followed by GPT-4o annotations that generate captions with human check.
\textit{(c) Questions Design.} For videos annotated with physical principles, we generate physics-related questions using both manual design and GPT-4o, following predefined rules. An automated filter and manual review processes eliminate irrelevant questions.
\textit{(d) File Organization.} The remaining valid questions are categorized by task, sub-task, and ability type by human experts.
\textit{(e) Quality Check.} The organized dataset undergoes a human review to ensure that the questions are physical world relevant, rely on all input information, are not grounded in common sense, and are accurately categorized with clear questions and corresponding answers.
Due to space limitations, the collection guidelines are provided in Appendix~\ref{app:protocol}.

\subsection{Can VLMs Understand the Physical World}\label{exp:main}
To assess whether VLMs can understand the physical world, we evaluated 75 representative VLMs on PhysBench and found that a significant performance gap remains between VLMs and human-level understanding.
The primary results are presented in Table~\ref{main_experiment}, while detailed analyses of sub-task performance and ability types across the four task categories are provided in Appendix~\ref{exp_bench_details}.

\textbf{Setup}.
Our evaluation was conducted under three configurations: (a) \textit{Image VLMs}, which support only single-image input (e.g., LLaVA-1.5 and BLIP-2); (b) \textit{Video VLMs}, designed for video comprehension (e.g., Chat-UniVi and PLLaVA); and (c) \textit{General VLMs}, which support multiple images and interleaved inputs (e.g., VILA-1.5 and GPT-4o). It is important to note that the data used for evaluating setups (a) and (b) is a subset of PhysBench test subset with interleaved QA pairs removed, whereas setup (c) was evaluated on the full dataset.
For most models, we followed the standard protocol outlined in VLMEvalKit~\cite{contributors2023opencompass}, setting the temperature to 0. For models that do not support multiple images as input, we employed two methods: the \textit{merge} method, where video frames are concatenated into a single image~\citep{fu2024blink, zhang2024task, Jiang2024MANTISIM}, and the \textit{seq} method, where video frames are input sequentially as individual images. Notably, only models using the \textit{seq} setup can handle interleaved text-image sequences. For details on VLM prompts and hyperparameters, see Appendix~\ref{app:setup}.

\begin{figure}[!th]
    \centering
    \vspace{-2mm}
    \begin{subfigure}{0.4\textwidth}
        \centering
        \includegraphics[width=0.999\linewidth]{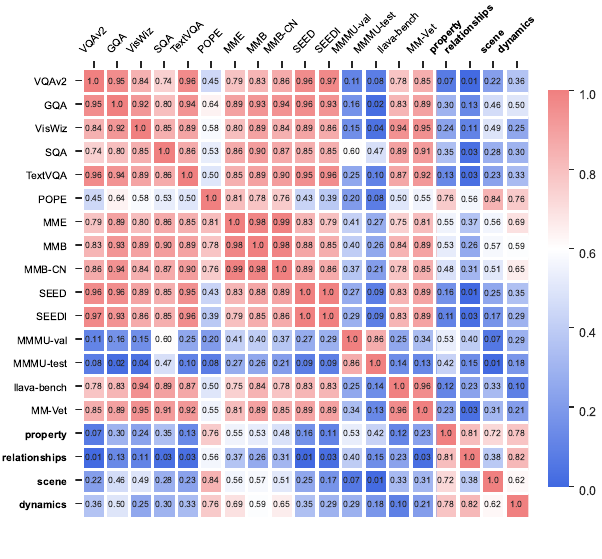}
    \end{subfigure}
    \hspace{.5mm}
    \begin{subfigure}{0.56\textwidth}
        \centering
        \includegraphics[width=0.999\linewidth]{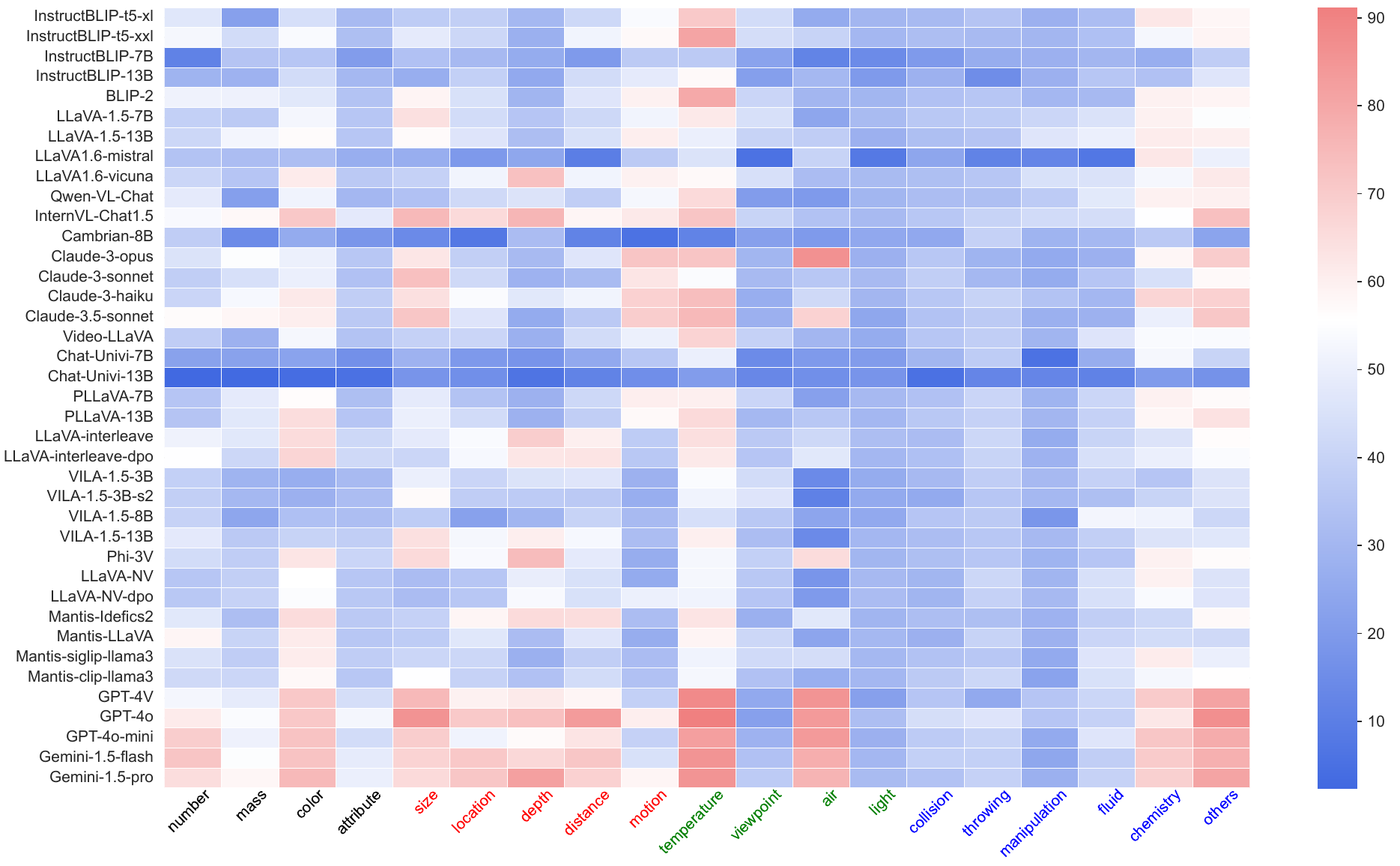}
    \end{subfigure}
    \vspace{-3mm}
    \caption{(a) Correlation map between 4 tasks in PhysBench and 15 other vision-language benchmarks. (b) The visualization of model performance across 19 sub-tasks is presented, where different colors represent the respective categories. The four colors, from left to right, represent physical object property, physical object relationships, physical scene, and physical-based dynamics.}\label{fig:main_vis}
\end{figure}

\begin{table}[th!]
    \setlength{\tabcolsep}{3pt}
    \centering
    \scalebox{0.78}{
\begin{tabular}{lccccccc} \hline
    & Size & Format & \faGear Property &  \faLocationArrow Relationships & \faTree Scene & \faFlask Dynamics & \textbf{Avg}\\\hline
    Random Choice&-&-&25.00&25.00&25.00&25.00&25.00\\\hline
    Human&-&-&97.10&95.67&94.91&95.68&95.87\\\hline
\multicolumn{7}{c}{Image VLM} \\\hline
InstructBLIP-t5-xl ~\citep{dai2024instructblip} &   4B   & merge& 35.35    & 36.67  & 37.45   & 35.95   &  36.24     \\
InstructBLIP-t5-xxl ~\citep{dai2024instructblip}  &   12B   &  merge & 41.11   & 38.47  & \underline{37.89}     & 36.42    & 38.51         \\
InstructBLIP-7B~\citep{dai2024instructblip}& 7B &  merge & 21.94 & 29.00                       & 19.53                           & 27.45                       & 23.82 \\
InstructBLIP-13B~\citep{dai2024instructblip}  & 13B &merge& 31.69 & 33.19 & 23.13 & 30.64  & 29.94 \\
BLIP-2~\citep{li2023blip} &  12B  &merge&41.70 & 40.83  & 36.25	 & 36.93& 38.61\\
LLaVA-1.5-7B~\citep{liu2023improved}&7B&merge&38.44&41.53&\textbf{38.60}&42.69&40.09\\
LLaVA-1.5-13B~\citep{liu2023improved}&13B &merge&41.31& 42.50& 34.40&\underline{44.38}& 40.45\\
LLaVA1.6-mistral~\citep{liu2024llavanext}&7B&merge&29.77&22.22&8.54&20.58&20.30\\
LLaVA1.6-vicuna~\citep{liu2024llavanext}&7B&merge& 40.26&\underline{59.72}&\textbf{38.60}&42.65&\underline{42.28}\includegraphics[scale=0.07]{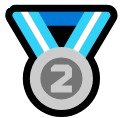}\\
Qwen-VL-Chat~\citep{bai2023qwen}&9B&merge&35.97&43.33&26.47&41.27&35.63\\
InternVL-Chat1.5~\citep{chen2024far}&26B&merge&\textbf{53.08}&\textbf{70.14}&37.01&\textbf{44.78}&\textbf{47.51}\includegraphics[scale=0.07]{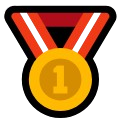}\\
Cambrian-8B~\citep{tong2024cambrian}&8B&merge&23.27&17.92&23.02&29.29&24.61\\ 
\hdashline
Claude-3-opus~\citep{anthropic_claude_models} &-&merge&41.97&40.97&30.63&36.50&37.00\\
Claude-3-sonnet~\citep{anthropic_claude_models} &-&merge&37.86&40.00&32.23&36.89&36.18 \\
Claude-3-haiku~\citep{anthropic_claude_models} &-&merge&43.28&53.33&30.06&39.93&39.44 \\
Claude-3.5-sonnet~\citep{anthropic_claude_models} &-&merge&\underline{46.46}&41.11&27.89&37.60&38.05\\ \hline
\multicolumn{7}{c}{Video VLM} \\\hline
Video-LLaVA~\citep{lin2023video}&7B&seq&36.82&\underline{36.11}&\underline{33.69}&\underline{40.52}&37.04 \\
Chat-Univi-7B~\citep{jin2023chatunivi}&7B&seq&19.28&20.97&18.86&28.46&22.19\\
Chat-Univi-13B~\citep{jin2023chatunivi}&13B&seq&4.30&11.53&15.67&11.47&10.36\\
PLLaVA-7B~\citep{xu2024pllava}&7B&seq&\textbf{38.02}&35.83&\textbf{36.34}&39.89&\textbf{37.94}\includegraphics[scale=0.07]{figures/icon/gold_medal.png}\\
PLLaVA-13B~\citep{xu2024pllava}&13B&seq&\textbf{39.91}&\textbf{38.33}&31.52&\textbf{40.76}&\underline{37.70}\includegraphics[scale=0.07]{figures/icon/sliver_medal.png}\\ \hline
\multicolumn{7}{c}{General VLM + Interleaved data} \\\hline
LLaVA-interleave~\citep{li2024llavanextinterleavetacklingmultiimagevideo}&7B&seq&47.23&44.62&35.64&37.21&41.00\\
LLaVA-interleave-dpo~\citep{li2024llavanextinterleavetacklingmultiimagevideo}&7B&seq& 47.97&42.67&33.73& 38.78&40.83\\
VILA-1.5-3B~\citep{lin2023vila}&3B&seq&32.40&33.02&34.84&35.78&34.11\\
VILA-1.5-3B-s2~\citep{lin2023vila}&3B&seq&33.14&30.26&35.72&33.00&33.07 \\
VILA-1.5-8B~\citep{lin2023vila}&8B&seq&33.41&29.88&30.85& 35.91&32.85 \\
VILA-1.5-13B~\citep{lin2023vila}&13B&seq& 40.53 & 40.15 & 31.96&36.07&37.15  \\
Phi-3V~\citep{abdin2024phi}&4B&seq&43.67&37.92&34.93&36.92&38.42\\
LLaVA-NV~\citep{zhang2024llavanextvideo}&7B&seq&38.33&30.83&34.00&37.17&35.42\\
LLaVA-NV-dpo~\citep{zhang2024llavanextvideo}&7B&seq&38.83&44.31&33.86&37.21&37.43\\
Mantis-Idefics2~\citep{Jiang2024MANTISIM}&8B&seq&41.97&41.44&29.53&36.56&37.39\\
Mantis-LLaVA~\citep{Jiang2024MANTISIM}&7B&seq&44.48&30.45&36.25&34.73&36.69\\
Mantis-siglip-llama3~\citep{Jiang2024MANTISIM}&8B&seq&42.47&32.78&\textbf{36.83}&37.51&37.64\\
Mantis-clip-llama3~\citep{Jiang2024MANTISIM}& 8B&seq&40.61&35.11&32.45&38.36&36.92\\ 
\hdashline
GPT-4V~\citep{achiam2023gpt}&-&seq&49.59&45.77&26.34&42.15&41.26\\
GPT-4o~\citep{achiam2023gpt}&-&seq&56.91&\textbf{64.80}&30.15&\textbf{46.99}&\textbf{49.49}\includegraphics[scale=0.07]{figures/icon/gold_medal.png}\\
GPT-4o-mini~\citep{achiam2023gpt}&-&seq&53.54&44.24&30.59&\underline{42.90}&43.15\\
Gemini-1.5-flash~\citep{team2023gemini}& -  &seq&  \textbf{57.41}     &  52.24    & 34.32         & 40.93& 46.07 \\
Gemini-1.5-pro~\citep{team2023gemini} & -  &seq&  \underline{57.26} & \underline{63.61}  & \underline{36.52}   &41.56&\underline{49.11}\includegraphics[scale=0.07]{figures/icon/sliver_medal.png}\\\hline
  \end{tabular}
    }
    \vspace{-3mm}
    \caption{\textbf{Evaluation results for 39 VLMs.} The evaluation of General VLMs is based on the data from Video and Image VLM evaluations, with the addition of interleaved data. 
    ``Seq`` refers to sequential input of frames of videos, while ``merge`` refers to merging video frames into a single image.
    \textbf{Bold} indicates the best result, and \underline{underline} indicates the second best in each group.
    }\label{main_experiment}
    \vspace{-4mm}
\end{table}

\textbf{VLMs exhibit a limited understanding of the physical world}. Our evaluation indicates that most models achieve an average accuracy of approximately 40\%, which is significantly below human-level performance. Even the best-performing model, GPT-4o, attains only 49.49\% accuracy, underscoring the substantial gap between current VLMs and true comprehension of the physical world. As shown in Figure~\ref{fig:main_vis}(b), considerable room for improvement remains, particularly in tasks related to physical scene understanding and physics-based dynamics.

\begin{figure}[!th]
    \centering
    \vspace{-2mm}
    \begin{subfigure}{0.42\textwidth}
        \centering
        \includegraphics[width=0.87\linewidth]{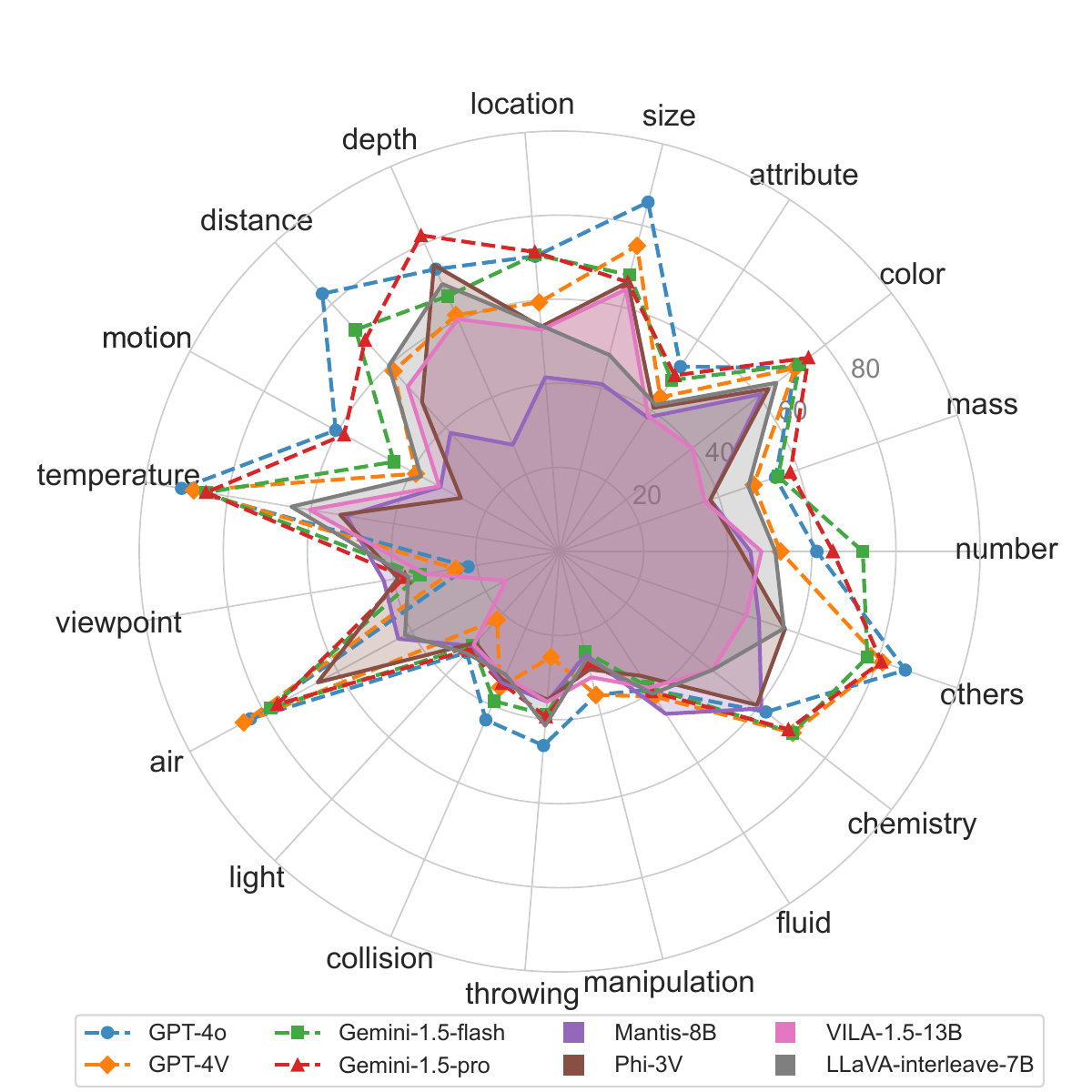}
    \end{subfigure}
    \hspace{.5mm}
    \begin{subfigure}{0.51\textwidth}
        \centering
        \includegraphics[width=\linewidth]{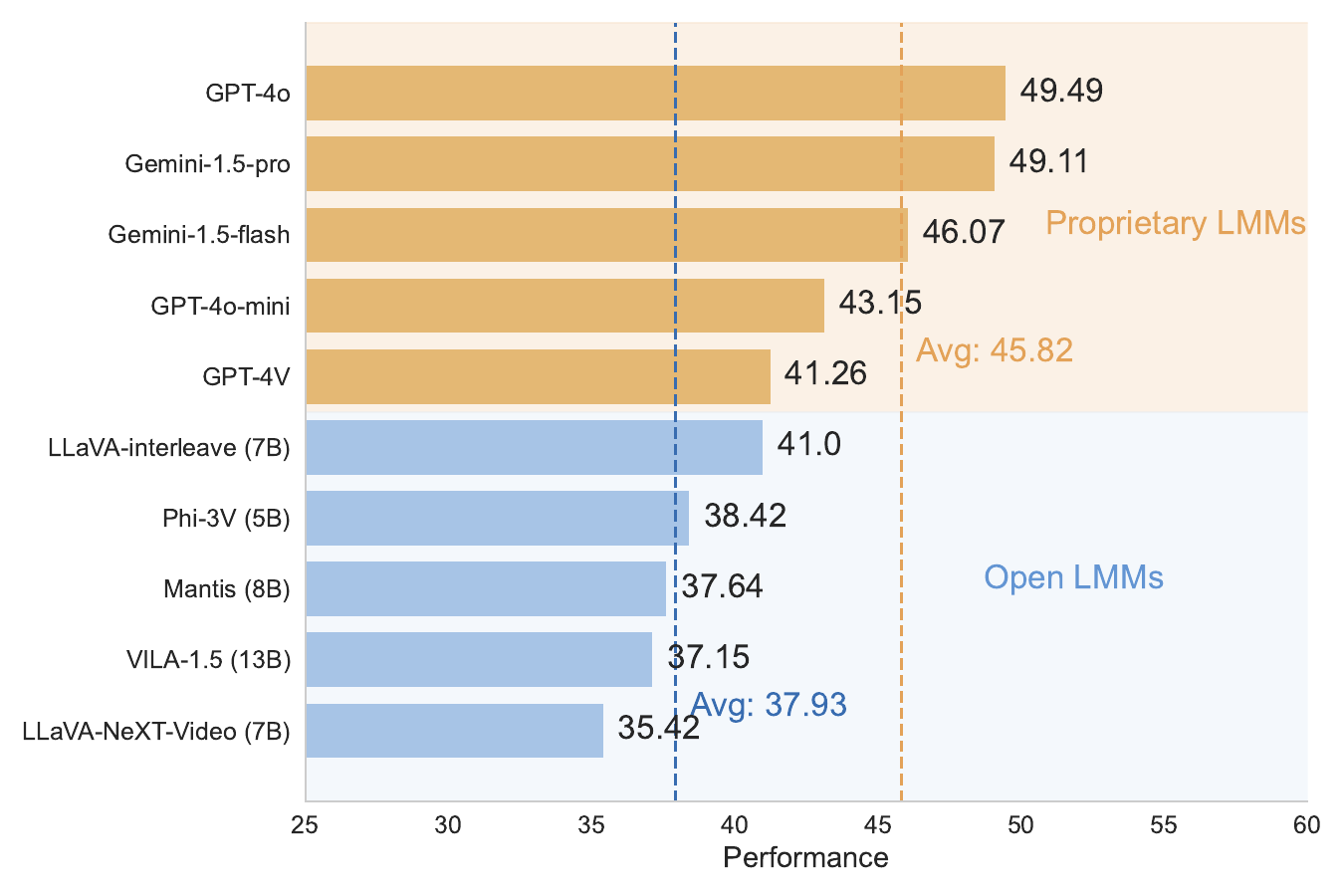}
    \end{subfigure}
    \vspace{-2mm}
    \caption{(a) The performance of 8 representative open-source General VLMs across 19 sub-tasks in PhysBench, which support interleaved inputs. The closer it is to the circular boundary, the better. (b) The overall performance of those 8 VLMs. Closed-source models generally perform better.}
    \label{fig:intro}
    \vspace{-6mm}
\end{figure}

\textbf{Closed-source models generally perform better}. As shown in Figure~\ref{fig:intro}(b), the GPT series and Gemini-1.5 models significantly outperform open-source models. Notably, GPT-4 surpasses the best open-source model, LLaVA-interleave, by 20.7\%, indicating a substantial gap between open-source and closed-source models. However, we did not observe a clear advantage with Claude, a finding that aligns with results from other benchmarks~\citep{cao2024visual, wu2024vsp}.

\subsection{Why Do VLMs Struggle with Physical World Understanding}\label{exp:task_analysis}
To further investigate why VLMs struggle with physical world understanding, we analyzed PhysBench and discovered that it differs significantly from common VQA tasks. Additionally, we found that the performance of larger model size or more training data does not result in clear improvements on PhysBench, which may be \textit{due to a lack of physical world knowledge in the training data}. Furthermore, we found that many errors stem from this deficiency; when we augmented the models with physical world knowledge, their performance improved. This further suggests that the gap between VLMs and physical world understanding may be attributed to limitations in the training data.

\textbf{Physical world understanding differs significantly from common VQA tasks}.
To assess the relationships between our tasks and other VLM benchmarks, we adopted the methodology proposed by \citep{tong2024cambrian, fang2024exploring} to construct a correlation map, as shown in Figure~\ref{fig:main_vis}(a). Details on the construction of the correlation map are provided in Appendix~\ref{app:c_map}.
Our analysis reveals that PhysBench differs significantly from traditional VLM benchmarks, exhibiting closer alignment with POPE~\citep{li2023evaluating} in tasks such as hallucination detection, while also showing that performance does not consistently improve with increased data or model scale.

\begin{figure}[!b]
    \centering
    \vspace{-4mm}
    \begin{subfigure}{0.32\textwidth}
        \centering
        \includegraphics[width=\linewidth]{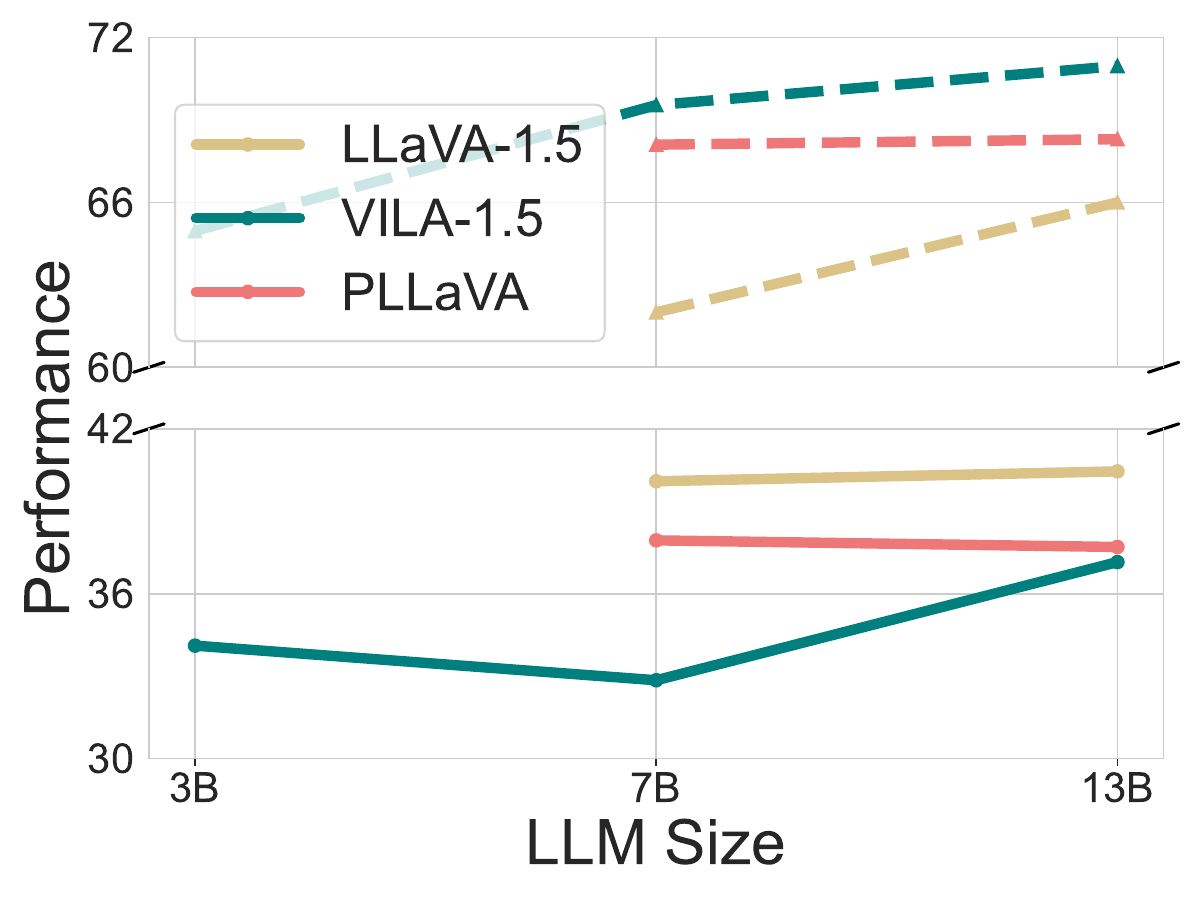}
    \end{subfigure}
    \hspace{.5mm}
    \begin{subfigure}{0.32\textwidth}
        \centering
        \includegraphics[width=\linewidth]{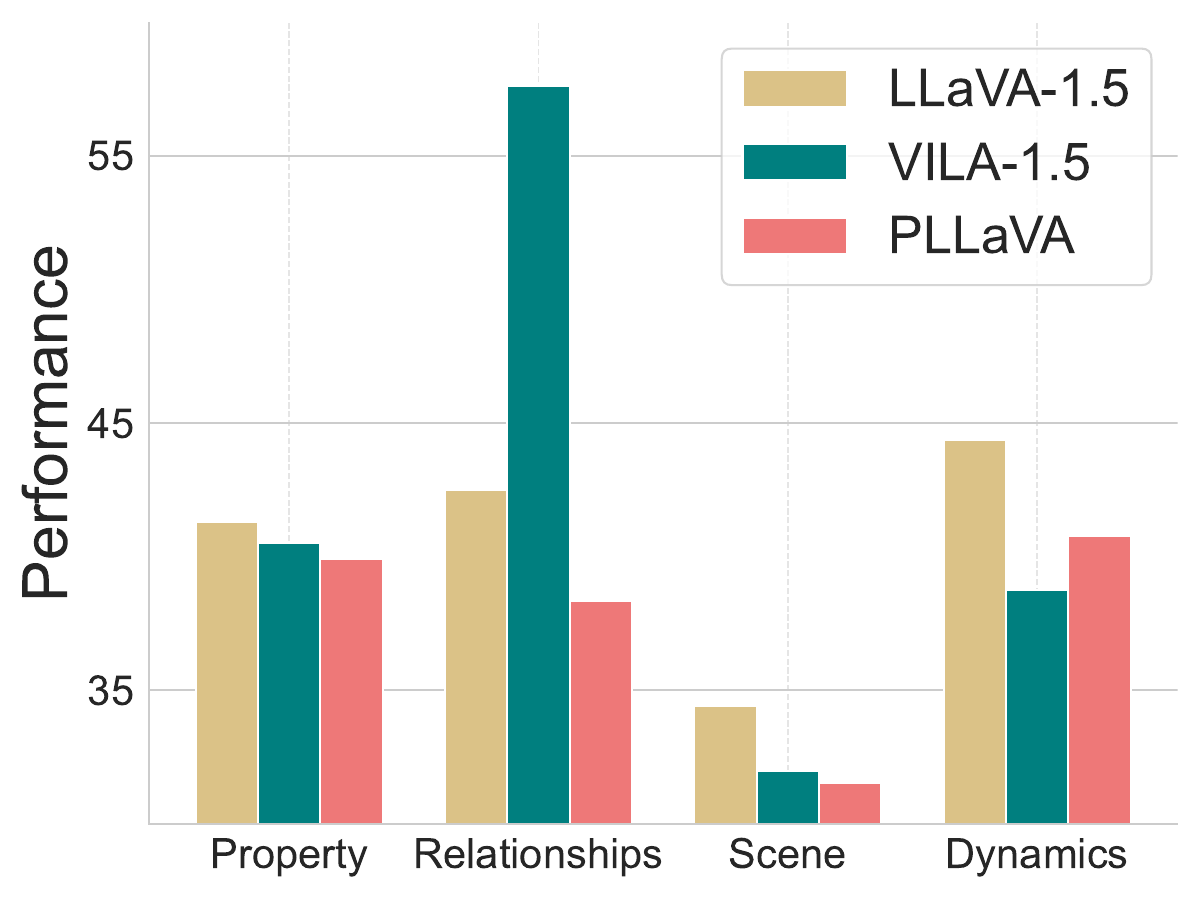}
    \end{subfigure}
        \hspace{.5mm}
    \begin{subfigure}{0.32\textwidth}
        \centering
        \includegraphics[width=\linewidth]{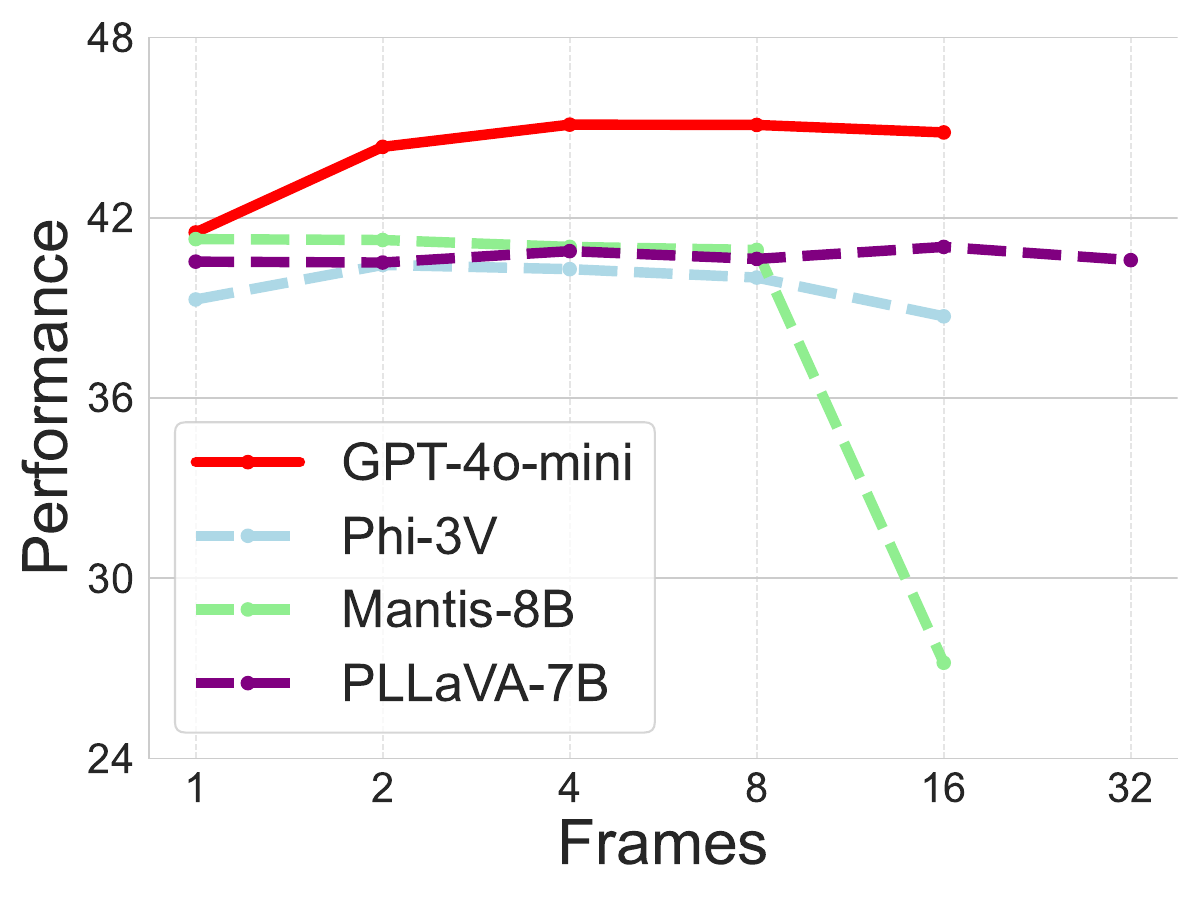}
    \end{subfigure}
    \vspace{-2mm}
    \caption{(a) Model size scalability.
    The solid line shows the average performance across 14 common QA tasks (Table~\ref{exp:c_map}), while the dashed line represents PhysBench results.
    (b) Data scalability. VILA and PLLaVA expand upon LLaVA's architecture by utilizing more data.
    (c) Frame scalability.}\label{fig:scaletable}
    \vspace{-6mm}
\end{figure}

\textbf{VLMs's physical world understanding ability does not scale with model size, data, or frames}.
\textit{(1) Model Size Scalability}. 
Figure~\ref{fig:scaletable}(a) shows that increasing model size using the same dataset significantly enhances performance on common QA tasks. However, this improvement does not extend to PhysBench, where gains are limited or even negative. For instance, while VILA-1.5's performance improves by 7.1\% on common QA tasks when scaling from 3B to 7B parameters, it decreases by 3.8\% on PhysBench.
\textit{(2) Data Scalability}.
As shown in Figure~\ref{fig:scaletable}(b), scaling up the dataset offers limited benefits for physical comprehension. PLLaVA and VILA-1.5, larger-data variants of LLaVA-1.5, exhibit minimal improvement or even a decline in performance on PhysBench compared to LLaVA-1.5. Analysis of the additional data (Appendix~\ref{app:word_stat}) reveals it is predominantly descriptive, focusing on content description rather than enhancing physical understanding. Nevertheless, VILA-1.5's spatial reasoning abilities have significantly improved, aligning with trends observed in other benchmarks~\citep{yu2023mmvet, li2023seed}.
\textit{(3) Frame Scalability}.
Figure~\ref{fig:scaletable}(c) indicates that the three open-source models are insensitive to the number of frames, performing similarly to single-frame inputs, with performance sometimes decreasing as frames increase. This suggests that current models cannot effectively utilize multi-frame information. Notably, increasing the number of frames led Mantis to frequently fail to follow instructions or refuse to answer, and expanding beyond eight frames did not yield further improvements.

\textbf{Perceptual and knowledge gaps constitute the majority of errors}.
To investigate the poor performance of VLMs on PhysBench, we randomly selected 500 questions and obtained explanations from three models—GPT-4o, Phi-3V, and Gemini-1.5-flash. Expert annotators classified the root causes of the mispredictions into six categories: perception errors, reasoning errors, lack of knowledge, refusal to answer, failure to follow instructions, and annotation errors in the dataset. 
The distribution of these error types is shown in Figure~\ref{fig:dist_error}, with selected cases and detailed analyses provided in Appendix~\ref{append:error_case}. 
The error distribution reveals that perceptual errors account for 37\%, 40\%, and 45\% of the mistakes made by GPT-4o, Gemini-1.5-flash, and Phi-3V, respectively, while lack of knowledge constitutes 34\%, 35\%, and 23\% of errors for these models. This analysis suggests that perceptual errors and knowledge gaps are the primary sources of mispredictions, indicating that while the models are adept at extracting information from text and visual inputs, their physical world understanding and complex reasoning abilities remain limited.

\begin{figure}[th!]
	\centering  
	\vspace{-2mm}
	\includegraphics[width=0.96\linewidth]{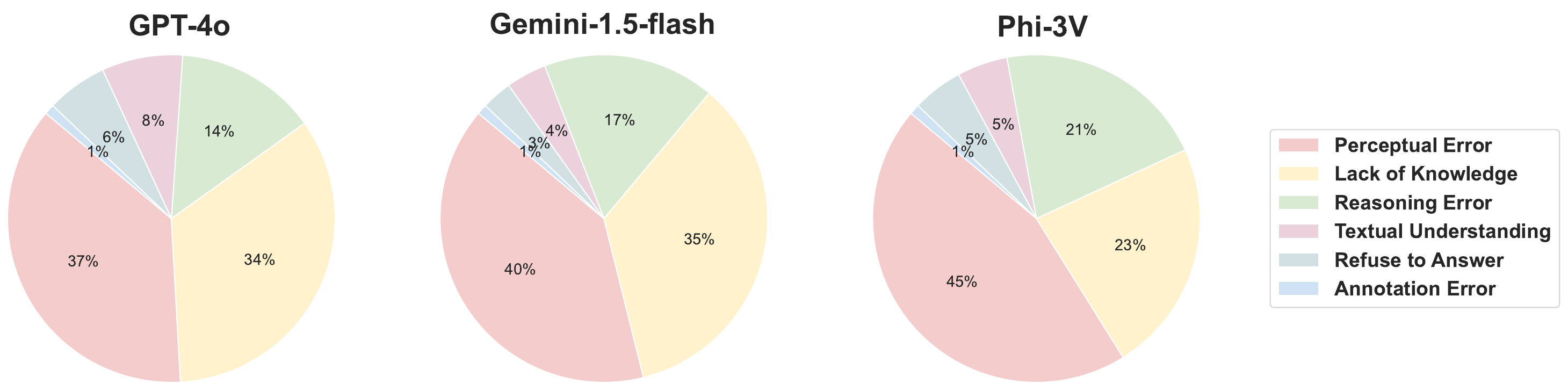}
	\vspace{-2mm}
    	\caption{Distribution of error types for GPT-4V, Gemini-1.5-flash, Phi-3V.}
	\label{fig:dist_error}  
	\vspace{-2mm}
\end{figure}

\textbf{Can VLMs transfer physical world knowledge?}
Our error analysis revealed that inadequate physical world knowledge and reasoning capabilities were key contributors to the models' poor performance. To investigate whether introducing additional examples could enhance performance, we conducted tests on 200 entries of PhysBench, pairing each with a similar example. These additional examples were incorporated through fine-tuning or in-context learning. As shown in Figure~\ref{fig:combine_three}(b), the performance improvements after adding physical world knowledge examples indicate that VLMs can transfer physical knowledge to some extent. This suggests that the original data’s lack of physical world knowledge was a significant factor in the models' suboptimal performance.
\section{PhysAgent}\label{sec: experiment}
Recognizing perceptual inaccuracies and knowledge gaps as key sources of error shown in Section~\ref{exp:task_analysis}, we introduce PhysAgent in Section~\ref{sec:phsagent} to improve VLMs' understanding of the physical world by integrating vision foundation models for enhanced perception and incorporating physical knowledge memory. 
To verify whether enhancing VLMs' physical understanding facilitates the deployment of embodied agents, we conducted five embodied agent tasks as detailed in Section~\ref{exp:robot}.

\subsection{How to Enhance VLMs for Physical World Understanding}\label{sec:phsagent}
We propose PhysAgent, a novel framework that integrates knowledge memory and vision foundation models to enhance physical world understanding in VLMs. 
This framework is inspired by our findings in Section~\ref{exp:task_analysis}, where we identified perceptual errors and insufficient knowledge as the primary causes of mistakes in VLMs. 
To address these shortcomings, we establish a \textit{knowledge memory} that provides prior physical world knowledge and rules. Additionally, we utilize vision \textit{foundational models} namely Depth Anything~\citep{depth_anything_v2}, SAM~\citep{kirillov2023segment}, and GroundingDINO~\citep{liu2023grounding} to achieve enhanced visual perception.
These models enable us to identify object types and spatial locations, and further acquire information about objects' dynamics through VLM reasoning or retrieval from memory. They also help solve problems that VLMs cannot address, such as estimating depth and numerical distances.
Unlike prior physical reasoning models that are confined to specific tasks and struggle to adapt to natural language queries, our method aims to fully leverage the reasoning and generalization capabilities of VLMs. Experiments on PhyBench show that PhysAgent improves performance by 18.4\% on GPT-4o. 

\begin{figure}[th!]
	\centering  
	\vspace{-1mm}
	\includegraphics[width=0.99\linewidth]{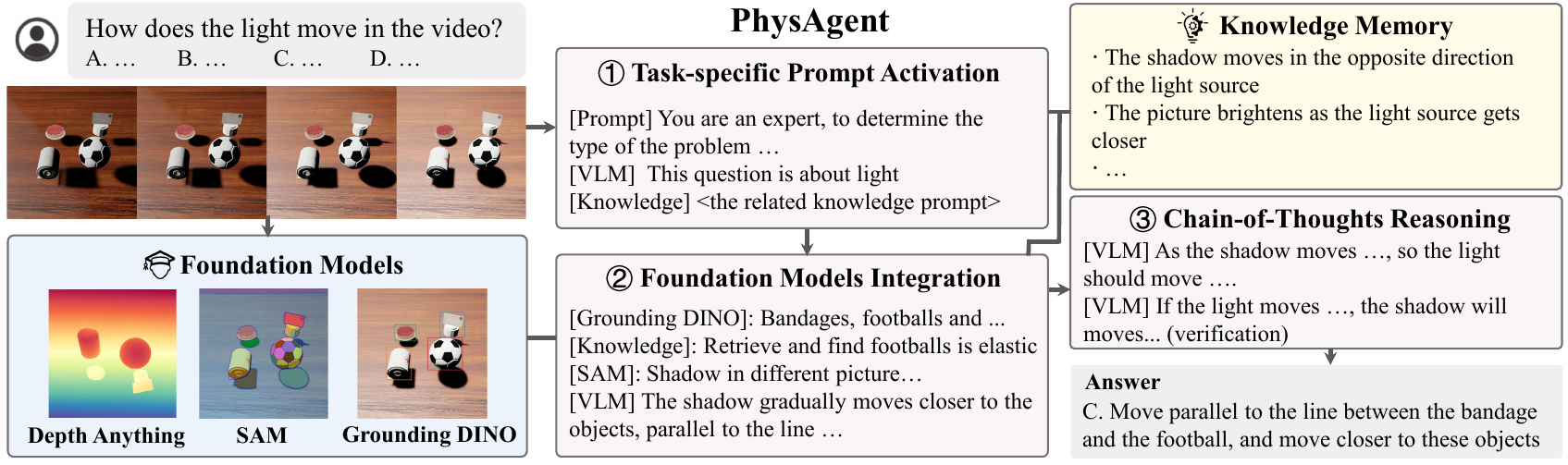}
	\vspace{-2mm}
    	\caption{Architecture of PhysAgent. 
     PhysAgent employs a three-step reasoning process to address the problem: activating task-specific prompts, integrating foundation models, and reasoning.}
	\label{fig:physagent_arch}  
	\vspace{-1mm}
\end{figure}

As illustrated in Figure~\ref{fig:physagent_arch}, given a question, PhysAgent follows three key steps:
(1) \textit{Task-specific Prompt Activation}: PhysAgent first classifies the question (manually or automatically) and activates task-specific prompts, incorporating relevant physical knowledge for different tasks. For instance, for a question about light, it retrieves knowledge on the relationship between light source movement and shadow direction to assist the VLMs.
(2) \textit{Foundation Models Integration}: PhysAgent processes the foundation model's outputs, leveraging VLM reasoning capabilities. For example, it identifies objects in the scene using GroundingDINO and retrieves relevant attributes from the knowledge memory.
(3) \textit{Chain-of-Thoughts Reasoning}: PhysAgent then engages in chain-of-thought reasoning, conducting a self-verification step to ensure logical consistency before providing the final answer.

\textbf{Baselines}. We utilized three prompt methods: Chain of Thought (CoT)~\citep{kojima2023largelanguagemodelszeroshot}, Desp-CoT~\citep{wu2023rolechainofthoughtcomplexvisionlanguage}, and Pure Language Reasoning (PLR), in addition to an oracle method, ContPhy~\citep{zheng2024contphy}, which served as our baseline. Detailed descriptions of the prompt strategies and the implementation of ContPhy are provided in Appendix~\ref{app:prompt_str} and Appendix~\ref{app:vipergpt}. 

\begin{figure}[!t]
    \centering
    \vspace{-9mm}
    \begin{subfigure}{0.28\textwidth} 
        \centering
        \fontsize{5.8pt}{\baselineskip}\selectfont 
        \renewcommand\tabcolsep{1.05pt} 
        \renewcommand\arraystretch{0.65} 
        \scalebox{0.88}{
            \begin{tabular}{lcccc}
                \hline
                & Property & Relationships & Scene & Dynamics \\\hline
                Phi-3V       & 43.6   & 37.9    & 34.9        & 36.9      \\\hdashline
                + CoT        & 42.5   & 34.5    & 29.8        & 36.7      \\
                + Desp-CoT   & 42.7   & 35.3    & 36.2        & 34.9      \\
                + PLR        & 38.6   & 32.6    & 36.7        & 34.0      \\
                + \textbf{PhysAgent}        & \textbf{44.5}   & \textbf{47.0}    & \textbf{38.6}        & \textbf{37.1}     \\\hline
                ContPhy      & 52.1   & 52.9    & 37.2        & 42.8      \\\hline
                GPT-4o       & 56.9   & 64.8    & 30.1        & 46.9      \\\hdashline
                + CoT        & 58.6   & 70.5    & 36.0        & 47.7      \\
                + Desp-CoT   & 57.7   & 64.0    & 36.9        & 46.2      \\
                + PLR        & 51.5   & 45.6    & 31.1        & 33.6      \\
                + \textbf{PhysAgent}        & \textbf{58.4}   & \textbf{84.2}    & \textbf{45.0}        & \textbf{51.3}    \\\hline  
            \end{tabular}
        }
        \caption{}
    \end{subfigure} 
    \hfill
    \begin{subfigure}{0.36\textwidth}
        \centering
        \includegraphics[width=\linewidth]{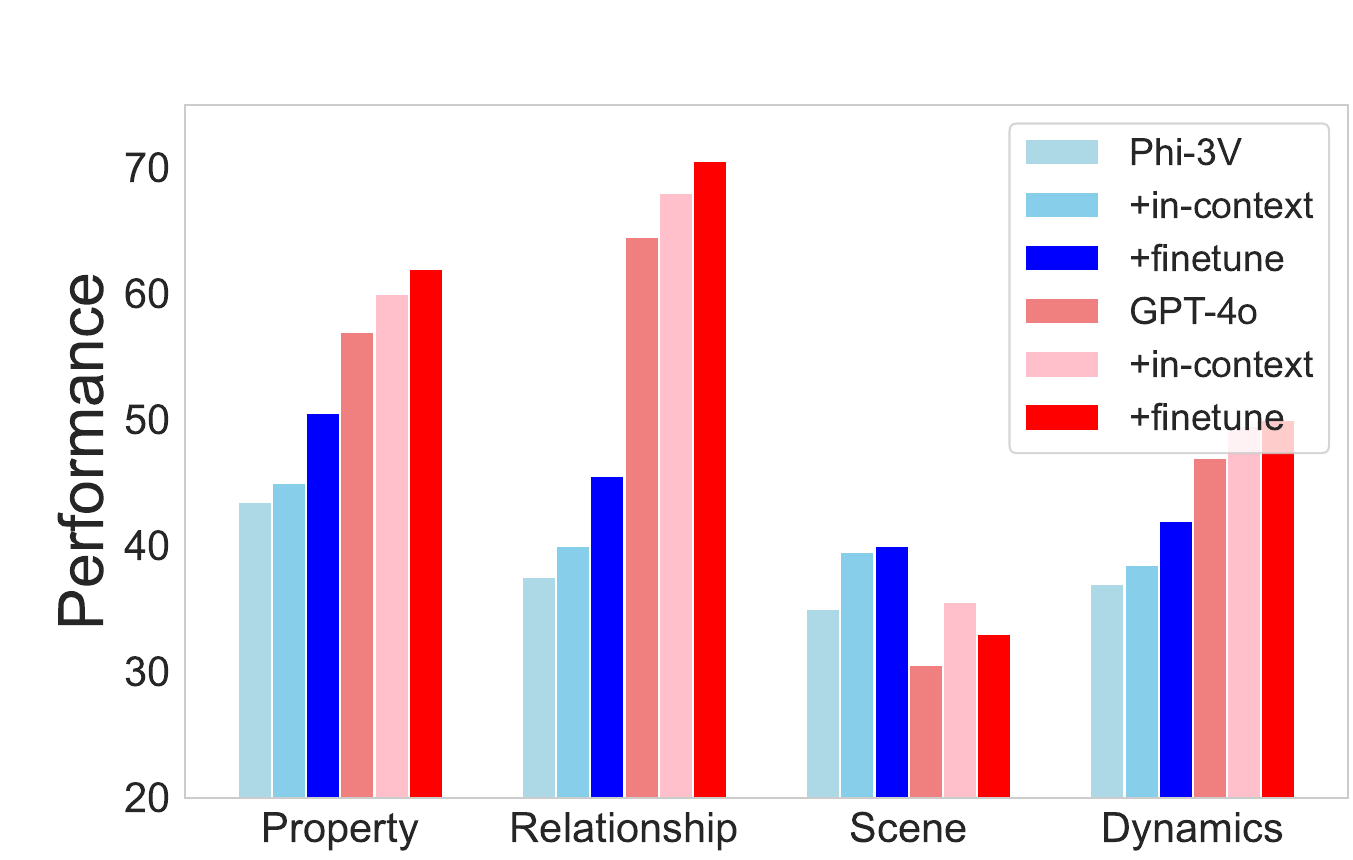}
        \caption{}
    \end{subfigure}
    \hfill
    \begin{subfigure}{0.33\textwidth}
        \centering
        \fontsize{5.8pt}{\baselineskip}\selectfont 
        \renewcommand\tabcolsep{1.05pt} 
        \renewcommand\arraystretch{0.5} 
        \scalebox{0.85}{
            \begin{tabular}{lccccc}
                \toprule
                & Affordance & Force & Color & Location & Tool \\
                \midrule
                \multicolumn{6}{l}{\quad\textbf{MOKA}} \\
                reasoning error$\downarrow$ & 0.3 & 0.5 & 0.1 & 0.1 & 0.3 \\
                execution error$\downarrow$ & 0.1 & 0.3 & 0.2 & 0.2 & 0.3 \\
                success$\uparrow$         & 0.6 & 0.2 & 0.7 & 0.7 & 0.4 \\
                \midrule
                \multicolumn{6}{l}{\quad\textbf{+ PhysAgent}} \\
                reasoning error$\downarrow$ & 0.0 & 0.2 & 0.0 & 0.1 & 0.2 \\
                execution error$\downarrow$ & 0.2 & 0.3 & 0.2 & 0.2 & 0.3 \\
                success$\uparrow$          & 0.8 & 0.5 & \textbf{0.8} & 0.7 & 0.5 \\
                \midrule
                \multicolumn{6}{l}{\quad\textbf{+ Fine-tune}} \\
                reasoning error$\downarrow$ & 0.0 & 0.1 & 0.1 & 0.0 & 0.0 \\
                execution error$\downarrow$ & 0.1 & 0.3 & 0.1 & 0.2 & 0.3 \\
                success$\uparrow$          & \textbf{0.9} & \textbf{0.6} & \textbf{0.8} & \textbf{0.8} & \textbf{0.7} \\
                \bottomrule
            \end{tabular}
        }
        \caption{}
    \end{subfigure}
    \vspace{-2mm}
    \caption{(a) Performance comparison of various methods. (b) Analysis of physical world knowledge transfer. (c) Performance evaluation across five embodied tasks as described in Figure~\ref{fig:embodied_agent}.}
    \label{fig:combine_three}
    \vspace{-4mm}
\end{figure}

\textbf{Results}.
The results in Figure~\ref{fig:combine_three}(a), lead to the following conclusions: 
\textit{(1) Prompting methods are unstable, and using pure language yields catastrophic results}. As observed, the CoT strategy has minimal impact, while both Desp-CoT and PLR show a decline in performance. This suggests that descriptive prompts are not particularly effective for addressing the questions, implying that our dataset requires a deeper understanding of the videos or images to answer accurately.
\textit{(2) ContPhy even worsens performance}. In three out of four tasks, ContPhy underperforms compared to its base model, GPT-4o, due to suboptimal module invocation and limited flexibility in its logical templates, which struggle to adapt to diverse scenarios. Additionally, ContPhy relies on models like RCNN to process visual information instead of directly leveraging GPT-4o, leading to potential information loss and subsequent performance degradation.
\textit{(3) PhysAgent consistently improves zero-shot performance}, notably achieving a 49.5\% improvement for GPT-4o in Scene. Compared to the CoT, Desp-CoT, and PLR prompting strategies, our method demonstrates consistent improvements.

\subsection{Can Physical World Understanding Help in Embodied Applications}\label{exp:robot}
Despite gaining significant attention in recent years for their strong generalization capabilities, VLMs as embodied agents~\citep{liu2024moka} still exhibit fundamental operational errors during physical world interactions. In this section, we investigate whether enhancing the physical world perception abilities of VLMs can improve their performance in downstream embodied agent tasks.

\begin{figure}[!th]
  \vspace{-1mm}
 \centering
 \begin{subfigure}[c]{0.46\textwidth} 
     \centering
     \fontsize{7.3pt}{\baselineskip}\selectfont
     \renewcommand\tabcolsep{1.2pt}
     \renewcommand\arraystretch{0.9}
     \scalebox{0.99}{
     \begin{tabular}{lp{5cm}}\hline
     \textbf{Affordance} & Grasp pot, knife, spoon, monitor, tennis racket.      \\
     \textbf{Force} & Grasp fragile items (egg, ripe persimmon), soft items (jelly, plastic cup), and rigid objects (iron ball). \\
     \textbf{Color}      & Grasp the specific color cube.   \\
     \textbf{Location}     & Grab the cube in specific location and move it to the plate. \\
     \textbf{Tool}       & Grasp specific tools, depending on the problem.  \\\hline
     \end{tabular}
     }
     \caption{}
 \end{subfigure} 
 \hfill
 \begin{subfigure}[c]{0.53\textwidth}
     \centering
     \includegraphics[width=0.96\linewidth]{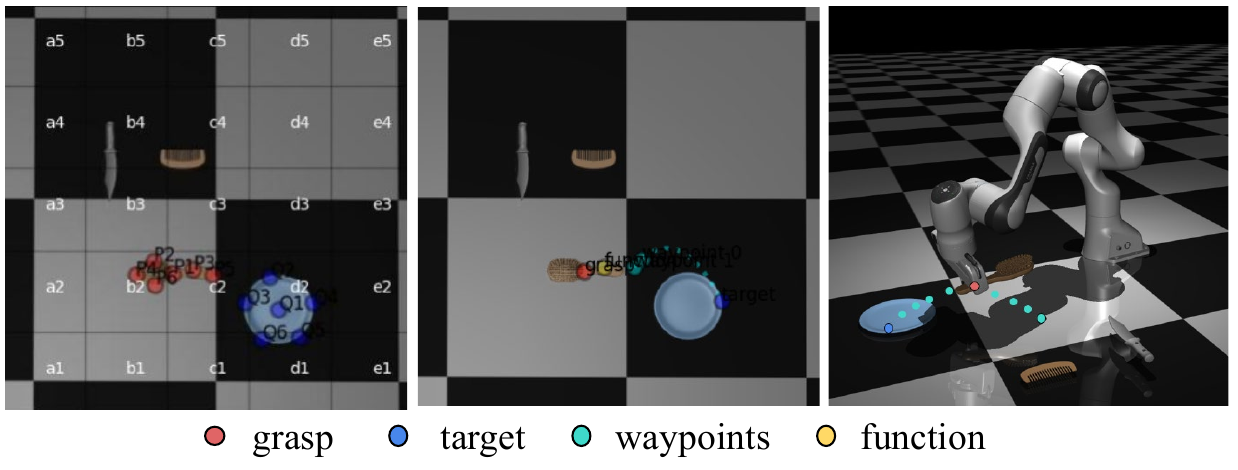}
     \caption{}
 \end{subfigure}

\vspace{-2mm}
\caption{(a) Description of each of the testing tasks. (b) Marked observation, predicted affordances, and motion in MOKA. MOKA leverages a VLM to generate key points and waypoints, and then converts these affordance representations into executable motions for the robotic arm.}
\label{fig:embodied_agent}
\end{figure}

To evaluate embodied agents, we designed five fundamental manipulation tasks as shown in Figure~\ref{fig:embodied_agent}(a).
The specifics of these tasks, along with the corresponding testing methods and language instructions, can be found in Appendix~\ref{app_robot_task_descri}.
These tasks require the agents to possess a basic understanding of spatial relations and the physical properties of objects.
Specifically, we utilized MuJoCo~\citep{todorov2012mujoco} and the 7-DoF Franka Emika robotic arm from Menagerie~\citep{menagerie2022github}, building our simulation platform based on MOKA~\citep{liu2024moka} as the embodied agent approach to test these embodied tasks. The VLM we used in these tasks is GPT-4o.

As illustrated in Figure~\ref{fig:embodied_agent}(b), MOKA prompts the VLM to generate key points and additional attributes for affordance representation based on free-form language instructions and visual observations of the environment.
Since the five tasks we tested were relatively basic and did not require decomposition into subtasks, we could directly invoke the VLM in a question-answering format to address the operational challenges. This approach ensures seamless compatibility between the pipelines of PhysAgent and MOKA. Once the key points and waypoints were obtained from the VLM, MOKA converted these affordance representations into executable motions for the robotic arm.
To evaluate the impact of enhanced physical-world understanding on embodied tasks, we applied two methods to MOKA's VLM: (1) fine-tuning it with PhysBench, and (2) employing PhysAgent to zero-shot assist in reasoning about affordance representations.

As shown in Figure~\ref{fig:combine_three}(c), we observe consistent improvements after fine-tuning with a subset of PhysBench, indicating that the benchmark's data is of high quality and suitable for use as demonstration data in open-world robotics tasks. Additionally, PhysAgent consistently yields stable zero-shot gains across all five tasks, with significant progress observed in the force task in Figure~\ref{fig:embodied_agent}(a).

\section{Conclusion}\label{sec: conclusion}
In conclusion, we introduce PhysBench, a benchmark designed to assess Vision-Language Models' understanding of the physical world. Through experiments on 75 models, we identified significant gaps in physical world understanding, particularly in open-source models, due to inadequate training data. To address this, we developed PhysAgent, a novel framework that improves physical reasoning by 18.4\% on GPT-4o. Additionally, we demonstrated the utility of our dataset and approach in robotic tasks, helping to advance the understanding of the physical world in machine intelligence.

\textbf{Statement}. We provide a detailed discussion of limitations, broader impacts, ethical considerations, and reproducibility in Appendix~\ref{app:statement}.

\clearpage
\vspace{-4mm}
\subsubsection*{Acknowledgments}
Jiageng Mao and Yue Wang acknowledge funding supports from Toyota Research Institute, Dolby, and Google DeepMind. Yue Wang is also supported by a Powell Faculty Research Award. 

\vspace{3mm}

\bibliography{main}
\bibliographystyle{iclr2025_conference}
\appendix
\newpage
\addtocontents{toc}{\protect\setcounter{tocdepth}{3}}
\hypersetup{linkcolor=black}
{\small \tableofcontents} 
\hypersetup{linkcolor=red}
\newpage

\section{Detailed Dataset Collection Process}~\label{app:collection_process}
\subsection{Simulation}
We use ~\citep{blender2018blender} as our simulation platform. We utilized \numobj objects and \numhdr HDR images to generate simulated videos and images. During each simulation, we concurrently save images or videos of depth, normal, and albedo, as well as the corresponding configuration files, which include the position, angle, movement ,and other properties of the object light source.

\textbf{Image generation}. In addition to ambient lighting, we employed two point lights and one sunlight. To ensure data diversity, the positions of the camera and the arrangement of objects (ensuring no overlap and that all objects are captured by the camera) were randomized to some extent. Drawing from the object attribute annotation methods described in Newton~\citep{wang2023newton}, we cleaned and re-annotated our data to develop a comprehensive table of objects and their attributes. Utilizing this table, we delineate the relational semantics between different objects and the corresponding queries. Following the approach in BLINK~\citep{fu2024blink}, each object in the simulated images is demarcated with a bounding box rather than being explicitly mentioned in the text, thereby enhancing the evaluation of the model’s image comprehension capabilities.

Despite imposing considerable constraints on our code and meticulously annotating the 3D assets, there remains the possibility of minor object overlaps or incomplete captures of objects by the camera. To ensure that objects in the data are clearly identifiable, we employed GroundingDINO~\citep{liu2023grounding} for detection. We only accept images where the labels detected by GroundingDINO match exactly in content and quantity with the generated labels. This process also provides us with the bounding boxes of objects for subsequent annotation. During the later stages of annotation, manual inspection of the images is conducted to ensure accuracy. To improve the detection success rate of GroundingDINO and reduce the probability of false detections, we set the box\_threshold to 0.2 and the text\_threshold to 0.2. Specifically, these parameter settings were obtained through a grid search, with detailed results presented in Table~\ref{tab:net-search}.

\textbf{Videos with varying lighting conditions}. We used only one point light source and arranged objects on a plane to render shadows. The variations in lighting include three aspects: the color of the light, the position of the light source, and the intensity of the light. In terms of the light source position, the movement involves translations along the x, y, and z axes. To avoid ambiguity in lateral directions~\citep{du2024embspatial}, during the dataset generation, the movement questions are typically framed in terms of moving along the line connecting two objects rather than simply asking for the direction of movement.

\textbf{Videos with varying camera conditions}. We used the same lighting and other configurations as in the image generation process. During video recording, we randomly altered the camera's position or shooting angle to capture the videos.

\textbf{Fulid}. We used assets from ContPhy~\citep{zheng2024contphy} and Unity~\citep{haas2014history} to generate videos across four types: fluid, rope, cloth, and ball, with 350, 250, 200, and 200 videos respectively. The videos were then manually annotated.

\subsection{Web}
For web data collection, we primarily use predefined topics (e.g., gases) to retrieve relevant videos or images from the internet (such as middle school physics experiments). After filtering and cleaning the data, we proceed with annotation. Additionally, we leverage large language models (LLMs) to generate suitable descriptions of physics-related concepts, which we then use to search for corresponding videos, followed by further cleaning and annotation.

In addition to the network data collection process described in Section~\ref{sec3:2}, we employ the following methods to gather data.

\textbf{Unsplash.} We use high-quality and high-resolution images from Unsplash~\cite{ali2023unsplash}. 57,859 images are downloaded, and finally we use only about 6,000 images.

\textbf{Manipulation}. We sampled approximately 500 videos from DROID~\citep{khazatsky2024droid}, Ego4D~\citep{grauman2022ego4d}, and MimicPlay~\citep{wang2023mimicplay}, providing detailed annotations to generate QA pairs categorized under object-manipulation tasks. The primary focus of these questions is to determine the appropriate sequence of actions based on given instructions, which are derived from the original datasets' descriptions of actions. Figure~\ref{fig:example_15}'s both first and second examples provide an example of this task. First, we filtered the videos to select those with a strong alignment between the instructions and the visuals, ensuring that the videos were clear, unambiguous, and matched the instructions well. Next, we identified 3-4 keyframes from these videos. The task involved sorting these key frames in the correct order to execute the instructions properly. Additionally, we used FunKPoint~\citep{lai2021functional} to annotate the affordance~\citep{gibson2014ecological} points in individual images from the original dataset. Specific examples of these annotations can be found in Figure~\ref{fig:example_15}'s third and fourth examples.

\textbf{nuScenes}. We cropped and annotated videos from the nuScenes~\citep{nuscenes2019} mini and test datasets, ultimately obtaining 1,356 QA pairs for spatial movement tasks. We categorized the questions into types such as left turn, straight, and right turn, and included arrows on the images to indicate the direction. The questions asked participants to identify which image they might see based on the indicated direction, as illustrated in Figure~\ref{fig:example_1}.

\textbf{Visual Prompt}. In certain tasks, we utilized Visual Prompts~\citep{fu2024blink}, and through experimentation, we identified an alternative annotation method, detailed in Appendix~\ref{e_o_v_p}. For tasks using Visual Prompts, we set the image size to $1024\times1024$ pixels. In this scale, we standardized the Visual Prompt to a red circle with a 30-pixel radius and white text options with a font size of 65 pixels. The positions of the options' centers in the dataset are recorded in the following format:
\begin{mycase}{light_grey}
{
            "A": [
                734,
                922
            ],
            "B": [
                202,
                898
            ],
            "C": [
                343,
                115
            ],
            "D": [
                410,
                559
            ]
        }
\end{mycase}
\textbf{Visual Correspondences}. Drawing inspiration from~\cite{fu2024blink, sarlin2020superglue}, we also annotated a portion of the corresponding point data using visual prompts. Specific examples can be found in Figure~\ref{fig:case02}.

\subsection{Real-world}
We also collected some real-world videos and images, primarily covering sub-tasks related to light, camera, and physical dynamics such as collisions. An iPhone 13 Pro Max was used as the recording device, and all images are in RGBD format.
\section{Data Annotation Protocol}\label{app:protocol}
\subsection{General Guidelines}\label{data_guidance}
As previously discussed, there is a significant gap in existing benchmarks, which primarily assess vision-language models (VLMs) based on descriptive tasks without adequately addressing their physical perception and reasoning abilities. To bridge this gap, our benchmark, PhysBench, is designed to provide a comprehensive evaluation framework for physical perception, integrating visual understanding with the assessment of physical properties, spatial relationships, and dynamic phenomena. This approach aims to advance AI systems toward more general-purpose capabilities in real-world physical environments. Our benchmark follows the guidelines outlined below for data collection:

\begin{itemize}
    \item \textbf{General Principles:} 
    \begin{itemize}
        \item Annotations must be accurate, consistent, and adhere to a high standard of academic rigor.
        \item It covers multiple tasks and topics to mirror real-world applications.
        \item It incorporates diverse visual contexts and physics knowledge to foster a well-rounded evaluation.
        \item It offers varying levels of challenge to effectively probe and uncover the potential limitations of current models.
        \item It provides robust evaluation settings for deterministic assessments.
    \end{itemize}
    \item \textbf{Specific Instructions:}
    \vspace{1pt}
        \begin{itemize}
            \item All questions must contain one or more images.
            \item All questions should be written in English.
            \item All questions should meet the college-level difficulty.
            \item Questions should not be ambiguous and must be answerable with one of the given options.
            \item Clearly categorize each question.
            \item Annotate all fields, including the question, answer options and other things that follow the format requirement.
        \end{itemize}
    
    \item \textbf{Review Process:} Ensure that every annotation undergoes a peer review to maintain high standards and minimize errors.
\end{itemize}
Annotations such as physical properties, spatial relationships, dynamic interactions, and environmental factors are also collected, providing detailed examples that demonstrate the physical perception and reasoning capabilities of the models for further analysis and usage.

\subsection{Data Format and Structure}\label{app_data_format}
\begin{figure}[th!]
	\centering  
	\vspace{-1.5mm}
	\includegraphics[width=1.0\linewidth]{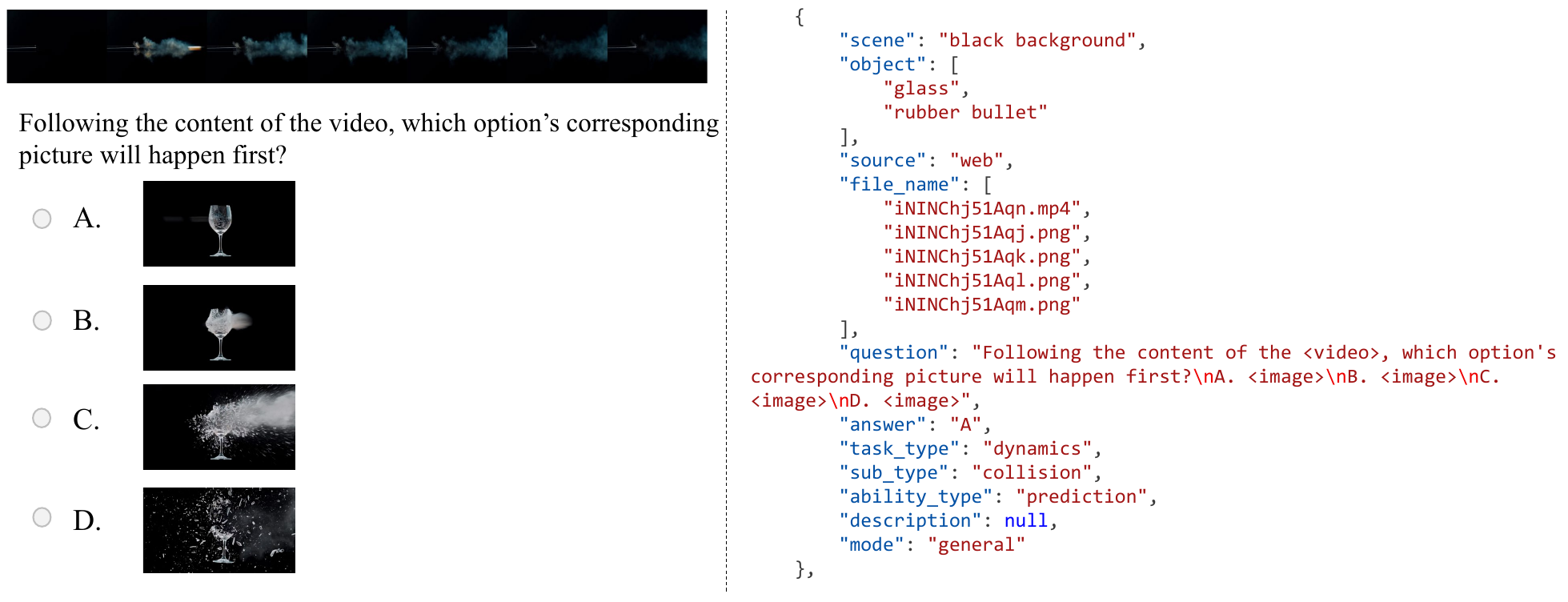}
	\vspace{-1.5mm}
    	\caption{A question-answer pair case in PhysBench and its JSON representation.}
	\label{fig:dataformat}  
	\vspace{-3mm}
\end{figure}

Detailed examples of annotated question examples as shown in Figure~\ref{fig:dataformat} are provided in the guidance to serve as a reference for the annotators.
\begin{itemize}
    \item \textbf{JSON File Format:} The structured JSON format will include fields for number, question type, question text, answer options (for multiple-choice), correct answer, question difficulty, and explanation (if there exists).
    
    \item \textbf{Naming Conventions:}
    \vspace{1pt}
        \begin{itemize}
            \item Each collected sample will be stored in a separate JSON file following a standard naming rule: \textbf{subject\_\{Number\}}.json
            \item Image Files: \textbf{image\_\{QuesNum\}\_\{ImageNum\}}.png
        \end{itemize}
\end{itemize}

\subsection{Quality Control and Validation}
\begin{itemize}
\item A secondary review team will rigorously vet annotations for quality and adherence to guidelines.
\item Regular audits of random samples from the dataset will be conducted to ensure sustained quality and consistency.
\item Periodic training sessions will be held to update annotators on best practices and any changes in annotation guidelines.
\item Feedback mechanisms will be established to promptly address and rectify any identified errors or inconsistencies in the annotations.
\end{itemize}

\subsection{Handling Ambiguities}
Instances of ambiguity or unclear data should be flagged for detailed review. These instances will be collaboratively examined during team meetings to establish a standardized approach for annotation.

\subsection{Ethical Considerations}
\begin{itemize}
    \item \textbf{Copyright and Licensing:} Adherence to copyright and licensing regulations is strictly enforced. Data from sources that prohibit copying or redistribution will be explicitly avoided.
    \item \textbf{Data Privacy:} Compliance with privacy laws and ethical standards in data handling is paramount. Annotators must avoid collecting questions that contain any private information.
    \item \textbf{Ethical Data Usage:} All data collection and usage must respect ethical guidelines. This includes avoiding biased or harmful content and ensuring that the datasets promote fairness and inclusivity.
\end{itemize}
\subsection{Data Contamination Considerations}
The risk of data contamination can be mitigated by assigning annotators to carefully select questions that extend beyond straightforward queries with easily accessible answers. It is essential that tasks rely on provided videos or images for answers rather than the common knowledge of large language models. This approach is beneficial for creating benchmarks that genuinely test the model's ability to comprehend and synthesize information from diverse and challenging sources.
\subsection{Annotation Platform}
\begin{figure}[th!]
	\centering  
	\vspace{-1.5mm}
	\includegraphics[width=1.0\linewidth]{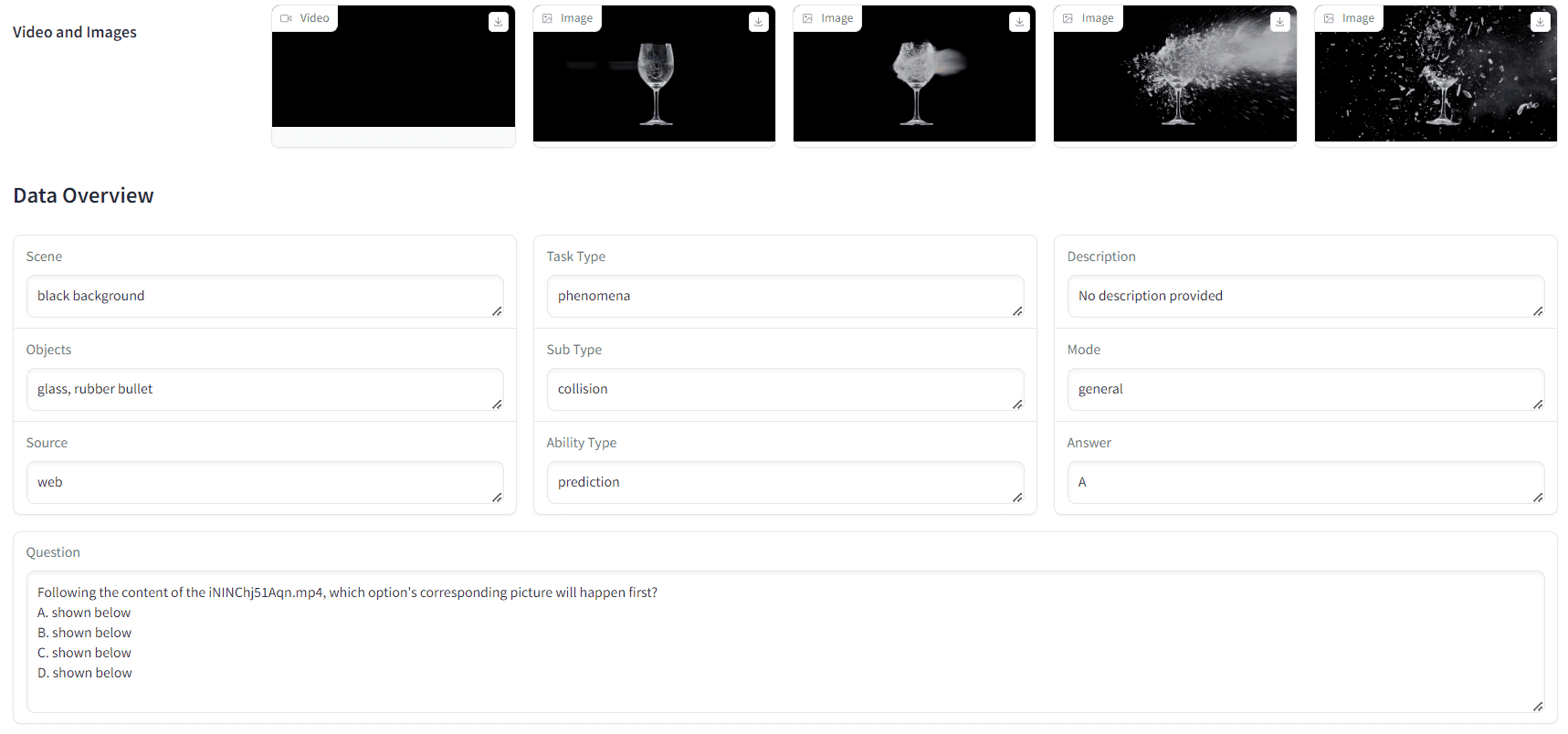}
	\vspace{-1.5mm}
    	\caption{Annotation Platform.}
	\label{fig:ann_plat}  
	\vspace{-3mm}
\end{figure}

We developed a GUI-based annotation platform, as shown in Figure~\ref{fig:ann_plat}, designed to assist human experts in the data annotation process. Through this program, experts can easily view various media content, such as videos and images, and perform annotations and edits directly within an intuitive interface. The streamlined layout enhances the user experience, ensuring that experts can complete annotation tasks efficiently and accurately, thereby improving the quality and efficiency of the annotations. The purpose of this tool is to simplify the complex annotation workflow, reduce manual effort, and make the annotation process more efficient.

\subsection{Benchmark Preparation and Release}~\label{exp_bench_release}
For convenience, PhysBench-test consists of 10,002 entries, organized into 19 subclasses, as the test set, and 200 entries as the validation set for parameter choosing (PhysBench-val)  since the answers for the PhysBench will not be publicly available and are hosted similarly to ~\cite{yue2023mmmu}. We also present 89,998 entries for further research. The experimental results presented in this paper, unless otherwise specified, are based on the test set.
Importantly, the answer labels for the remaining set will not be publicly released to prevent data leakage, and we will maintain an online evaluation platform. It should be noted that the scores we refer to for PhysBench are based on the entire test dataset, including the validation split.

To ensure that each source dataset is well represented in the validation split and that the distribution of sub-task types and ability types in the validation set is similar to that of the entire dataset, we adopted the following sampling strategy:
1. Randomly sample questions to ensure that the distribution of sub-task and ability types in the validation set matches the full dataset.
2. Randomly sample the remaining questions from each source dataset based on its proportion in the entire dataset.

Additionally, we conducted several quality checks to address any potential errors.

\subsection{More Details of the Annotation Pipeline}
\begin{figure}[h!]
	\centering  
	\vspace{-1.5mm}
	\includegraphics[width=1.0\linewidth]{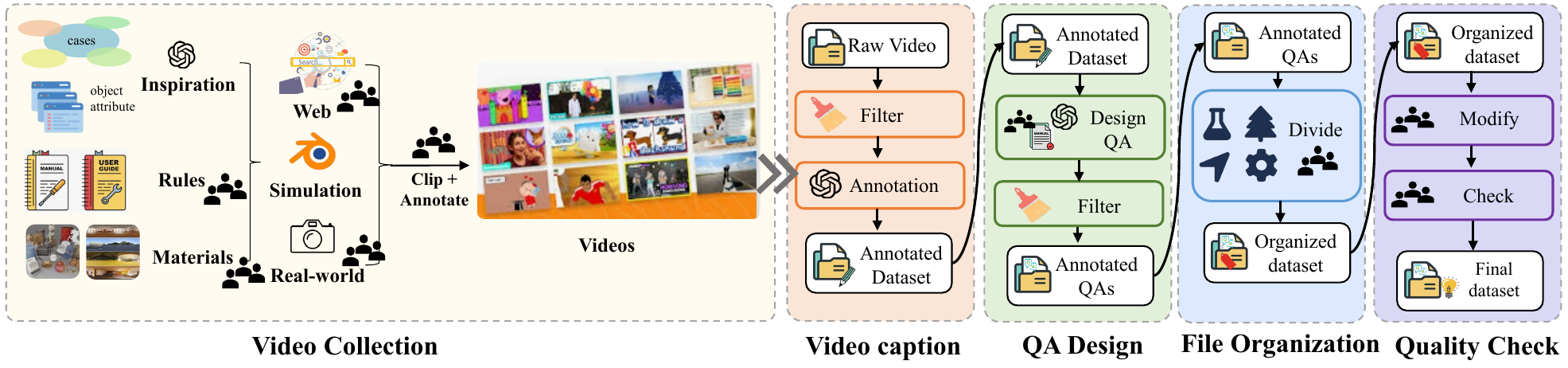}
	\vspace{-3mm}
    	\caption{
        \textbf{Annotation Pipeline.} Icon \includegraphics[scale=0.08]{figures/icon/human_icon.png} indicate stages involving human participation.
        }\label{fig:anno_pipe}
	\vspace{-3mm}
\end{figure}

As described in Section~\ref{sec3:2}, all questions were manually annotated by graduate students in STEM fields and subsequently refined through a rigorous review process. The detailed workflow is depicted in Figure~\ref{fig:anno_pipe}, where the icon \includegraphics[scale=0.08]{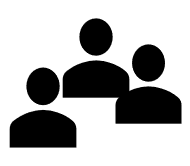} denotes stages involving human participation. The annotators were divided into three groups, each comprising six individuals. During the annotation process, either a GUI interface similar to Figure~\ref{fig:ann_plat} or direct editing of JSON files was employed. The first group was responsible for video collection, the final group handled quality checks, and the intermediate group completed the remaining steps. A comprehensive explanation of this process is provided below.

\textbf{Video Collection}. Videos and images were sourced through web searches, simulations, and real-world recordings. The collection process utilized predefined simulation rules, LLM-guided queries, and other strategies to identify relevant content. Specifically, for data that could be simulated, we generated it using pre-written simulation scripts, as described in Section~\ref{app:collection_process}. Additionally, we conducted YouTube searches for videos, leveraging captions generated by GPT to guide annotators in collecting videos. We also sought compilations on topics like "interesting physical phenomena." For real-world recordings, we pre-designed scenarios and objects to capture. 
Given the complexity of collecting such data, we employed GPT as heuristic tools to expand the search scope and enrich the dataset's diversity. Finally, annotators curated the videos by trimming them to retain only relevant segments and provided detailed physical descriptions, such as the direction of shadow movement or the causes of observed events.

\textbf{Question Design}. In the previous step, we utilized GPT to generate annotations for videos. This approach was adopted after observing that directly inputting a video and prompting GPT to generate question-answer pairs in an in-context learning format often led to suboptimal adherence to instructions. Instead of focusing on generating questions, GPT tended to explain the video content. Additionally, inputting videos directly consumed more tokens, increasing computational costs. Furthermore, annotating videos with captions was considered beneficial for subsequent research leveraging our dataset. 
Given the complexity of collecting such data, we employed GPT as a heuristic strategy to broaden the scope of search and enhance the diversity of our dataset. GPT generated questions using an in-context learning approach, with the templates provided in Appendix~\ref{app:llm_prompt}. Notably, all final questions were curated and verified by human annotators, with GPT serving only as a reference.

\textbf{File Organization}. We presented annotators with examples and detailed classification criteria, as outlined in Appendix~\ref{app:task_description} and Appendix~\ref{app:task_examples}. Annotators were then instructed to categorize tasks accordingly and structure each task in the JSON format shown in Figure~\ref{fig:dataformat}.

\textbf{Quality Check}. After organizing the dataset, we conduct a two-stage review process to ensure its quality. The dataset undergoes human verification to confirm that the questions are relevant to the physical world, rely on all provided input information, are not solely based on common sense, and are accurately categorized with clear questions and corresponding answers. The first stage involves filtering and refining the questions, while the second stage focuses on thorough validation to ensure data accuracy and consistency.

Given the difficulty of acquiring such data, where expressing spe-
cific properties often requires multiple images, we undertook a five-step process, spending a total of \textbf{4,000 hours} on annotation.

\section{Detailed Task Description}\label{app:task_description}
This section primarily introduces the definitions of the various capabilities and sub-tasks. Specific examples can be found in Appendix~\ref{app:task_examples}. As illustrated in Figure~\ref{fig:abstract}, humans not only accurately comprehend visual content but also draw upon knowledge to explain and reason about the scenes they observe. This is the primary goal of most VQA datasets. Building on this foundation, several studies have focused on the commonsense understanding of the geometric relationships and properties within the 3D world. However, the real physical world encompasses not only 3D geometric relationships but also includes object properties (e.g., elasticity, ductility), physical object relationships (e.g., velocity, acceleration, depth), physical scene and environmental factors understanding (e.g., temperature, camera parameters, lighting conditions), and the principles and mechanisms of physical dynamics (e.g., the sequence of collisions, energy transfer, fluid flow, simple physical and chemical reactions, optical and electromagnetic dynamics). To address this gap, we propose PhysBench, which aims to bring Vision-Language Models closer to achieving spatial intelligence~\citep{gupta2021embodied, shen2021igibson}.
\subsection{Ability Description}\label{app:task_description_ability}
Table~\ref{tab:ability_type} presents the description of capability types for PhysBench. The tasks are categorized into four types: physical properties, physical object relationships, physical scene, and physics-based dynamics. Each task type corresponds to different capability types such as recognition, comparison, prediction, judgment, reasoning, and perception, with detailed descriptions provided for each. Specifically, the physical properties task type includes the recognition of object properties and comparisons between the physical properties of different objects. The physical object relationships task type distinguishes between static and dynamic relationships of objects. The physical scene understanding task type involves the prediction, judgment, reasoning, and perception of changes in environmental conditions. The physics-based dynamics task type evaluates the ability to predict, judge, reason, and perceive various physics-based dynamics or phenomena.
\begin{table}[th!]
    \small
    \centering
    \caption{Ability Type Description for PhysBench.}
    \label{tab:ability_type}
    \resizebox{0.99\columnwidth}{!}{
        \begin{tabular}{lcp{8cm}}
        \toprule
        \textbf{Task type} & \textbf{Ability type} & \textbf{Description}\\ \hline
        
        \multirow{2}{*}{Physical Object Property} 
        & Identify &  Identify a physical property of an object.\\
        & Comparison & Comparison of the same physical property between different objects, or changes in a specific property over time.
        \\ \hline

        \multirow{2}{*}{Physical Object Relationships}               
        & Static        & The static spatial properties of an object. \\
        & Dynamic       & The dynamic spatial properties of an object, meaning the spatial properties are changing dynamically.\\ \hline
        
        \multirow{4}{1cm}{Physical Scene Understanding} 
        & Prediction    & Predict what might happen if a certain environmental condition is changed, or what will happen next. \\
        & Judgment      & Judge what will be different if a certain environmental condition is modified. \\
        & Reasoning     & Explain the occurrence reason or condition of the environment.  \\
        & Perception    & Understand what environment condition change has occurred and what its definition is, or determine the environmental attributes that caused the phenomenon to occur. \\\hline

        \multirow{4}{1cm}{Physics-based Dynamics} 
        & Prediction    & Predict what might happen if a certain object or object attribute in the video is changed, or what will happen next. \\
        & Judgment      & Judge what will change if a certain attribute is modified, or, for example, the sequence in which actions occur. \\
        & Reasoning     & Explaining the occurrence reason or condition of phenomena.  \\
        & Perception    & Understand what phenomenon has occurred and what its definition is, or determine the physical attributes that caused the phenomenon to occur. \\\hline
        \toprule
        \end{tabular}
        }
\end{table}

\subsection{Physical Object Property Sub-task}
Table~\ref{tab:sub_type_pp} describes the sub-tasks for object types in PhysBench. The object types are divided into four subcategories: number, mass, color, and attributes. Specifically, the “number” subcategory involves the count of certain objects or changes in their quantity; the “mass” subcategory focuses on approximate changes in object mass, mass estimation, or mass comparison; the “color” subcategory describes the color of objects or color changes; and the “attributes” subcategory covers object characteristics such as rigidity, fluidity, gas, hardness, malleability, elasticity, smoothness, and sharpness.

Notably, our dataset imposes more rigorous evaluations on conventional attributes like mass and color. For instance, in the case of counting tasks, previous works~\citep{kafle2017analysis, yi2019clevrer, chen2022comphy} typically only require identifying the number of objects in an image. In contrast, our dataset often ties quantities to specific object attributes (e.g., "How many objects of a certain color are outside the plate?" or "How many objects are not blurred by the camera?"). 
\begin{table}[th!]
    \small
    \centering
    \caption{Sub-task Description for Physical Object Property Type in PhysBench.}
    \label{tab:sub_type_pp}
    \resizebox{0.79\columnwidth}{!}{
        \begin{tabular}{lp{8cm}}
        \toprule
        \textbf{Sub type} & \textbf{Description} \\
        \toprule
        Number& The number of certain objects or changes in the number of objects.\\
        Mass&  Approximate changes in mass, mass estimation, or mass comparisons.\\
        Color&  Color of an object or changes in color.\\
        Attribute & The attributes or types of objects, such as rigid body, fluid, gas, stiffness, malleability, elasticity, smoothness, sharpness, \textit{etc.} \\
        \toprule
        \end{tabular}}
\end{table}

\subsection{Physical Object Relationships Sub-task}
Table~\ref{tab:sub_type_spa} presents the descriptions of physical object relationships sub-tasks in PhysBench. The relationships types are divided into five subcategories: size, location, depth, distance, and movement. Specifically, the “size” subcategory relates to the dimensions of an object; the “location” subcategory describes the absolute and relative positions of objects, including tasks related to object localization and spatial information processing; the “depth” subcategory focuses on an object’s depth relative to the camera or depth comparisons between different objects; the “distance” subcategory involves the comparison or estimation of distances between objects as well as their absolute size; and the “movement” subcategory addresses the analysis of movement direction, changes in speed, and changes in acceleration.
\begin{table}[th!]
    \small
    \centering
    \caption{Sub-task Description for Physical Object Relationships Type in PhysBench.}
    \label{tab:sub_type_spa}
    \resizebox{0.79\columnwidth}{!}{
        \begin{tabular}{lp{8cm}}
        \toprule
        \textbf{Sub type} & \textbf{Description} \\
        \toprule
        Size& The size of the object.\\
        Location&Positional relationships (absolute and relative), including directly or indirectly locating objects and other tasks involving spatial information. \\ 
        Depth& Depth of an object relative to the camera or depth comparisons between different objects.\\
        Distance& Comparison or estimation of distances between objects or their absolute sizes.\\
        Motion& Motion, velocity, acceleration and the direction of movement, changes in speed, or changes in acceleration.\\
        \toprule
        \end{tabular}}
\end{table}

\subsection{Physical Scene Understanding Sub-task}
Table~\ref{tab:sub_type_env} describes the sub-tasks for physical scene understanding types in PhysBench. The physical scene understanding types are divided into four subcategories: temperature, camera, gas, and light. Specifically, the “temperature” subcategory involves temperature and its fluctuations, as well as dynamics caused by these changes; the “camera” subcategory focuses on changes in camera position and the resulting effects; the “gas” subcategory covers conditions of the gas environment, such as high pressure, low pressure, or vacuum states; and the “light” subcategory describes the color tone of the light source (warm or cool), changes in the position of the light source, light intensity, and the nature of the light source (point or surface).
\begin{table}[th!]
    \small
    \centering
    \caption{Sub-task Description for Physical Scene Understanding Type in PhysBench.}
    \label{tab:sub_type_env}
    \resizebox{0.79\columnwidth}{!}{
        \begin{tabular}{lp{8cm}}
        \toprule
        \textbf{Sub type} & \textbf{Description} \\
        \toprule
        Temperature& The temperature and its changes, as well as phenomena caused by temperature fluctuations.\\
        Viewpoint& The position of the camera and its changes, along with phenomena caused by shifts in camera position.\\
        Air& The air environment encompasses conditions such as air pressure, humidity, airflow direction, and intensity.\\
        Light& Includes the color tone of the light source (warm or cool), changes in the position of the light source, its intensity, and the nature of the light source (point or surface).\\
        \toprule
        \end{tabular}}
\end{table}

\subsection{Physics-based Dynamics Sub-task}
Table~\ref{tab:sub_type_phe} describes the sub-tasks for physics-based dynamics types in PhysBench. The physics-based dynamics types are divided into six subcategories: collision, throwing, manipulation, fluid, chemistry, and other physics-based dynamics. Specifically, the “collision” subcategory includes physics-based dynamics such as friction between objects, collisions, and explosions; the “throwing” subcategory involves physical world dynamics and phenomena such as throwing and falling; the “manipulation” subcategory focuses on the manipulation of deformable objects and affordance-based sequence arrangements; the “fluid” subcategory covers fluid motion, shapes, and other fluid-related dynamics; the “chemistry” subcategory involves basic chemical reactions or other dynamics or phenomena related to chemistry in the physical world; and the “other” subcategory includes various other dynamics related to the physical world.

\begin{table}[th!]
    \small
    \centering
    \caption{Sub-task Description for Physics-based Dynamics Type in PhysBench}
    \label{tab:sub_type_phe}
    \resizebox{0.79\columnwidth}{!}{
        \begin{tabular}{lp{8cm}}
        \toprule
        \textbf{Sub type} & \textbf{Description} \\
        \toprule
        Collision& Friction between objects, collisions, explosions, and similar dynamics.\\
        Throwing & Throwing, falling, and other physical world dynamics.\\
        Manipulation & Manipulation of deformable objects and affordance-based sequence arrangements.\\
        Fluid& Fluid motion, shapes, and other fluid-related dynamics. \\
        Chemistry&  Basic chemical reactions or other dynamics involving chemical knowledge in the physical world. \\
        Others & Other dynamics related to the physical world. \\
        \toprule
        \end{tabular}}
\end{table}

All of our questions are designed as multiple-choice, with only one correct answer among the four options. Referring to BLINK~\citep{fu2024blink}, the visual prompt is set as a red circle with a 10px size, which has been found to be the most suitable.
\section{More Dataset Analysis}\label{app_more_data_stat}
\subsection{Global Statics}
\begin{figure}[th!]
    \centering  
    \vspace{-1.5mm}
    \includegraphics[width=0.97\linewidth]{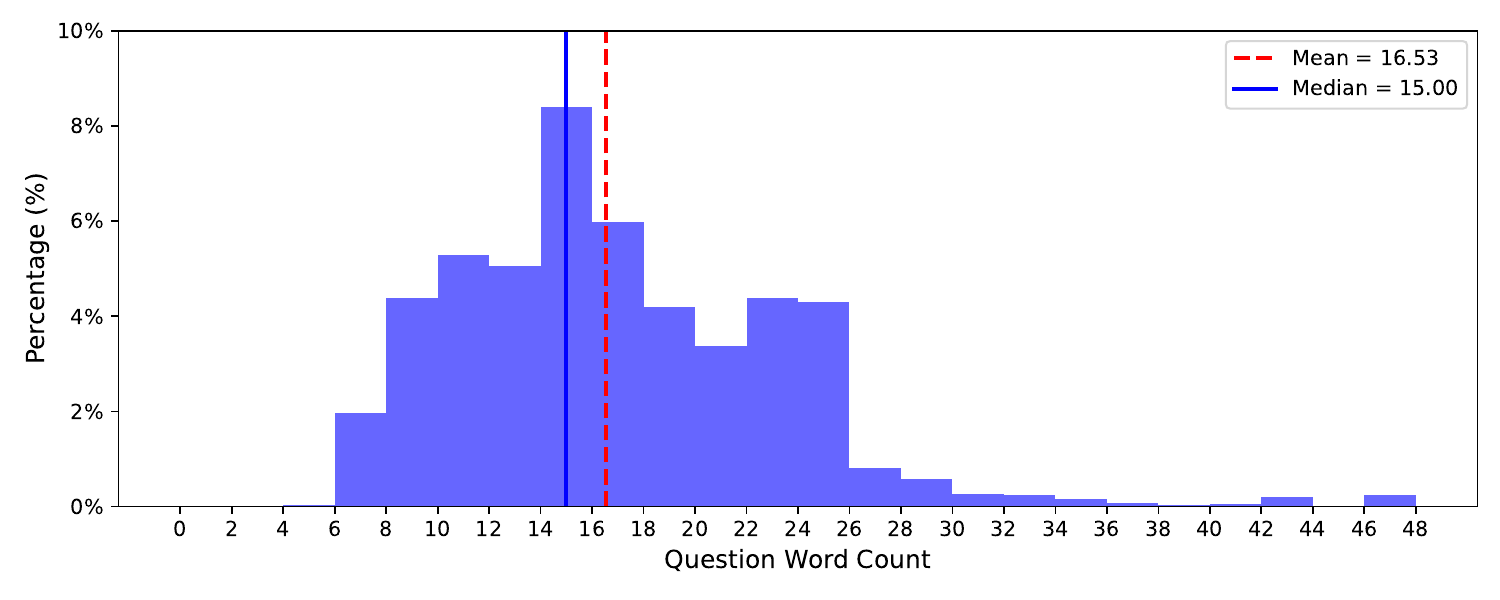}
    \vspace{-4.5mm}
    \caption{The distribution of the number of words per question in PhysBench. Questions with a length greater than 48 are categorized as 47 for visualization simplicity.}
    \label{fig:question_len}  
    \vspace{-2mm}
\end{figure}

\textbf{Question Distribution.}
Figure~\ref{fig:question_len} further elucidates the distribution of word counts, highlighting the diverse patterns of
questions. The average word of the questions within PhysBench is 16.53, while the maximum number of words in a question reaches 48.

\begin{figure}[th!]
    \centering  
    \vspace{-1.5mm}
    \includegraphics[width=0.97\linewidth]{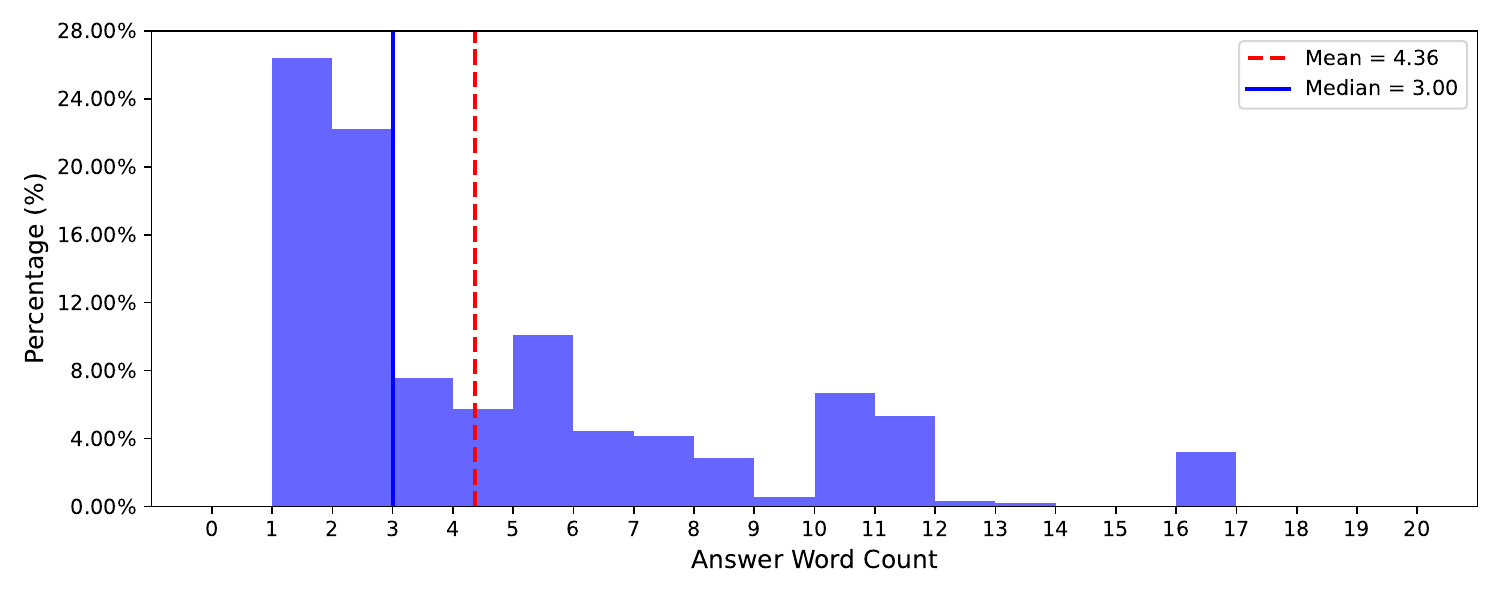}
    \vspace{-4.5mm}
    \caption{The distribution of the number of words per question in PhysBench. Options with a length greater than 20 are categorized as 20 for visualization simplicity.}
    \label{fig:answer_len}  
    \vspace{-2mm}
\end{figure}

\textbf{Option Distribution.}
Figure~\ref{fig:answer_len} further elucidates the distribution of word counts, highlighting the diverse patterns of
options. The average number of words in the options within PhysBench is 4.36, while the maximum number of words in a question reaches 20. It is worth noting that an “option” here refers to the text following a choice, such as “A. Point A,” where the character length is 2.

\textbf{Image Resolution.} The resolution of most images falls within the 1024-2048 range, accounting for approximately 79.7\% of the total images. Only four images in the dataset have a resolution below 256. The minimum resolution is 183, while the maximum is 7201.

\textbf{Video Resolution.} Similarly, the majority of videos have a resolution between 1024-2048, covering about 98.6\% of the total videos. Only 1.3\% of videos have a resolution below 1024. The highest video resolution is 1920, and the lowest is 370.

\begin{figure}[th!]
    \centering  
    \vspace{-1.5mm}
    \includegraphics[width=1.0\linewidth]{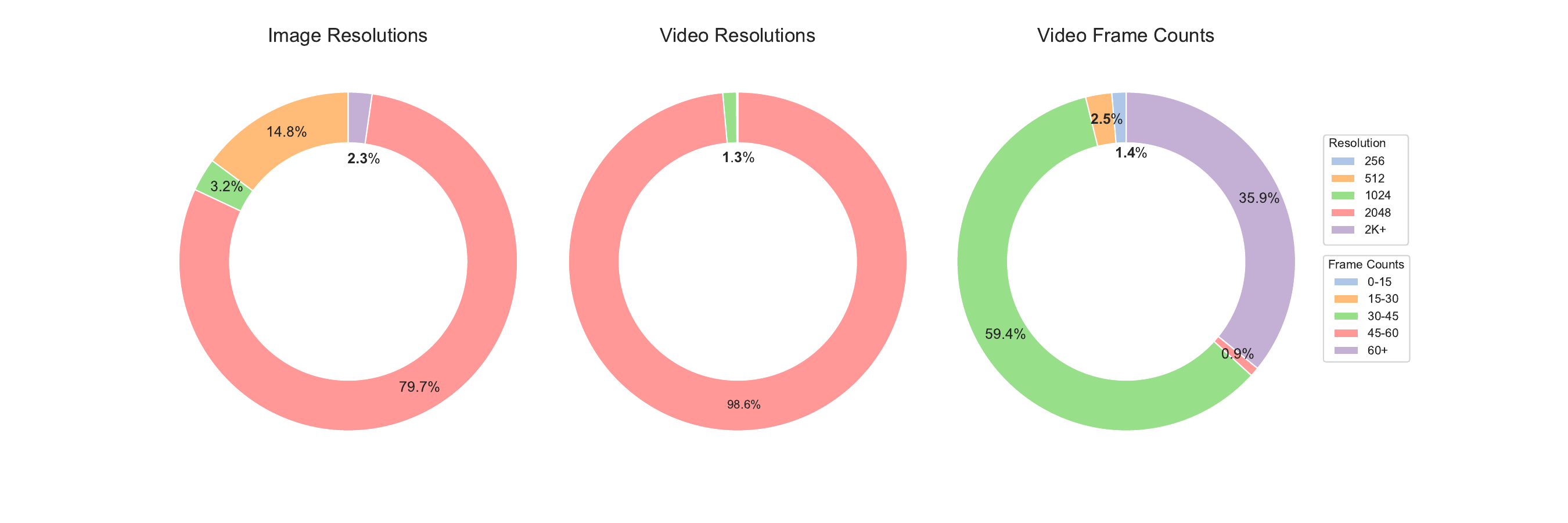}
    \vspace{-3.5mm}
    \caption{The distribution charts for image and video resolution, as well as video frame counts. From left to right: the distribution of image resolution, the distribution of video resolution, and the distribution of video frame counts.}
    \label{fig:image_video_res_pie}  
    \vspace{-3mm}
\end{figure}

\textbf{Video Frames.} Considering that VLMs typically extract keyframes when processing videos—generally selecting 6-8 frames—the total number of frames in the videos doesn't need to be large. However, this doesn't imply that our videos are limited to 6–8 frames. The frame counts are primarily distributed in the ranges of 30–45 frames and 60+ frames, accounting for 59.4\% and 35.9\% of the total, respectively. The distribution charts for image and video resolution, as well as video frame counts, can be seen in Figure~\ref{fig:image_video_res_pie}.

\subsection{Word Statics and Word Cloud}\label{app:word_stat}
\begin{figure}[th!]
    \centering  
    \vspace{-1.5mm}
    \includegraphics[width=1.0\linewidth]{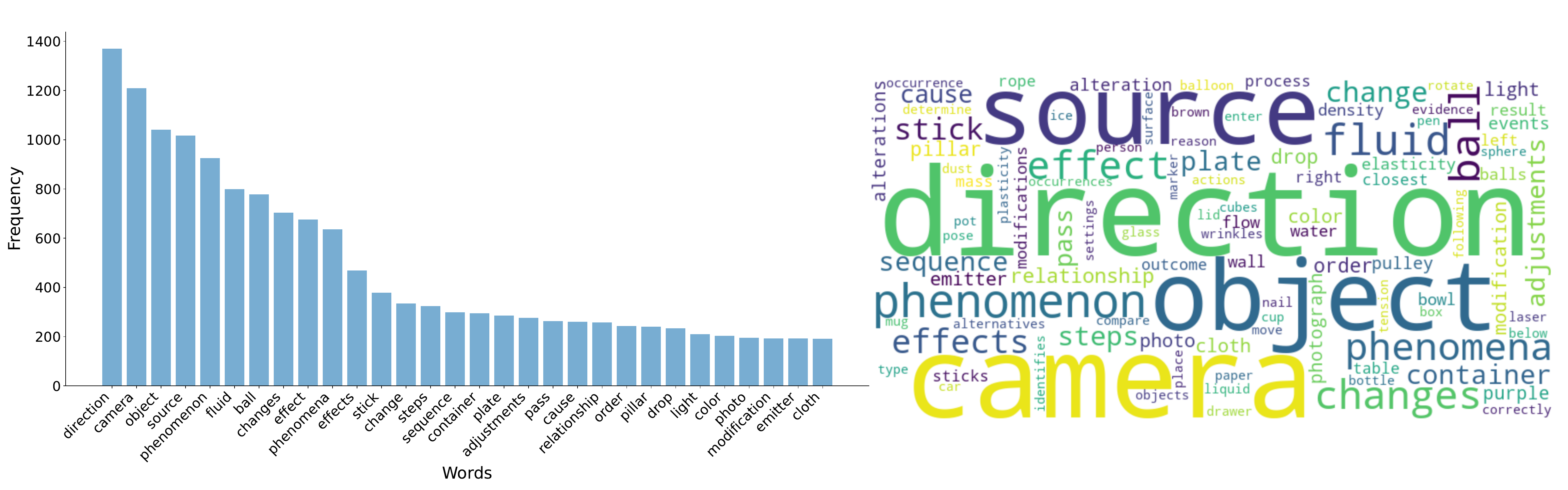}
    \vspace{-1.5mm}
    \caption{Word Statics and Word Cloud for PhysBench.}
    \label{fig:word_physbench}  
    \vspace{-3mm}
\end{figure}

\begin{figure}[th!]
    \centering  
    \vspace{-1.5mm}
    \includegraphics[width=1.0\linewidth]{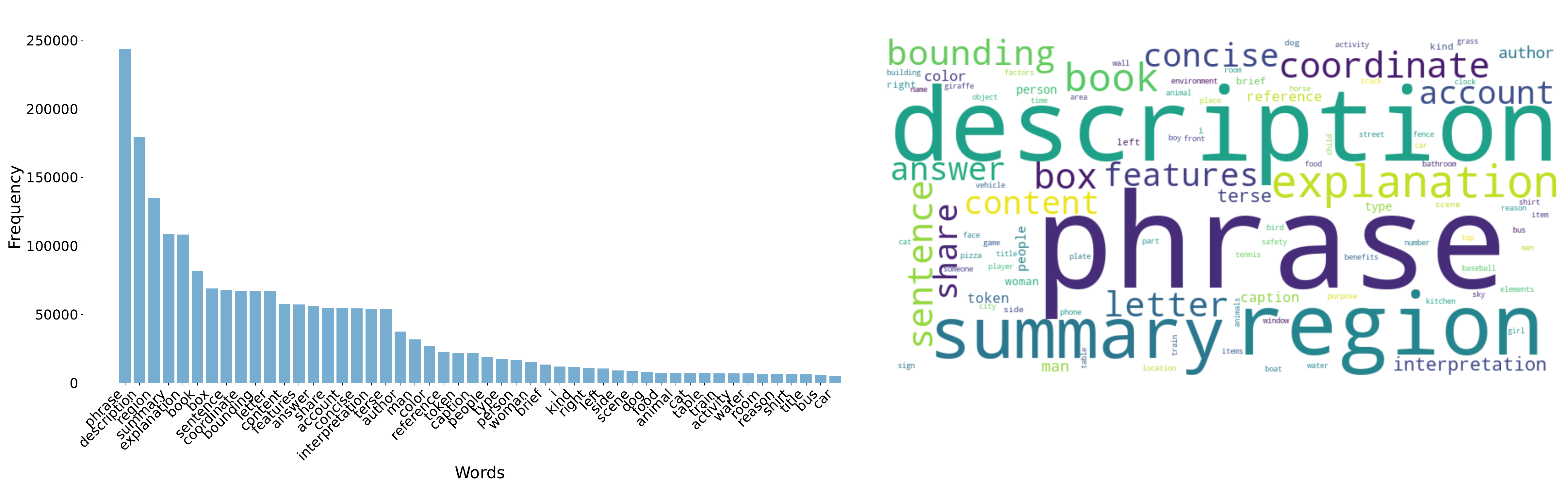}
    \vspace{-1.5mm}
    \caption{Word Statics and Word Cloud for LLaVA-1.5-13B Training Data.}
    \label{fig:word_llava}  
    \vspace{-3mm}
\end{figure}

We created a word cloud and word frequency chart for our dataset, as shown in Figure~\ref{fig:word_physbench}. It reveals a significant presence of terms related to physical world perception, such as “direction,” “camera,” “phenomenon,” “effects,” “relationship,” and “light.” Additionally, we generated a word cloud and word frequency chart for the \textbf{LLaVA-1.5-13B} training data, which includes 595K pretraining samples and 665K instruction-tuning samples, as illustrated in Figure~\ref{fig:word_llava}.

\textbf{VILA-1.5-13B}. In the pretraining data, CC3M~\citep{sharma2018conceptual} and COYO~\citep{byeon2022coyo} 25M primarily consist of image captions, while MMC4~\citep{zhu2024multimodal} 25M includes webpage descriptions, none of which significantly contribute to the model's understanding of the physical world. Therefore, we mainly focus on the instruction-tuning data. The LLaVA-1.5-SFT-665K instruction-tuning data has already been used in LLaVA-1.5-13B, so it is not considered further. Additionally, FLAN~\citep{wei2021finetuned} consists purely of text data and is thus also excluded from consideration. 

\begin{figure}[th!]
    \centering  
    \vspace{-1.5mm}
    \includegraphics[width=1.0\linewidth]{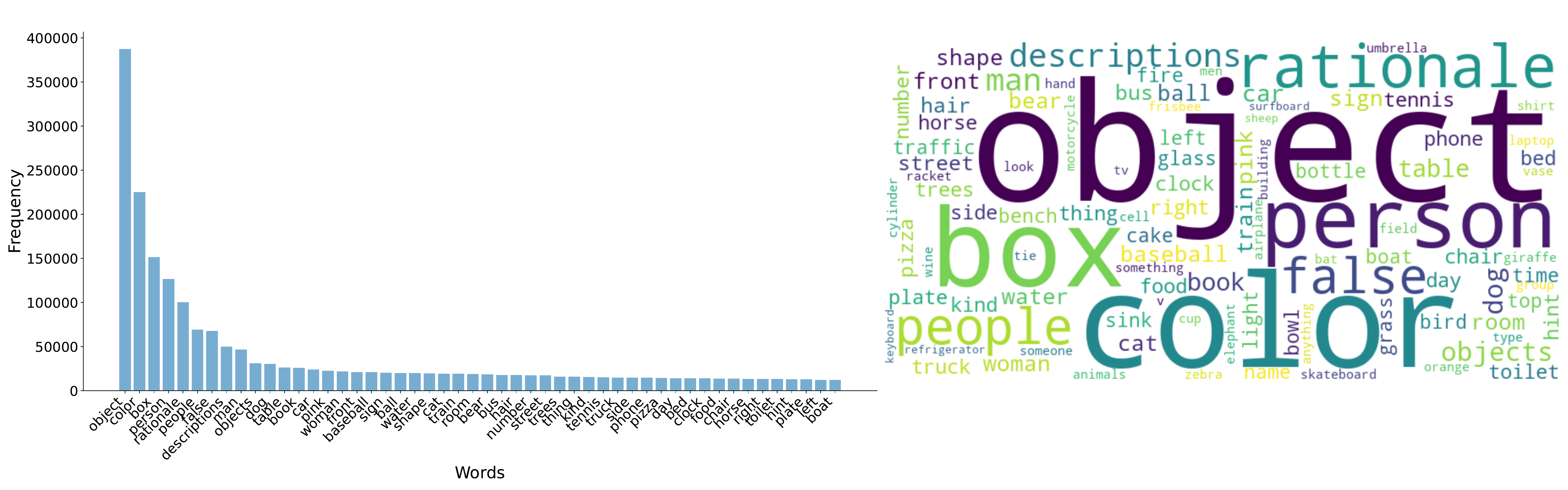}
    \vspace{-1.5mm}
    \caption{Word Statics and Word Cloud for VILA-1.5-13B Training Data.}
    \label{fig:word_vila}
    \vspace{-3mm}
\end{figure}

The data used for VILA-1.5-13B includes the following:
\begin{itemize}
    \item Captioning: Image Paragraph Captioning~\citep{krause2017hierarchical}, MSR-VTT~\citep{xu2016msr}, TextCaps~\citep{sidorov2020textcaps}
    \item Reasoning: CLEVR~\citep{johnson2017clevr}, NLVR~\citep{suhr2017corpus}, VisualMRC~\citep{tanaka2021visualmrc}
    \item Translation: Multi30k~\citep{elliott2016multi30k}
    \item VQA: ActivityNet-QA~\citep{yu2019activitynet}, DocVQA~\citep{mathew2021docvqa}, GQA~\citep{hudson2019gqa}, iVQA~\citep{yang2021just}, MSRVTT-QA~\citep{xu2017video}, MSVD-QA~\citep{xu2017video}, OCR-VQA~\citep{mishra2019ocr}, ST-VQA~\citep{biten2019scene}, ViQuAE~\citep{lerner2022viquae}, VQAv2~\citep{goyal2017making}, Visual Dialog~\citep{das2017visual}
\end{itemize}
According to the guidelines from the VILA repository, the aforementioned instruction-tuning data is all included in M$^3$IT~\citep{li2023m3it}. Therefore, we use M$^3$IT to analyze the training data for VILA-1.5. The final word frequency statistics and word cloud for the VILA-1.5-13B training data are shown in Figure~\ref{fig:word_vila}.

\textbf{PLLaVA-13B}. PLLaVA-13B is based on LLaVA-Next~\citep{liu2024llavanext}, with LLaVA-Next incorporating additional data from ShareGPT-4V~\citep{chen2023sharegpt4v} compared to LLaVA-1.5-13B. The main enhancement from LLaVA-Next to PLLaVA-13B is the addition of 783k instructional video-to-text tuning data, enabling LLaVA-Next to handle video input. Specifically, the training data are sourced from the dataset used in VideoChat2~\citep{li2023mvbench}, which includes data for various video understanding tasks, such as 27k conversation videos from VideoChat~\citep{li2023videochat} and Video-ChatGPT~\citep{maaz2023videochatgpt}, 80k classification task data from Kinetics~\citep{kay2017kinetics} and SthSthV2~\citep{goyal2017sthsthv2}, 450k captioned data from Webvid~\citep{bain2021webvid}, YouCook2~\citep{zhou2018youcook2}, TextVR~\citep{wu2023textvr}, and VideoChat, 117 reasoning data points from NextQA~\citep{xiao2021nextqa} and CLEVRER~\citep{yi2019clevrer}, and 109k annotated question-answering samples from Webvid, TGIF~\citep{li2016tgif}, and Ego4D~\citep{grauman2022ego4d}.

\begin{figure}[th!]
    \centering  
    \vspace{-1.5mm}
    \includegraphics[width=1.0\linewidth]{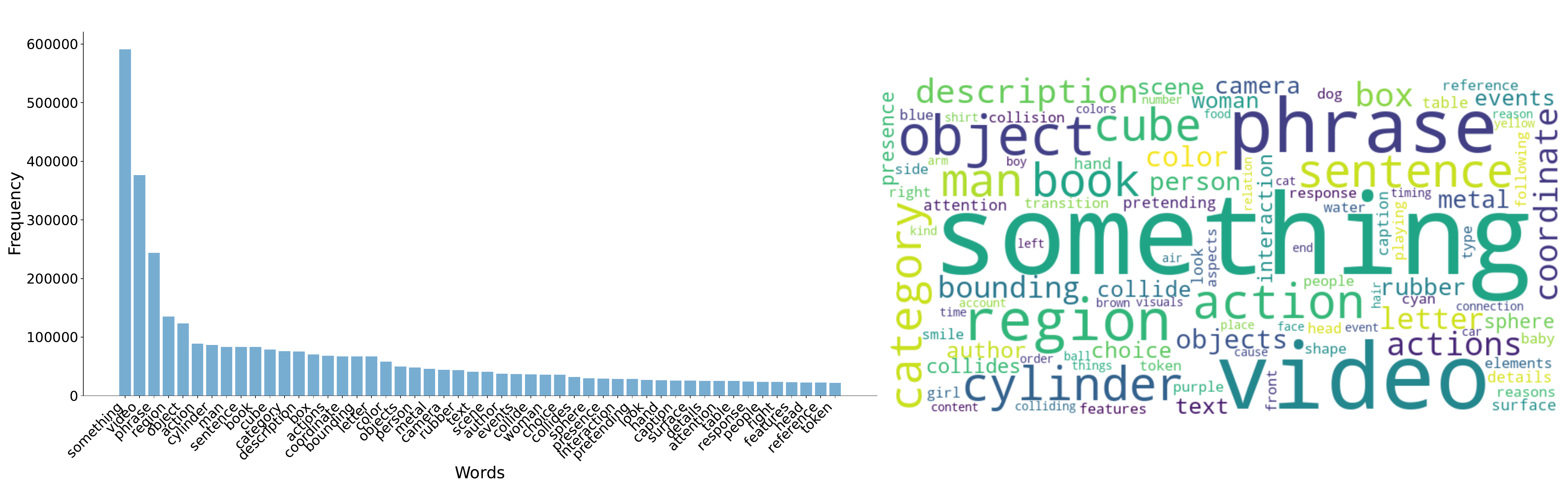}
    \vspace{-1.5mm}
    \caption{Word Statics and Word Cloud for PLLaVA-13B Training Data.}
    \label{fig:word_pllava}
    \vspace{-3mm}
\end{figure}

We used ShareGPT-4V and downloaded all the training data from the PLLaVA repository, creating a word cloud and word frequency chart of the training data, as shown in Figure~\ref{fig:word_pllava}.


\begin{figure}[th!]
    \centering
    \vspace{-1.5mm}
    \begin{minipage}[b]{0.328\linewidth}
        \centering
        \includegraphics[width=\linewidth]{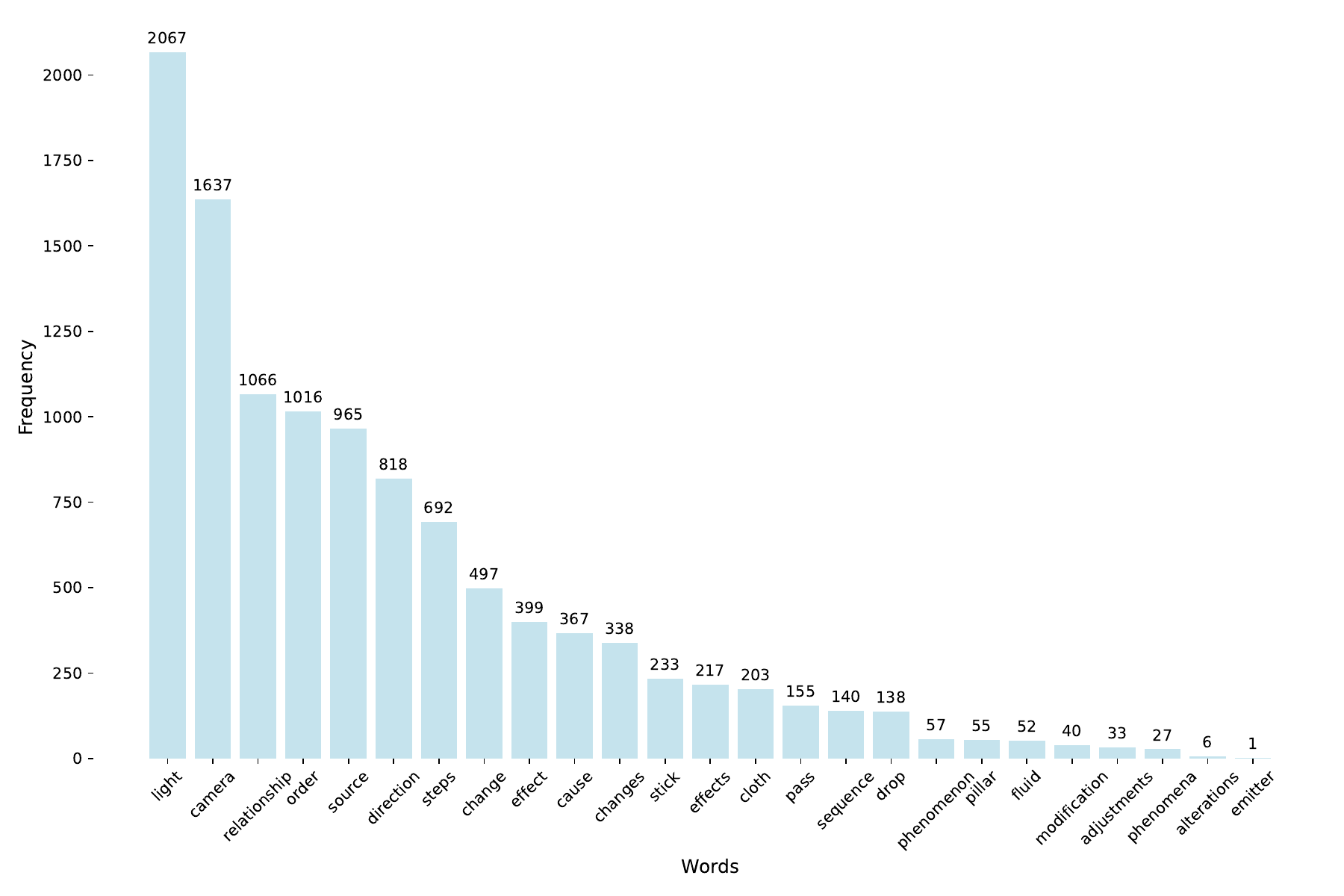}
    \end{minipage}
    \begin{minipage}[b]{0.328\linewidth}
        \centering
        \includegraphics[width=\linewidth]{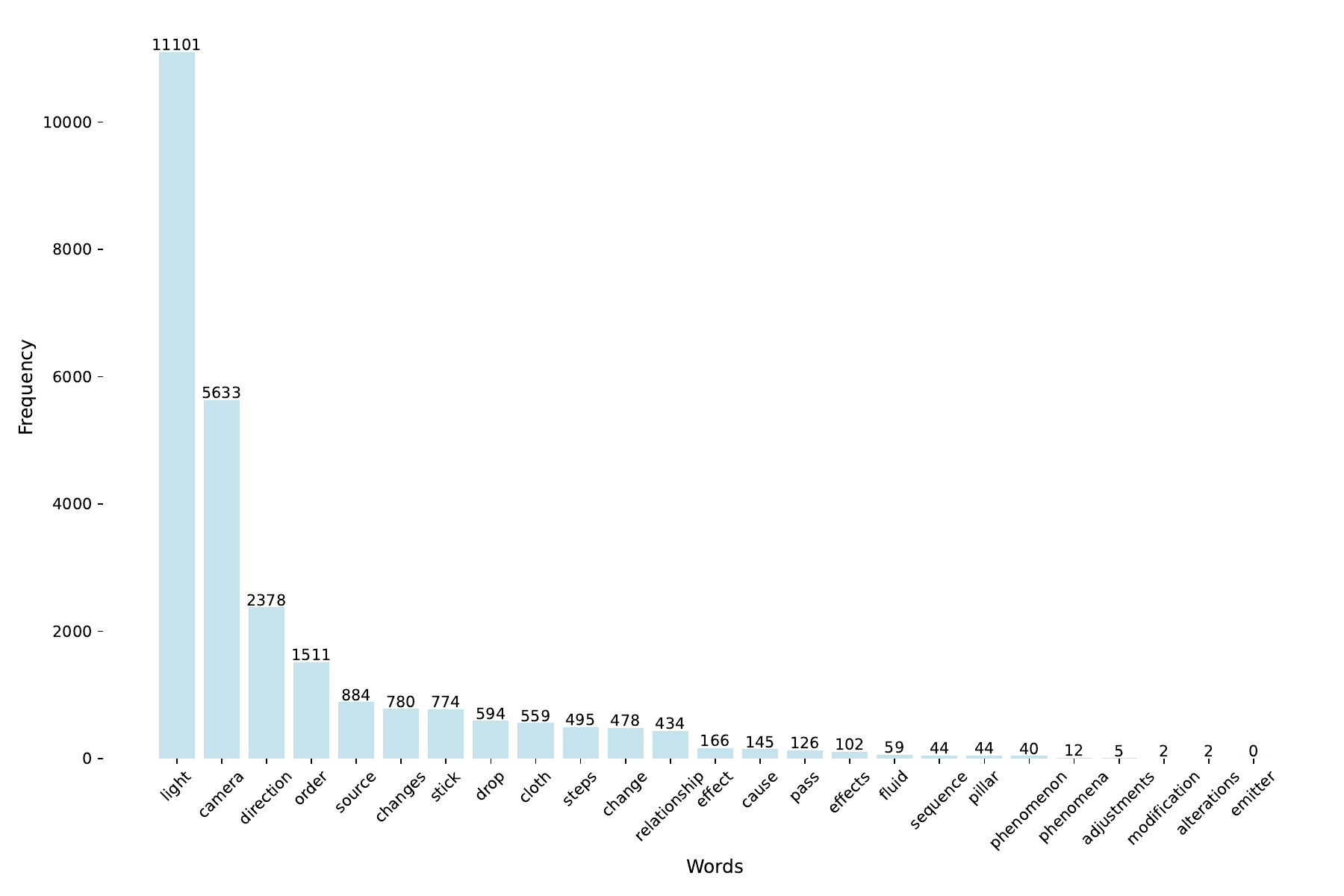}
    \end{minipage}
    \begin{minipage}[b]{0.328\linewidth}
        \centering
        \includegraphics[width=\linewidth]{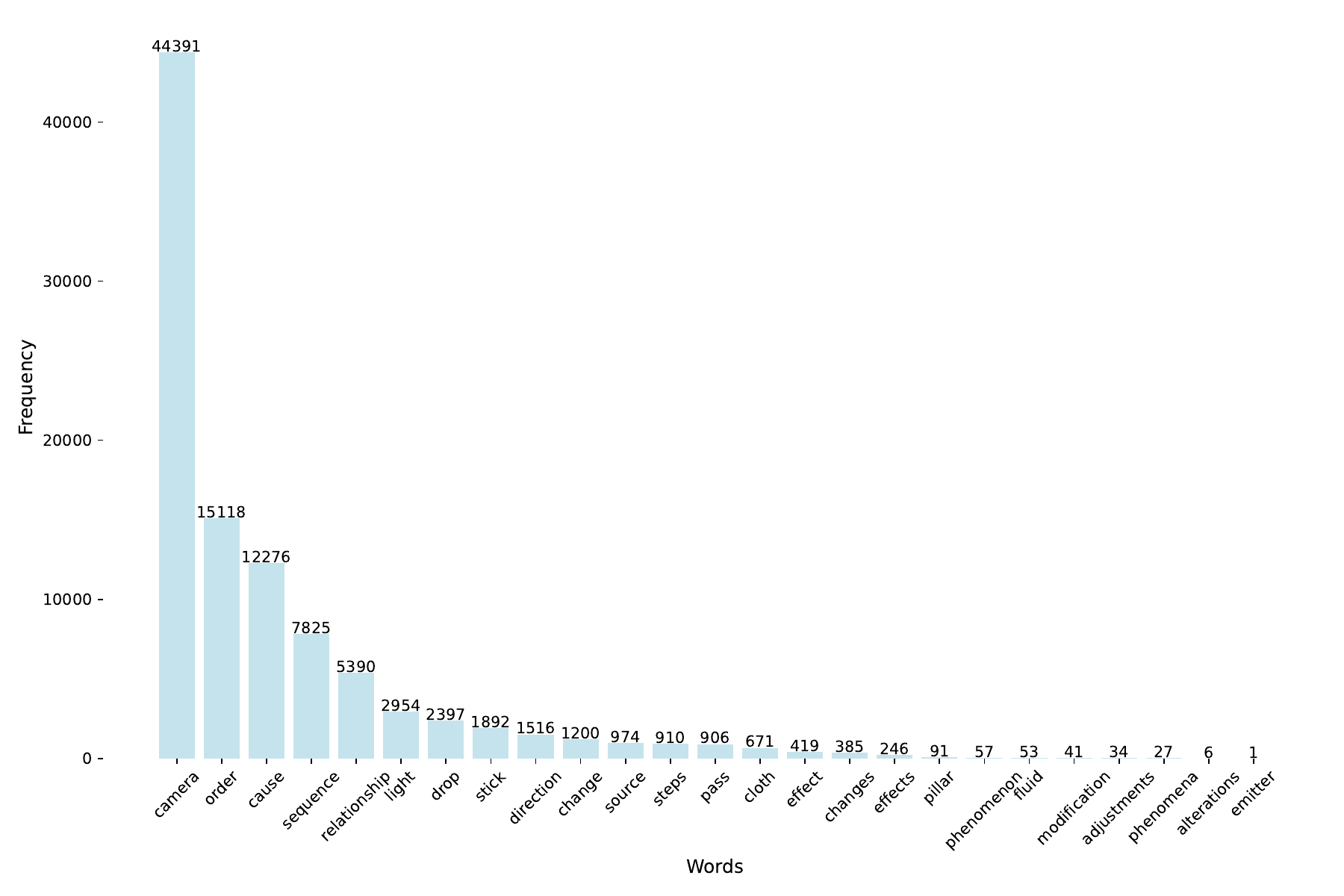}
    \end{minipage}
    \vspace{-1.5mm}
    \caption{The frequency of common terms in PhysBench within the training data of the LLaVA-1.5-13B, VILA-1.5-13B, and PLLaVA-13B models.}
    \label{fig:word_our_in_llava_series}
    \vspace{-3mm}
\end{figure}

Observing Figure~\ref{fig:word_llava}~\ref{fig:word_vila}~\ref{fig:word_pllava}, we can see that the most frequent words in the training data of LLaVA-1.5-13B, VILA-1.5-13B, and PLLaVA-13B are terms like 'description', 'phrase', 'summary', 'region', and similar words. Keywords such as 'direction' appear much less frequently. We have listed the frequency of several key terms from our dataset in the training data of LLaVA-1.5-13B, VILA-1.5-13B, and PLLaVA-13B, as shown in Figure~\ref{fig:word_our_in_llava_series}. Although PLLaVA-13B includes words like 'collides' and 'camera', they are primarily used to describe phenomena without addressing the underlying physical mechanisms, which may explain why PLLaVA-13B shows no substantial improvement.

\section{More Details on the Setup}\label{app:setup}
\subsection{Prompt for LLMs}\label{app:llm_prompt}
\textbf{Video Cases Prompt}. The general system prompt for LLMs to generate video or image cases is as follows: 

\begin{mycase}{light_grey}
You are an assistant that communicates only in JSON and is an expert in physics. Do not write normal text.

Your response should be in JSON format with the following string fields:

1.  `video description`: A description of the video.

2.  `question`: A multiple-choice question based on the video's content.

3.  `answer`: A response to the question, with one of the values being `A`, `B`, `C`, or `D`.

For the same `video description`, you may provide multiple question-answer pairs, which should be presented in a list format.

Important notes:
1. The `question` should not contain details from the `video description`.
2. The relationship between each question-answer pair is one of juxtaposition, not progression.

The following is a conversation between the user and the assistant. Ensure that the assistant's answers are as rich and informative as possible.
\end{mycase}

Then we will the below sentence as the first turn chat and let the LLMs give the more cases in the latter turns.
We replace the examples provided in the first turn to enable GPT to generate more diverse responses. Notably, some of the examples generated by GPT do not fully align with the data standards defined in Appendix~\ref{app:protocol}. We use these examples solely as references to collect more diverse data. However, as described in Section~\ref{sec3:2}, the subsequent data annotation undergoes a rigorous five-step manual process to ensure high data quality.  

The process of generating VQA templates is similar: questions are generated based on the captions of the input video or image. However, QA templates are more focused, targeting specific types of relationships, such as those related to spatial positioning.

\begin{mycase}{light_grey}
    \textbf{\{"role": "system", "content": (Contents of the color block above)\}}\\
    \textbf{\{"role": "user", "content": "give me some cases"\}}\\
    \textbf{\{"role": "assistant", "content":} 
    \begin{itemize}
        \item \textbf{Video Description:} A ball thrown horizontally from a cliff
        \begin{itemize}
            \item \textbf{Questions:}
            \begin{enumerate}
                \item What causes the ball to fall downward?\\
                \qquad A. Air resistance\\
                \qquad B. Gravitational force\\
                \qquad C. Magnetism\\
                \qquad D. Elastic collision
                \item Which of the following affects how far the ball travels horizontally?\\
                \qquad A. Mass of the ball\\
                \qquad B. Initial velocity of the ball\\
                \qquad C. Color of the ball\\
                \qquad D. Shape of the ball
                \item What happens if the ball is thrown with a higher initial velocity?\\
                \qquad A. It travels farther horizontally\\
                \qquad B. It travels shorter horizontally\\
                \qquad C. It stops mid-air\\
                \qquad D. It moves vertically upward
            \end{enumerate}
            \item \textbf{Answers:} [B, B, A]
        \end{itemize}
        
        \item \textbf{Video Description:} A feather falling in a vacuum chamber
        \begin{itemize}
            \item \textbf{Questions:}
            \begin{enumerate}
                \item What happens if the vacuum chamber is filled with air?\\
                \qquad A. The feather falls faster\\
                \qquad B. The feather falls at the same speed\\
                \qquad C. The feather floats\\
                \qquad D. The feather rises upward
                \item What would happen to the action if it happened in a strong magnetic field?\\
                \qquad A. It would fall faster\\
                \qquad B. It would float\\
                \qquad C. It would fall slower\\
                \qquad D. It would remain unaffected
            \end{enumerate}
            \item \textbf{Answers:} [A, D]
        \end{itemize}
    \end{itemize}

    \textbf{\}},\\
    \textbf{\{"role": "user", "content": "give me some cases"\}}
\end{mycase}

\textbf{Caption}. video or image caption system prompt is as follows:
\begin{mycase}{light_grey}
You are an outstanding video annotator and physicist.

Your task is to provide a detailed description of a given video based on the user's input video and a brief description of the video.

The description should include, but is not limited to, the following aspects:

1. The content of the video.

2. The principles behind the phenomena observed in the video.

3. The properties of certain objects within the video, such as estimated size, mass, color, and type.

4. Various spatial information, including relative and absolute positions, sizes, distances, and movements.

Ensure that the descriptions are thorough, precise, and reflect your expertise in physics.
\end{mycase}

\textbf{Answer Extraction}. The prompt used to instruct GPT-4o is illustrated  as follows:
\begin{mycase}{light_grey}
The following sentences contain answers (one of A, B, C, D) and corresponding analysis.

Your role is to find the answer.

Please return only one of the four letters: A, B, C, or D.

The sentences are:
\end{mycase}

\subsection{3D Assets}
\begin{figure}[th!]
	\centering  
	\vspace{-1.5mm}
	\includegraphics[height=4.15cm]{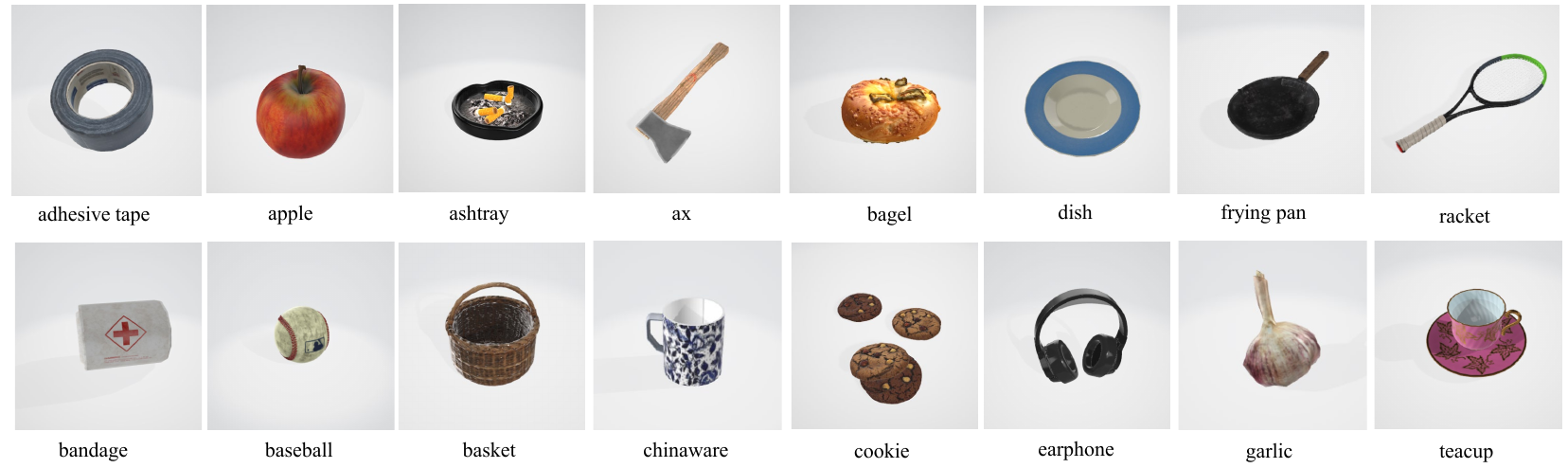}
	\vspace{-1.5mm}
    	\caption{\textbf{3D Assets Sample}. We have 678 objects and 470 HDR environment in total for simulation.}
	\label{fig:3d_assets}  
	\vspace{-3mm}
\end{figure}

The overview of the 3D assets is shown in Figure~\ref{fig:3d_assets}. After defining the object attribute table, we selected 678 objects and annotated their attributes accordingly. For each object, we then adjusted its size and position to ensure that multiple objects (typically 4-5 in our dataset) do not overlap and remain clearly visible. These objects were subsequently used for simulations.

\subsection{Model Hyperparameters}\label{app:hyper}
In our experiments, we conducted comparisons with some of the most recent and representative MLLMs in the following. In our experiments, we conducted comparisons with some of the most recent and representative MLLMs in the following. We divided the models into three categories: \textbf{Image VLM}, \textbf{Video VLM}, and \textbf{General VLM}, according to whether they support only one image input, support a video input (also supports image), or support interlaced video and multiple images.

Following~\cite{zhang2024task}, for Image VLMs, we concatenate multiple frames from a video (defaulting to 8 frames) into a single image, arranged from left to right and bottom to top, before inputting it into the model. Thus, both Image VLMs and Video VLMs were tested on question-answer pairs involving only a single image or video, which represent a subset of PhysBench. In contrast, General VLMs, which support interleaved image-text sequences, were tested on the full PhysBench test split.

For most open-source models, we used the hyperparameter torch\_dtype = torch.float16. However, for some models that only support torch\_dtype = torch.bfloat16, we used torch\_dtype = torch.bfloat16 for those cases. Additionally, for other parameter settings, we generally followed the configurations provided in the original papers, their code repositories, or the examples from Hugging Face.

\subsubsection{Image VLM}
The following is a description of the specific models reviewed, as well as the specific parameter configurations, can be found in Table~\ref{tab:model_params_image_only}.

\begin{itemize}
\item
\textbf{BLIP-2}~\citep{li2023blip} employs a dual-stage strategy to bridge the modality gap effectively, utilizing a lightweight Q-Former pre-trained on 129 million image-text pairs. In the first stage, the model initiates the learning of vision-language representations by leveraging a frozen image encoder, ViT-g/14 from EVA-CLIP~\citep{fang2023eva}. In the subsequent stage, a frozen LLM, FlanT5~\citep{chung2024scaling}, is employed to facilitate vision-to-language generative learning. This innovative approach enables effective zero-shot instructed image-to-text generation. The tested version is BLIP2-t5-xxl (we call it as BLIP-2 in this paper).

\item
\textbf{InstructBLIP}~\citep{dai2024instructblip} is derived from the pre-trained BLIP-2 model~\citep{li2023blip}, which integrates a ViT-g/14 image encoder, a Vicuna~\citep{vicuna2023}, and a Q-Former that bridges these two components. During the vision-language instruction tuning process, only the Q-Former is fine-tuned, utilizing data from 13 distinct visual question-answering datasets. The tested version is InstructBLIP-t5-xl (we call it as InstructBLIP-t5-xl in this paper), InstructBLIP-t5-xxl (we call it as InstructBLIP-t5-xxl in this paper), InstructBLIP-vicuna-7b (we call it as InstructBLIP-7B in this paper) and InstructBLIP-vicuna-13b (we call it as InstructBLIP-13B in this paper). It is worth noting that both InstructBLIP-vicuna-7b and InstructBLIP-vicuna-13b provided no responses to a small part of questions. For these instances, we disregarded these questions and did not assign any scores.

\item 
\textbf{LLaVA-1.5}~\cite{liu2023improved}. LLaVA~\citep{liu2024visual} establishes a connection between the visual encoder ViT-L/14 from CLIP~\citep{radford2021learning} and the language decoder LLaMA~\citep{touvron2023llama} using a lightweight, fully connected (FC) layer. Initially, the system trains this FC layer with 595K image-text pairs, while keeping both the visual encoder and LLM static. Subsequently, LLaVA fine-tunes both the FC layer and the LLM using a dataset of 158K instructional vision-language pairs. LLaVa-1.5~\citep{liu2023improved} is an enhanced version of LLaVA, trained on additional datasets. The tested version is LLaVa-1.5-7b and LLaVa-1.5-13b (we call it as LLaVA-1.5-7B and LLaVA-1.5-13B in this paper).

\item 
\textbf{Qwen-VL-Chat}~\citep{bai2023qwen} introduces a series of large-scale vision-language models (LVLMs) designed to perceive and understand both text and images. The Qwen-VL series is built upon the Qwen-LM~\citep{qwen} foundation, augmented with visual capabilities through a meticulously designed visual receptor, input-output interface, and a three-stage training pipeline using a multilingual multimodal cleaned corpus. These models excel in image description, question-answering, grounding, and text-reading by aligning image-caption-box tuples. The series includes Qwen-VL and Qwen-VL-Chat, both of which achieve state-of-the-art performance on a broad range of visual-centric benchmarks and real-world dialog benchmarks, demonstrating their superiority over existing vision-language chatbots. The tested version is Qwen-VL-Chat, as Qwen-VL-Chat have stronger instruction fellow ability than Qwen-VL.

\item
\textbf{LLaVA-NeXT}~\citep{liu2024llavanext} is the latest iteration in the LLaVA series~\citep{liu2024visual}, building upon the foundation of LLaVA-1.5~\citep{liu2023improved}. This new model enhances visual reasoning, OCR, and world knowledge capabilities. It increases input image resolution to four times more pixels, supporting resolutions up to 672x672, 336x1344, and 1344x336. LLaVA-NeXT features an improved visual instruction tuning data mixture, further enhancing its visual reasoning and OCR capabilities. Additionally, it supports better visual conversations across various scenarios and applications, demonstrating improved world knowledge and logical reasoning. Despite these enhancements, LLaVA-NeXT maintains the minimalist design and data efficiency of its predecessor, utilizing less than 1M visual instruction tuning samples. The tested versions are LLaVA-NeXT-mistral-7b (LLaVA1.6-mistral) and LLaVA-NeXT-vicuna-7b (LLaVA1.6-vicuna), with the base models being Mistral-7b~\citep{jiang2023mistral} and Vicuna-7b~\citep{vicuna2023}, respectively (we call it as LLaVA1.6-mistral and LLaVA1.6-vicuna in this paper).

\item
\textbf{InternVL-Chat-V1-5}~\citep{chen2024far} is an open-source multimodal large language model with enhanced visual understanding through a continuous learning vision encoder, dynamic high-resolution image processing, and a high-quality bilingual dataset. The tested version is InternVL-Chat-V1-5-quantable (we call it as InternVL-Chat1.5), as the GPU memory size restriction of 40G A6000.

\item
\textbf{Cambrian-1}~\citep{tong2024cambrian} is a family of multimodal large language models (MLLMs) designed with a vision-centric approach. This model series addresses the gap between language models and visual representation learning by thoroughly evaluating various visual representations. Cambrian-1 introduces the Spatial Vision Aggregator (SVA), a dynamic and spatially-aware connector that integrates high-resolution vision features with large language models (LLMs) via cross-attention layers~\citep{dai2024instructblip} while reducing the number of tokens. The tested version is Cambrian-1-8b (we call it as Cambrian-8B in this paper).

\item
\textbf{MiniCPM-V}~\citep{yao2024minicpmvgpt4vlevelmllm}. MiniCPM-V2 is a robust multimodal large language model designed for efficient end-side deployment. The model is constructed by integrating SigLip-400M~\citep{zhai2023sigmoid} and MiniCPM-2.4B~\citep{hu2024minicpm}, connected through a perceiver resampler. The latest iteration, MiniCPM-V2.5, further improves upon its predecessors. Built on SigLip-400M and Llama3-8B-Instruct with a total of 8B parameters, MiniCPM-V2.5 demonstrates significant performance enhancements over MiniCPM-V2. The most recent and advanced model in the MiniCPM-V series, MiniCPM-V2.6, is built on SigLip-400M and Qwen2-7B~\citep{qwen}, also with a total of 8B parameters. MiniCPM-V2.6 shows substantial performance improvements over MiniCPM-Llama3-V2.5 and introduces new capabilities for multi-image and video understanding.
\item
\textbf{MolmoE}~\citep{deitke2024molmo}. MolmoE is a family of open vision-language models developed by the Allen Institute for AI. The Molmo models are trained on PixMo, a dataset of 1 million highly curated image-text pairs, and exhibit state-of-the-art performance among multimodal models of similar size, while remaining fully open-source. MolmoE-1B is a multimodal Mixture-of-Experts large language model (LLM) with 1.5B active and 7.2B total parameters, based on OLMoE-1B-7B-0924~\citep{muennighoff2024olmoeopenmixtureofexpertslanguage}. It closely matches the performance of GPT-4V on both academic benchmarks and human evaluations, achieving state-of-the-art performance among similarly-sized open multimodal models. Molmo 7B-O, based on OLMo-7B-1024 (a preview of the next generation of OLMo models), utilizes OpenAI CLIP as its vision backbone, performing between GPT-4V and GPT-4o on both academic benchmarks and human evaluations. Molmo 7B-D, based on Qwen2-7B and also using OpenAI CLIP as its vision backbone, performs similarly, sitting between GPT-4V and GPT-4o in both academic benchmarks and human evaluations. It powers the Molmo demo available at molmo.allenai.org. Finally, Molmo 72B, based on Qwen2-72B with OpenAI CLIP as the vision backbone, achieves the highest academic benchmark score and ranks second in human evaluations, trailing GPT-4o by only a small margin.
\item 
\textbf{Xinyuan-VL}~\citep{cylingo2024xinyuanvl2b}. Xinyuan-VL-2B is a high-performance multimodal large model developed by the Cylingo Group for end-user applications. It is fine-tuned based on Qwen2-VL-2B~\citep{Qwen2VL} and trained on over 5 million multimodal data samples, with additional training on a small amount of plain text data.
\item 
\textbf{Aquila-VL}~\citep{gu2024infinitymmscalingmultimodalperformance}. The Aquila-VL-2B model is a VLM trained using the LLava framework. The Qwen2.5-1.5B~\citep{Qwen2VL} model serves as the LLM, while siglip-so400m-patch14-384 is employed as the vision tower.
The model was trained on our custom-built Infinity-MM dataset, which consists of approximately 40 million image-text pairs. This dataset combines open-source data collected from the internet with synthetic instruction data generated using open-source VLM models.
\item 
\textbf{DeepSeek-VL}~\citep{lu2024deepseekvl}. DeepSeek-VL is an open-source Vision-Language (VL) model designed for real-world applications involving vision and language understanding. It demonstrates broad multimodal capabilities, enabling the processing of logical diagrams, web pages, formula recognition, scientific literature, natural images, and embodied intelligence in complex scenarios. The models we tested, with sizes of 1B and 7B parameters, are respectively sourced from deepseek-ai/deepseek-vl-1.3b-chat and deepseek-ai/deepseek-vl-7b-chat.

\textbf{PaliGemma 2}~\citep{PaliGemma} represents a significant advancement in vision-language modeling, building upon its predecessor by incorporating the sophisticated Gemma 2 architecture. Following the approach of PaLI-3~\citep{chen2023pali3visionlanguagemodels}, the PaliGemma model family utilizes open-source components, specifically combining the SigLIP vision model with Gemma 2 language models~\citep{team2024gemma}. This multimodal system effectively processes both visual and textual inputs to generate multilingual outputs. The architecture has been optimized to achieve superior fine-tuning performance across a comprehensive range of vision-language tasks, including image and short video captioning, visual question answering, optical character recognition, object detection, and instance segmentation. The model variants include paligemma2-10b-ft-docci-448, paligemma2-3b-ft-docci-448.

\item 
\textbf{Claude}~\citep{anthropic_claude_models} is a large language model developed by Anthropic with a strong focus on safety, interpretability, and alignment. It has been designed to align with human intent and values, using various methodologies like reinforcement learning from human feedback (RLHF) to ensure that the model behaves as intended. Claude integrates mechanisms to reduce harmful or biased outputs and is optimized for interactive dialogue across a range of domains. In this paper, the tested version is Claude-3-opus, Claude-3-sonnet, Claude-3-haiku, Claude-3.5-sonnet. Except for Claude-3-haiku, where images were used in their original size, the images for the other three models were resized to 128$\times$128 due to cost considerations. Notably, both Claude-3-sonnet and Claude-3.5-sonnet demonstrated poor adherence to instructions, often failing to provide direct answers as requested in the prompt, instead offering explanations alongside the options. Therefore, we used GPT-4o to extract the answers for scoring, with the prompt provided in Appendix~\ref{app:llm_prompt}.
\end{itemize}

\begin{table}[th!]
\small
\centering
\begin{tabular}{p{0.24\linewidth}  p{0.69\linewidth}}
\toprule
\textbf{Model} & \textbf{Generation Setup} \\
\midrule
BLIP-2 & torch\_dtype = torch.float16, max\_new\_tokens = 200\\ \midrule
InstructBLIP-t5-xl & torch\_dtype = torch.float16, max\_new\_tokens = 200 \\ \midrule
InstructBLIP-t5-xxl &torch\_dtype = torch.float16, max\_new\_tokens = 200 \\ \midrule
InstructBLIP-7B &torch\_dtype = torch.float16, max\_new\_tokens = 200 \\ \midrule
InstructBLIP-13B &torch\_dtype = torch.float16, max\_new\_tokens = 200 \\ \midrule
LLaVA-1.5-7B& torch\_dtype = torch.float16, max\_new\_tokens = 200, do\_sample = False\\ \midrule
LLaVA-1.5-13B&torch\_dtype = torch.float16, max\_new\_tokens = 200, do\_sample = False \\ \midrule
Qwen-VL-Chat& torch\_dtype = torch.bfloat16 \\ \midrule
LLaVA1.6-mistral &torch\_dtype = torch.float16, max\_new\_tokens = 200, do\_sample = False \\ \midrule
LLaVA1.6-vicuna & torch\_dtype = torch.float16, max\_new\_tokens = 200, do\_sample = False\\ \midrule
\multirow{2}{*}{InternVL-Chat1.5} & torch\_dtype = torch.bfloat16, max\_new\_tokens = 512, num\_beams = 1, do\_sample = False\\ \midrule
\multirow{2}{*}{Cambrian-8B} & torch\_dtype = torch.float16, max\_new\_tokens = 512, temperature = 0, num\_beams = 1,  use\_cache = True\\ \midrule
\multirow{2}{*}{MiniCPM-V2} & dtype = torch.float32, context = None, sampling = False, temperature = 0.1, max\_new\_token = 10 \\\midrule
\multirow{2}{*}{MiniCPM-V2.5} & dtype = torch.float32, context = None, sampling = False, temperature = 0.1, max\_new\_token = 10 \\\midrule
\multirow{2}{*}{MiniCPM-V2.6} & dtype = torch.float32, context = None, sampling = False, temperature = 0.1, max\_new\_token = 10 \\\midrule
MolmoE-1B & dtype = torch.float32, stop\_strings = stop\_strings=\texttt{"<|endoftext|>"}, dtype = 'auto', max\_new\_tokens = 200 \\ \midrule
MolmoE-7B-O & dtype = torch.float32, stop\_strings = stop\_strings=\texttt{"<|endoftext|>"}, dtype = 'auto', max\_new\_tokens = 200 \\ \midrule
MolmoE-7B-D & dtype = torch.float32, stop\_strings = stop\_strings=\texttt{"<|endoftext|>"}, dtype = 'auto', max\_new\_tokens = 200 \\ \midrule
MolmoE-72B & dtype = torch.float32, stop\_strings = stop\_strings=\texttt{"<|endoftext|>"}, dtype = 'auto', max\_new\_tokens = 200 \\ \midrule
Xinyuan-VL & max\_new\_tokens = 128, resize = (1024, 1024) \\ \midrule
Aquila-VL & do\_sample = False, temperature = 0, max\_new\_tokens = 4096 \\ \midrule
DeepSeek-VL-1B & \\ dtype = torch.bfloat16, max\_new\_tokens = 10, do\_sample=False, use\_cache=True \\ \midrule
DeepSeek-VL-7B & \\ dtype = torch.bfloat16, max\_new\_tokens = 10, do\_sample=False, use\_cache=True \\ \midrule
Claude-3-opus & max\_new\_tokens = 1000, temperature = 0\\ \midrule
Claude-3-sonnet & max\_new\_tokens = 1000, temperature = 0\\ \midrule
Claude-3-haiku & max\_new\_tokens = 1000, temperature = 0\\ \midrule
Claude-3.5-sonnet & max\_new\_tokens = 1000, temperature = 0\\
\bottomrule
\end{tabular}
\caption{Generating parameters for Image VLM. Parameters not explicitly stated indicate the use of the model's default system settings.}
\label{tab:model_params_image_only}
\end{table}
\subsubsection{Video VLM}\label{app:setup_video_vlm}
The following is a description of the specific models reviewed, as well as the specific parameter configurations, can be found in Table~\ref{tab:model_params_image_video}.
\begin{itemize}
\item
\textbf{Chat-Univi}~\citep{jin2023chatunivi} is a unified vision-language model capable of comprehending and engaging in conversations involving both images and videos through a unified visual representation. Chat-Univi employs a set of dynamic visual tokens to uniformly represent images and videos, enabling it to efficiently utilize a limited number of visual tokens to capture the spatial details necessary for images and the comprehensive temporal relationships required for videos. This approach is supported by a multiscale representation that allows the model to perceive both high-level semantic concepts and low-level visual details. The tested version is Chat-Univi-7B and Chat-Univi-13B.

\item
\textbf{Video-LLaVA}~\citep{lin2023video}is a sophisticated multi-modal framework designed to empower large language models (LLMs) with the ability to understand both visual and auditory content in videos. Unlike previous models that handle either visual or audio signals independently, Video-LLaVA integrates both to achieve comprehensive video comprehension. The tested version is Video-LLaVA-7b (we call it as Video-LLaVA in this paper).

\item
\textbf{PLLaVA}~\citep{xu2024pllava}(Pooling LLaVA) is an advanced video understanding model designed to extend image-based models to video, enabling dense video caption generation. The model employs a simple yet powerful pooling module to bridge image-finetuned Vision-Language Models (Vision-LLM) into the video domain, achieving significant performance improvements in video captioning tasks. The tested versions are PLLaVA-7B and PLLaVA-13B. Notably, since the model uses num\_frame as a parameter during loading, our approach for processing images involved duplicating the image for the num\_frames input.
\end{itemize}

We also evaluated VideoChatGPT~\citep{maaz2023video}, Video-LLaMA~\citep{damonlpsg2023videollama}, and VideoChat2~\citep{li2024mvbench}, and observed that these models exhibit poor instruction-following capabilities. Despite repeatedly prompting them to provide options, the models consistently responded with descriptions of the video or image content rather than addressing the questions posed. As a result, we did not include specific evaluation metrics for these models in our study.

\begin{table}[th!]
\small
\centering
\begin{tabular}{p{0.24\linewidth}  p{0.69\linewidth}}
\toprule
\textbf{Model} & \textbf{Generation Setup} \\
\midrule
\multirow{2}{*}{Video-LLaVA} & torch\_dtype = torch.float16, max\_new\_tokens = 1024, temperature = 0.1, use\_cache = True, do\_sample = True \\ \midrule
\multirow{3}{*}{Chat-Univi-7B} & torch\_dtype = torch.float16, max\_new\_tokens = 10, temperature = 0, num\_beams = 1,  use\_cache = True, do\_sample = False, top\_p = None, output\_scores = True, return\_dict\_in\_generate = True, length\_penalty = 1 \\ \midrule
\multirow{3}{*}{Chat-Univi-13B} & torch\_dtype = torch.float16, max\_new\_tokens = 10, temperature = 0, num\_beams = 1,  use\_cache = True, do\_sample = False, top\_p = None, output\_scores = True, return\_dict\_in\_generate = True, length\_penalty = 1 \\ \midrule
\multirow{2}{*}{PLLaVA-7B} &  conv\_mode = conv\_eval\_videoqabench, max\_new\_tokens = 256,  do\_sample = False\\ \midrule
\multirow{2}{*}{PLLaVA-13B} & conv\_mode = conv\_eval\_videoqabench, max\_new\_tokens = 256,  do\_sample = False\\
\bottomrule
\end{tabular}
\caption{Generating parameters for Video VLM. Parameters not explicitly stated indicate the use of the model's default system settings.}
\label{tab:model_params_image_video}
\end{table}
\subsubsection{General VLM}
The following is a description of the specific models reviewed, as well as the specific parameter configurations, can be found in Table~\ref{tab:model_params_general}, ~\ref{tab:model_params_general2}.

\begin{itemize}
\item
\textbf{LLaVA-NeXT-Interleave}~\citep{li2024llavanextinterleavetacklingmultiimagevideo} is a multimodal large language model that extends LLaVA's capabilities to handle multi-image, video, and 3D data. By using an interleaved data format, it unifies single-image and multi-modal tasks, enabling it to transfer knowledge across different scenarios. The model is trained on the M4-Instruct dataset, which includes over a million samples spanning multi-image, video, and 3D tasks. The tested versions are llava-interleave-qwen-7b-hf (we call it as LLaVA-interleave) and  llava-interleave-qwen-7b-dpo-hf (we call it as LLaVA-interleave-dpo in this paper).

\item
\textbf{VILA-1.5}~\citep{lin2023vila} is a multimodal visual language model (VLM) designed to handle both multi-image and video understanding tasks. It is pretrained on large-scale interleaved image-text data, which enhances its ability to perform tasks such as video reasoning and in-context learning. The tested versions are VILA-1.5-3B, VILA-1.5-3B-s2, VILA-1.5-8B and VILA-1.5-13B.

\item
\textbf{NVILA}~\citep{liu2024nvila} is a family of open Vision-Language Models (VLMs) optimized for efficient video and multi-image understanding. We enhance VILA's architecture through a 'scale-then-compress' approach, increasing spatial and temporal resolutions before compressing visual tokens, enabling efficient processing of high-resolution images and long videos. Through systematic optimization across training, fine-tuning, and deployment phases, NVILA achieves comparable or superior performance to leading VLMs while reducing training costs by 4.5×, fine-tuning memory by 3.4×, and latency by up to 2.8×. The code and models are publicly available for reproducibility.

\item
\textbf{Phi-3V}~\citep{abdin2024phi}. Phi-3 is a family of open AI models developed by Microsoft, designed to be the most capable and cost-effective small language models (SLMs) available. Phi-3 models outperform other models of the same size and even those in the next size up across various language, reasoning, coding, and math benchmarks. Phi-3V is the VLM based on Phi-3. The tested version is Phi-3V-128k (we call it as Phi-3V in this paper).
\item
\textbf{Phi-3.5V}~\citep{phi-3.5-vision}. Phi-3.5V, released on November 15, 2024, is an advanced version of Phi-3V. This state-of-the-art, lightweight multimodal model is built on datasets comprising synthetic data and curated, publicly available web content, with a focus on high-quality, reasoning-rich data in both text and vision. As part of the Phi-3 model family, the multimodal version supports a context length of 128K tokens. The model has undergone a comprehensive enhancement process, including supervised fine-tuning and direct preference optimization, to ensure accurate adherence to instructions and robust safety measures.
\item
\textbf{mPLUG-Owl3}~\citep{ye2024mplugowl3longimagesequenceunderstanding}. 
Here’s a refined version of your sentence: mPLUG-Owl3 is a state-of-the-art multimodal large language model designed to address the challenges of understanding long image sequences. mPLUG-Owl3 introduces \textit{Hyper Attention}, a method that enhances the speed of visual sequence processing in multimodal large language models by sixfold, enabling the handling of sequences up to eight times longer. Simultaneously, mPLUG-Owl3 maintains exceptional performance across single-image, multi-image, and video tasks. The tested versions are mPLUG-Owl3-1B (mPLUG-Owl3-1B-241014), mPLUG-Owl3-2B (mPLUG-Owl3-2B-241014) and mPLUG-Owl3-7B (mPLUG-Owl3-7B-241101).

\item 
\textbf{InternVL2}~\citep{wang2024mpo}. InternVL 2.0 is the latest iteration in the InternVL series of multimodal large language models. It includes a range of instruction-tuned models, with parameter sizes ranging from 1 billion to 108 billion.
In comparison to state-of-the-art open-source multimodal large language models, InternVL 2.0 outperforms most open-source alternatives and demonstrates competitive performance that rivals proprietary commercial models. Its capabilities include document and chart comprehension, infographics question answering, scene text understanding and OCR tasks, scientific and mathematical problem-solving, as well as cultural understanding and integrated multimodal processing.
InternVL 2.0 is trained with an 8K context window and incorporates training data consisting of long texts, multiple images, and videos. This training significantly enhances its ability to process and understand these types of inputs, surpassing the capabilities of InternVL 1.5~\citep{chen2023internvl}. For larger model variants, we implement a merge-based approach rather than sequential processing for video data to optimize GPU memory consumption, as demonstrated in Table~\ref{main_experiment_new}.

\item
\textbf{InternVL 2.5}~\citep{gao2024mini} was released on December 9, 2024, representing an advanced iteration in the multimodal large language model (MLLM) series. While maintaining the core architecture of InternVL 2.0, it introduces significant enhancements in training strategies, testing methodologies, and data quality. The model preserves the "ViT-MLP-LLM" paradigm established by its predecessors, InternVL 1.5 and 2.0. This new version integrates a newly incrementally pre-trained InternViT with various pre-trained LLMs, including InternLM 2.5 and Qwen 2.5, utilizing a randomly initialized MLP projector. Consistent with previous implementations, InternVL 2.5 employs a pixel unshuffle operation, reducing visual tokens to one-quarter of their original count, and adopts a dynamic resolution strategy similar to InternVL 1.5, processing images in 448×448 pixel tiles. A significant advancement since InternVL 2.0 is the expansion of capabilities to include multi-image and video data processing. For larger model variants, we implement a merge-based approach rather than sequential processing for video data to optimize GPU memory consumption, as demonstrated in Table~\ref{main_experiment_new}.

\item 
\textbf{LLaVA-NeXT-Video}~\citep{zhang2024llavanextvideo} is a multimodal large language model designed to excel in video understanding tasks through zero-shot modality transfer. Trained primarily on image data, it demonstrates impressive performance on video tasks by leveraging deep learning models with DPO training and AI feedback. The model supports various deployment scenarios, from cloud environments to edge devices, making it highly versatile. It is part of the broader LLaVA-NeXT suite, focused on advancing visual-language integration. The tested version are 
LLaVA-NeXT-Video-7B-Qwen and LLaVA-NeXT-Video-7B-Qwen-dpo, we call them as LLaVA-NV and LLaVA-NV-dpo respectively in this paper.

\item
\textbf{Mantis}~\citep{Jiang2024MANTISIM} is a multimodal large language model designed for interleaved multi-image tasks. Built on the LLaMA-3 architecture, it excels in co-reference, reasoning, comparison, and temporal understanding. Mantis uses the Mantis-Instruct dataset, containing 721K examples, to train on various multi-image skills. It achieves state-of-the-art performance on some interleaved benchmarks, while maintaining strong single-image performance on par with CogVLM~\citep{wang2023cogvlm}. The tested version are Mantis-Idefics2, Mantis-LLaVA, Mantis-siglip-llama3 and Mantis-clip-llama3.

\item
\textbf{GPT}~\citep{achiam2023gpt}. GPT-4V is an advanced multimodal model that extends GPT-4's capabilities with integrated vision processing, allowing it to understand and generate text based on visual inputs. GPT-4o is an optimized variant designed for better performance in language tasks while maintaining lower computational requirements. GPT-4o-mini is a more lightweight version of GPT-4o, designed for deployment in resource-constrained environments while still providing strong performance in language understanding and generation tasks. The version of GPT-4V used is GPT-4-turbo. Notably, due to token length limitations, all images were resized to 512$\times$512 before being input into GPT. Our testing was conducted around mid-August 2024, using the most advanced model available at that time. The remaining models were also tested during this period.

\item
\textbf{Gemini-1.5}~\citep{team2023gemini}. Gemini-1.5-Flash and Gemini-1.5-Pro are advanced multimodal large language models, each designed with distinct capabilities for different performance needs. Gemini-1.5-Flash emphasizes fast processing and efficient memory usage, making it ideal for tasks requiring speed on devices with limited resources. Our testing of Gemini-1.5-Flash and Gemini-1.5-Pro was conducted around mid-August 2024.
\end{itemize}

\begin{table}[th!]
\small
\centering
\begin{tabular}{p{0.24\linewidth}  p{0.69\linewidth}}
\toprule
\textbf{Model} & \textbf{Generation Setup} \\
\midrule
LLaVA-interleave &torch\_dtype = torch.float16, max\_new\_tokens = 10, do\_sample = False\\ \midrule
LLaVA-interleave-dpo &torch\_dtype = torch.float16, max\_new\_tokens = 10, do\_sample = False \\ \midrule
\multirow{2}{*}{VILA-1.5-3B} & torch\_dtype = torch.float16, max\_new\_tokens = 4000, temperature = 0.1, num\_beams = 1,  use\_cache = False, do\_sample = False, top\_p = None \\ \midrule
\multirow{2}{*}{VILA-1.5-3B-s2} &  torch\_dtype = torch.float16, max\_new\_tokens = 4000, temperature = 0.1, num\_beams = 1,  use\_cache = False, do\_sample = False, top\_p = None \\ \midrule
\multirow{2}{*}{VILA-1.5-8B} &  torch\_dtype = torch.float16, max\_new\_tokens = 4000, temperature = 0.1, num\_beams = 1,  use\_cache = False, do\_sample = False, top\_p = None \\ \midrule
\multirow{2}{*}{VILA-1.5-13B} &  torch\_dtype = torch.float16, max\_new\_tokens = 4000, temperature = 0.1, num\_beams = 1,  use\_cache = False, do\_sample = False, top\_p = None \\ \midrule
\multirow{2}{*}{Phi-3V} & torch\_dtype = torch.float16, max\_new\_tokens = 500, temperature = 0, do\_sample = False, length\_penalty = 1, repetition\_penalty = 1 (When in error analysis the max\_new\_tokens is set to 5000)\\ \midrule
\multirow{2}{*}{Phi-3.5V} & torch\_dtype = torch.float16, max\_new\_tokens = 500, temperature = 0, do\_sample = False, length\_penalty = 1, repetition\_penalty = 1 (When in error analysis the max\_new\_tokens is set to 5000)\\ \midrule
LLaVA-NV &  torch\_dtype = torch.float16, max\_new\_tokens = 10\\ \midrule
LLaVA-NV-dpo & torch\_dtype = torch.bfloat16, max\_new\_tokens = 10 \\ \midrule
Mantis-Idefics2 & torch\_dtype = torch.bfloat16, max\_new\_tokens = 10,  do\_sample = False\\ \midrule
\multirow{2}{*}{Mantis-LLaVA} & torch\_dtype = torch.bfloat16, max\_new\_tokens = 1, num\_beams = 1, do\_sample = False, length\_penalty = 1, repetition\_penalty = 1\\ \midrule
\multirow{2}{*}{Mantis-siglip-llama3} & torch\_dtype = torch.bfloat16, max\_new\_tokens = 1, num\_beams = 1, do\_sample = False \\ \midrule
\multirow{2}{*}{Mantis-clip-llama3} & torch\_dtype = torch.bfloat16, max\_new\_tokens = 1, num\_beams = 1, do\_sample = False \\ \midrule
mPLUG-Owl3-1B &  max\_new\_tokens = 100, decode\_text = True\\ \midrule
mPLUG-Owl3-2B &  max\_new\_tokens = 100, decode\_text = True\\ \midrule
mPLUG-Owl3-7B &  max\_new\_tokens = 100, decode\_text = True\\ \midrule
\multirow{2}{*}{InternVL2-1B} &  dtype = torch.bfloat16, max\_new\_tokens = 10, do\_sample = False, max\_num = 12\\ \midrule
\multirow{2}{*}{InternVL2-2B} &  dtype = torch.bfloat16, max\_new\_tokens = 10, do\_sample = False, max\_num = 12\\ \midrule
\multirow{2}{*}{InternVL2-4B} &  dtype = torch.bfloat16, max\_new\_tokens = 10, do\_sample = False, max\_num = 12\\ \midrule
\multirow{2}{*}{InternVL2-8B} &  dtype = torch.bfloat16, max\_new\_tokens = 10, do\_sample = False, max\_num = 12\\ \midrule
\multirow{2}{*}{InternVL2-26B} &  dtype = torch.bfloat16, max\_new\_tokens = 10, do\_sample = False, max\_num = 1\\ \midrule
\multirow{2}{*}{InternVL2-40B} &  dtype = torch.bfloat16, max\_new\_tokens = 10, do\_sample = False, max\_num = 1\\ \midrule
\multirow{2}{*}{InternVL2-76B} &  dtype = torch.bfloat16, max\_new\_tokens = 10, do\_sample = False, max\_num = 1\\ \midrule
\multirow{2}{*}{GPT-4o} & max\_new\_tokens = 300, temperature = 0, seed = 42 (When in error analysis the max\_new\_tokens is set to 2000) \\ \midrule
GPT-4o-mini & max\_new\_tokens = 300, temperature = 0, seed = 42 \\ \midrule
GPT-4V & max\_new\_tokens = 300, temperature = 0, seed = 42 \\ \midrule
Gemini-1.5-flash & stream = True \\ \midrule
Gemini-1.5-pro &  stream = True\\
\bottomrule
\end{tabular}
\caption{Generating parameters for General VLM. Parameters not explicitly stated indicate the use of the model's default system settings.}
\label{tab:model_params_general}
\end{table}

\begin{table}[th!]
\small
\centering
\begin{tabular}{p{0.24\linewidth}  p{0.69\linewidth}}
\toprule
\textbf{Model} & \textbf{Generation Setup} \\
\midrule
\multirow{2}{*}{InternVL2.5-1B} &dtype = torch.bfloat16, max\_new\_tokens = 10, do\_sample = False, max\_num = 12\\ \midrule
\multirow{2}{*}{InternVL2.5-2B} &dtype = torch.bfloat16, max\_new\_tokens = 10, do\_sample = False, max\_num = 12\\ \midrule
\multirow{2}{*}{InternVL2.5-4B} &dtype = torch.bfloat16, max\_new\_tokens = 10, do\_sample = False, max\_num = 12\\ \midrule
\multirow{2}{*}{InternVL2.5-8B} &dtype = torch.bfloat16, max\_new\_tokens = 10, do\_sample = False, max\_num = 12\\ \midrule
\multirow{2}{*}{InternVL2.5-26B} &dtype = torch.bfloat16, max\_new\_tokens = 10, do\_sample = False, max\_num = 1\\ \midrule
\multirow{2}{*}{InternVL2.5-38B} &dtype = torch.bfloat16, max\_new\_tokens = 10, do\_sample = False, max\_num = 1\\ \midrule
\multirow{2}{*}{InternVL2.5-78B} & dtype = torch.bfloat16, max\_new\_tokens = 10, do\_sample = False, max\_num = 1\\ 
NVILA-8B & default\\ 
NVILA-15B & default\\ 
NVILA-Lite-8B & default\\ 
NVILA-Lite-15B & default\\ 
\bottomrule
\end{tabular}
\caption{Generation parameters for more models (continued from Table~\ref{tab:model_params_general}). All other parameters not specified here use the default model configurations.}
\label{tab:model_params_general2}
\end{table}
\subsection{Prompt for VLM test}
Referring to~\cite{liu2024visual}, during testing, we appended an end prompt to each question-answer pair (i.e., the value corresponding to the "question" key in the Figure~\ref{fig:dataformat}). The end prompt is as follows:
\begin{mycase}{light_grey}
Answer with the option's letter from the given choices directly. You can only answer one letter from A, B, C, or D.
\end{mycase}
Consistent with~\cite{zhang2024task}, when asking video-related questions to Image VLMs, we prepend the prompt with the following statement:
\begin{mycase}{light_grey}
This is a series of images sampled at equal intervals from the beginning to the end of a video. Based on the series of images, output the best option for the question.
\end{mycase}
For Video VLMs, due to their weaker instruction-following capabilities, we provided additional prompts during testing to guide the models in selecting the correct answer rather than simply describing the video content. (Despite this, many models still responded with video descriptions rather than answering the questions, as noted in Appendix~\ref{app:setup_video_vlm}.) For Video LLMs, if the question involves only a single image input, we prepend the following statement:
\begin{mycase}{light_grey}
Based on the image, output the best option for the question. You must only output the option.
\end{mycase}
Add the last line in the prompt:
\begin{mycase}{light_grey}
The best choice option is:
\end{mycase}
If the input involves only a single video, we prepend the following statement:
\begin{mycase}{light_grey}
This is a series of images sampled at equal intervals from the beginning to the end of a video. Based on the series of images, answer the question. Based on the video, output the best option for the question. You must only output the option.
\end{mycase}

\subsection{Reference Datasets Summary}
Our dataset was entirely annotated by humans and underwent two rounds of rigorous filtering and screening. The data sources we utilized, along with their respective uses, are detailed in Table~\ref{tab:data_ref}.
\begin{table}[th!]
    \centering
    \caption{Datasets Reference for our PhysBench. 'None' indicates that the dataset does not explicitly state which license is being used.}
    \label{tab:data_ref}
    \resizebox{0.99\columnwidth}{!}{
        \begin{tabular}{lcp{8cm}}
        \toprule
        \textbf{Dataset} & \textbf{License} & \textbf{Description of usage} \\
        \toprule
        Unsplash~\cite{ali2023unsplash}& \url{https://unsplash.com/license}& We crawled some images for annotation and labeling.\\
        ContPhy~\cite{zheng2024contphy} & CC-BY 4.0 & We used his code to generate some of the simulation data.\\
        ChronoMagic-Bench~\cite{yuan2024chronomagic} &Apache 2.0 & As source data, we annotated some QA. \\
        \multirow{2}{*}{DROID~\cite{khazatsky2024droid}} &  \multirow{2}{*}{CC-BY 4.0}  &  As source data, we annotated some QA data belonging to physics-based dynamics--manuplation type. \\
        \multirow{2}{*}{Ego4D~\cite{grauman2022ego4d}} & \multirow{2}{*}{CC-BY 4.0}  & As source data, we annotated some QA data belonging to physics-based dynamics--manuplation type.  \\     
        \multirow{2}{*}{MimicPlay~\cite{wang2023mimicplay}} & \multirow{2}{*}{CC-BY 4.0} & As source data, we annotated some QA data belonging to physics-based dynamics--manuplation type. \\     
        \multirow{2}{*}{nuScenes~\cite{nuscenes2019}}      & \multirow{2}{*}{CC BY-NC-SA 4.0} & As source data, we annotated some QA data belonging to spatital movement type.\\     
        \multirow{2}{*}{Netwon~\cite{wang2023newton}}   & \multirow{2}{*}{CC-BY 4.0} & As reference data for us to create the attribute table for objects.  \\
        \multirow{2}{*}{FunKPoint~\cite{lai2021functional}} & \multirow{2}{*}{None} &As source data, we annotated some QA data belonging to physics-based dynamics--manuplation type.\\
        \multirow{2}{*}{PhotoTourism~\cite{SSS:2006}} & \multirow{2}{*}{None} & As source data, we annotated some QA data belonging to physical scene understanding--camera type.\\
        \toprule
        \end{tabular}}
\end{table}

\subsection{Prompt Strategies}\label{app:prompt_str}

In Section~\ref{sec:phsagent}, we explore several prompting strategies:Chain of Thought (CoT)~\citep{kojima2023largelanguagemodelszeroshot} with the prompt "Let's think step-by-step!"; Desp-CoT~\citep{wu2023rolechainofthoughtcomplexvisionlanguage}, which begins with an image description prompt; and Pure Language Reasoning (PLR), similar to Mathvista~\citep{lu2024mathvista}. The specific prompt content can be found in Appendix~\ref{app:prompt_str}. Following the settings of~\cite{kojima2022large, chen2024m3cotnovelbenchmarkmultidomain}, we extract the final generated answer through GPT-4o-mini.
In this section, we provide the prompts used to implement these three strategies. Note that for Phi-3V:

(1.1) The prompt for CoT~\citep{kojima2023largelanguagemodelszeroshot} is:  
\begin{mycase}{light_grey}
Let's think step by step! Start by selecting the correct option's letter from the given choices, then provide a detailed explanation of your thought process.
\end{mycase}
(1.2) The prompt for Desp-CoT~\citep{wu2023rolechainofthoughtcomplexvisionlanguage} is:
\begin{mycase}{light_grey}
Each image or video is followed by a description, which you can refer to. Answer with the option's letter from the given choices directly.
\end{mycase}
(1.3) The prompt for PLR is:  
\begin{mycase}{light_grey}
Each image or video is replaced by description. Answer with the option's letter from the given choices directly.
\end{mycase}
For GPT-4o:  

(2.1) The prompt for CoT~\citep{kojima2023largelanguagemodelszeroshot} is: 
\begin{mycase}{light_grey}
Let's think step by step! Start by selecting the correct option's letter from the given choices, then provide a detailed explanation of your thought process.
\end{mycase}
(2.2) The prompt for Desp-CoT~\citep{wu2023rolechainofthoughtcomplexvisionlanguage} is: 
\begin{mycase}{light_grey}
Each image or video is followed by a description, which you can refer to. Answer with the option's letter from the given choices directly.
\end{mycase}
(2.3) The prompt for PLR is:  
\begin{mycase}{light_grey}
Each image or video is replaced by description. Answer with the option's letter from the given choices directly.
\end{mycase}
We used the following prompts to generate the corresponding descriptions for each model (for image descriptions, “video” is replaced with “image” where applicable):
\begin{mycase}{light_grey}
Give me the description of this video.
\end{mycase}

\subsection{ViperGPT Implementation}\label{app:vipergpt}
Following the methodology described in ContPhy~\citep{zheng2024contphy}, since the code for its model, ContPro, has not been released, we implemented an oracle neural-symbolic model using ViperGPT~\citep{suris2023vipergpt}, which we refer to as ContPhy, and evaluated it on the PhysBench. Similar to ContPhy~\citep{zheng2024contphy}, we decomposed the question-answering task into four main modules: video perception, physical simulation, program parser, and symbolic execution.

Given a raw video, the video perception module detects objects and their associated static attributes using the MASK-RCNN detector~\citep{he2017mask}. The physical simulator takes point clouds as input and predicts object dynamics across various scenarios using dynamic prediction models~\citep{li2018learning, sulsky1995application}. The program parser, powered by a large language model (we use GPT-4o), translates the question query into executable Python programs. Based on the detected object attributes and predicted dynamics, the symbolic executor runs the programs to derive the answer to the question.

Since neuro-symbolic approaches rely on pre-trained domain-specific modules to extract objects' static attributes, physical properties, and dynamic trajectories, this method is not suitable for the complex and diverse tasks in PhysBench. For instance, in ContPhy, simulating four types of tasks—rope, cloth, ball, and fluid—requires pre-training three models: MASK R-CNN~\citep{he2017mask} based on Detectron2~\citep{wu2019detectron2}, DPI-Net~\citep{li2018learning}, and Material Point Method (MPM)~\citep{sulsky1995application}. Given the variety of tasks in PhysBench, where even individual tasks within certain subcategories can differ significantly, it is impractical to design and call a specific module for every task. Therefore, we designed an Oracle model for the QA pairs generated through specific simulations to ensure that the tasks are similar enough to allow the use of consistent logical processing templates following ~\cite{zheng2024contphy}, while GPT-4o directly provided answers for the remaining questions without invoking additional modules or running Python code.

Following the approach in ContPhy, we use ResNet-50~\citep{he2016deep} as the backbone for MASK R-CNN to densely detect object locations in each frame and associate static attributes such as color and material. We use the default config from Detectron2, while the number of classes is different across scenarios. Specifically, the batch size is $16$ for $8$ GPUs thus each mini-batch has $2$ images per GPU. We train the model for about $10k$ iterations, with a learning rate of $0.01$. For image size, we keep the original resolution.

We fine-tune the network using the training set data from all simulations. For scenes involving fluid and soft-body dynamics, we adopt MPM, while DPI-Net is used for other tasks. The training data for these two modules is similar to that of MASK R-CNN, and the overall training procedure is comparable to ContPhy. After extracting objects' static attributes, physical properties, and dynamic trajectories, and parsing the natural language query into an executable program, we run the program using the object states as input and produce the predicted answer. 
Due to the rich diversity and versatile resources of our dataset, it is challenging to ensure that templates perfectly fit all predefined modules. For example, some videos become significantly distorted when rescaled into a square format, and in certain videos, fluids are difficult to extract as points due to their similar color to the environment. For the specific training parameters of MPM and DPI-Net, we have kept them consistent with ContPhy.

\subsection{Human Performance}
To evaluate human performance on PhysBench, we recruited 12 graduate students in STEM fields and provided monetary compensation for their participation. Each question was assigned to all annotators. 
To ensure the quality of the results, we followed the methodology of MathVista~\citep{lu2024mathvista} by implementing qualification questions during participant recruitment. These questions tested basic knowledge of physical world concepts, and only those who answered the qualification questions correctly were deemed eligible for the study. 
Given the large number of questions in PhysBench, the testing was divided into 10 sessions, delivered as online questionnaires with no time constraints for completion. The average score across all participants was used as the final measure of human performance.
\section{More Experiments Results}~\label{app:more_data}
\subsection{GroundingDINO Configuration}
GroundingDINO has two key parameters: box\_threshold and text\_threshold. The box\_threshold parameter is used to filter predictions based on the confidence level of the detected bounding boxes, ensuring that only boxes with confidence scores above this threshold are retained. On the other hand, the text\_threshold parameter filters predictions based on their relevance to the input text prompt, retaining only those that meet or exceed this threshold.

It is important to note that the filtered phrases may contain a significant number of duplicates. To address this, we consider a prediction successful only if the set of output phrases matches exactly with the set of input phrases, ensuring both completeness and accuracy. The detailed experimental results are presented in Table~\ref{tab:net-search}.
\begin{table}[th!]
    \small
    \centering
    \caption{\textbf{GroundingDINO Accuracy with different box threshold and text threshold}. The number of samples tested is 1000.}
    \label{tab:net-search}
    \resizebox{0.79\columnwidth}{!}{
    \begin{tabular}{c|ccccccc}
        \hline
        \raisebox{-0.5ex}[0pt][0pt]{\tiny{box threshold}} \raisebox{0.5ex}[0pt][0pt]{\tiny{text threshold}} &        0.1 & 0.2 & 0.3 & 0.4 & 0.5 & 0.6 & 0.7 \\
        \hline
             0.1 &   0.015 &       0 &       0 &       0 &       0 &       0 &       0 \\
             0.2 &   0.015 &   0.075 &   0.005 &   0.005 &       0 &       0 &       0 \\
             0.3 &    0.01 &    0.03 &    0.06 &    0.01 &   0.005 &       0 &       0 \\
             0.4 &   0.005 &   0.015 &       4 &       4 &       0 &       0 &       0 \\
             0.5 &   0.005 &   0.005 &       0 &       0 &       0 &       0 &       0 \\
             0.6 &   0.005 &       0 &       0 &       0 &       0 &       0 &       0 \\
             0.7 &   0.005 &       0 &       0 &       0 &       0 &       0 &       0 \\
        \hline
    \end{tabular}
        }
\end{table}
\subsection{Effect of Visual Prompting}\label{e_o_v_p}
\begin{figure}[th!]
	\centering  
	\vspace{-1.5mm}
	\includegraphics[height=3.95cm]{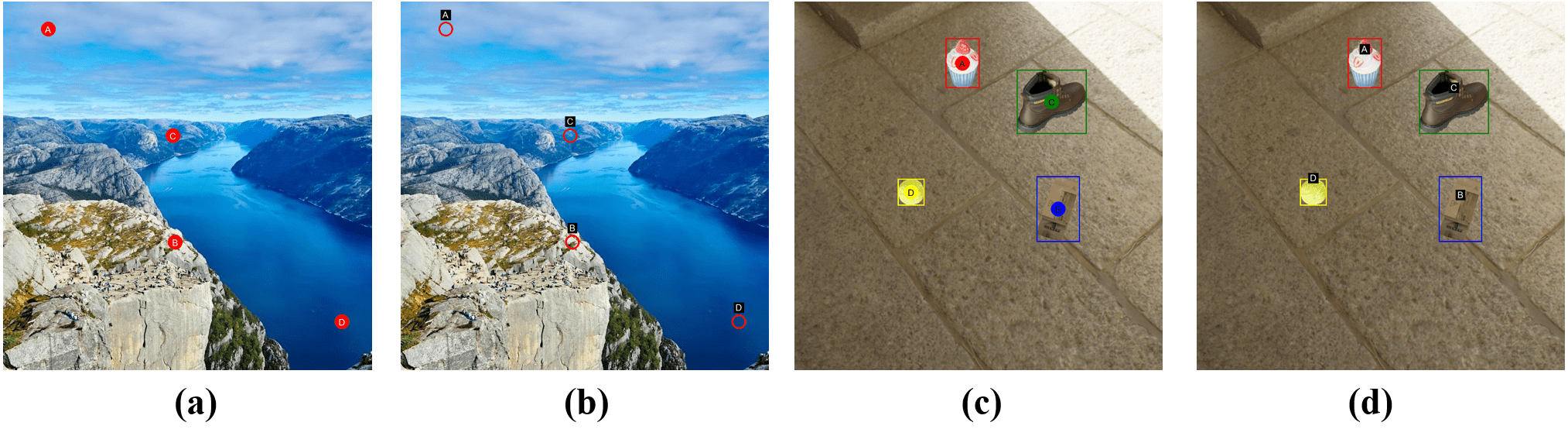}
	\vspace{-1.5mm}
    	\caption{\textbf{Several different labeling methods}. (a) (b) are two methods for labeling depth. (c) (d) are two labeling methods for attributes. Attribute also try (a)(b)}
	\label{fig:case01}  
	\vspace{-3mm}
\end{figure}

To enhance the VLM's responsiveness to visual prompts, we assessed its sensitivity to these prompts. The images used in our tests were all 1024$\times$1024 pixels. We employed two different annotation methods for depth and attribute sub-tasks, as illustrated in Figure~\ref{fig:case01}, drawing on BLINK~\citep{fu2024blink} for reference. 
For depth, the default color for circles is red. Method (a) follows the approach introduced in the BLINK paper, while the second method aligns exactly with the instances in BLINK-eval. Additionally, the prompts for both methods are largely similar to those in the BLINK-eval examples. Similarly, for attributes, we experimented with two annotation methods, denoted as (c) and (d) in Figure~\ref{fig:case01}. However, since the white text doesn't look very clear in the yellow checkbox in the first method, the yellow color was selected.

The test images comprise 1000 samples, with a uniform circle radius of 20 pixels and a font size of 25 pixels. The detailed experimental results are presented in Table~\ref{tab:visual_prompt_exp_0}. For samples (a) and (b), the text prompts require selecting the location or object indicated by the prompt, while samples (c) and (d) involve identifying a bounding box or point. For instance, answers for (c) and (d) might be "A. The object enclosed in the red box at point A" using both color and letter identifiers.

Based on the experimental results, we find that the outcomes for (a) and (b) are similar in terms of depth and attribute, though (a) performs slightly better. For attributes, (c) and (d) yield results close to random, likely because the tested models cannot perceive color differences in the bounding boxes, thus failing to answer effectively. However, (a) and (b) significantly improve accuracy. Therefore, we ultimately decided to use the annotation method in (a).
\begin{table}[th!]
    \small
    \centering
    \caption{\textbf{Comparison of Visual Prompts for Depth and Attribute Annotation}. }
    \label{tab:visual_prompt_exp_0}
    \resizebox{0.79\columnwidth}{!}{
        \begin{tabular}{lcccccc}
        \hline
                     & \multicolumn{2}{c}{Depth} & \multicolumn{4}{c}{Attribute} \\
                     & Prompt (a)   & Prompt (b)  & Prompt (a)   & Prompt (b)  & Prompt (c)  &  Prompt(d)      \\ \hline  
        LLaVA-1.5-7b &  27.3       &    27.2     &  31.5   & 29.9  & 25.3 & 25.2  \\
        LLaVA-1.5-13b&  21.2       &    20.2     &  41.1    & 38.5  & 26.0 & 25.6  \\ 
        Phi-3V-128k  &  78.2       &    75.8     &  41.2   & 40.1   & 24.8 & 25.1  \\
        VILA-1.5-8b  &  19.9       &    20.3     &  27.9  &  26.1       & 25.9& 24.7 \\ \hline 
        \end{tabular}
    }
\end{table}

Building on the findings of the previous experiment, we further tested the impact of varying circle sizes using the annotation method, which showed the best model performance. 
The corresponding results are detailed in Table~\ref{tab:visual_prompt_exp_1}. Additionally, the text size was consistently 5 pixels larger than the center of the circle. Based on the previous experiment, we adopted the annotation method (a) for all subsequent annotations. Therefore, we used method (a) to test depth, with a sample size of 1000 images. 

\begin{figure}[th!]
	\centering  
	\vspace{-1.5mm}
	\includegraphics[height=2.55cm]{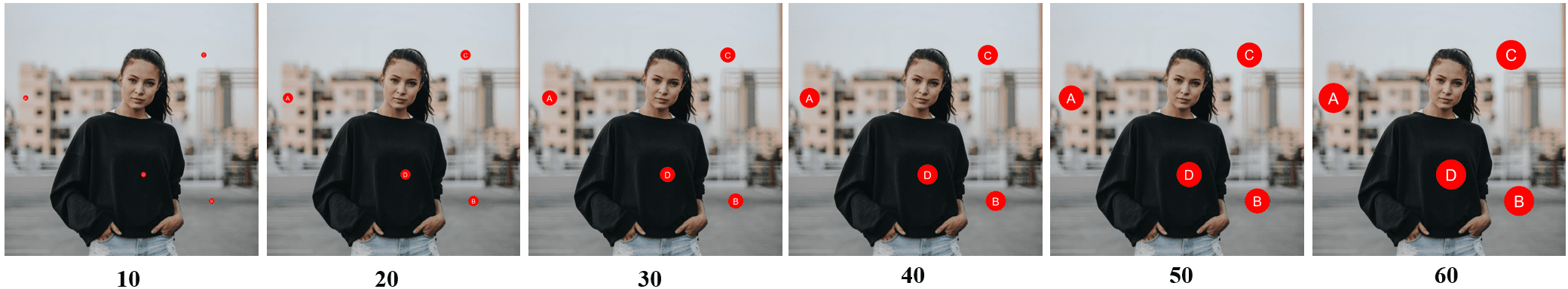}
	\vspace{-1.5mm}
    	\caption{Visual Prompt with Different Size.}
	\label{fig:case02}  
	\vspace{-3mm}
\end{figure}

We have tested 6 scales as shown in Figure~\ref{fig:case02} and the results indicate that for LLaVA-1.5, larger sizes generally yield better results. For Phi-3V and VILA-1.5, the optimal radius is 30 pixels, although performance does not show a significant relationship on either side of 30 pixels. Most models show low sensitivity to size variations, except for LLaVA-1.5-7b. To balance performance and aesthetics, we ultimately selected a radius size of 30 pixels.

\begin{table}[th!]
    \small
    \centering
    \caption{\textbf{Effects of different visual prompt circle radius on Depth task performance}. }
    \label{tab:visual_prompt_exp_1}
    \resizebox{0.58\columnwidth}{!}{
        \begin{tabular}{lcccccc} \hline
                      & 10 & 20 & 30 & 40 & 50 & 60 \\ \hline
        LLaVA-1.5-7b  &  19.2  & 27.3   &  27.3  & 28.4   &  30.9   &  31.2    \\
        LLaVA-1.5-13b &  19.9  & 21.2   & 24.9   & 25.2  & 25.8  &    27.9  \\
        Phi-3V-128k   &  74.7  &  78.2  &  78.8  &  76.8  &  73.1   &    73.0     \\
        VILA-1.5-8b   &  20.3  &  19.9  &  26.9  & 25.3   & 26.7 & 22.2 \\ \hline 
        \end{tabular}
    }
\end{table}

\subsection{More PhysBench Results}\label{exp_bench_details}
The performance of 39 models across 8 ability categories in PhysBench is presented in Table~\ref{tab:app_exp_main_ability}. Note that the ability classifications for the physical scene understanding and physics-based dynamics categories are identical, though their content differs slightly; these are combined here for clarity. Detailed descriptions of the ability classifications can be found in Appendix~\ref{app:task_description_ability}, while the remaining 36 newly tested models are listed in Appendix~\ref{app:latex}.

The details for Ability Type, Physical Object Property, Physical Object Relationships, Physical Scene Understanding, and Physics-based Dynamics can be found in Table~\ref{tab:app_exp_main_ability},Table~\ref{tab:app_exp_main_pp},Table~\ref{tab:app_exp_main_spa},Table~\ref{tab:app_exp_main_env},Table~\ref{tab:app_exp_main_phe}, respectively.

\begin{table}[th!]
    \setlength{\tabcolsep}{3pt}
    \centering
    \scalebox{0.75}{
\begin{tabular}{lcccccccc} \hline
\textbf{Model} & \textbf{Identify} & \textbf{Comparison} & \textbf{Static} & \textbf{Dynamic} & \textbf{Perception} & \textbf{Prediction} & \textbf{Judgment} & \textbf{Reasoning} \\\hline
    
\multicolumn{9}{c}{Image VLM} \\\hline
InstructBLIP-t5-xl~\cite{dai2024instructblip}   & 48.30 & 25.39 & 36.38 & 35.76 & 40.06 & 34.07 & 42.74 & 35.83 \\
InstructBLIP-t5-xxl~\cite{dai2024instructblip}  & 57.63 & 28.40 & 33.21 & 53.64 & 41.90 & 32.62 & 47.58 & 36.28 \\
InstructBLIP-7B~\cite{dai2024instructblip}      & 25.02 & 19.31 & 30.98 & 24.50 & 32.21 & 23.74 & 29.91 & 18.28 \\
InstructBLIP-13B~\cite{dai2024instructblip}     & 39.49 & 25.70 & 32.46 & 35.33 & 35.88 & 27.12 & 33.62 & 20.18 \\
BLIP-2~\cite{li2023blip}                        & 55.97 & 30.78 & 36.01 & 53.64 & 42.51 & 33.24 & 48.15 & 34.29 \\
LLaVA-1.5-7B~\cite{liu2023improved}             & 47.78 & 31.26 & 40.30 & 45.70 & 47.40 & 39.90 & 45.01 & 37.38 \\
\hline
\end{tabular}
    }
    \caption{\textbf{Evaluation Results for 39 VLMs Categorized by Ability Dimensions}. PhysBench includes two evaluation dimensions: ability and task. Table~\ref{main_experiment} presents results categorized by the task-type dimension.}
    \label{tab:app_exp_main_ability}
\end{table}

\begin{table}[th!]
    \setlength{\tabcolsep}{3pt}
    \centering
    \scalebox{0.93}{
\begin{tabular}{lcccc}\hline
\textbf{Model}                & \textbf{Number} & \textbf{Mass} & \textbf{Color} & \textbf{Attribute} \\\hline
\multicolumn{5}{c}{Image VLM} \\\hline
InstructBLIP-t5-xl~\cite{dai2024instructblip}   & 47.14 & 22.18 & 44.33 & 31.37 \\
InstructBLIP-t5-xxl~\cite{dai2024instructblip}  & 52.86 & 43.00 & 54.00 & 33.22 \\
InstructBLIP-7B~\cite{dai2024instructblip}      & 11.23 & 34.93 & 35.57 & 20.66 \\
InstructBLIP-13B~\cite{dai2024instructblip}     & 29.12 & 28.67 & 43.33 & 30.77 \\
BLIP-2~\cite{li2023blip}                        & 51.30 & 50.85 & 47.33 & 34.63 \\
LLaVA-1.5-7B~\cite{liu2023improved}             & 38.99 & 36.86 & 52.67 & 35.42 \\
\hline  
\end{tabular}
    }
    \caption{
    Evaluation Results for 39 VLMs in PhysBench Physical Object \faGear Property Sub-task (the fifth last column of Table~\ref{main_experiment}).
    }
    \label{tab:app_exp_main_pp}
\end{table}

\begin{table}[th!]
    \setlength{\tabcolsep}{3pt}
    \centering
    \scalebox{0.93}{
   \begin{tabular}{lccccc} 
\hline
\textbf{Model}                & \textbf{Size} & \textbf{Location} & \textbf{Depth} & \textbf{Distance} & \textbf{Motion} \\\hline
\multicolumn{6}{c}{Image VLM} \\\hline
InstructBLIP-t5-xl~\cite{dai2024instructblip}   & 37.50 & 38.72 & 30.53 & 41.67 & 53.93 \\
InstructBLIP-t5-xxl~\cite{dai2024instructblip}  & 48.21 & 41.35 & 27.72 & 51.67 & 57.30 \\
InstructBLIP-7B~\cite{dai2024instructblip}      & 33.93 & 30.92 & 26.35 & 20.00 & 37.08 \\
InstructBLIP-13B~\cite{dai2024instructblip}     & 26.79 & 39.10 & 25.26 & 43.33 & 48.31 \\
BLIP-2~\cite{li2023blip}                        & 58.93 & 44.74 & 29.47 & 46.67 & 59.55 \\
LLaVA-1.5-7B~\cite{liu2023improved}             & 64.29 & 43.23 & 35.09 & 41.67 & 51.69 \\
\hline
\end{tabular}
    }
    \caption{Evaluation Results for 39 VLMs in PhysBench Object  \faLocationArrow Relationships Sub-task (the forth last column of Table~\ref{main_experiment}).
    }
    \label{tab:app_exp_main_spa}
\end{table}

\begin{table}[th!]
    \setlength{\tabcolsep}{3pt}
    \centering
    \scalebox{0.93}{
\begin{tabular}{lcccc}
\hline
\textbf{Model}                & \textbf{Temperature} & \textbf{Viewpoint} & \textbf{Air} & \textbf{Light} \\\hline
\multicolumn{5}{c}{Image VLM} \\\hline
InstructBLIP-t5-xl~\cite{dai2024instructblip}   & 70.59 & 43.26 & 26.00 & 30.03 \\
InstructBLIP-t5-xxl~\cite{dai2024instructblip}  & 80.88 & 43.36 & 40.00 & 30.30 \\
InstructBLIP-7B~\cite{dai2024instructblip}      & 36.76 & 22.94 & 12.00 & 14.74 \\
InstructBLIP-13B~\cite{dai2024instructblip}     & 57.35 & 21.61 & 30.61 & 19.86 \\
BLIP-2~\cite{li2023blip}                        & 79.41 & 40.90 & 30.00 & 29.48 \\
LLaVA-1.5-7B~\cite{liu2023improved}             & 61.76 & 44.58 & 24.00 & 31.76 \\
\hline
\end{tabular}
    }
    \caption{
    Evaluation Results for 39 VLMs in PhysBench Physical \faTree Scene Understanding Sub-task (the third last column of Table~\ref{main_experiment}).}
    \label{tab:app_exp_main_env}
\end{table}

\begin{table}[th!]
    \setlength{\tabcolsep}{3pt}
    \centering
    \scalebox{0.87}{
\begin{tabular}{lcccccc}
\hline
\textbf{Model}                & \textbf{Collision} & \textbf{Throwing} & \textbf{Manipulation} & \textbf{Fluid} & \textbf{Chemistry} & \textbf{Others} \\\hline
\multicolumn{7}{c}{Image VLM} \\\hline
InstructBLIP-t5-xl~\cite{dai2024instructblip}   & 31.49 & 34.68 & 27.89 & 32.41 & 62.16 & 58.16 \\
InstructBLIP-t5-xxl~\cite{dai2024instructblip}  & 32.72 & 31.35 & 29.90 & 33.33 & 52.70 & 58.84 \\
InstructBLIP-7B~\cite{dai2024instructblip}      & 19.97 & 26.60 & 27.96 & 29.94 & 27.03 & 38.10 \\
InstructBLIP-13B~\cite{dai2024instructblip}     & 28.26 & 15.44 & 27.89 & 35.65 & 33.78 & 46.60 \\
BLIP-2~\cite{li2023blip}                        & 33.64 & 35.39 & 30.40 & 31.17 & 59.46 & 58.84 \\
LLaVA-1.5-7B~\cite{liu2023improved}             & 36.25 & 37.77 & 41.21 & 45.37 & 59.46 & 55.10 \\
\hline
\end{tabular}
    }
    \caption{Evaluation Results for 39 VLMs in PhysBench Physics-based \faFlask Dynamics Sub-task (the second last column of Table~\ref{main_experiment}).}
    \label{tab:app_exp_main_phe}
\end{table}

\clearpage 
\subsection{PhysBench-val Results}\label{exp_bench_val}
\begin{table}[th!]
    \setlength{\tabcolsep}{3pt}
    \centering
    \scalebox{0.78}{
\begin{tabular}{lccccccc} \hline
    & Size & Format & \faGear Property &  \faLocationArrow Relationships & \faTree Scene & \faFlask Dynamics & \textbf{Avg}\\\hline
    Random Choice&-&-&25.00&25.00&25.00&25.00&25.00\\\hline
\multicolumn{7}{c}{Image VLM} \\\hline
InstructBLIP-t5-xl ~\cite{dai2024instructblip} &   4B   & merge& 40.54&47.62&45.95&44.44&44.69\\
InstructBLIP-t5-xxl ~\cite{dai2024instructblip}  &   12B   &  merge & 48.65&52.38&43.24&49.21&48.60 \\
InstructBLIP-7B~\cite{dai2024instructblip}& 7B &  merge & 28.57&30.95&36.36&35.48&33.14 \\
InstructBLIP-13B~\cite{dai2024instructblip}  & 13B &merge& 30.56&32.50&60.00&36.51&37.80\\
BLIP-2~\cite{li2023blip} &  12B  &merge&51.35&52.38&37.84&46.03&46.93\\
\hline
  \end{tabular}
    }
    \vspace{-2mm}
    \caption{\textbf{Evaluation results for 39 vision-language models in PhysBench-val.} Note that the evaluation of General VLMs is based on the data from Video and Image VLM evaluations, with the addition of interleaved data, meaning that the full test dataset of PhysBench is being assessed. In this context, “seq" refers to the sequential input of images after frame selection from videos, while "merge" refers to merging video frames into a single image.
    }
    \vspace{-4mm}
\end{table}
\clearpage

\subsection{Embodied Tasks Detailed Description}\label{app_robot_task_descri}
\begin{figure}[th!]
    \centering
    \vspace{-1.5mm}

    \hspace{0.05\linewidth}
    \begin{minipage}[b]{0.225\linewidth}
        \includegraphics[width=\linewidth]{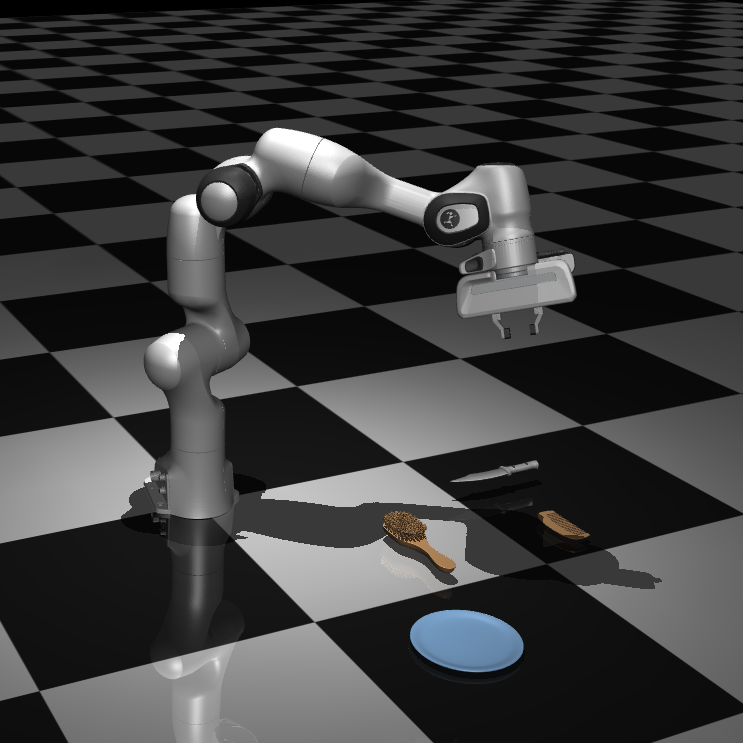}
        \caption*{(a)}
    \end{minipage}
    \hfill
    \begin{minipage}[b]{0.225\linewidth}
        \includegraphics[width=\linewidth]{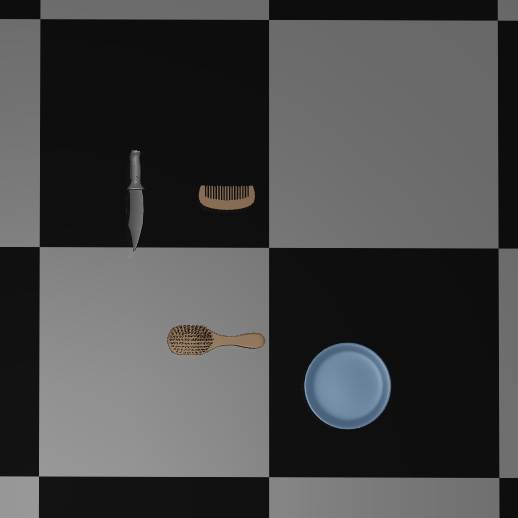}
        \caption*{(b)}
    \end{minipage}
    \hfill
    \begin{minipage}[b]{0.225\linewidth}
        \includegraphics[width=\linewidth]{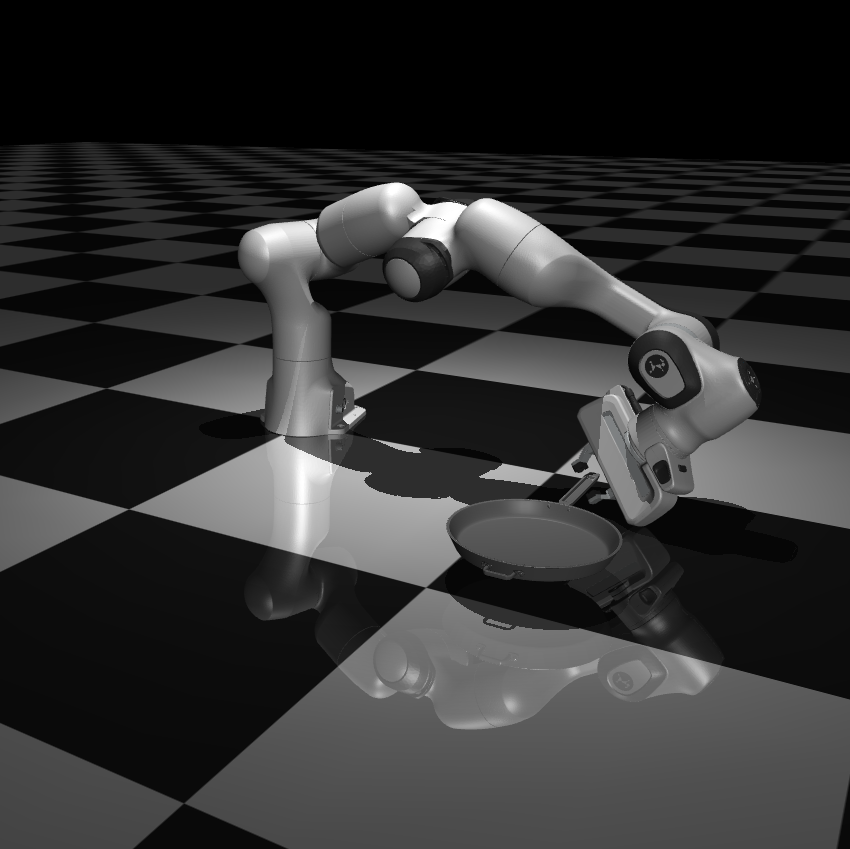}
        \caption*{(c)}
    \end{minipage}
    \hspace{0.05\linewidth}

    \vspace{1mm}

    \begin{minipage}[b]{0.225\linewidth}
        \includegraphics[width=\linewidth]{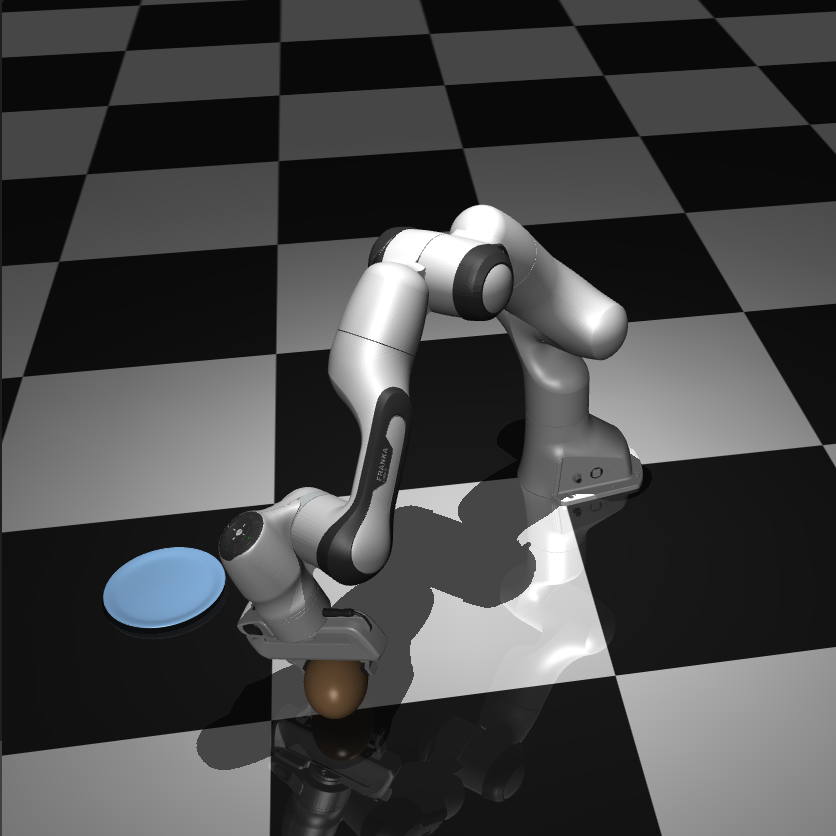}
        \caption*{(d)}
    \end{minipage}
    \hfill
    \begin{minipage}[b]{0.225\linewidth}
        \includegraphics[width=\linewidth]{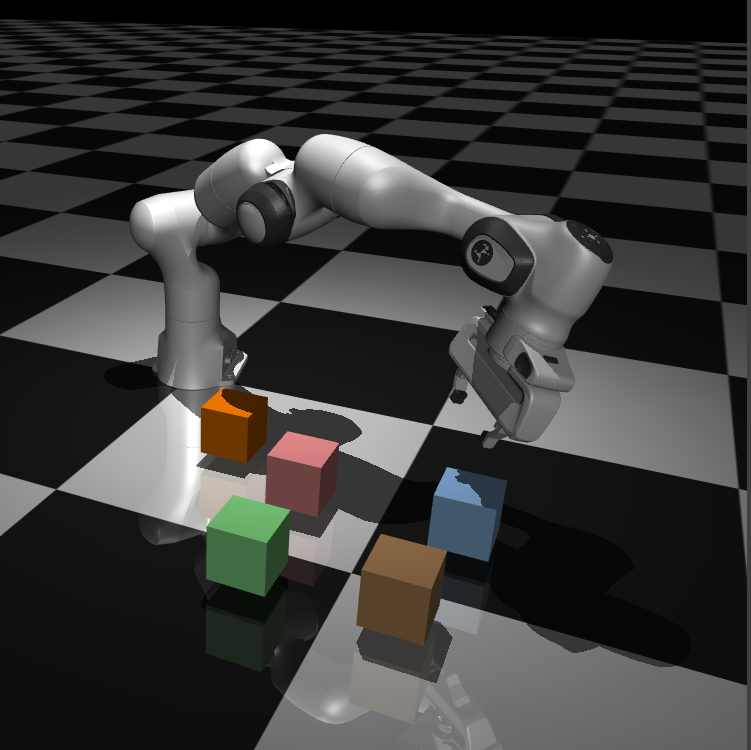}
        \caption*{(e)}
    \end{minipage}
    \hfill
    \begin{minipage}[b]{0.225\linewidth}
        \includegraphics[width=\linewidth]{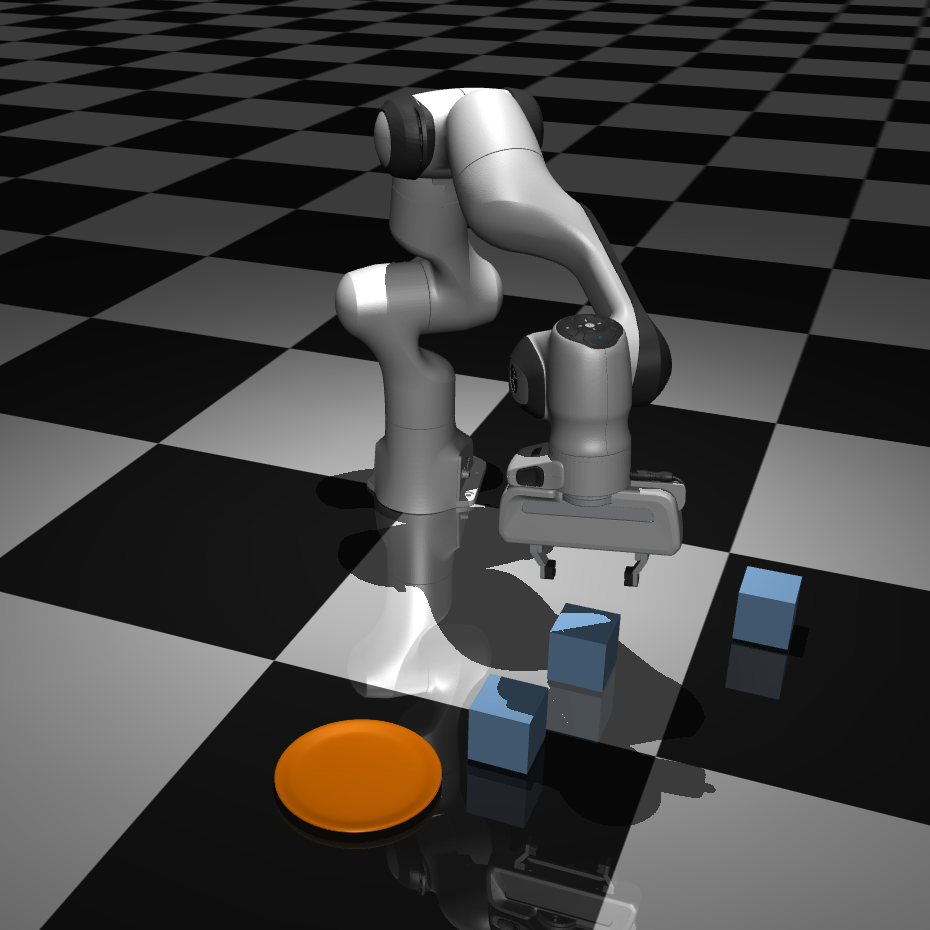}
        \caption*{(f)}
    \end{minipage}
    \hfill
    \begin{minipage}[b]{0.225\linewidth}
        \includegraphics[width=\linewidth]{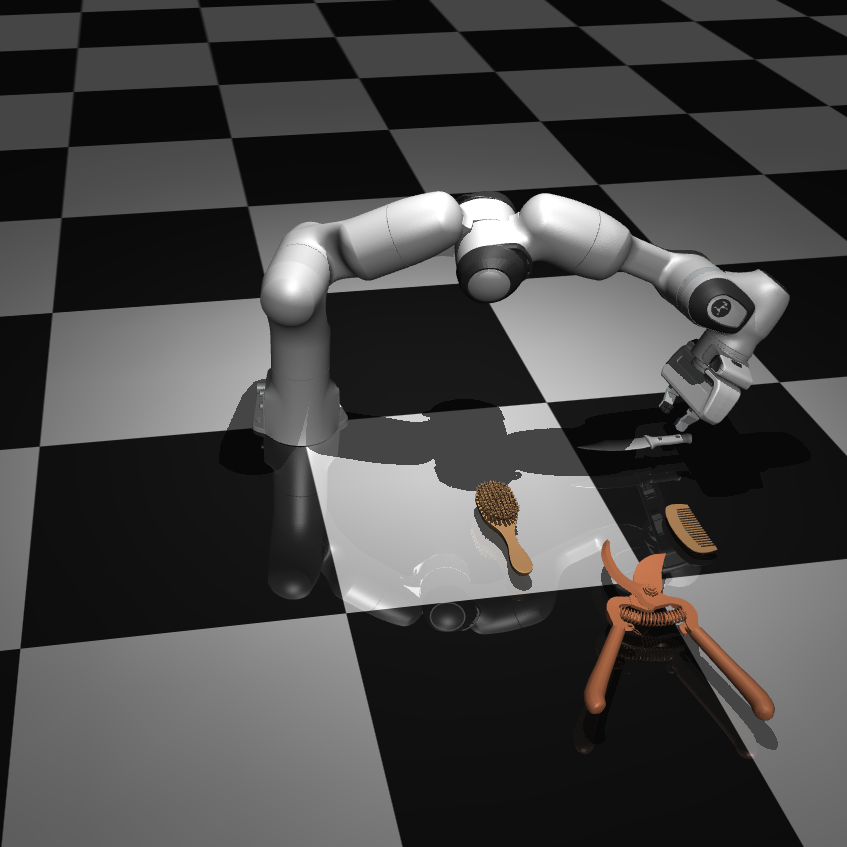}
        \caption*{(g)}
    \end{minipage}

    \vspace{-4mm}
    \caption{(a) Overview of the simulation platform. (b) Top-down view of the area. (c) Affordance test: Testing whether the robotic arm correctly grasps the object. (d) Force test: Testing whether the robotic arm can properly grasp deformable, fragile, and rigid objects. (e) Color test: Testing whether the robotic arm can pick up the correct colored object among identical ones. (f) Number test: Testing whether the robotic arm can grasp a specific number of objects. (g) Tool test: Testing whether the robotic arm can select the correct tool given a specific scenario.}
    \label{fig:robotics_task}
    \vspace{-3mm}
\end{figure}

To further validate the effectiveness of our data and method, we built a simulation platform using MuJoCo~\citep{todorov2012mujoco} and the Franka Emika Panda from Menagerie~\citep{menagerie2022github}, and conducted tests on five embodied tasks. Using the proposed mark-based visual prompting technique with GroundedSAM~\citep{ren2024grounded} and farthest point sampling~\citep{qi2017pointnet}, MOKA~\citep{liu2024moka} converts affordance reasoning into a series of visual question-answering problems that pre-trained VLMs can solve. The setup matches MOKA, and the general visual setup can be seen in Figure~\ref{fig:robotics_task}(a)(b). Our tabletop environment only has one top-down camera, which is the primary camera used in
MOKA to capture  RGBD images. For each task, we report the number of successes out of 10 trials following the setting of~\cite{liu2024moka}.

We present the specific language instructions given to the VLM in Figure~\ref{fig:embodied_instruct}. It is important to note that, for each task, we only provide a single example. For instance, while we tested five objects in the Affordance task—pot, knife, spoon, monitor, and tennis racket—we only used the tennis racket as an example here. For each task, we report the success rate over 10 trials, following the MOKA protocol. 
"Specifically, the test content and evaluation methods for each task are as follows:  
(1) Affordance test: This test evaluates whether the robotic arm can correctly grasp various objects. Figure~\ref{fig:robotics_task}(c) illustrates the robotic arm successfully grasping a pot. In total, we tested the grasping ability on 10 items, including a pot, knife, spoon, spatula, monitor, tennis racket, phone, and others. Specifically, we tested 5 objects—pot, knife, spoon, monitor, and tennis racket—attempting to grasp each object twice.
(2) Force Test: This test evaluates the robotic arm's capacity to properly grasp deformable, fragile, and rigid objects. Due to simulation constraints, the evaluation was based on the robotic arm's output metrics. We tested fragile items (e.g., egg, ripe persimmon), soft items (e.g., jelly, plastic cup), and rigid objects (e.g., iron ball), with two attempts per object. It should be noted that the simulation system models all objects as rigid bodies, meaning that breakage during grasping is not depicted. Furthermore, while using the MOKA system to control the Panda robotic arm, we were unable to directly manipulate the gripper's size. Instead, we provided the VLM with approximate object dimensions, allowing the VLM to determine the necessary gripping force to evaluate success or failure."
(3) Color test: This test evaluates whether the robotic arm can accurately pick up the correctly colored object from a set of identical items. As shown in Figure~\ref{fig:robotics_task}(e), the blocks are identical except for their color, and the task requires selecting the object of the designated color. We tested five colors—blue, pink, brown, green, and orange—conducting two trials for each color.
(4) Location test: This test evaluates whether the robotic arm can correctly grasp objects at specific locations. The goal is to ensure that the robotic arm accurately grasps the required objects based on their positions. Specifically, we tested with three blocks, requiring the arm to grasp the middle block (4 attempts), the block farthest from the plate (3 attempts), and the block closest to the plate (3 attempts).
(5) Tool test: This test assesses whether the robotic arm can select the appropriate tool for a given task. For instance, as shown in Figure~\ref{fig:robotics_task}(g), the task is: "If you need to cut a watermelon, which tool should you grasp?" In this scenario, the robotic arm is expected to grasp the fruit knife. In total, we posed 5 questions, with 2 attempts per question, requiring the robotic arm to select and grasp different target tools for each task.

On the other hand, these five tasks require minimal consideration of height (or depth) information, making the evaluation more fundamental. For MOKA's waypoints selected from free space, their height must be explicitly specified for accurate deprojection into 3D space, as they are not anchored to any objects. For this reason, in typical tabletop manipulation scenarios, MOKA primarily focuses on cases where the waypoints are at the same height as the target point.

\begin{figure}[!th]
     \centering
     \scalebox{0.99}{
     \begin{tabular}{lp{9cm}}\hline 
     \textbf{Affordance} & Grasp the tennis racket.      \\
     \textbf{Force} & Grasp the egg. \\
     \textbf{Color}      & Grasp the blue cube.   \\
     \textbf{Location}     & Grab the block farthest from the plate and move it to the plate. \\
     \textbf{Tool}       & Grasp the tools for cutting a watermelon.  \\\hline
     \end{tabular}
     }
     \caption{The language description of the testing tasks.}\label{fig:embodied_instruct}
\end{figure}

We present an overview of our implementation of MOKA in Algorithm~\ref{alg:overview}. Our experiments aim to enhance the VLM’s ability to understand the physical world and validate its impact on downstream embodied agent tasks. Specifically, we employ two methods to improve the VLM: first, fine-tuning the VLM using PhysBench, and second, incorporating the PhysAgent to assist during VLM inference.
It is worth noting that the five tasks we address are relatively fundamental, unlike the complex multi-action tasks described in the MOKA paper, which require hierarchical decomposition from high- to low-level actions. In our case, the tasks can be executed directly without such decomposition."
\begin{algorithm}[H]
    \caption{MOKA Pipeline}
    \begin{algorithmic}[1]\label{alg:overview}
      \STATE \textbf{Input:} Vision-language Model $\mathcal{M}$, Task instruction $l$, text prompt for low-level reasoning $p_{low}$ and initial observation $s$
      \STATE Get observation $s$ from the top-down camera
      \STATE Propose keypoint and waypoint candidates and get annotated image $f(s_k)$
      \STATE Query $\mathcal{M}$ for low-level motion reasoning, obtain $y_{low} = \mathcal{M}([p_{low}, l, f(s)])$
      \STATE Execute $y_{low}$ on the robot in simulation
    \end{algorithmic}
\end{algorithm}

\subsection{Correlation Map}\label{app:c_map}
Following the approach of~\cite{tong2024cambrian, fang2024exploring, fang2025kaa}, we used the Pearson correlation coefficient to construct a relationship matrix. The data used to build this matrix can be found in Table~\ref{exp:c_map}.

\begin{table}[th!]
    \setlength{\tabcolsep}{3pt}
    \centering
    \scalebox{0.65}{
    \begin{tabular}{lcccccccccccccc}\hline
     & VQAv2 & GQA & VisWiz & SQA & TextVQA & POPE & MME & MMB & MMB\small{CN} & SEED & SEEDI & MMMU\small{val} & MMMU\small{test} & LLaVA-bench \\ \hline
    LLaVA-1.5-7B & 78.5 & 62.0 & 50.0 & 66.8 & 58.2 & 85.9 & 1510.7 & 64.3 & 58.3 & 61.5 & 67.0 & 33.2 & 31.1 & 63.4 \\
    LLaVA-1.5-13B & 80.0 & 63.3 & 53.6 & 71.6 & 61.3 & 85.9 & 1531.3 & 67.7 & 63.6 & 62.4 & 68.2 & 36.4 & 33.6 & 70.7 \\
    InstructBLIP-7B & 61.1 & 49.2 & 34.5 & 60.5 & 50.1 & 78.8 & 1210.1 & 36.0 & 23.7 & 53.4 & 58.8 & 32.9 & 30.6 & 60.9 \\
    InstructBLIP-13B & 62.3 & 49.5 & 33.4 & 63.1 & 50.7 & 78.9 & 1212.8 & 42.0 & 25.0 & 55.2 & 61.7 & 35.7 & 33.8 & 58.2 \\
    Qwen-VL-Chat & 78.2 & 57.5 & 38.9 & 68.2 & 61.5 & 85.6 & 1487.5 & 60.6 & 56.7 & 58.2 & 65.4 & 35.9 & 32.9 & 64.1 \\
    VILA-1.5-3B & 80.4 & 61.5 & 53.5 & 69.0 & 60.4 & 85.9 & 1442.4 & 63.4 & 52.7 & 60.9 & 67.9 & 33.3 & 30.8 & 75.9 \\
    VILA-1.5-3B-s2 & 79.8 & 61.4 & 61.3 & 69.6 & 63.4 & 85.3 & 1431.7 & 62.8 & 52.2 & 60.0 & 66.4 & 32.8 & 31.3 & 76.7 \\
    VILA-1.5-8B & 80.9 & 61.9 & 58.7 & 79.9 & 66.3 & 84.4 & 1577.0 & 72.3 & 66.2 & 64.2 & 71.4 & 36.9 & 36.0 & 80.0 \\
    VILA-1.5-13B & 82.8 & 64.3 & 62.6 & 80.1 & 65.0 & 86.3 & 1569.6 & 74.9 & 66.3 & 65.1 & 72.6 & 37.9 & 33.6 & 80.8 \\
    BLIP-2 & 41.0 & 44.6 & 29.4 & 61.0 & 42.1 & 85.3 & 1293.8 & 44.0 & 27.0 & 46.4 & 49.7 & 35.4 & 34.0 & 56.2 \\ \hline
    \end{tabular}
    }
    \vspace{-2mm}
    \caption{The performance of the 10 models used to construct the correlation map across 15 other VLM benchmarks.}\label{exp:c_map}
    \vspace{-4mm}
\end{table}

The correlation presented in Figure~\ref{fig:main_vis}(a) illustrates the relationships between the four major categories in PhysBench and other tasks. Additionally, we provide a detailed correlation map between PhysBench and 15 other vision-language benchmarks in Figure~\ref{fig:data_cases_full_m-m} below.
\begin{figure}[h!]
	\centering  
	\vspace{-2mm}
	\includegraphics[width=0.85\linewidth]{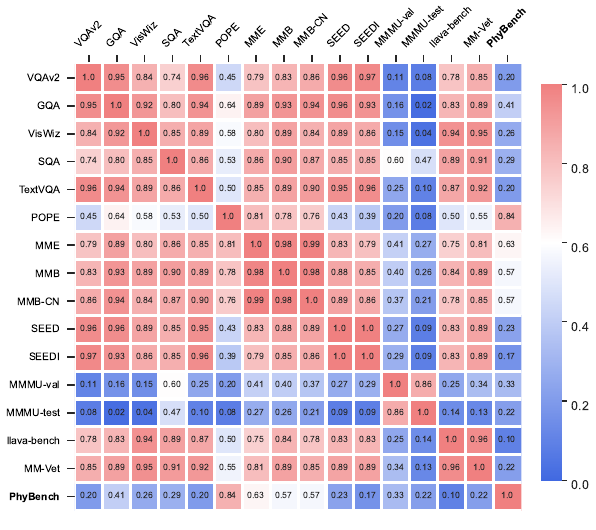}
	\vspace{-6mm}
        \caption{Correlation map between PhysBench and 15 other vision-language benchmarks.} 
	\label{fig:data_cases_full_m-m}  
	\vspace{-2mm}
\end{figure}

\subsection{Performance on Related Benchmarks}
To further evaluate the contribution of our data and model to understanding the physical world, we conducted tests on three existing benchmarks related to physical-world perception. Notably, these benchmarks focus on specific aspects of physical-world perception, whereas our PhysBench provides a more comprehensive and holistic evaluation, as summarized in Table~\ref{table_benchmark_comparison_qa}. Furthermore, since these datasets were not originally designed for VLMs, we applied necessary preprocessing to adapt them for our evaluations. 

\textbf{Setup}.\textit{EmbSpatial}~\citep{du2024embspatial} is a benchmark designed to evaluate spatial understanding within embodied environments with source image come from  MP3D~\citep{chang2017matterport3d}, ScanNet~\citep{dai2017scannet} and AI2-THOR~\citep{kolve2017ai2}. We utilized its benchmark dataset for testing purposes. 
\textit{ContPhy}~\citep{zheng2024contphy} is a benchmark aimed at assessing visual models' capabilities in perceiving continuous physical phenomena and properties. It comprises four simulation systems based on Unity3D: Fluid Hourglass, Rope-Pulley System, Cloth Magic Trick, and Ball Playground. Since ContPhy primarily targets visual and physical models, we selected 200 items for each of the four categories (split evenly between property-based and dynamics-based tasks) and translate them into multiple-choice format suitable for VLMs. The question prompts were modified to better align with the answering capabilities of VLMs.
\textit{Physion++}~\citep{physion++} evaluates the impact of physical properties such as mass, friction, elasticity, and deformability on physical phenomena. It leverages the ThreeDWorld simulation platform~\citep{gan2020threedworld} to generate a series of videos, each paired with a corresponding question. The videos consist of an inference phase, where artificial systems can identify objects' mechanical properties, followed by a prediction phase, where the model must predict whether two specified objects will collide after the video ends. Physion++ is primarily designed for vision models such as ResNet~\citep{he2016deep} and VGG~\citep{simonyan2014very}, with answers in the form of binary classification (yes/no), derived by converting model outputs into probabilities. 
To adapt Physion++ for VLMs, we processed the videos and reformulated the questions into natural language prompts, adding necessary contextual hints. For example, if a video includes a transition phase, we explicitly include a statement like, "The black screen in the video marks a transition in objects or scenes" in the prompt. Due to the small size of the test set, we combined the train and test sets, resulting in 250 VQA pairs for evaluation. During fine-tuning with PhysBench, we modified the options to an open-ended format. The results, presented in Table~\ref{tab:threeother}, are reported in terms of accuracy. All parameter settings and configurations are kept consistent with those outlined in the main text.

\begin{table}[h]
    \centering
    \small
    \vspace{-1mm}
    \caption{Performance on Three Related Benchmarks.}\label{tab:threeother}
    \vspace{-2mm}
    \resizebox{0.87\textwidth}{!}{
        \begin{tabular}{lcccc}
        \hline
        & ContPhy-Property&ContPhy-Dynamics & Physion++ & EmbSpatial \\ \hline
        Phi-3V&49.25&43.25&68.80&55.91\\ 
        Phi-3V + finetune &64.50&62.75&82.20&66.95\\ 
        Phi-3V + PhysAgent & 52.00 &44.75&78.40&62.28 \\ \hline
        \end{tabular}
    }
    \vspace{-2mm}
\end{table}

\textbf{Results}. As presented in Table~\ref{tab:threeother}, leveraging PhysBench data for fine-tuning or in a zero-shot setting with PhysAgent leads to performance improvements across the benchmarks, particularly in Physion++, where improvements of 19.50\% and 9.6\% are observed, with fine-tuning achieving the most significant gains. These results highlight the effectiveness of our data and methods in enhancing the capability of Vision-Language Models to comprehend the physical world.
\section{More Related Works}~\label{app_related_work}

\textbf{Vision-Language Models}. Vision-Language Models (VLMs) are large language models that integrate visual modalities, such as images and videos, with language knowledge~\citep{Wu2023NExTGPTAM,huang2024one,bai2022lat,Zhan2024AnyGPTUM,dai2024instructblip, bai2024meissonic}. Notable models like BLIP-2~\citep{li2023blip} and LLaVA~\citep{liu2024visual} have advanced image-captioning datasets and visual instruction tuning, with LLaVA-Next further improving single-image performance at higher computational costs~\citep{liu2024llavanext}. Subsequent models, such as QwenVL~\citep{bai2023qwen}, CogVLM~\citep{wang2023cogvlm}, and Yi-VL~\citep{ai2024yi}, have followed a similar architecture to LLaVA.
As single-image and text interaction technologies continue to mature, many recent VLMs~\citep{alayrac2022flamingo,Peng2023Kosmos2GM,pan2024auto, lin2023vila} are now capable of handling complex visual tasks with interleaved images or videos, enabling VLMs to tackle more sophisticated tasks~\citep{lu2024mathvista, yu2024hallucidoctor} and paving the way for interactions with the real physical world.

\textbf{Vision-Language Benchmarks} VLMs~\citep{liu2024visual,achiam2023gpt,pan2024auto, yu2023visually, chow2024unified, run2024mmevol, li2024driving, li2024wolf} have inherited and advanced many intriguing features from text-only LMs. Benchmarks for VLMs have rapidly emerged to evaluate performance in areas such as image question answering~\citep{ying2024mmtbench}, in-context response~\citep{yu2023mmvet}, chart understanding~\citep{li2024mmsci}, and web comprehension~\citep{liu2024visualwebbench, zhou2023webarena}. Some benchmarks cover spatial relations understanding~\citep{li2023seed}, but often overlook the ability to devise complex spatial action plans based on physical world comprehension.
Recently, new benchmarks have also emerged that focus on the ability to understand multiple images in long contexts~\citep{zhang2024task, kil2024compbench,Jiang2024MANTISIM, wu2023q, li2024mvbench, ge2024demon24} and complex realistic environments~\citep{fu2024blink, bai2024humanedit}.
However, these benchmarks—whether based on answering questions from images, videos, or tables, or using visual prompts~\citep{fu2024blink, yu2024anyedit}—ultimately rely on responding to the content of the given images rather than the true perception of the physical world, thus falling short of achieving spatial intelligence~\citep{gupta2021embodied, yang2024thinking}.
\begin{table}[h]
    \centering
    \tiny
    \caption{\textbf{Comparison between PhysBench and other vision-language benchmarks}. In the format, I, T, V present text, image, and video. Annotated means annotate the existing dataset, like MSCOCO~\cite{karpathy2015deep}. LLaVA$^\text{Wd}$: LLaVA-Bench(In-the-Wild)-Detail~\cite{liu2024visual}. Reasoning means that it requires the VLMs' reasoning ability to answer the question.}
    \label{table_vlm_benchmark_comparison}
    \resizebox{\textwidth}{!}{
        \begin{tabular}{lcccccc}
        \toprule
        \textbf{Dataset} & \textbf{Size (k)} & \textbf{Format} & \textbf{Interleaving} & \textbf{Source} & \textbf{Domain} & \textbf{Reasoning}\\
        \midrule
        VQA-v2~\cite{goyal2017making}&1,105,904 &I+T&\crossmark&Annotated & Image content & \crossmark \\
        GQA~\cite{hudson2019gqa} &22,669,678 &I+T& \crossmark &Annotated & Image content & \crossmark \\
        VizWiz~\cite{gurari2018vizwiz} & 32,000 &I+T& \crossmark &Annotated & Image content & \crossmark \\
        TextVQA~\cite{singh2019towards} & 45,000 &I+T& \crossmark &Annotated & Image content & \crossmark \\
        OKVQA~\cite{okvqa} & 14,000 &I+T& \crossmark &Annotated & Image content & \crossmark \\
        SEED~\cite{li2023seed} & 19,000 &V+I+T& \crossmark &Annotated & Image and video content & \crossmark \\
        MMBench~\cite{liu2023mmbench} & 3,000 &I+T& \crossmark &Annotated & Image content & \crossmark \\
        MME~\cite{yin2023survey} & 1,297 &I+T& \crossmark &Annotated & Image content & \crossmark \\ 
        POPE~\cite{li2023evaluating} & 18,000 &I+T& \crossmark &Annotated & Image hallucination detection & \crossmark \\
        MM-Vet~\cite{yu2023mmvet} & 200 &I+T& \crossmark &Annotated & Image chat & \crossmark \\
        LLaVA$^\text{Wd}$~\cite{liu2024visual} & 60 &I+T& \crossmark &Annotated & Image chat & \crossmark \\
        SQA$^\text{I}$~\cite{lu2022learn} & 6,000 &I+T& \crossmark &Annotated & Image content & \crossmark \\
        NLVR2~\cite{suhr2018corpus} & 6,967 &I+T& \crossmark &Annotated & Image content & \crossmark \\
        MathVista~\cite{lu2024mathvista} & 6,141 &I+T& \crossmark &Annotated & Math & \checkmark \\
        BLINK~\cite{fu2024blink} &1,901 &I+T& \checkmark &Annotated, Chart & Visual prompt & \crossmark \\
        Mantis-eval~\cite{Jiang2024MANTISIM} & 217 &I+T& \checkmark &Annotated & Image chat & \crossmark \\
        Q-Bench~\cite{wu2023q} & 2,990 &I+T& \checkmark &Annotated & Image content & \crossmark \\
        MMMU~\cite{yue2023mmmu} & 11,500 &I+T& \checkmark &Annotated, Web, Textbook & Image content & \crossmark \\
        PhysBench &  10,002 &V+I+T& \checkmark &Annotated, Web, Simulation, Real-world & Physical property and dynamics & \checkmark \\
        \bottomrule
        \end{tabular}
    }
\end{table}

\textbf{Video Benchmarks}. 
With the growing interest in video understanding, the development of benchmarks for VLMs has become increasingly emphasized. In video comprehension, the research community has made significant strides, particularly for short videos. There are specialized benchmarks for temporal perception~\citep{yu2019activitynet, wu2024star}, action understanding~\citep{liu2024tempcompass,mangalam2024egoschema}, video classification~\citep{kay2017kinetics}, video reasoning~\citep{xiao2021next, xie2023funqa}, video captioning~\citep{miech2019howto100m, xu2016msr}, video question-answering~\citep{zhou2024mlvu, li2023llama, li2024mvbench}, long video comprehension~\citep{wu2024longvideobench, chandrasegaran2024hourvideo}, video generation~\citep{bansal2024videophy}, and interleaved video-text question-answering~\citep{wang2024sok}.
However, these works primarily focus on evaluating video content and do not explore the underlying mechanisms of video representation or address true physical world perception. Furthermore, during our experiments, we observed significant challenges with current video VLMs in following instructions and answering questions, as the models frequently output descriptions of the video rather than directly addressing the posed questions.

\textbf{Interleaved Vision-Language Benchmarks}.
VLMs are increasingly processing longer and more complex inputs. Along with this development, several benchmarks with interleaved inputs have emerged~\citep{li2023fine, wang2024muirbench, meng2024mmiumultimodalmultiimageunderstanding}. For example, SEED-Bench~\citep{li2023seed, li2024seed} focuses on video understanding, BLINK~\citep{fu2024blink} introduces visual prompts, and NLVR2~\citep{suhr2018corpus} differentiates between two images. However, all of these benchmarks still primarily assess content description based on images, without evaluating physical understanding or perception abilities. Furthermore, current interleaved benchmarks only involve images and text, while PhysBench is a dataset that interweaves video, image, and text inputs. A detailed comparison with the previously mentioned benchmarks and other vision-language benchmarks can be found in Table~\ref{table_vlm_benchmark_comparison}.

\textbf{Science-related Benchmarks}. 
In addition to the benchmarks related to physical world comprehension mentioned in Section~\ref{sec: realate}, there are also benchmarks that assess models' understanding through middle or university-level physics exam questions. SciQ~\citep{welbl-etal-2017-crowdsourcing}, ScienceQA~\citep{lu2022learn}, E-EVAL~\citep{hou2024eeval}, MMLU-STEM~\citep{hendrycks2020measuring}, and C-Eval-STEM~\citep{huang2023ceval} include some physics-related questions, but these datasets often allow questions to be answered simply by analyzing the provided images, lacking the complexity of reasoning and computational tasks. JEEBench~\citep{arora2023llms} requires multistep reasoning with physics knowledge but is limited in scope and purely text-based. SciBench~\citep{wang2024scibench}, OlympiadBench~\citep{he2024olympiadbench}, MathVista~\citep{lu2024mathvista}, and OCWCourses~\citep{lewkowycz2022solving} provide college-level physics questions.
However, these benchmarks mainly consist of homework and exam-style questions, focusing more on mathematical reasoning~\citep{zheng2024processbenchidentifyingprocesserrors} and general knowledge rather than true physical world comprehension. In contrast, our PhysBench is the first systematic and comprehensive question-answering benchmark specifically designed for understanding the real physical world.

\begin{table}[h]
    \centering
    \small
    \caption{A comparison between PhysBench and other physical understanding benchmarks not in question-answering format.}
    \label{table_benchmark_comparison}
    \resizebox{\textwidth}{!}{
\begin{tabular}{lcccccccccccr}\toprule
           & Property & Attribute & Location & Velocity & Temperature & Camera & Light & Collision & Manipulation & Fluid & Interleaved & Size \\\midrule
Physics 101~\cite{wu2016physics}& \crossmark    & \crossmark     & \crossmark    & \crossmark    & \crossmark       & \crossmark  & \crossmark & \checkmark    & \crossmark        & \crossmark & \crossmark        & 17,408                   \\
IntPhys~\cite{riochet2018intphys} & \checkmark   & \crossmark     & \crossmark    & \crossmark    & \crossmark       & \crossmark  & \crossmark & \crossmark     & \crossmark        & \crossmark & \crossmark        & 15,000                   \\
ESPRIT~\cite{rajani2020esprit}& \crossmark    & \crossmark     & \crossmark    & \crossmark    & \crossmark       & \crossmark  & \crossmark & \checkmark    & \crossmark        & \crossmark & \crossmark        & 2,441                    \\
CRAFT~\cite{ates2020craft}& \checkmark   & \crossmark     & \crossmark    & \crossmark    & \crossmark       & \crossmark  & \crossmark & \checkmark    & \crossmark        & \crossmark & \crossmark        & 57,000                   \\
CoPhy~\cite{baradel2019cophy}& \checkmark   & \crossmark     & \checkmark   & \crossmark    & \crossmark       & \crossmark  & \crossmark & \checkmark    & \crossmark        & \crossmark & \crossmark        & 216,000                  \\
PhysBench   & \checkmark   & \checkmark    & \checkmark   & \checkmark   & \checkmark      & \checkmark & \checkmark& \checkmark    & \checkmark       & \checkmark& \checkmark       & 10,002                  \\
  \bottomrule              
\end{tabular}
    }
\end{table}

\textbf{3D Scence VQA}. Recently, multi-modal 3D perception has garnered increasing attention due to its connection to the physical world, driving rapid advancements in the field. SQA3D~\citep{ma2022sqa3d} highlights the importance of “situations” within contextual understanding. EmbodiedScan~\citep{wang2023embodiedscan} and SceneVerse~\citep{jia2025sceneverse} expand the scope by collecting more scenes or annotating additional objects, scaling annotations to the millions. OpenEQA~\citep{OpenEQA2023}, SpatialRGPT-Bench~\citep{cheng2024spatialrgpt}, MMScan~\citep{lyu2024mmscan}, and MSNN~\citep{linghu2024multi} integrate comprehensive annotations, adapting the task into a VQA format. However, these studies primarily focus on geometric relationships, which represent only a subset of the broader understanding of the physical world, as discussed in Section~\ref{sec: realate}. Our work prioritizes a more holistic evaluation of the physical world perception capabilities of VLMs across four major task categories: Physical Object Properties, Physical Object Relationships, Physical Scene Understanding, and Physics-based Dynamics. Additionally, the spatial components in our dataset differ significantly from those in existing Spatial VQA datasets. While such benchmarks typically rely on 3D point cloud scenes or interleaved 2D images from multiple viewpoints, our dataset uses interleaved images to capture physical world dynamics, such as viewpoint rotations and the progression of physical phenomena.

\textbf{VLMs for Robotic Manipulation}. Recently, two main approaches have been proposed for applying Vision-Language Models (VLMs)~\citep{liu2024visual, achiam2023gpt, team2023gemini, mao2023gpt, mao2023language} to robotic manipulation: \textit{(a) directly generating actions}~\citep{kim24openvla, octo_2023, zawalski2024robotic, niu2024llarva, Zawalski24-ecot} and \textit{(b) employing VLMs as agents}~\citep{liu2024moka, google2024pivot, huang2023voxposer}. Approach (a) involves directly outputting actions, which requires extensive training—for instance, OpenVLA was trained using 64$\times$A100 GPUs—and produces embodied-specific actions that necessitate targeted fine-tuning. In contrast, approach (b) generates affordance representations through VLMs, which are subsequently converted into actions. This method requires less training and exhibits stronger generalization capabilities. However, it suffers from weaker perception capabilities in the physical world, leading to performance limitations~\citep{liu2024moka, mao2024learning}.
\section{More Examples}~\label{app:task_examples}
We use red color as \red{right answer}. We use red to indicate the correct answer (\red{correct answer}). It is important to note that, to reduce difficulty and facilitate evaluation, we employed a multiple-choice format rather than open-ended responses. For space-saving purposes, the example figures in the main text do not display the available options.
\subsection{Physical Object Property Sub-task}

\begin{table*}[th!]
\fontsize{9.0pt}{\baselineskip}\selectfont
\linespread{0.9}\selectfont
\begin{mycase}{light_pink}
\begin{question}{light_pink}
\includeimage{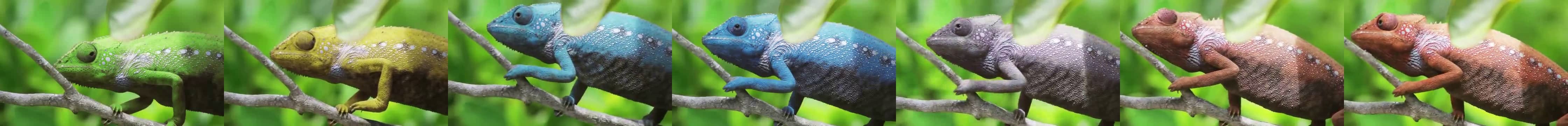}{5} 
The color of the original spots on the chameleon's body has not changed during the process of changing its color. What color are these spots?
\end{question}
\begin{answer}{light_pink}
A. Black B. Green \red{C. White} D. Blue
\end{answer}
\end{mycase}
\vspace{-2mm}
\captionof{figure}{Example for property color. Ability Type is identify.}
\vspace{-3mm}
\label{fig:example_1}
\end{table*}
\vspace{-4mm}
\begin{table*}[th!]
\fontsize{9.0pt}{\baselineskip}\selectfont
\linespread{0.9}\selectfont
\begin{mycase}{light_pink}
\begin{question}{light_pink}
\includeimage{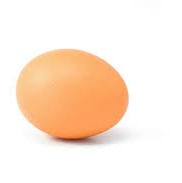}{7} 
The weight of the object about the size of a ping-pong ball is closest to which of the options?
\end{question}
\begin{answer}{light_pink}
A. 5g \quad   \red{B. 50g} \quad     C. 500g\quad   D. 5kg
\end{answer}
\begin{question}{light_pink}
\includeimage{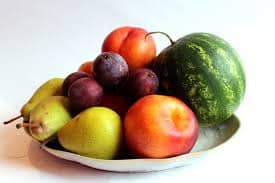}{7} 
Which fruit in the picture is most likely to be the heaviest individually?
\end{question}
\begin{answer}{light_pink}
\red{A. Watermelon}\quad B. Pear\quad C. Plum\quad D. Peach
\end{answer}
\begin{question}{light_pink}
\includeimage{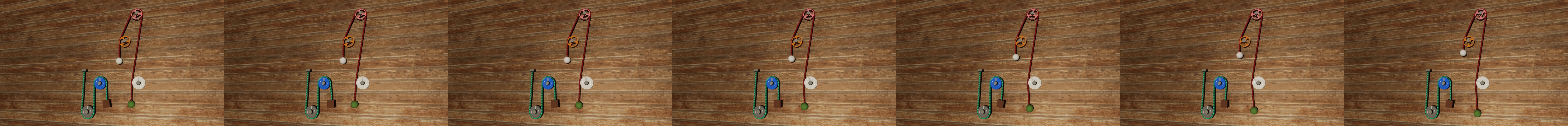}{5} 
What can be said about the mass of the brown cube compared to the white sphere?
\end{question}
\begin{answer}{light_pink}
A. The brown cube is less massive \quad B. The white sphere is less massive

C. They have the same mass \quad \red{D. Cannot answer}
\end{answer}
\end{mycase}
\vspace{-2mm}
\captionof{figure}{Three examples related to property mass, categorized by the following ability types: identification, comparison, and comparison.}
\vspace{-3mm}
\label{fig:example_2}
\end{table*}

\begin{table*}[th!]
\fontsize{9.0pt}{\baselineskip}\selectfont
\linespread{0.9}\selectfont
\begin{mycase}{light_pink}
\begin{question}{light_pink}
\includeimage{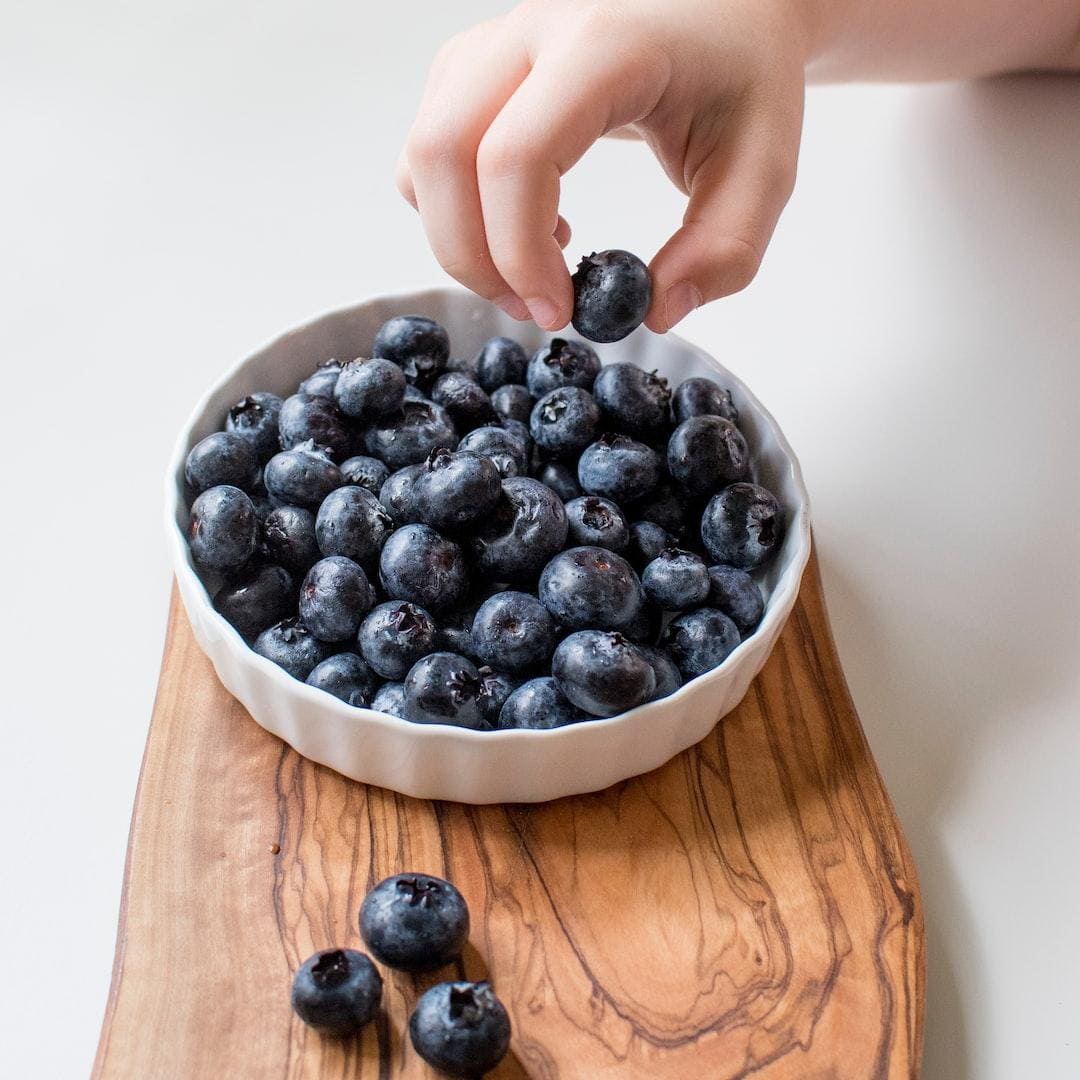}{7} 
How many blueberries are there outside the plate?
\end{question}
\begin{answer}{light_pink}
A. 1\quad B. 2\quad \red{C. 3}\quad D. 4
\end{answer}
\begin{question}{light_pink}
\includeimage{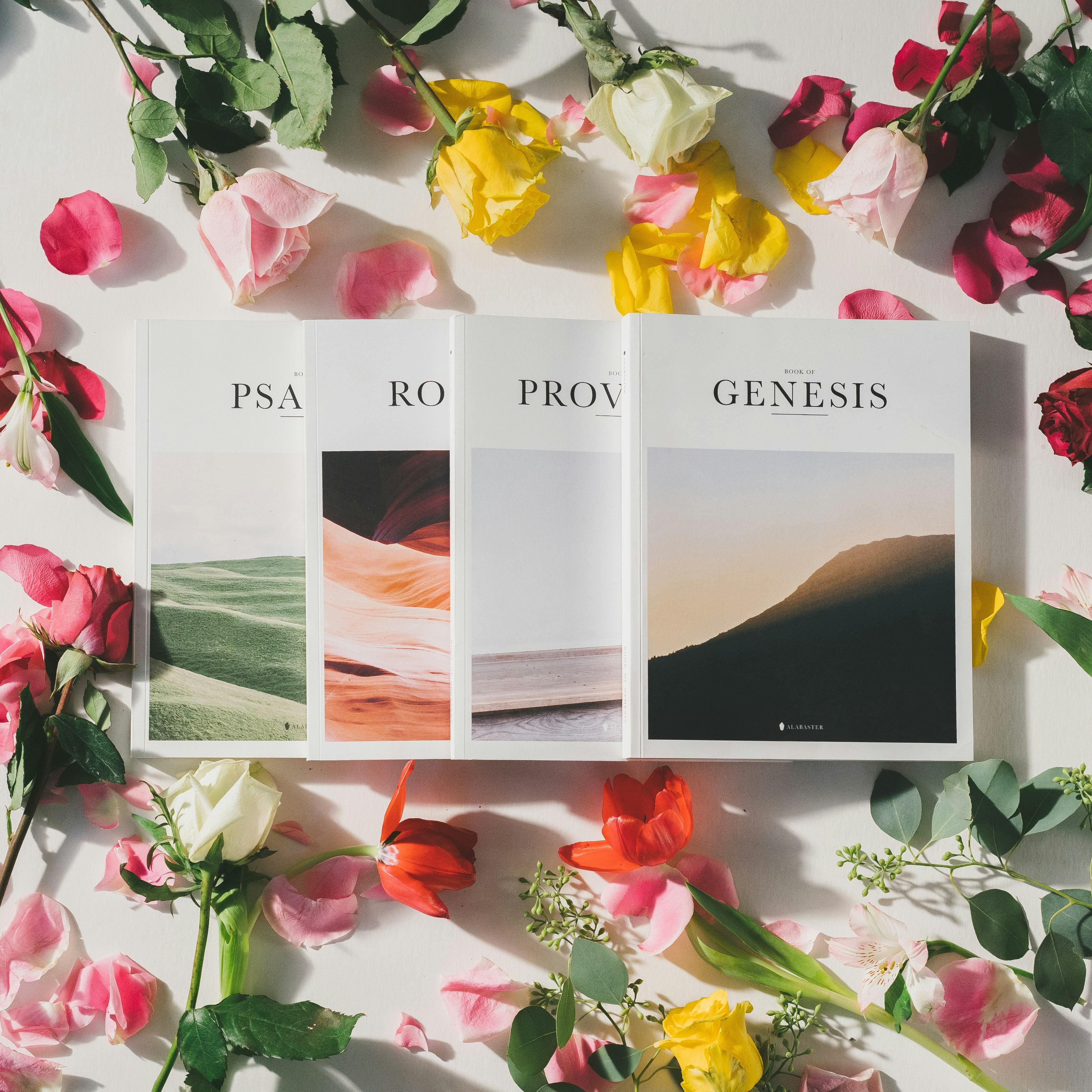}{7} 
How many of the books pictured have covers with titles beginning with P?
\end{question}
\begin{answer}{light_pink}
A. 1\quad \red{B. 2}\quad C. 3\quad D. 4
\end{answer}
\begin{question}{light_pink}
\includeimage{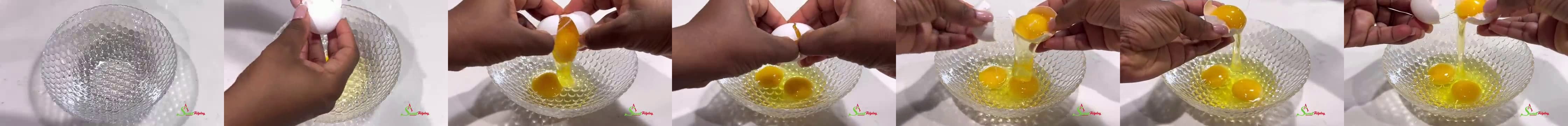}{5} 
Following the content of the <video>, How many eggs were broken?
\end{question}
\begin{answer}{light_pink}
A. 1\quad B. 2\quad \red{C. 3}\quad D. 4
\end{answer}
\begin{question}{light_pink}
\includeimage{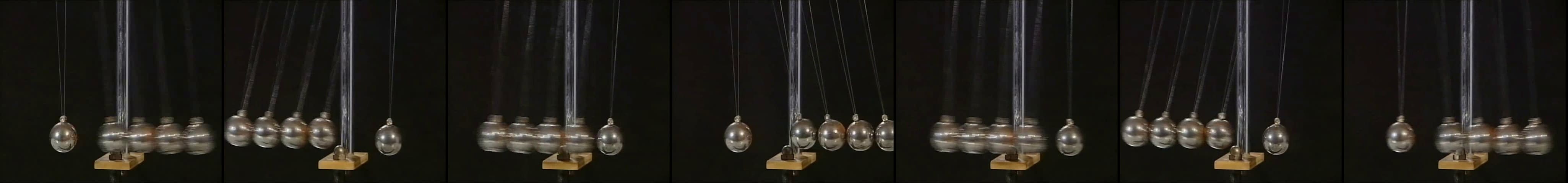}{7} 
How many balls are moving close together without separating?
\end{question}
\begin{answer}{light_pink}
A. 2\quad B. 3\quad \red{C. 4}\quad D. 6
\end{answer}
\end{mycase}
\vspace{-2mm}
\captionof{figure}{Four examples related to property number, ability types are all identification.}
\vspace{-3mm}
\label{fig:example_3}
\end{table*}
\begin{table*}[th!]
\fontsize{9.0pt}{\baselineskip}\selectfont
\linespread{0.9}\selectfont
\begin{mycase}{light_pink}
\begin{question}{light_pink}
\includeimage{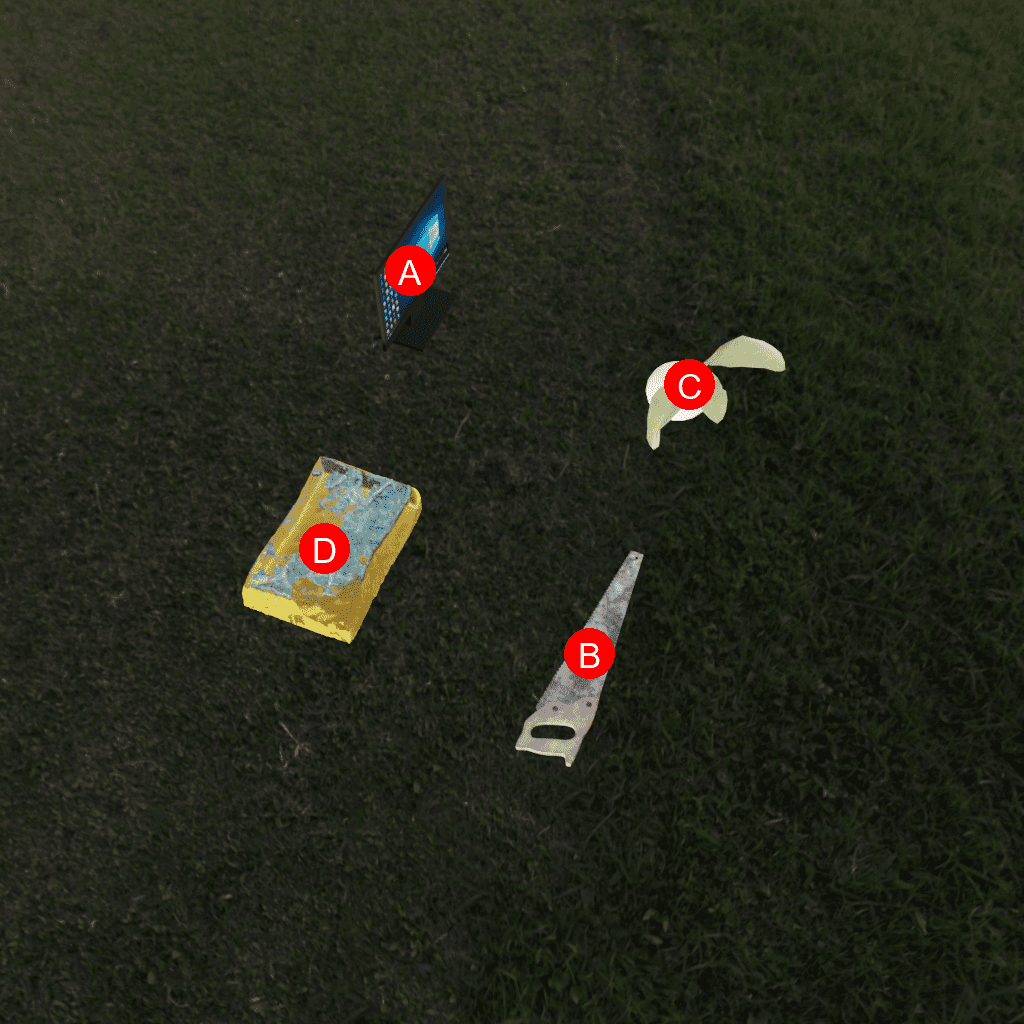}{7} 
In the photo, which point with option signifies the object with the most sharp?
\end{question}
\begin{answer}{light_pink}
A. Point A\quad \red{B. Point B}\quad C. Point C\quad D. Point D
\end{answer}
\begin{question}{light_pink}
\includeimage{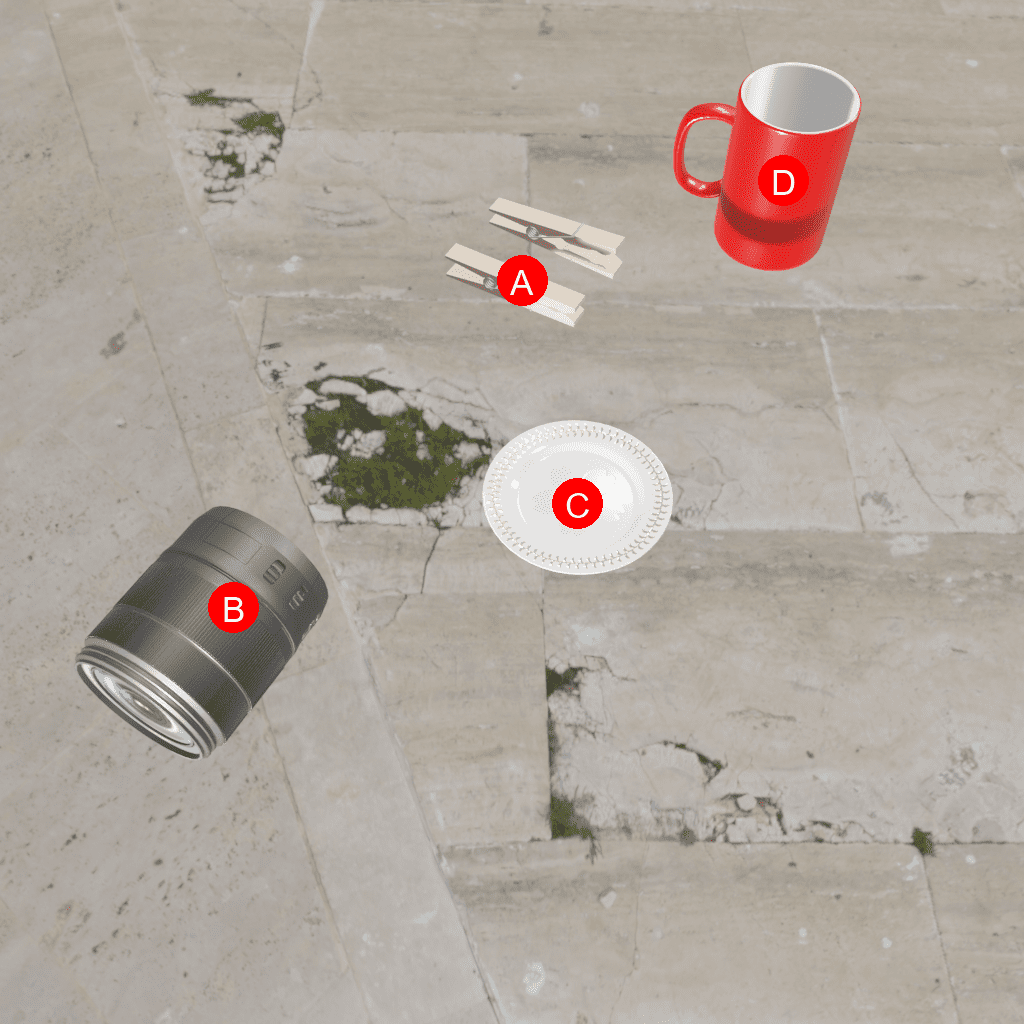}{7} 
Can you tell me which point with option in the image points to the object that has the least brittle?
\end{question}
\begin{answer}{light_pink}
\red{A. Point A}\quad B. Point B\quad C. Point C\quad D. Point D
\end{answer}
\begin{question}{light_pink}
\includeimage{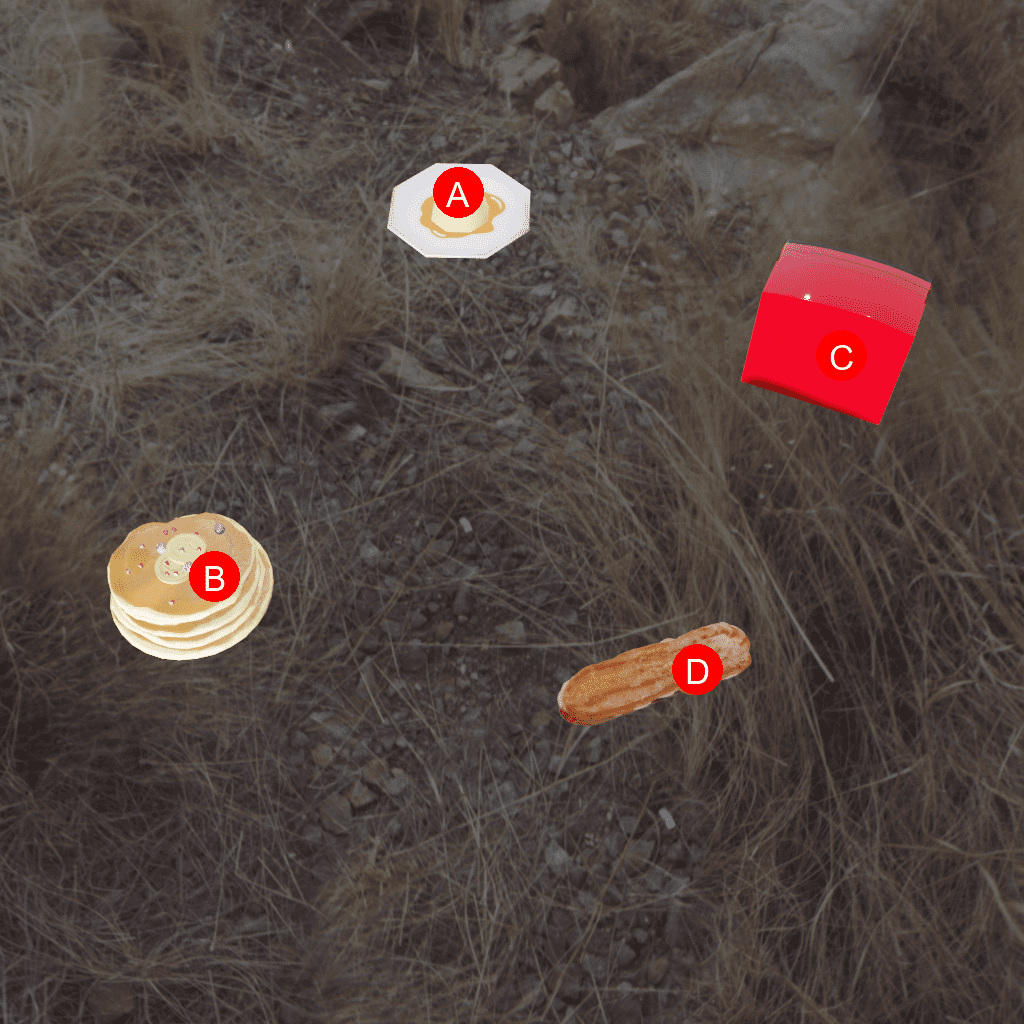}{7} 
Which point with option in the photograph pinpoints the object with the most stiff?
\end{question}
\begin{answer}{light_pink}
A. Point A\quad B. Point B\quad \red{C. Point C}\quad D. Point D
\end{answer}
\begin{question}{light_pink}
\includeimage{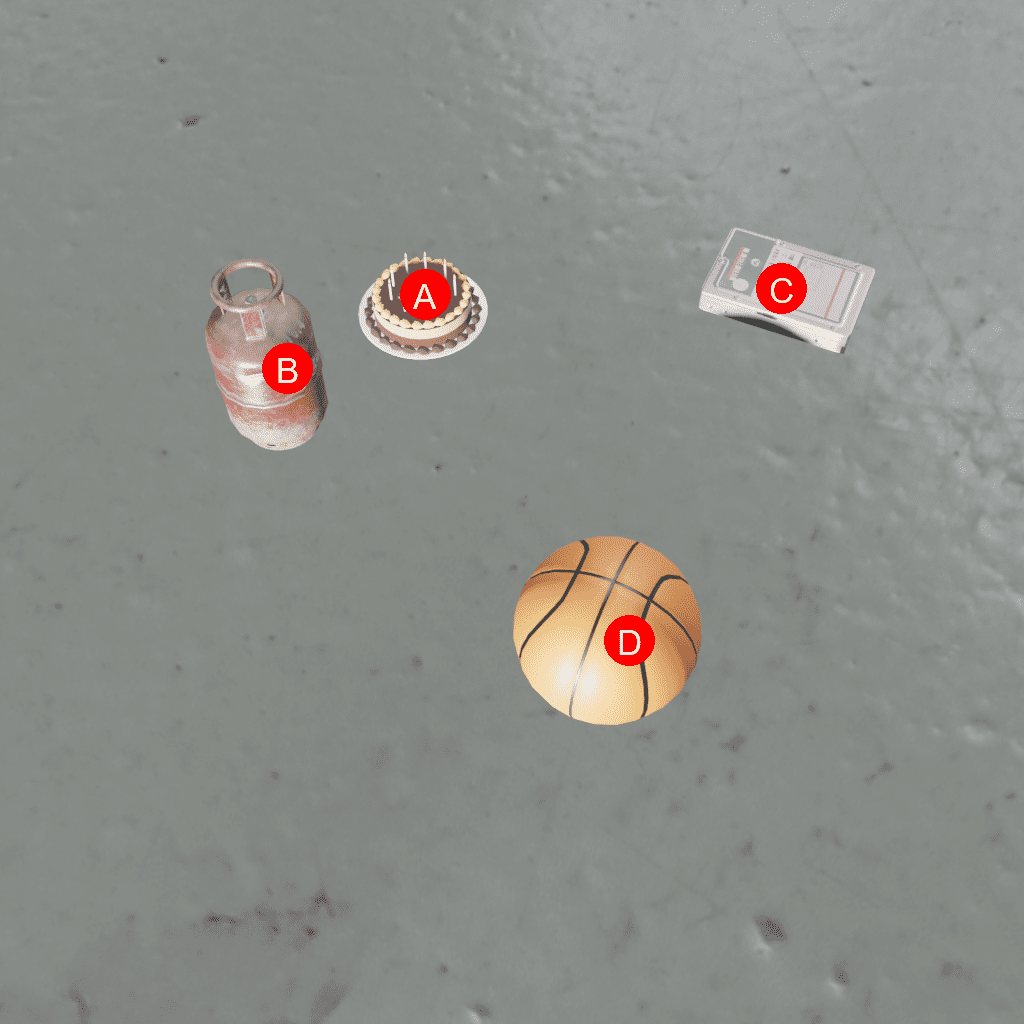}{7} 
Which point with option in the image marks the object that exhibits the most elastic?
\end{question}
\begin{answer}{light_pink}
A. Point A\quad B. Point B\quad C. Point C\quad \red{D. Point D}
\end{answer}
\begin{question}{light_pink}
\includeimage{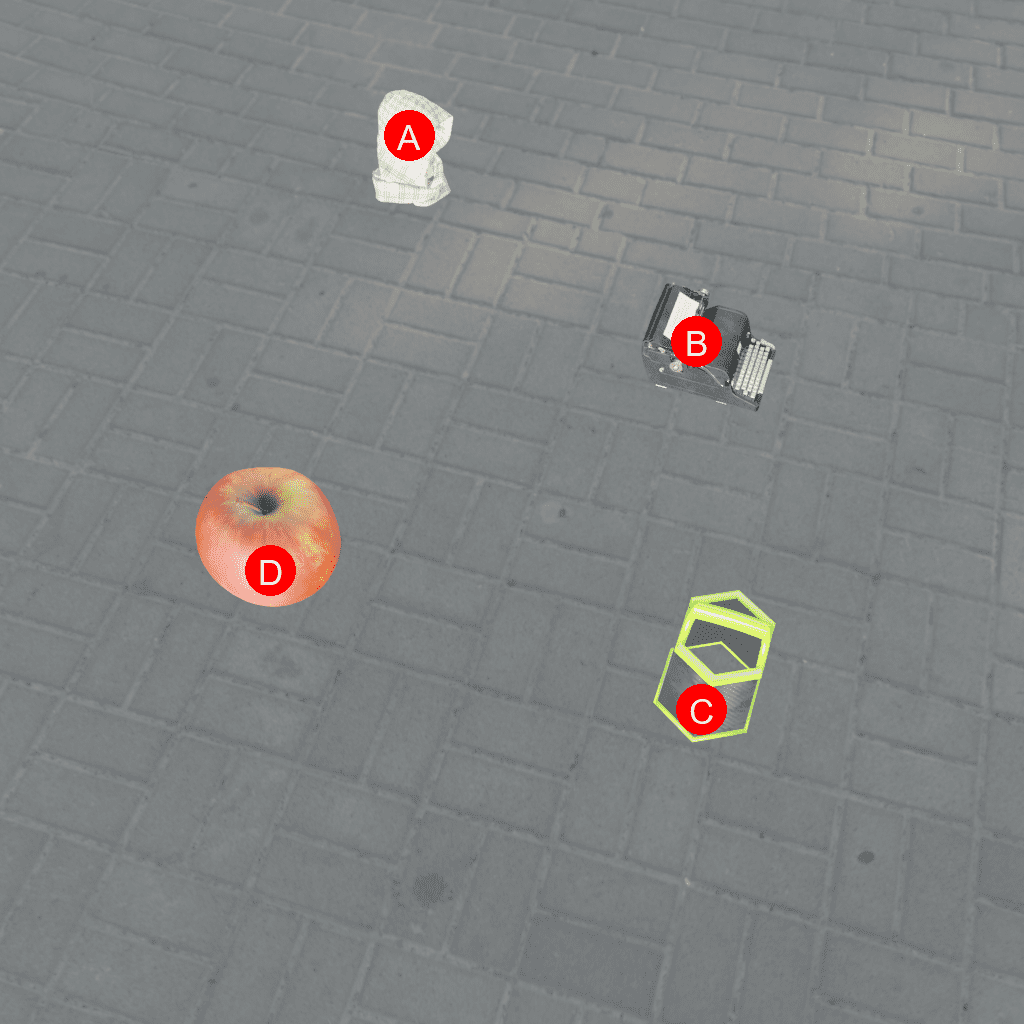}{7} 
Which point with option in the photograph captures the object that has the most malleable?
\end{question}
\begin{answer}{light_pink}
\red{A. Point A}\quad B. Point B\quad C. Point C\quad D. Point D
\end{answer}
\begin{question}{light_pink}
\includeimage{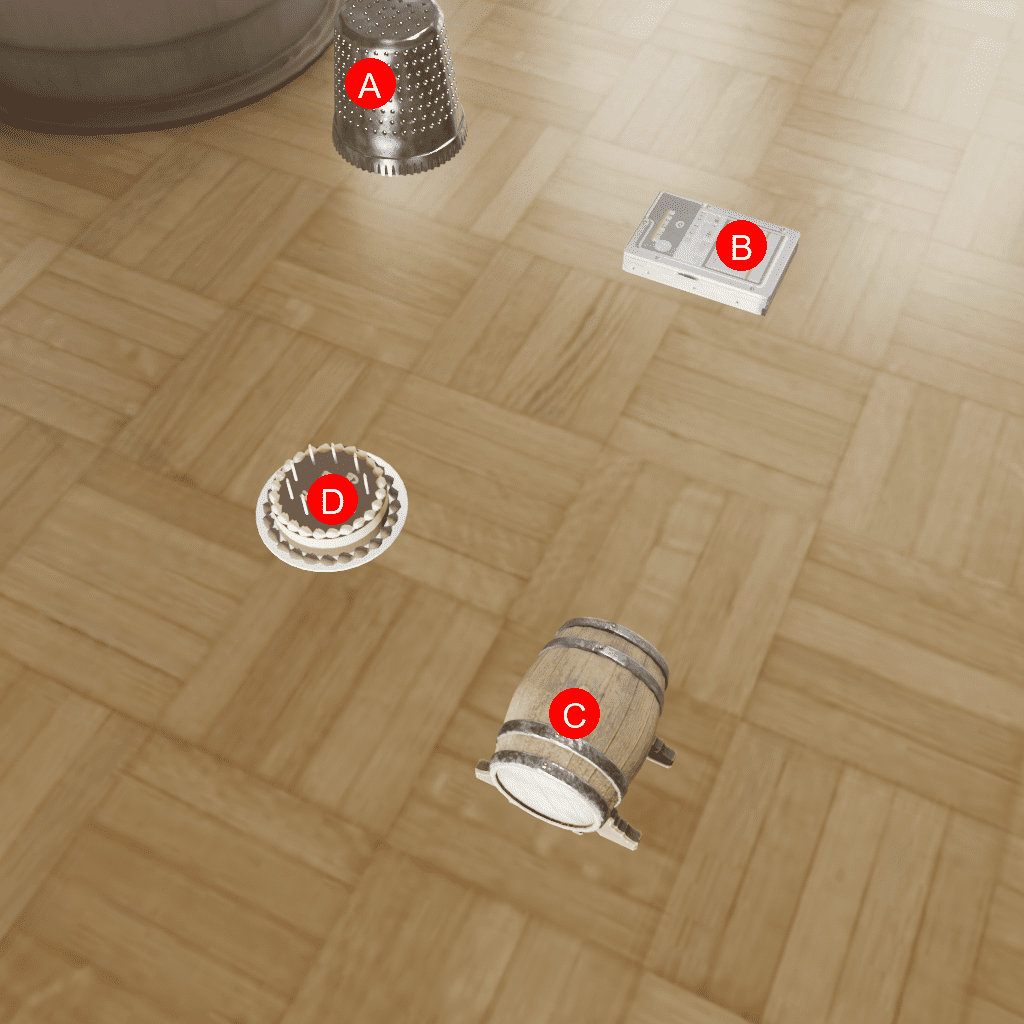}{7} 
Which point with option in the image isolates the object with the most soft?
\end{question}
\begin{answer}{light_pink}
A. Point A\quad B. Point B\quad C. Point C\quad \red{D. Point D}
\end{answer}
\end{mycase}
\vspace{-2mm}
\captionof{figure}{Six examples of property attributes include sharpness, brittleness, stiffness, elasticity, malleability, and softness.}
\vspace{-3mm}
\label{fig:example_4}
\end{table*}

\subsection{Physical Object Relationships Sub-task}

\begin{table*}[th!]
\fontsize{9.0pt}{\baselineskip}\selectfont
\linespread{0.9}\selectfont
\begin{mycase}{light_blue}
\begin{question}{light_blue}
\includeimage{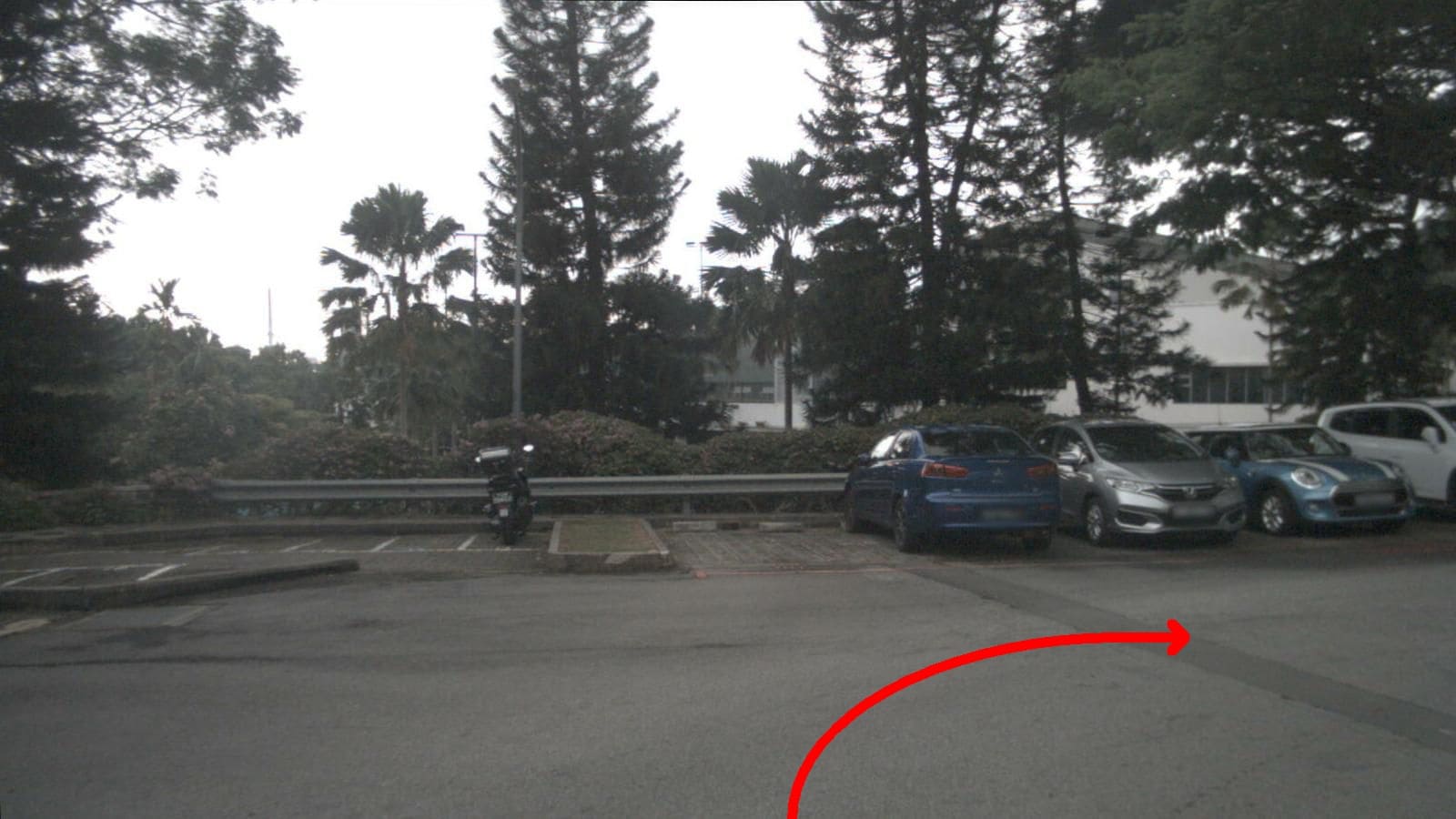}{6}Moving according to the arrow in the picture, which of the following options are you most likely to encounter?   
\end{question}

\begin{answer}{light_blue}
A. \includeimage{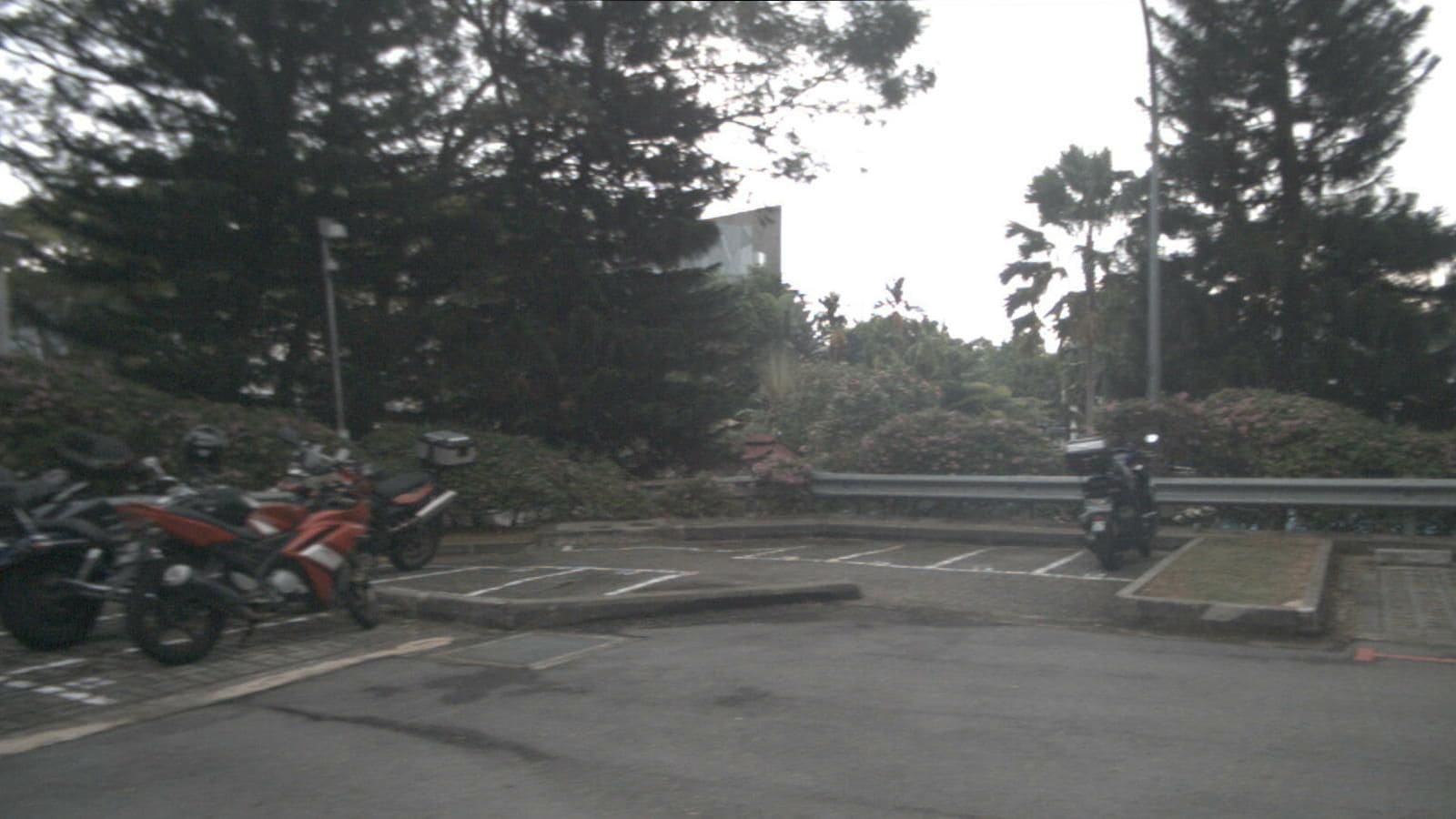}{6}\qquad
B. \includeimage{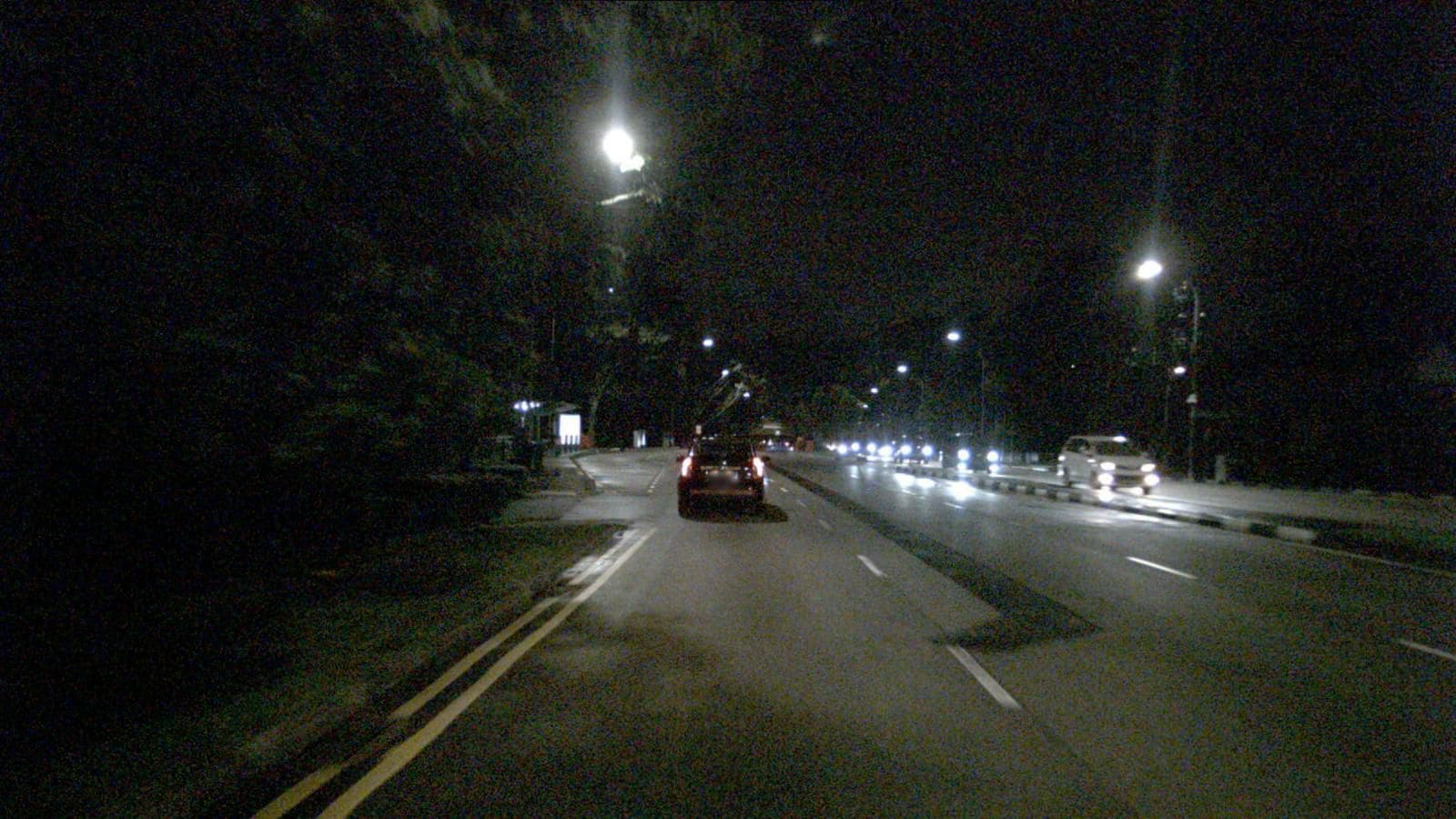}{6}\\
C. \includeimage{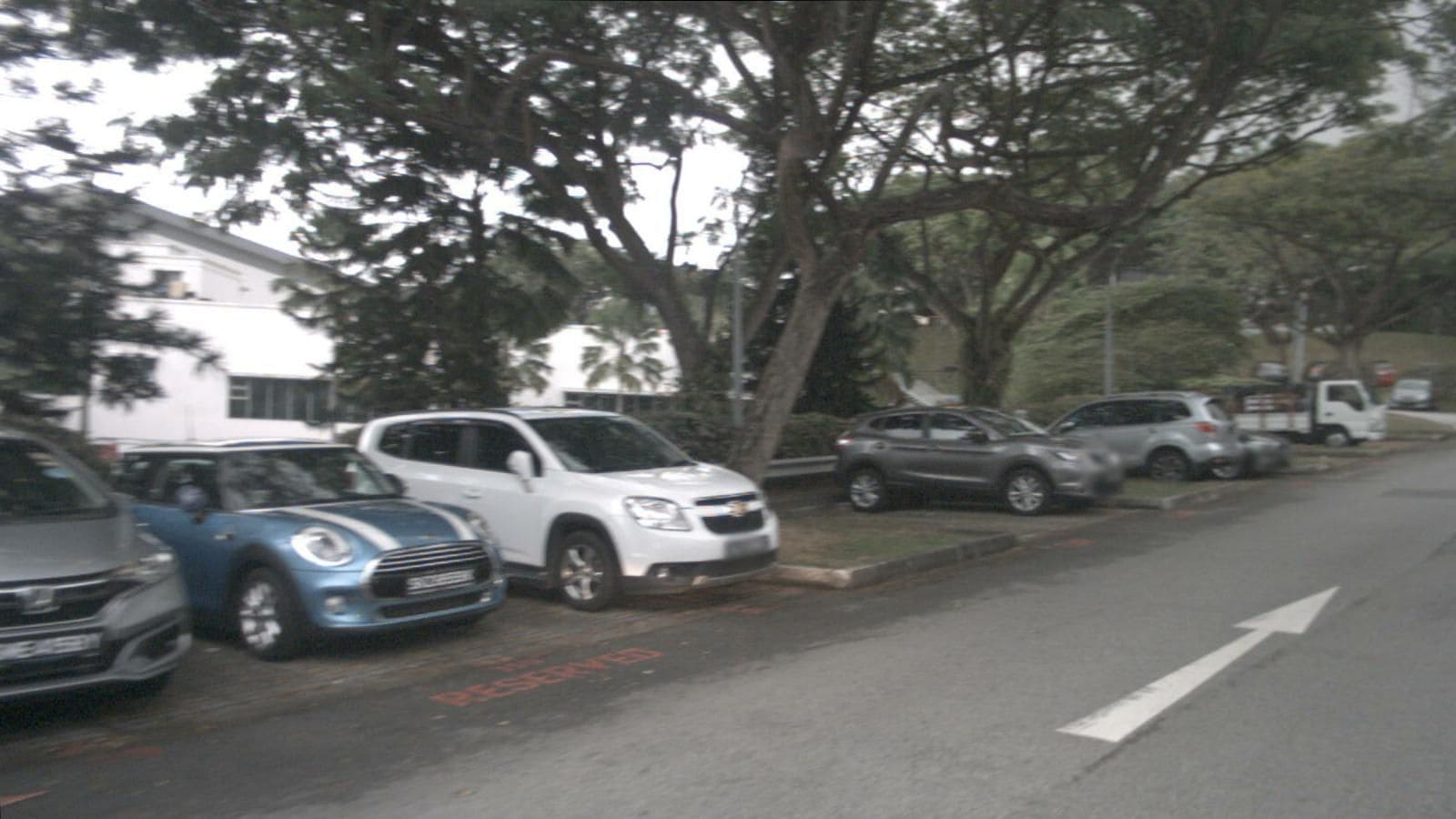}{6}\qquad
\red{D.} \includeimage{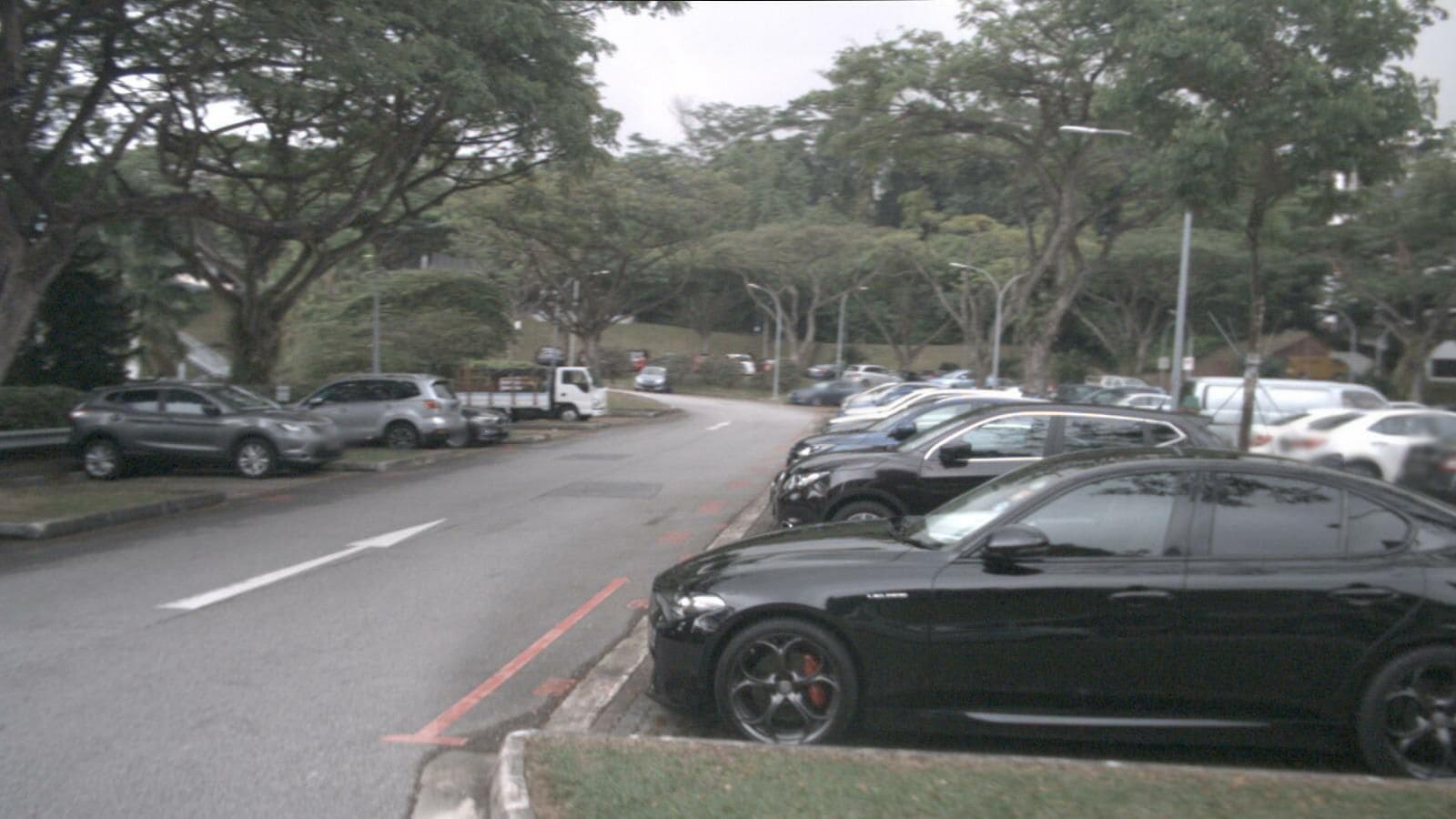}{6}
\end{answer}

\begin{question}{light_blue}
\includeimage{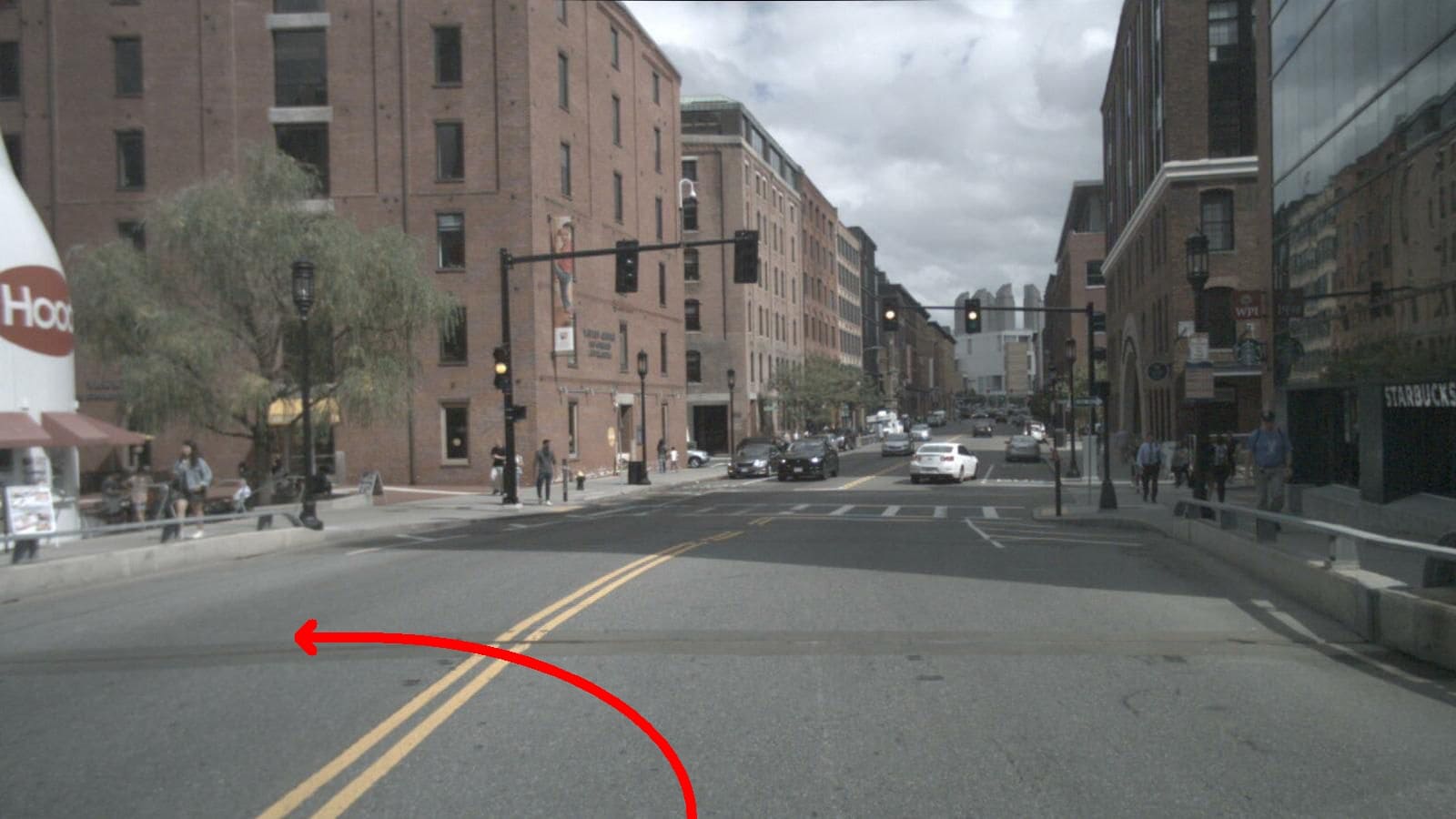}{6}Following the direction indicated by the arrow in the picture, which of the following options are you most likely to encounter? 
\end{question}
\begin{answer}{light_blue}
A. \includeimage{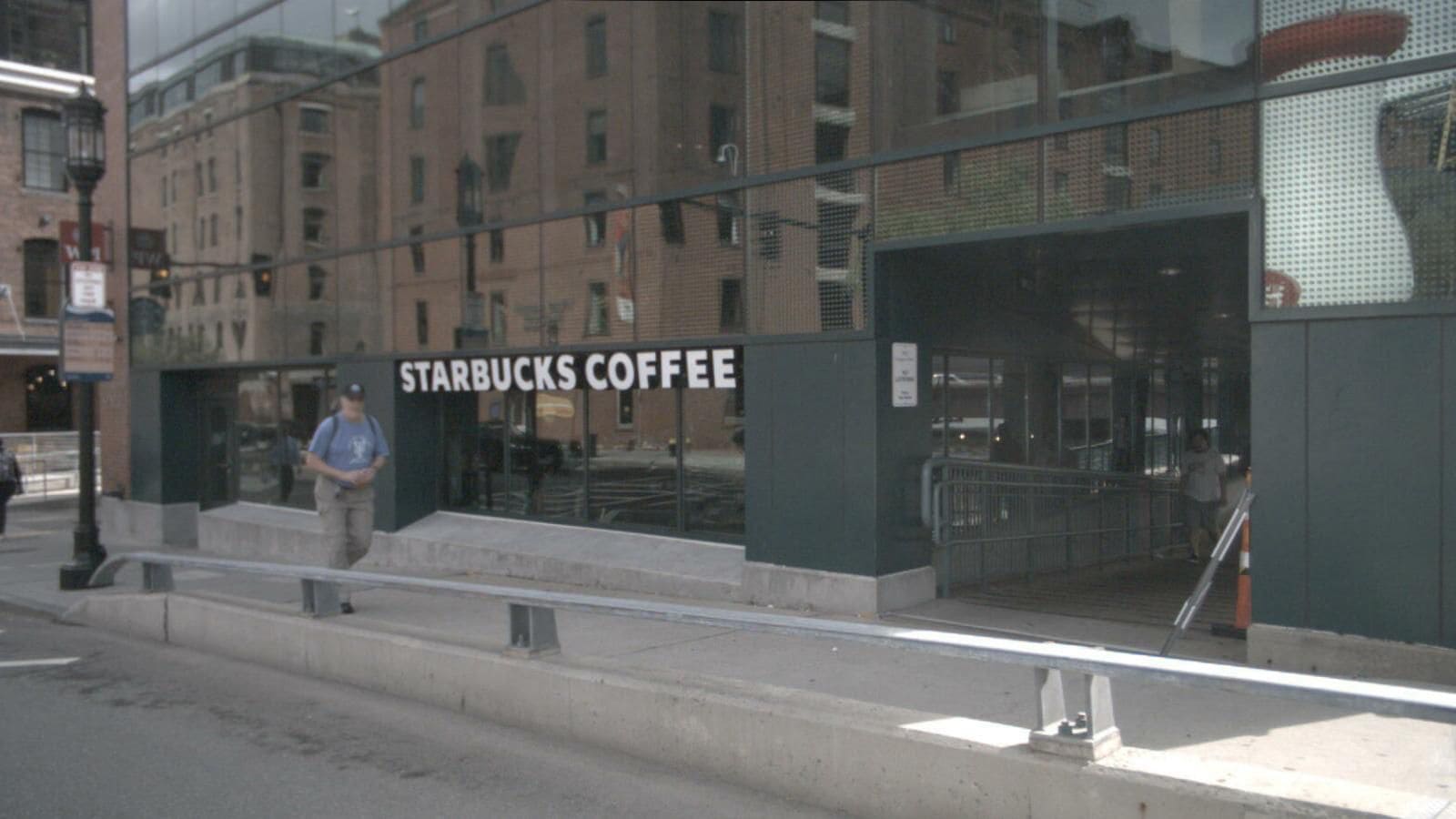}{6}\qquad
\red{B.} \includeimage{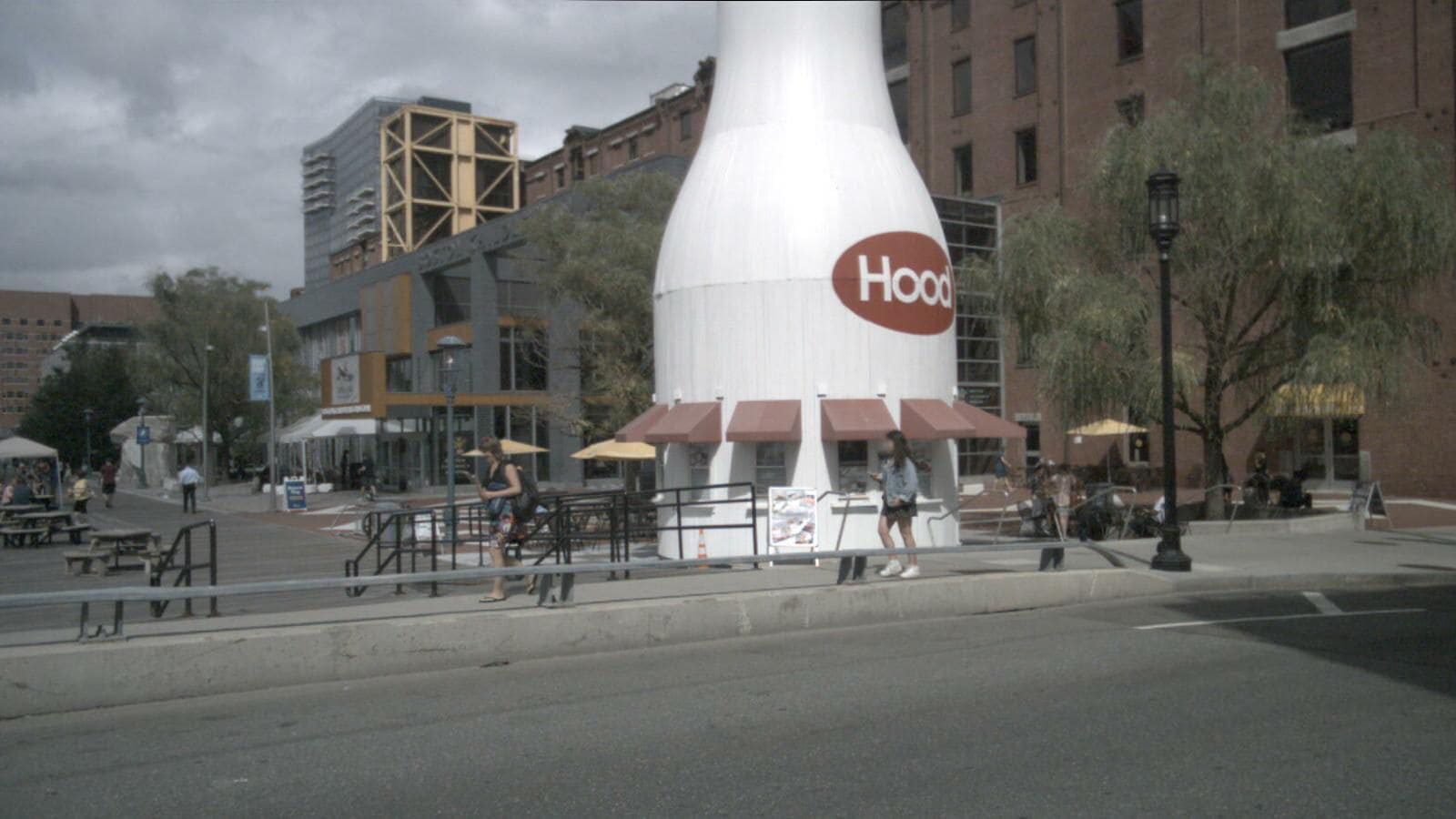}{6}\\
C. \includeimage{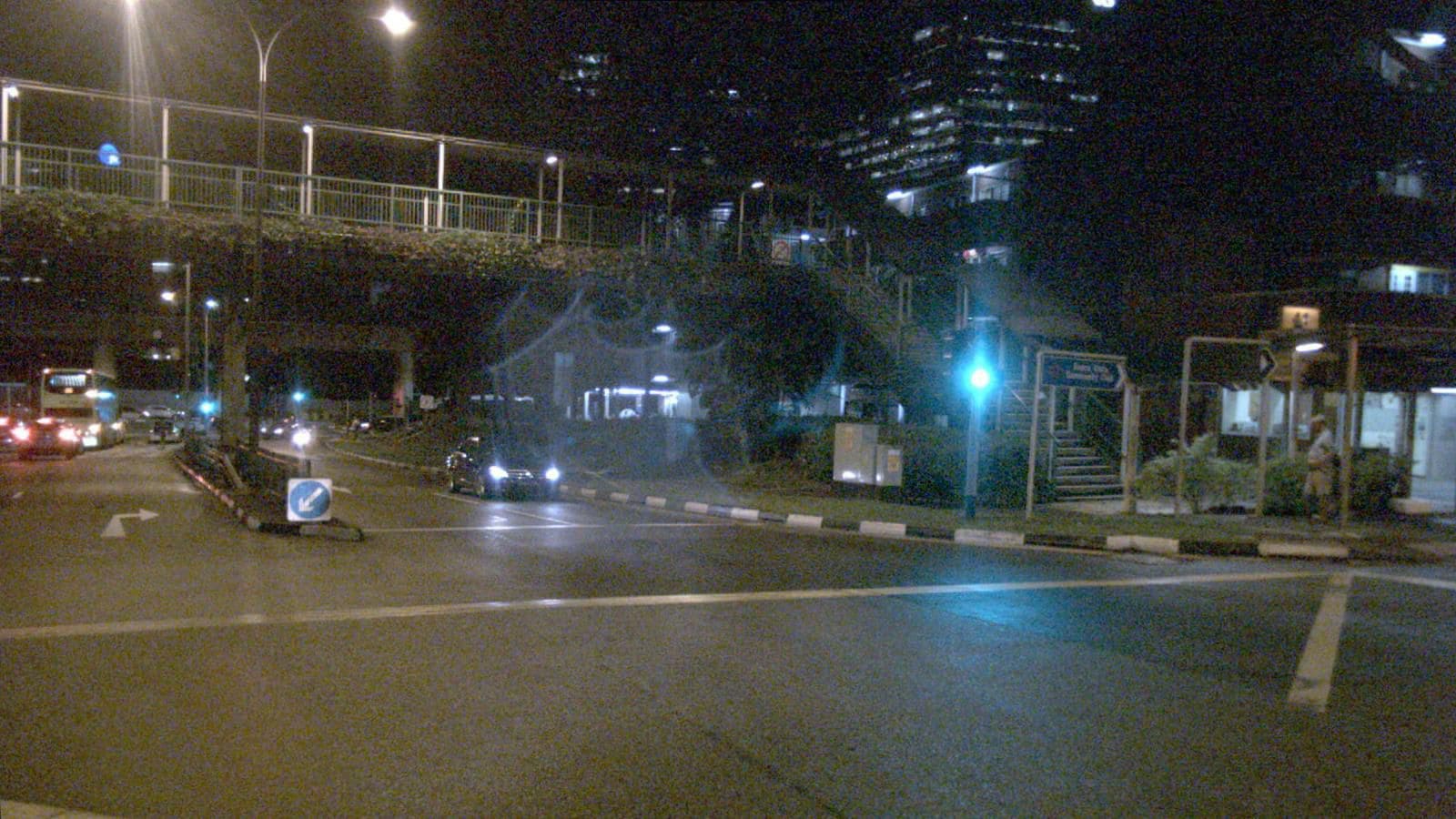}{6}\qquad
D. \includeimage{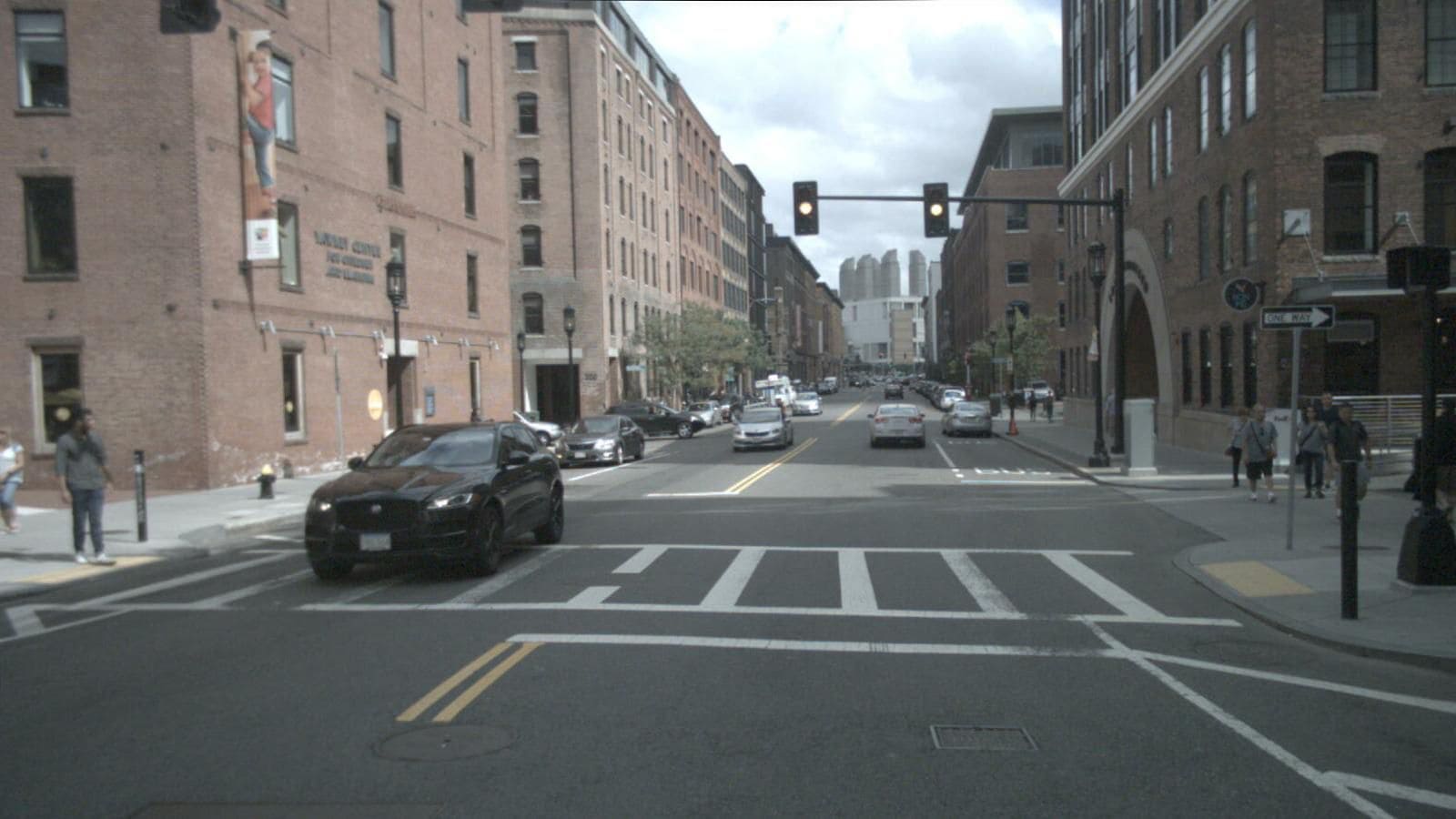}{6}
\end{answer}

\begin{question}{light_blue}
\includeimage{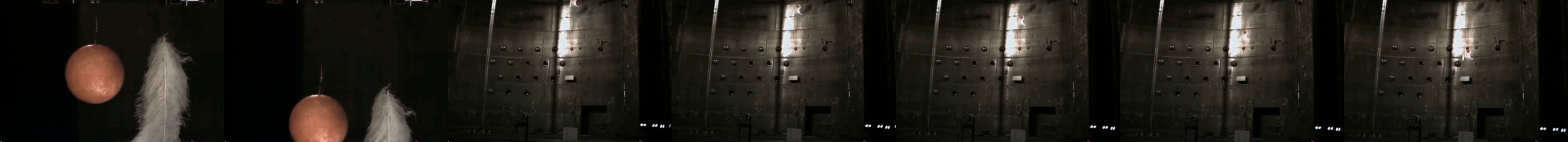}{5}
What is the relationship between the speeds of the two objects in the video after they begin to fall?
\end{question}
\begin{answer}{light_blue}
\red{A. Both objects have the same speed.}\\  
B. The egg consistently falls faster than the feather. \\ 
C. The feather consistently falls faster than the egg.\\  
D. Initially, the egg falls faster, but the feather eventually surpasses it.
\end{answer}
\end{mycase}
\vspace{-2mm}
\captionof{figure}{Three examples for relationships motion, categorized by the following ability types: static, static and dynamic.}
\vspace{-3mm}
\label{fig:example_5}
\end{table*}
\begin{table*}[th!]
\fontsize{9.0pt}{\baselineskip}\selectfont
\linespread{0.9}\selectfont
\begin{mycase}{light_blue}
\begin{question}{light_blue}
\includeimage{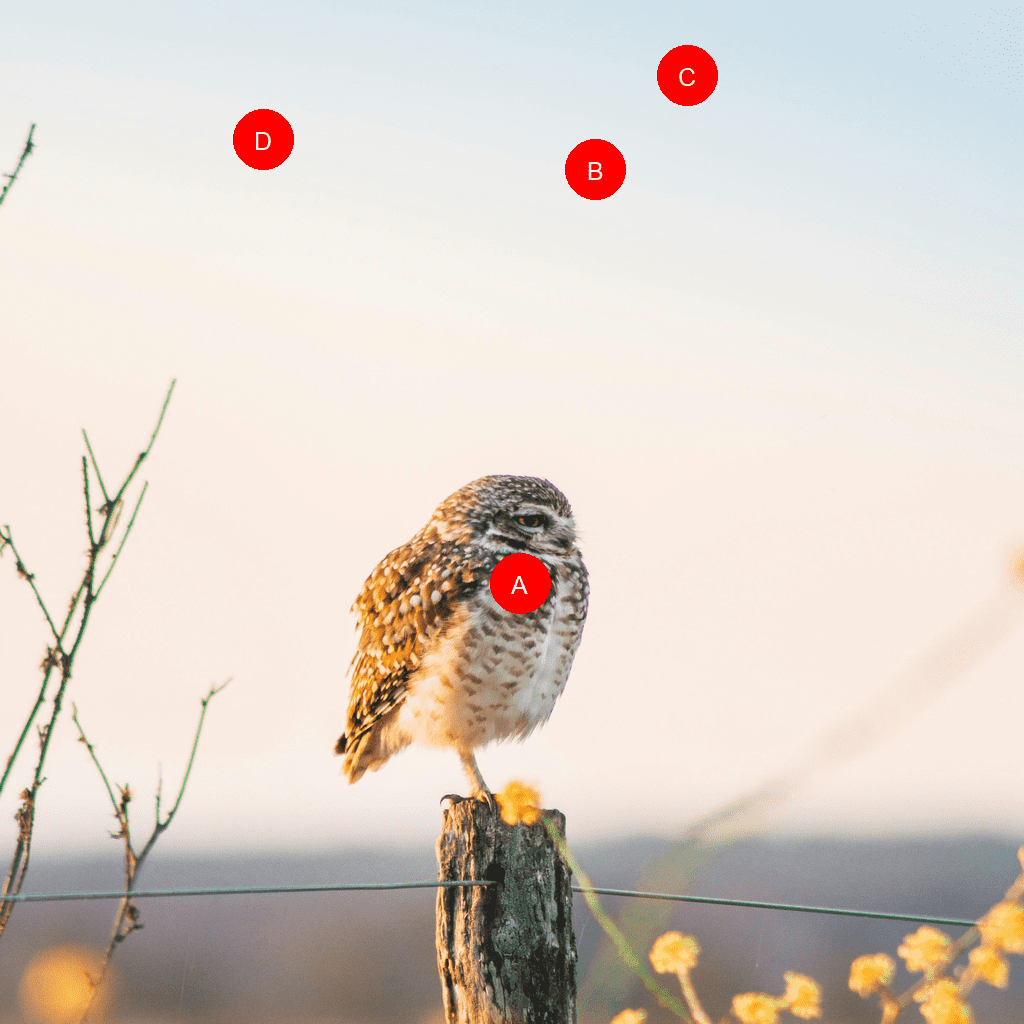}{7}Determine which point is nearest to the camera:
\end{question}
\begin{answer}{light_blue}
\red{A. Point A is nearest}\quad B. Point B is nearest\quad C. Point C is nearest\quad D. Point D is nearest 
\end{answer}
\end{mycase}
\vspace{-2mm}
\captionof{figure}{An example for relationships depth. Ability Type is static.}
\vspace{-3mm}
\label{fig:example_6}
\end{table*}
\begin{table*}[th!]
\fontsize{9.0pt}{\baselineskip}\selectfont
\linespread{0.9}\selectfont
\begin{mycase}{light_blue}
\begin{question}{light_blue}
\includeimage{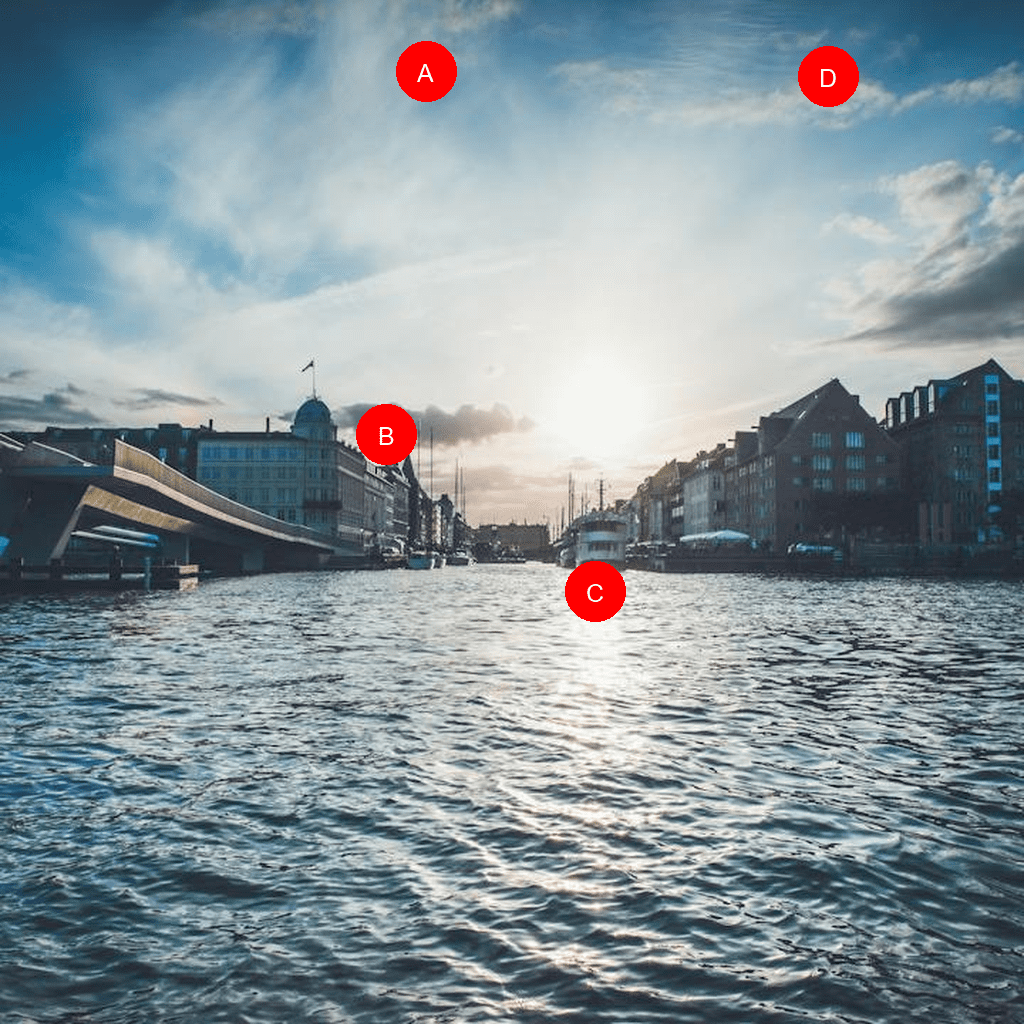}{7}Find the point that is closest to the river: 
\end{question}
\begin{answer}{light_blue}
A. Point A is the closest\quad B. Point B is the closest\quad
\red{C. Point C is the closest}\quad D. Point D is the closest
\end{answer}
\begin{question}{light_blue}
\includeimage{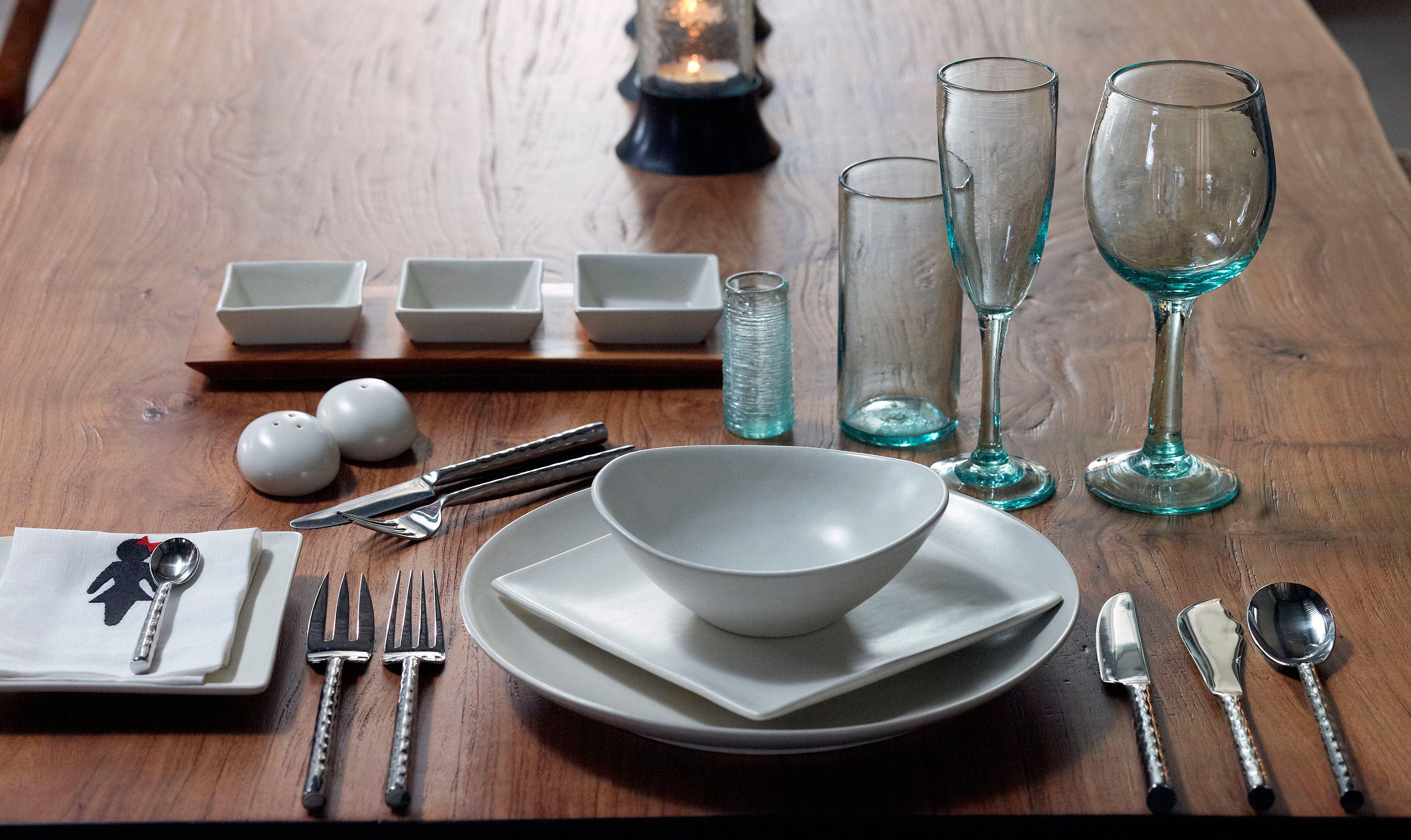}{7}What value is the distance between each of the three small bowls placed side by side closest to?
\end{question}
\begin{answer}{light_blue}
A.3mm\quad \red{B.3cm}\quad C.10cm\quad  D.20cm
\end{answer}
\begin{question}{light_blue}
\includeimage{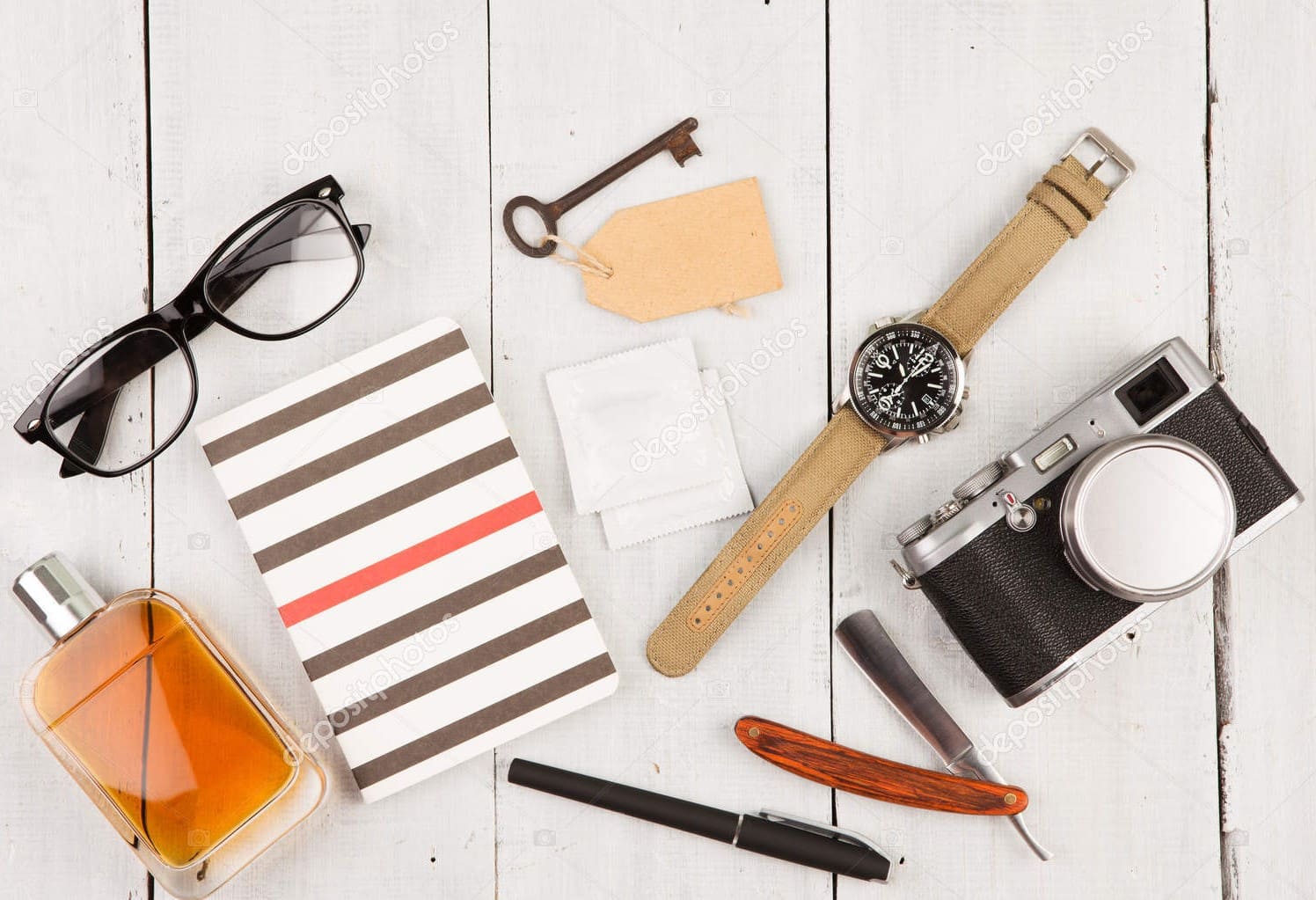}{7}What is farthest from the key in the picture?
\end{question}
\begin{answer}{light_blue}
\red{A. Pen}\quad   B. Camera\quad  C. Watch\quad  D. Glasses
\end{answer}
\end{mycase}
\vspace{-2mm}
\captionof{figure}{Three examples for relationships distance. Ability Type is static, static and dynamic.}
\vspace{-3mm}
\label{fig:example_7}
\end{table*}
\begin{table*}[th!]
\fontsize{9.0pt}{\baselineskip}\selectfont
\linespread{0.9}\selectfont
\begin{mycase}{light_blue}
\begin{question}{light_blue}
\includeimage{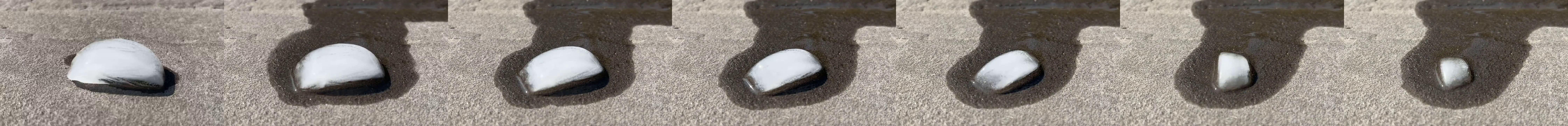}{5}
What is happening to the size of the ice cube in the video?
\end{question}
\begin{answer}{light_blue}
A. Increasing\quad \red{B. Decreasing}\quad C. No change\quad D. Increasing first and then decreasing
\end{answer}
\begin{question}{light_blue}
\includeimage{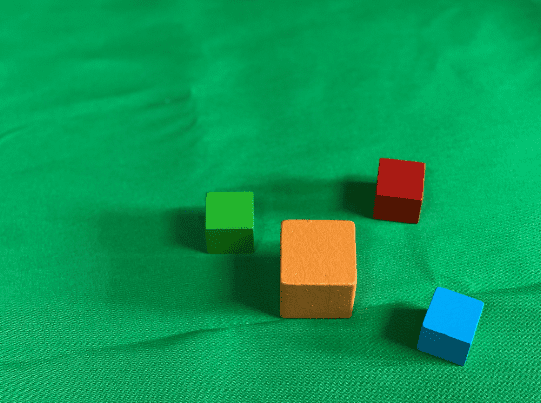}{7}
Which object is the biggest in volume?
\end{question}
\begin{answer}{light_blue}
A. red cube\qquad B. purple cube\qquad C. green cube\qquad \red{D. orange cube}
\end{answer}
\end{mycase}
\vspace{-2mm}
\captionof{figure}{Two examples for relationships size. Ability Type is dynamic and static.}
\vspace{-3mm}
\label{fig:example_8}
\end{table*}
\begin{table*}[th!]
\fontsize{9.0pt}{\baselineskip}\selectfont
\linespread{0.9}\selectfont
\begin{mycase}{light_blue}
\begin{question}{light_blue}
\includeimage{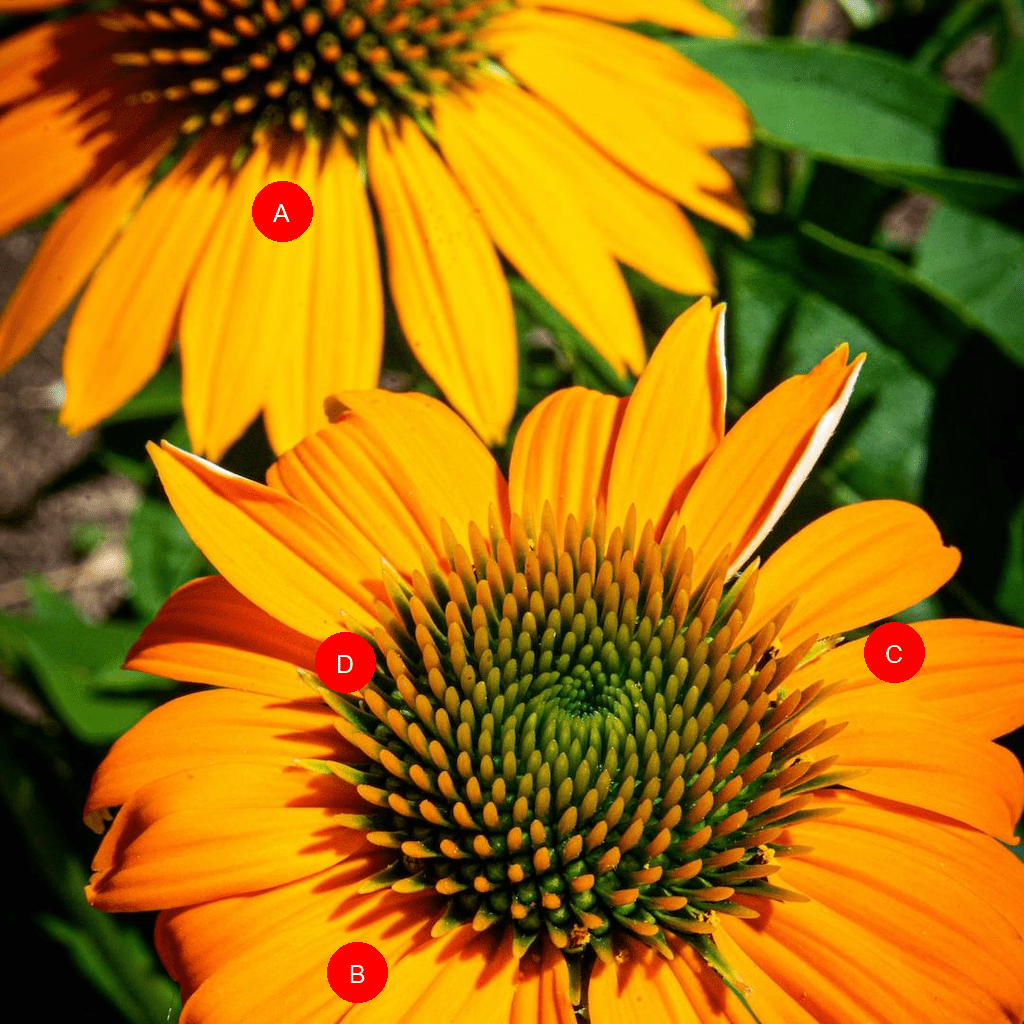}{7}Among these points, which one is not on the same sunflower as the other 3 points?
\end{question}
\begin{answer}{light_blue}
\red{A. Point A}\quad B. Point B\quad C. Point C\quad D.  Point D
\end{answer}
\begin{question}{light_blue}
\includeimage{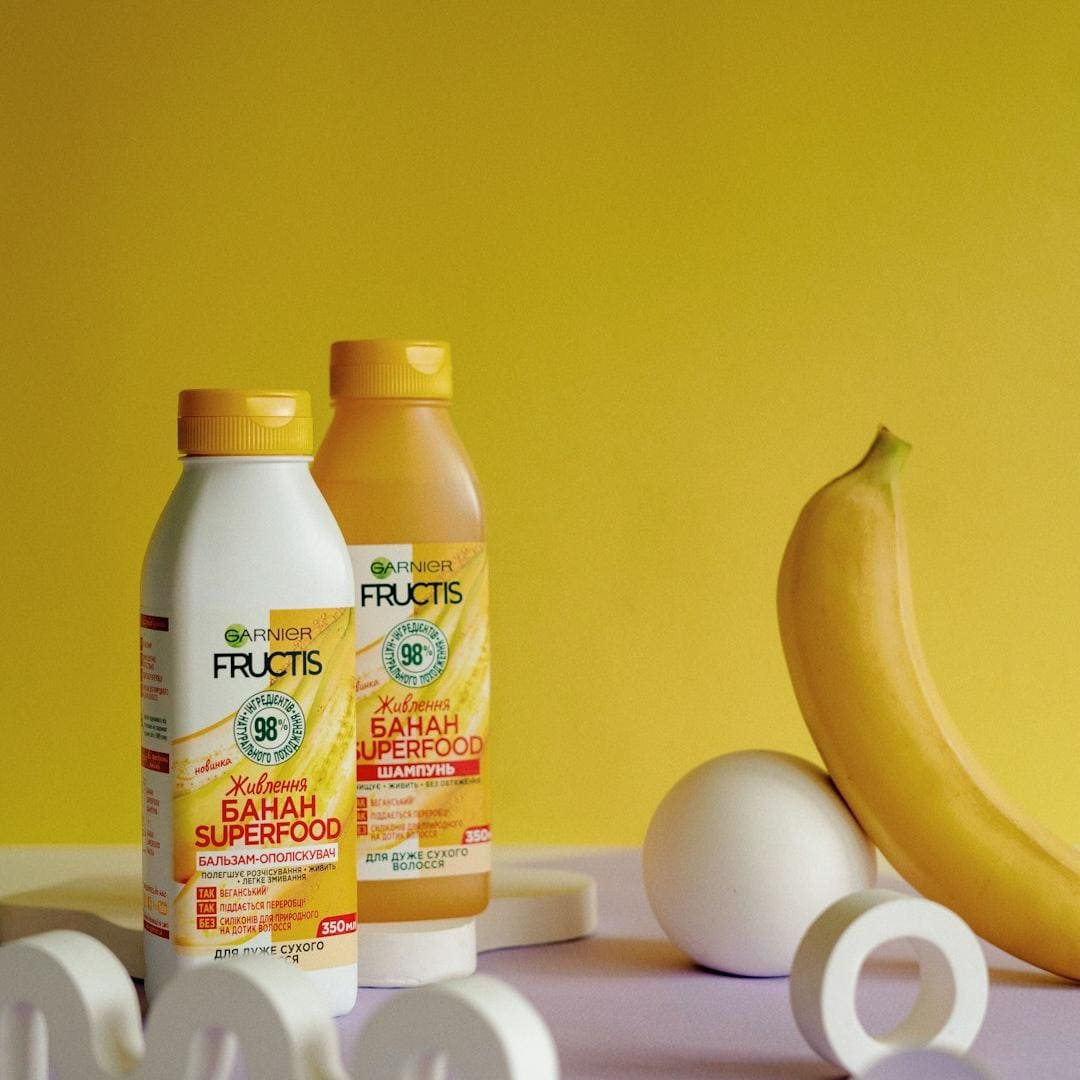}{7}What is the object above the egg?
\end{question}
\begin{answer}{light_blue}
A. Milk\quad B. Yogurt\quad \red{C. Banana}\quad  D. Egg
\end{answer}
\begin{question}{light_blue}
\includeimage{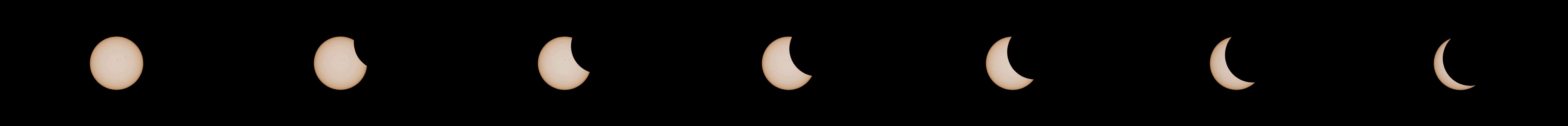}{5}
How does the moon pass through the sun in the video?
\end{question}
\begin{answer}{light_blue}
A. From the lower left corner to the upper right corner of the picture

B. From the upper left corner to the lower right corner of the picture

C. From the lower right corner to the upper left corner of the picture

\red{D. From the upper right corner to the lower left corner of the picture}
\end{answer}
\begin{question}{light_blue}
Which point with the option in the image\includeimage{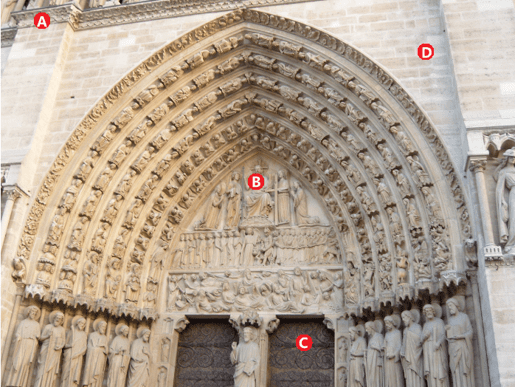}{5} is corresponds to the reference point P in the\includeimage{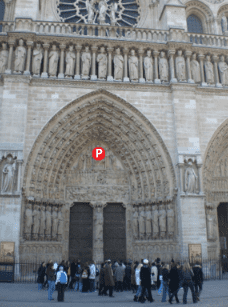}{5}
How does the moon pass through the sun in the video?
\end{question}
\begin{answer}{light_blue}
A. Point A\qquad
B. Point B\qquad
C. Point C\qquad
D. Point D
\end{answer}
\end{mycase}
\vspace{-2mm}
\captionof{figure}{Four examples for relationships location. Ability Type is static, static, dynamic and static.}
\vspace{-3mm}
\label{fig:example_9}
\end{table*}

\subsection{Physical Scene Understanding Sub-task}
\begin{table*}[th!]
\fontsize{9.0pt}{\baselineskip}\selectfont
\linespread{0.9}\selectfont
\begin{mycase}{light_green}
\begin{question}{light_green}
\includeimage{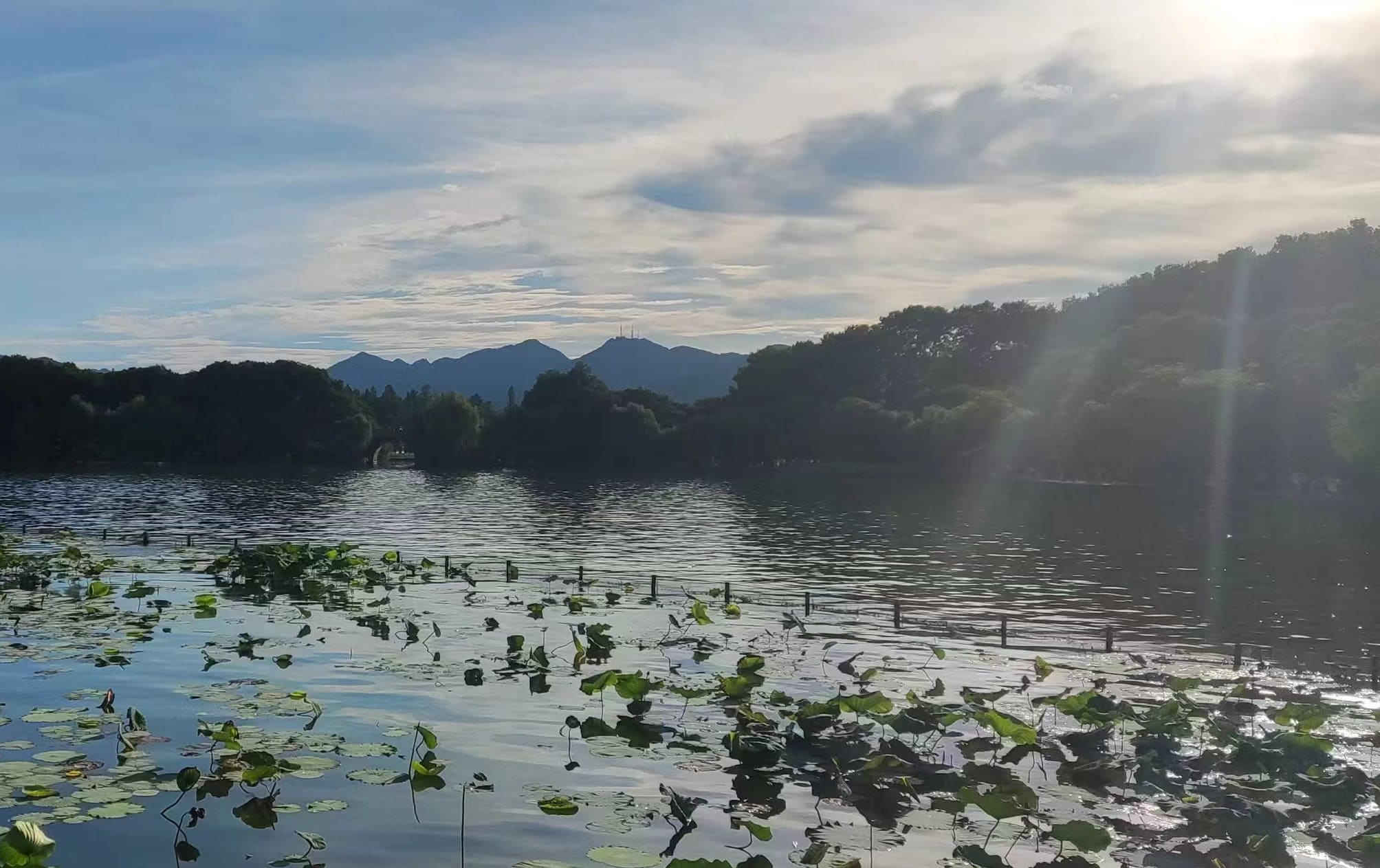}{6} Where is the light source in the picture?
\end{question}
\begin{answer}{light_green}
\red{A. In the clouds on the upper right of the screen}\\
B. In the lower left corner of the screen\\
C. A little above the center of the screen\\
D. In the exact center of the screen
\end{answer}
\begin{question}{light_green}
\includeimage{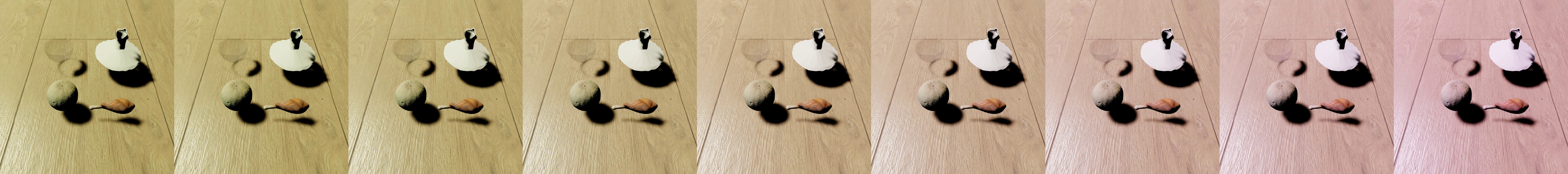}{7}
Reflecting on the events in the video, which of the following alterations to the light source is most likely to result in the phenomenon observed?
\end{question}
\begin{answer}{light_green}
A. The color of the light changes from yellow to pink\\
B. It's just that the light source is weaker and the light source position remains the same\\
C. Move parallel to the line between the drumstick and the ballet skirt\\
\red{D. It's just that the light source is stronger and the light source position remains the same}
\end{answer}
\begin{question}{light_green}
\includeimage{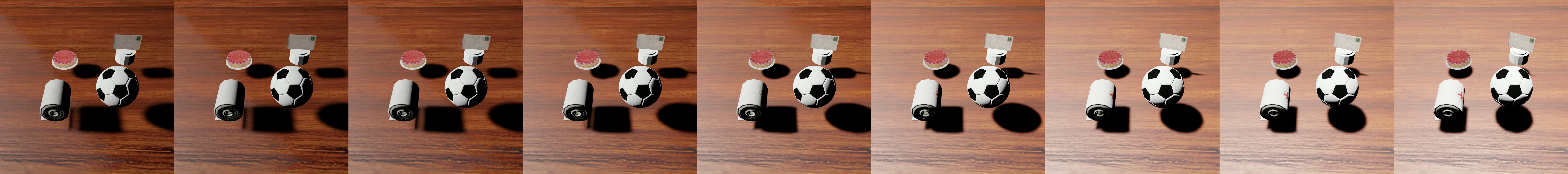}{7}
From the events in the video, which of the listed changes to the light source is most likely to have resulted in the observed phenomenon?
\end{question}
\begin{answer}{light_green}
\red{A. Move parallel to the line between the postcard and the cake}\\
B. The color of the light changes from orange to blue\\
C. The color of the light changes from lime yellow to green\\
D. The light source moves downward
\end{answer}
\begin{question}{light_green}
\includeimage{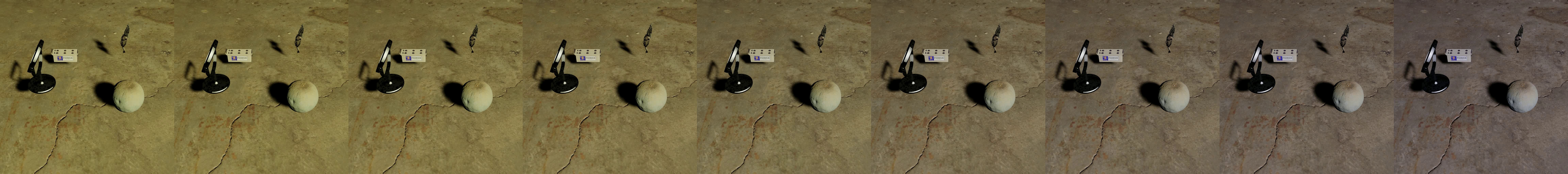}{7}
Taking into account the phenomena observed in the video, which of the following changes to the light source is most likely to have led to this result?
\end{question}
\begin{answer}{light_green}
A. It's just that the light source is stronger and the light source position remains the same\\
B. It's just that the light source is weaker and the light source position remains the same\\
C. The color of the light changes from red to purple\\
\red{D. The color of the light changes from yellow to blue}
\end{answer}
\end{mycase}
\vspace{-2mm}
\captionof{figure}{Four examples illustrating scene environmental lighting conditions. The corresponding ability types are perception, reasoning, reasoning, and reasoning.}
\vspace{-3mm}
\label{fig:example_10}
\end{table*}
\begin{table*}[th!]
\fontsize{9.0pt}{\baselineskip}\selectfont
\linespread{0.9}\selectfont
\begin{mycase}{light_green}
\begin{question}{light_green}
\includeimage{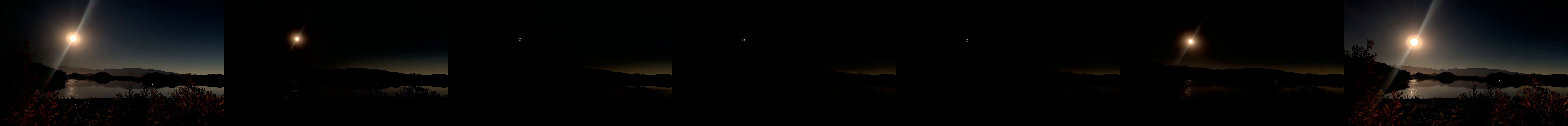}{5}
How does the brightness in the picture change?
\end{question}
\begin{answer}{light_green}
A. First it gets brighter, then it gets darker\\
\red{B. First it gets darker, then it gets brighter}\\
C. It keeps getting darker\\
D. t keeps changing
\end{answer}
\end{mycase}
\vspace{-2mm}
\captionof{figure}{Examples for scene environmental lighting conditions (Continued). Ability Type is judgement.}
\vspace{-3mm}
\label{fig:example_11}
\end{table*}

\begin{table*}[th!]
\fontsize{9.0pt}{\baselineskip}\selectfont
\linespread{0.9}\selectfont
\begin{mycase}{light_green}
\begin{question}{light_green}
\includeimage{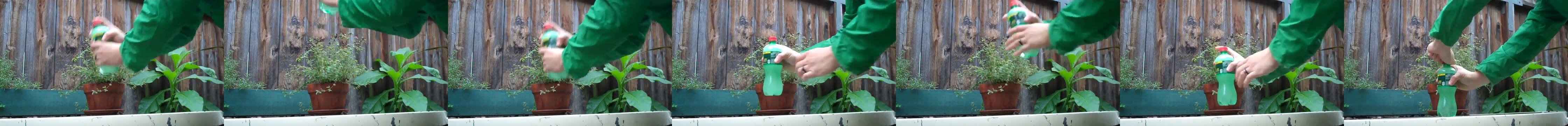}{5}
What happens to the gas pressure inside the bottle before open it?
\end{question}
\begin{answer}{light_green}
\red{A. Increases}\qquad B. Decreases\qquad C. Stays the same\qquad D. Varies randomly
\end{answer}
\end{mycase}
\vspace{-2mm}
\captionof{figure}{Example for scene environmental air conditions and the ability type of it is perception.}
\vspace{-3mm}
\label{fig:example_14}
\end{table*}
\begin{table*}[th!]
\fontsize{9.0pt}{\baselineskip}\selectfont
\linespread{0.9}\selectfont
\begin{mycase}{light_green}
\begin{question}{light_green}
\includeimage{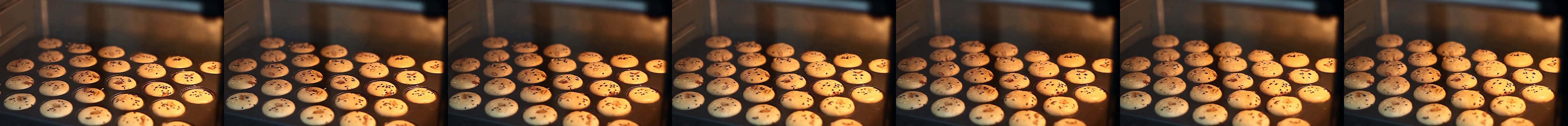}{5} What oven temperature range is typically used for the phenomenon shown in the video?
\end{question}
\begin{answer}{light_green}
A. Under 100°C\quad \red{B. Approximately 200℃}\quad C. Nearly 400℃\quad D. Over 600℃
\end{answer}
\begin{question}{light_green}
\includeimage{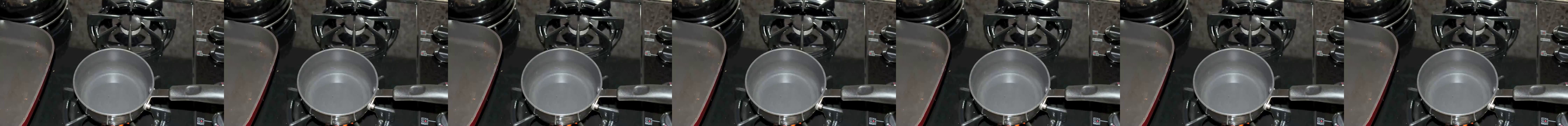}{5} What is the possible cause of the phenomenon in the video?
\end{question}
\begin{answer}{light_green}
A. Increased humidity\quad  
\red{B. Increased temperature}\quad 
C. Decreased temperature\quad 
D. Decreased humidity
\end{answer}
\end{mycase}
\vspace{-2mm}
\captionof{figure}{Two examples for scene temperature conditions and the ability types of them are all perception.}
\vspace{-3mm}
\label{fig:example_12}
\end{table*}
\begin{table*}[th!]
\fontsize{9.0pt}{\baselineskip}\selectfont
\linespread{0.9}\selectfont
\begin{mycase}{light_green}
\begin{question}{light_green}
\includeimage{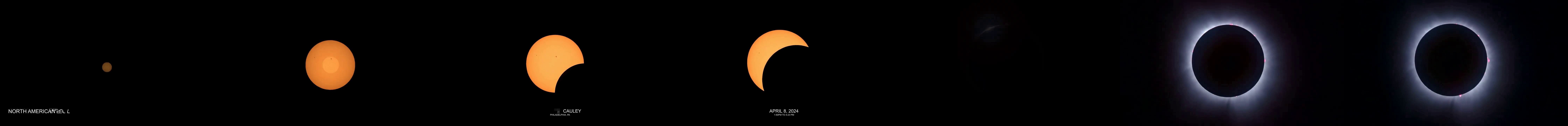}{5}
At the beginning of the video, how does the camera's focal length change? 
\end{question}
\begin{answer}{light_green}
A. Focus length remains unchanged\\
\red{B. Focus length increases}\\
C. Focus length decreases\\
D. Unknown
\end{answer}
\begin{question}{light_green}
\includeimage{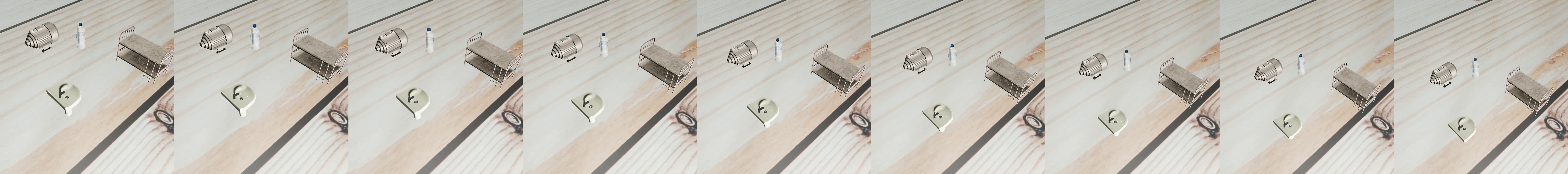}{7}
Based on the phenomenon in the video, which of the following camera changes can produce the effect in the video?
\end{question}
\begin{answer}{light_green}
A. Move parallel to the line between the motor and the wash basin\\
B. Move parallel to the line between the bunk bed and the wash basin\\
\red{C. The camera moves upward or downward} \\
D. The camera rotates along the horizontal axis (left or right)
\end{answer}
\begin{question}{light_green}
\includeimage{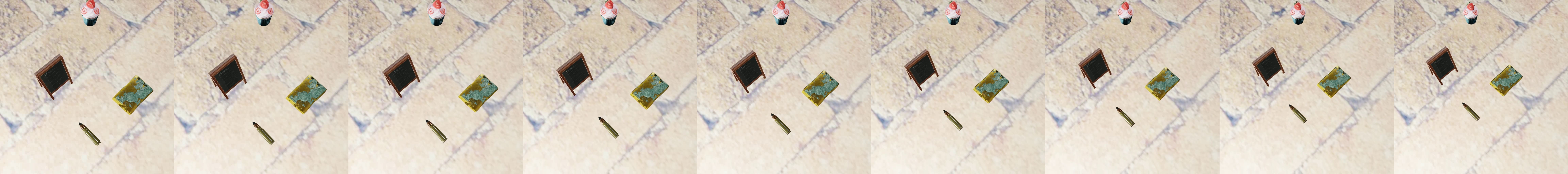}{7}
From the video, which of these camera changes could be responsible for the depicted phenomenon?
\end{question}
\begin{answer}{light_green}
\red{A. The camera is farther away from the objects}\\
B. Move parallel to the line between the cupcake and the sponge\\
C. The camera is closer to the objects\\
D. Move parallel to the line between the cupcake and the blackboard
\end{answer}
\begin{question}{light_green}
\includeimage{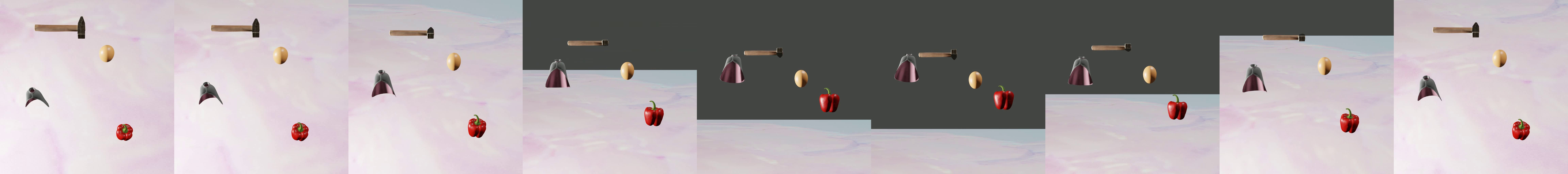}{7}
Based on the phenomenon in the video, which of the following camera changes can produce the effect in the video?
\end{question}
\begin{answer}{light_green}
\red{A. The camera rotates along the vertical axis (upside or downside).}\\
B. The camera is closer to the objects\\
C. The camera rotates along the horizontal axis (left or right).\\
D. The camera moves upward or downward
\end{answer}
\end{mycase}
\vspace{-2mm}
\captionof{figure}{Four examples illustrating scene viewpoint conditions. The corresponding ability types are perception, reasoning, reasoning and reasoning.}
\vspace{-3mm}
\label{fig:example_13}
\end{table*}

\subsection{Physics-based Dynamics Sub-task}

\begin{table*}[th!]
\fontsize{9.0pt}{\baselineskip}\selectfont
\linespread{0.9}\selectfont
\begin{mycase}{light_yellow}
\begin{question}{light_yellow}
Select the option that shows the correct procedure to put on a pair of gloves.

image \#1: \includeimage{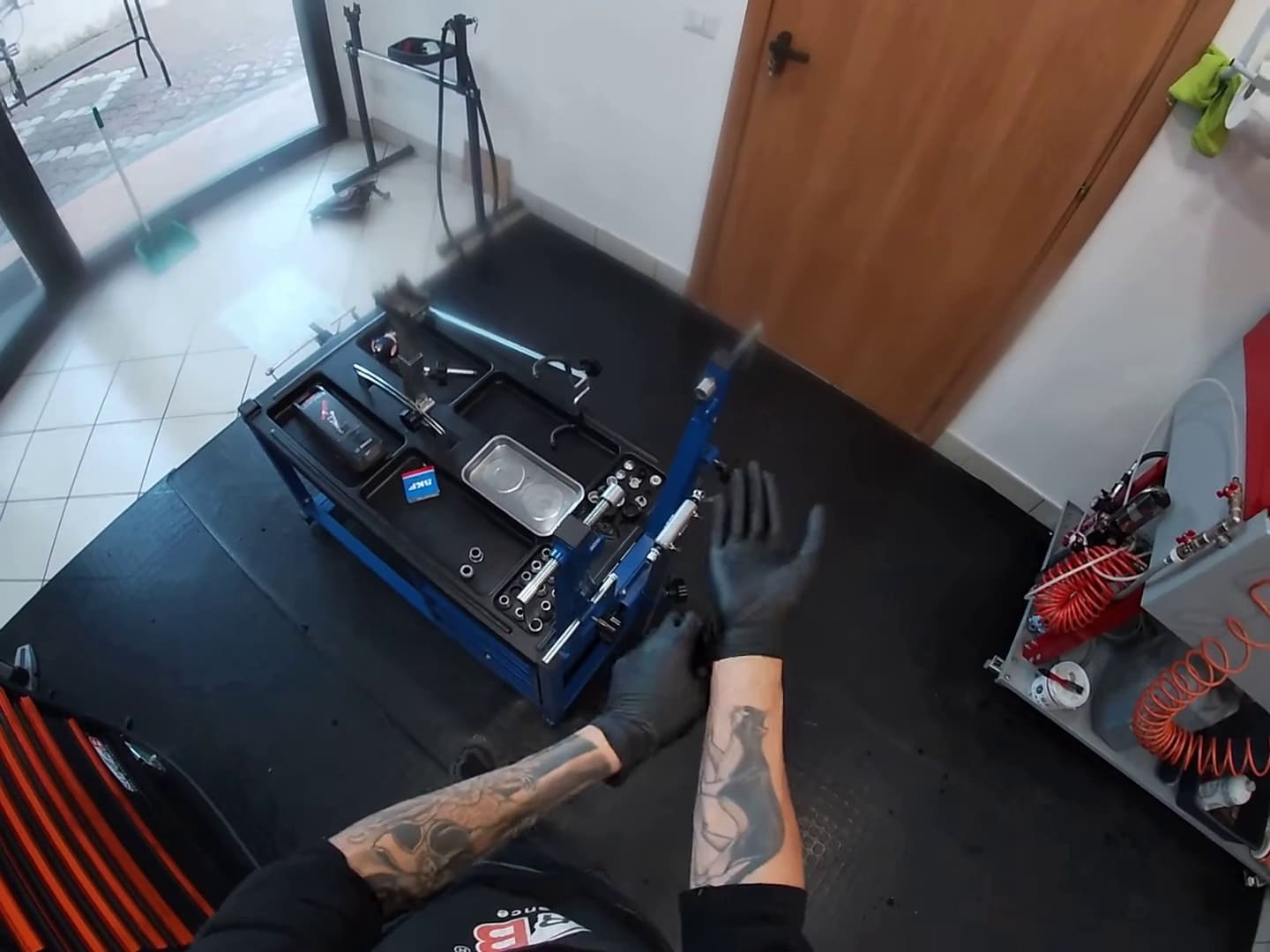}{6}
image \#2: \includeimage{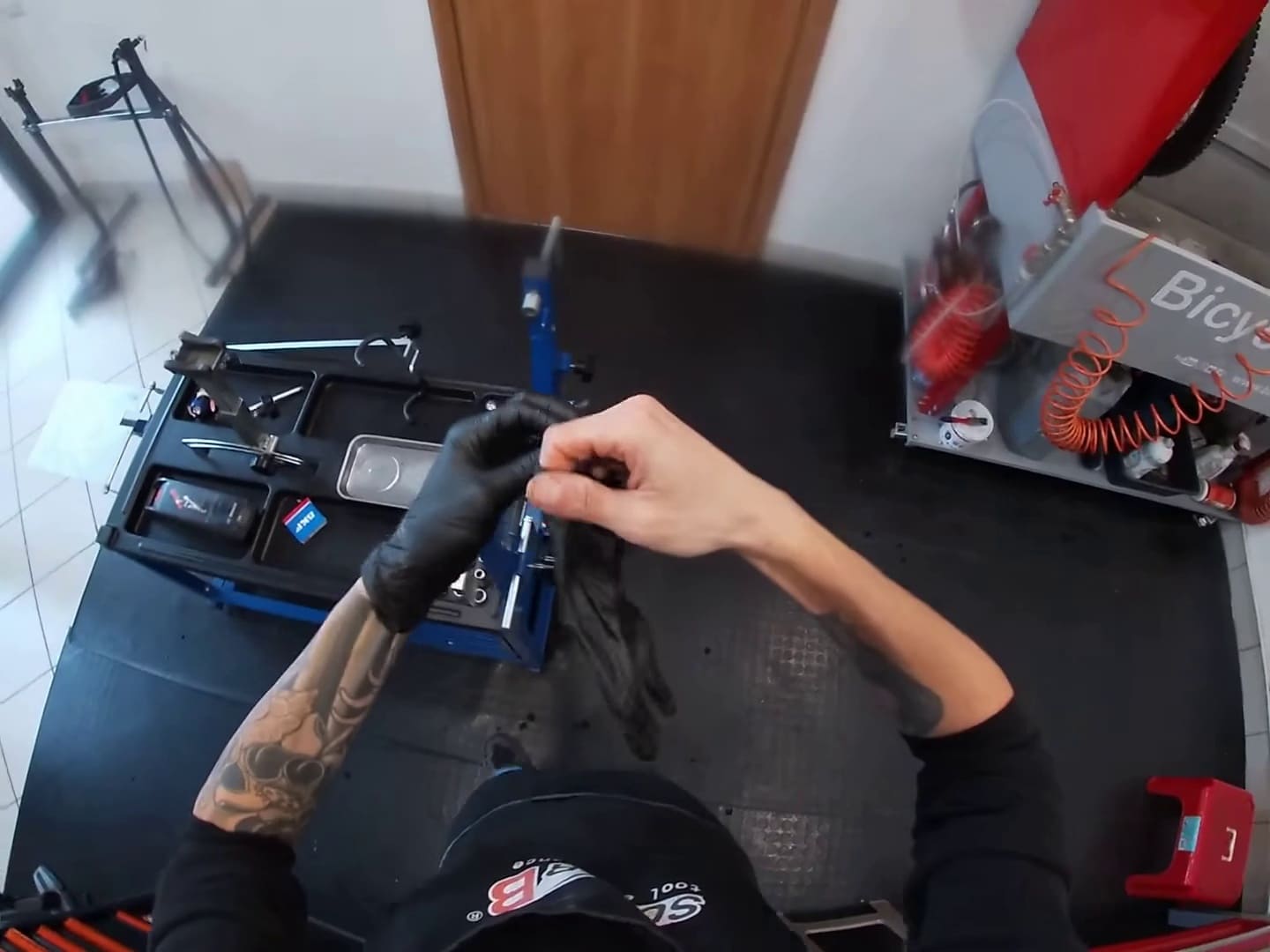}{6}
image \#3: \includeimage{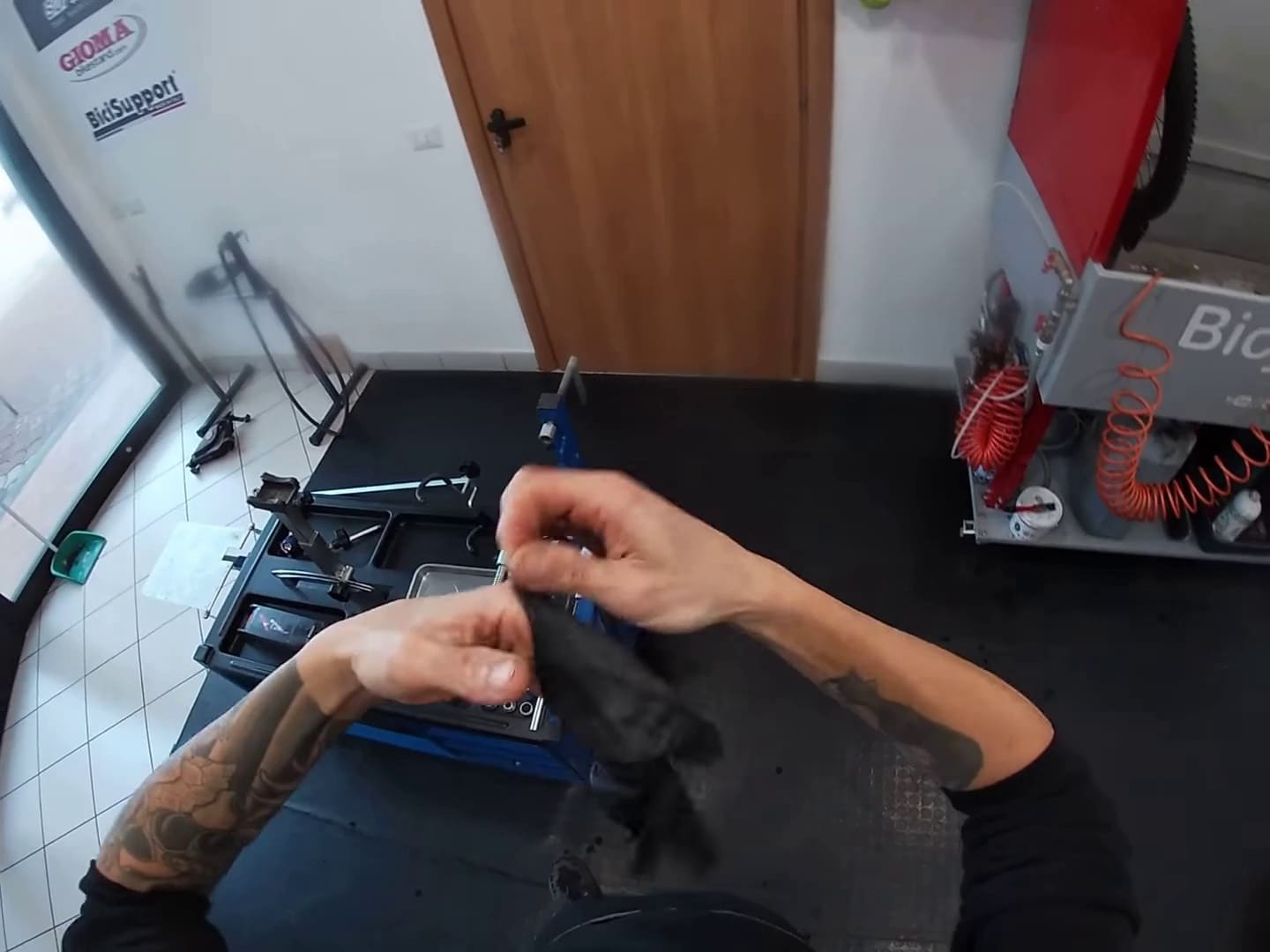}{6}
\end{question}
\begin{answer}{light_yellow}
A. 1 - 3 - 2\qquad
\red{B. 3 - 2 - 1}

C. 2 - 1 - 3\qquad
D. 1 - 2 - 3
\end{answer}

\begin{question}{light_yellow}
Which of the following options lists the steps in the correct sequence to put the carrot in the microwave?
image \#1: \includeimage{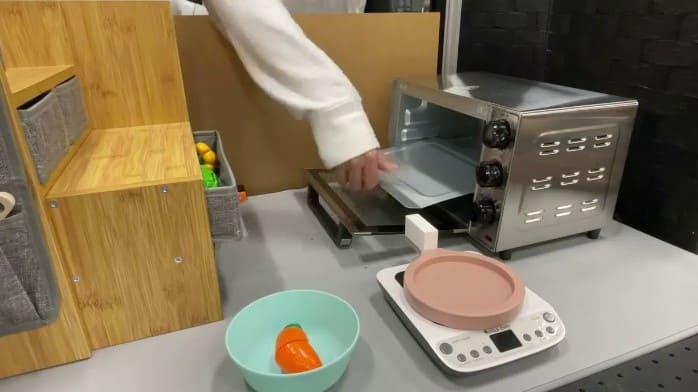}{6}
image \#2: \includeimage{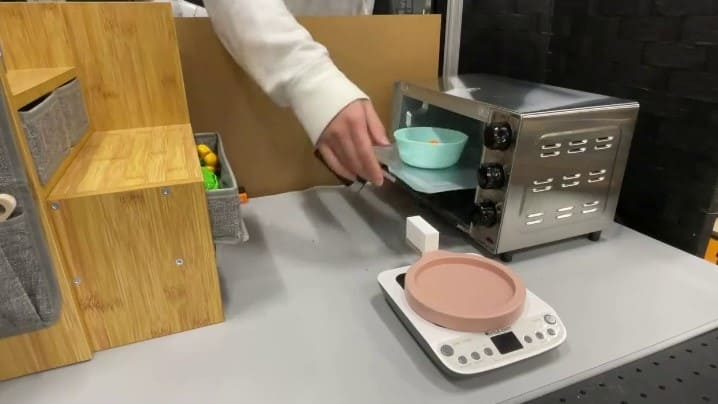}{6}

image \#3: \includeimage{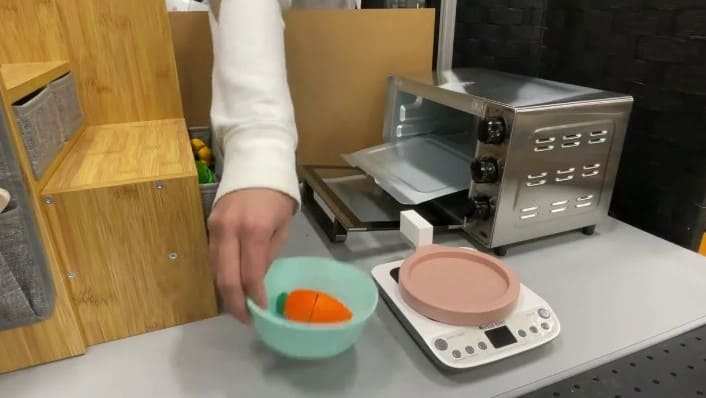}{6}
image \#4: \includeimage{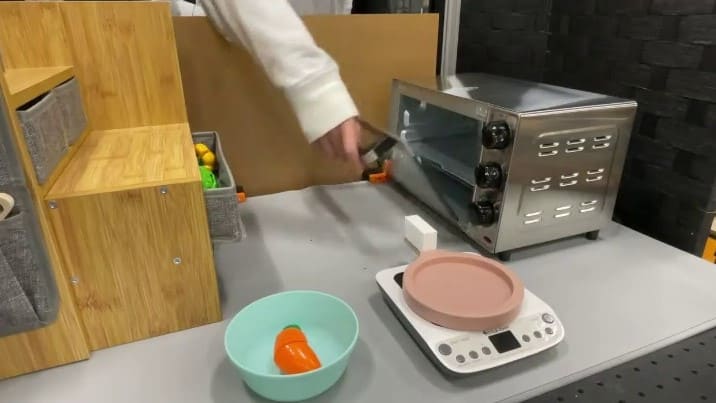}{6}
\end{question}
\begin{answer}{light_yellow}
A. 2 - 1 - 4 - 3\qquad
\red{B. 1 - 3 - 2 - 4}

C. 2 - 3 - 1 - 4\qquad
D. 2 - 4 - 3 - 1
\end{answer}
\begin{question}{light_yellow}
\includeimage{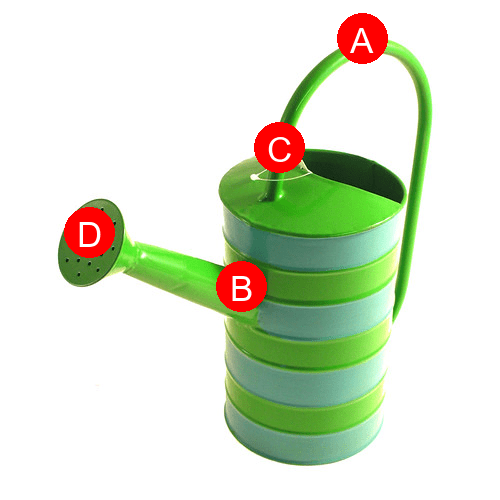}{7}To poke the watering can, which point is most suitable?
\end{question}
\begin{answer}{light_yellow}
\red{A.} \qquad B. \qquad C. \qquad D.
\end{answer}
\begin{question}{light_yellow}
\includeimage{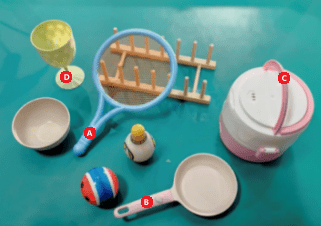}{7}In order to pick up the cup, which of the following color points has reasonable affordance?
\end{question}
\begin{answer}{light_yellow}
A. \qquad B. \qquad C. \qquad \red{D.}
\end{answer}
\begin{question}{light_yellow}
What operation is used to transform the object from Image \includeimage{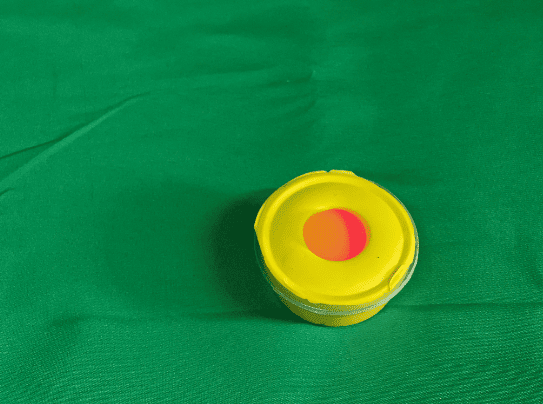}{7} to Image \includeimage{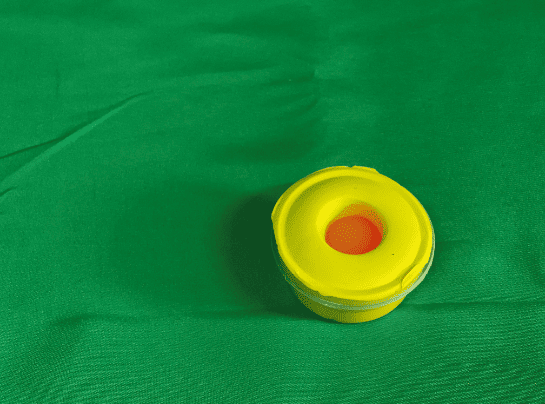}{7}?  
\end{question}
\begin{answer}{light_yellow}
A. Remove the small ball from the clay block.\\  
B. Add another ball to the clay block. \\ 
C. Air-dry the clay block.  \\
\red{D. Press the small ball into the clay block.}
\end{answer}
\end{mycase}
\vspace{-2mm}
\captionof{figure}{Five examples illustrating dynamics manipulation. The corresponding ability types are judgment, judgment, perception, perception and reasoning.}
\vspace{-3mm}
\label{fig:example_15}
\end{table*}
\begin{table*}[th!]
\fontsize{9.0pt}{\baselineskip}\selectfont
\linespread{0.9}\selectfont
\begin{mycase}{light_yellow}
\begin{question}{light_yellow}
\includeimage{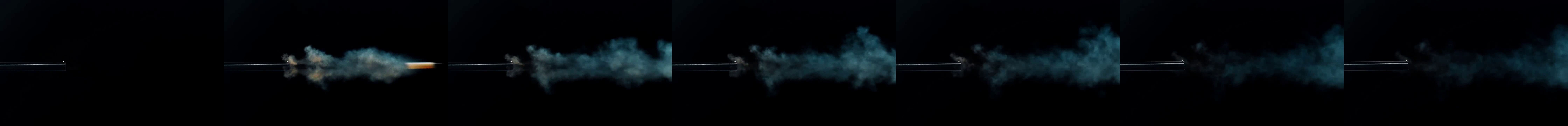}{5}
Following the content of the video, which option's corresponding picture will happen first?
\end{question}
\begin{answer}{light_yellow}
\red{A.} \includeimage{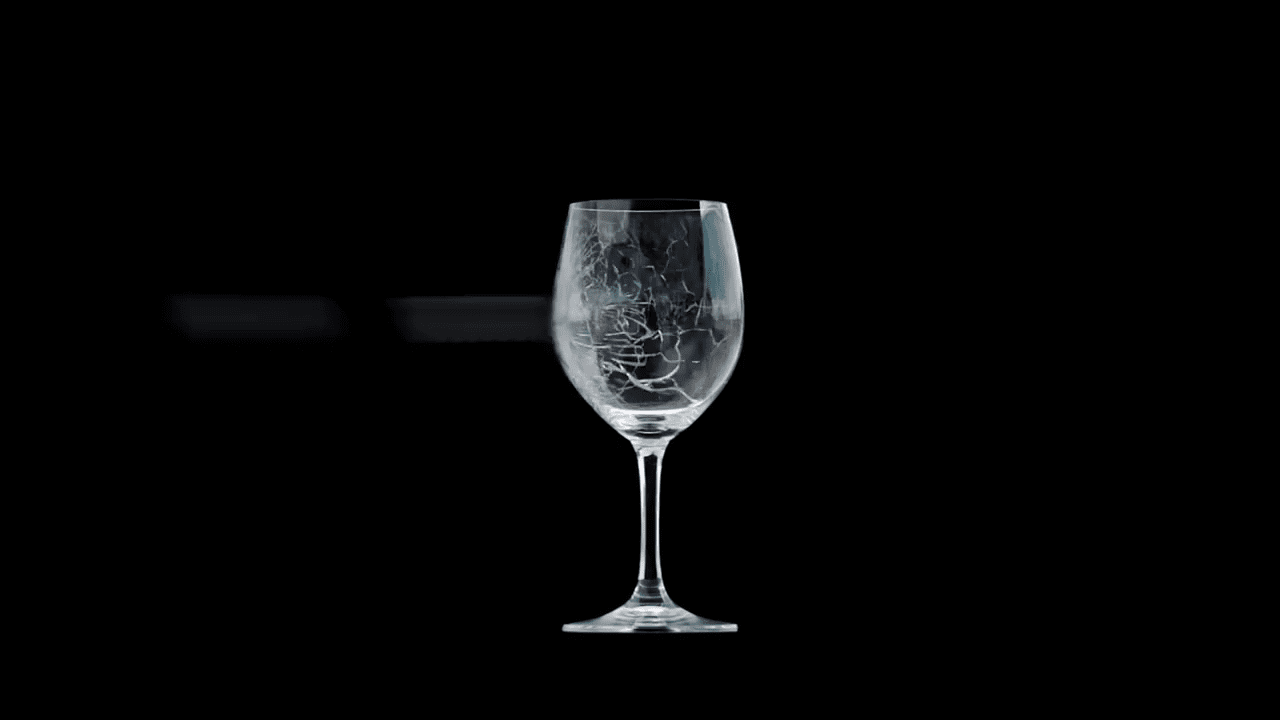}{5} 
B. \includeimage{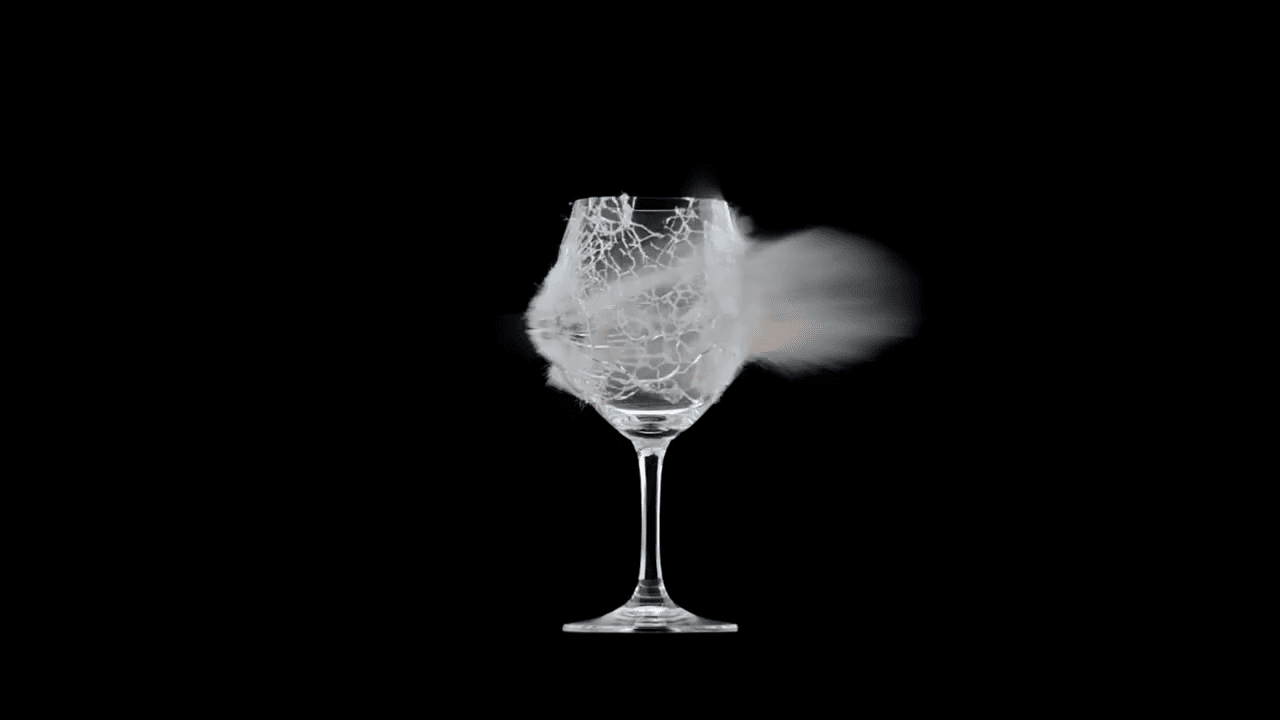}{5}
C. \includeimage{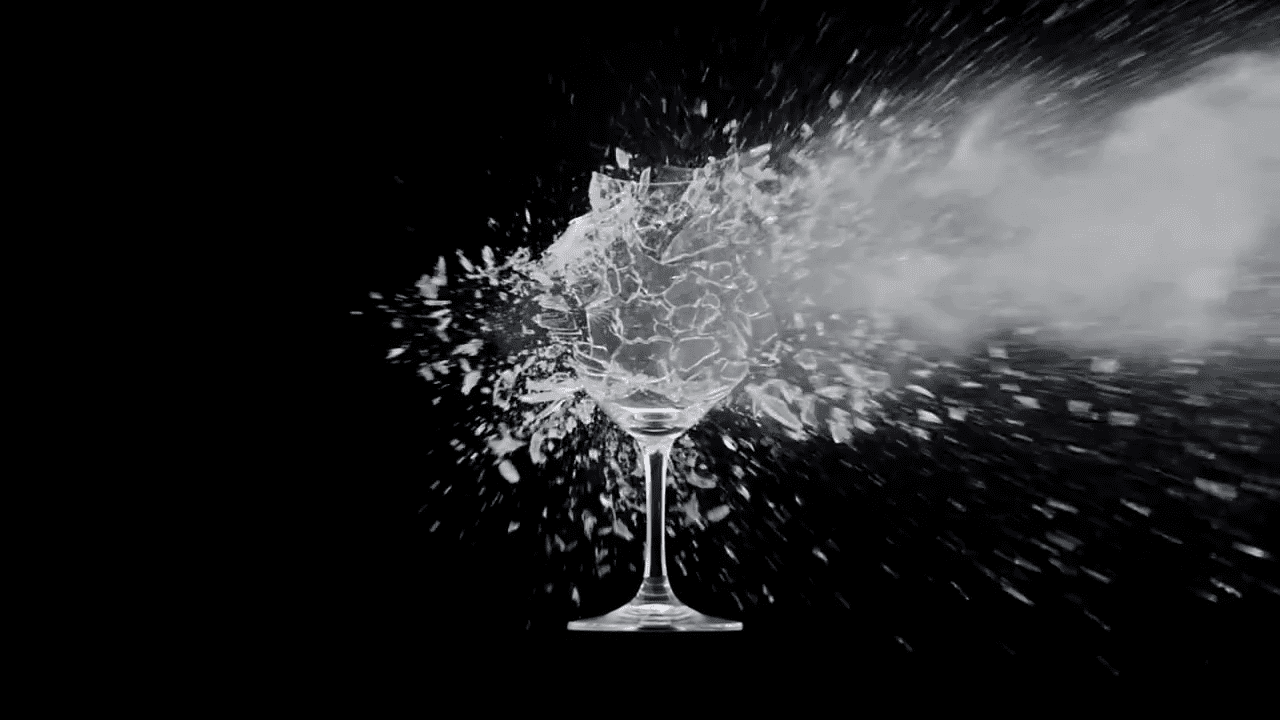}{5}
D. \includeimage{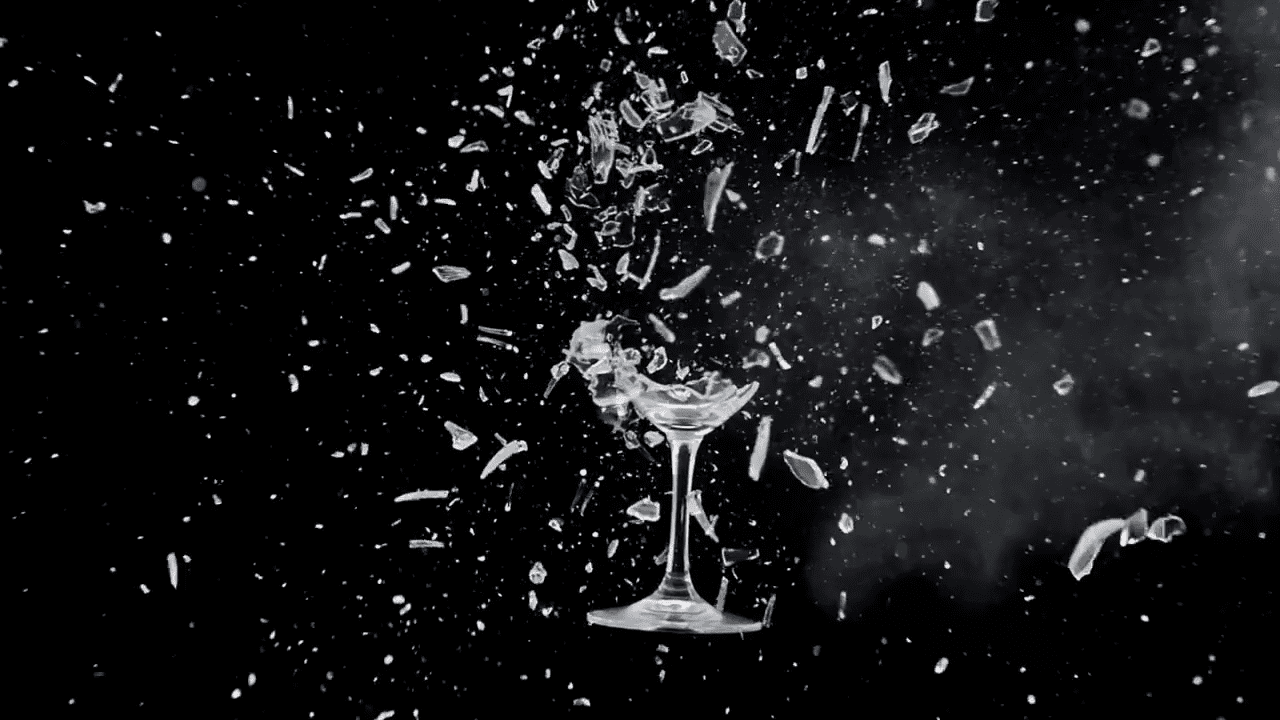}{5}
\end{answer}
\end{mycase}
\vspace{-2mm}
\captionof{figure}{Example for dynamics collision and the ability type of it is prediction.}
\vspace{-3mm}
\vspace{-3mm}
\label{fig:example_16}
\end{table*}

\begin{table*}[th!]
\fontsize{9.0pt}{\baselineskip}\selectfont
\linespread{0.9}\selectfont
\begin{mycase}{light_yellow}
\begin{question}{light_yellow}
\includeimage{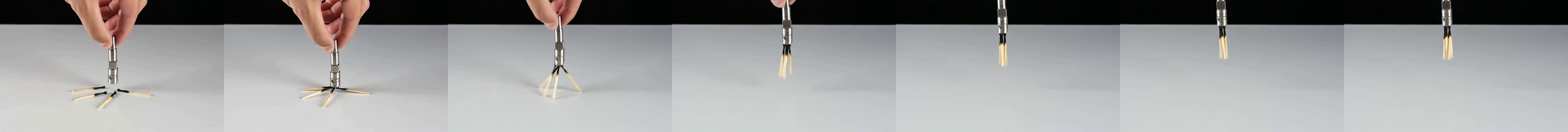}{5.3} Why are the phenomena happens in the video?
\end{question}
\begin{answer}{light_yellow}
A. Because they are lighter than air. \\
\red{B. Due to the presence of ferric oxide in the match heads.} \\
C. Because they are coated with a special glue.\\
D. Due to static electricity on the magnet.
\end{answer}
\end{mycase}
\vspace{-2mm}
\captionof{figure}{Example for dynamics chemistry and the ability type of it is perception.}
\vspace{-3mm}
\label{fig:example_20}
\end{table*}

\begin{table*}[th!]
\fontsize{9.0pt}{\baselineskip}\selectfont
\linespread{0.9}\selectfont
\begin{mycase}{light_yellow}
\begin{question}{light_yellow}
\includeimage{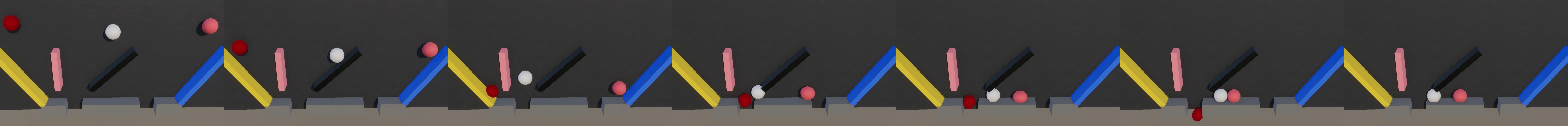}{5}
What will happen to the white ball?
\end{question}
\begin{answer}{light_yellow}
A. It will drop into the right pit.\quad
\red{B. It will not drop into the right pit.} \\
C. Its movement cannot be determined.\quad
D. It will drop into the left pit.
\end{answer}
\end{mycase}
\vspace{-2mm}
\captionof{figure}{Example for dynamics throwing and the ability type of it is prediction.}
\vspace{-3mm}
\vspace{-3mm}
\label{fig:example_17}
\end{table*}
\begin{table*}[th!]
\fontsize{9.0pt}{\baselineskip}\selectfont
\linespread{0.9}\selectfont
\begin{mycase}{light_yellow}
\begin{question}{light_yellow}
\includeimage{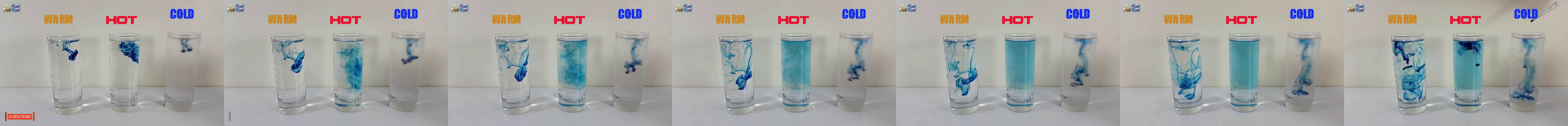}{5}
Which bottle contains the fastest diffusion rate?
\end{question}
\begin{answer}{light_yellow}
A. The one on the left\quad \red{B. The one in the middle}\quad C. The one on the right\quad D. Both are equally fast
\end{answer}

\begin{question}{light_yellow}
\includeimage{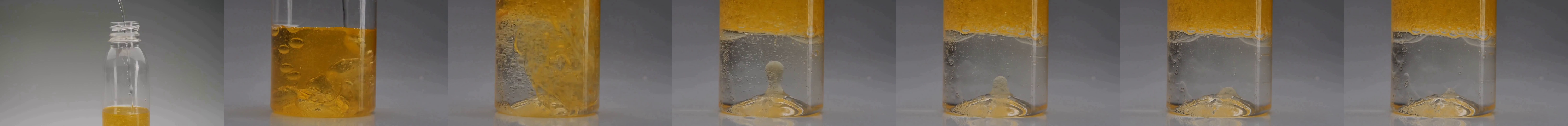}{5}
We already know that the liquid poured in is water, so what is the original yellow liquid?
\end{question}
\begin{answer}{light_yellow}
A. Orange juice\quad \red{B. Oil}\quad C. Beer\quad D. Urine
\end{answer}
\begin{question}{light_yellow}
\includeimage{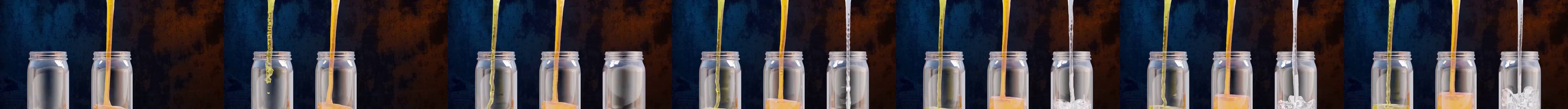}{4.5}
Which color object looks the most viscous in the video?
\end{question}
\begin{answer}{light_yellow}
A. Transparent liquid\quad B. Light yellow color liquid\quad \red{C. Thick yellow liquid}\quad D. Dark blue liquid
\end{answer}
\begin{question}{light_yellow}
\includeimage{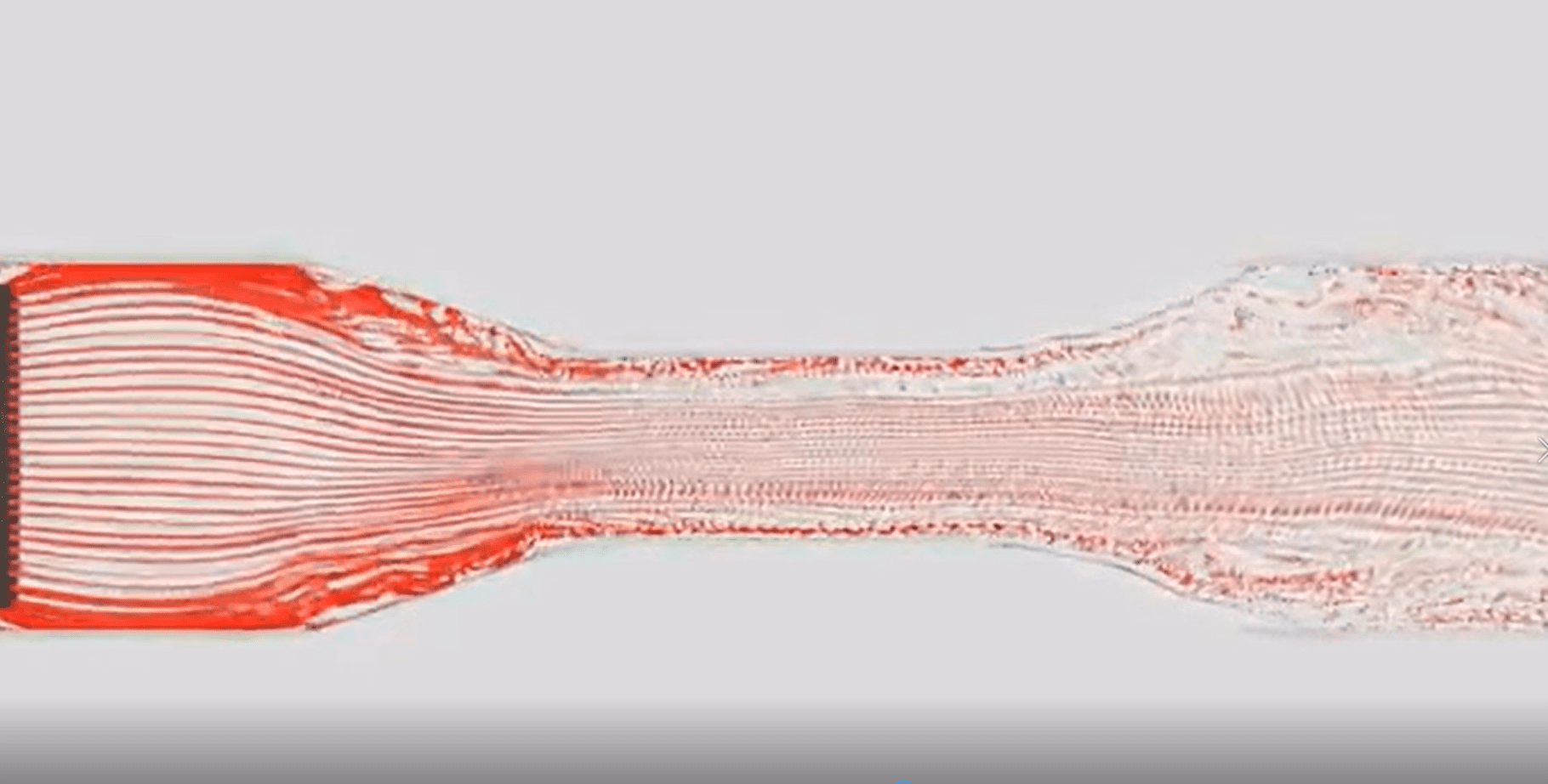}{7}
We already know that the red particles in the picture are liquid particles. In which area of the picture does the liquid flow fastest?
\end{question}
\begin{answer}{light_yellow}
A. All parts of the image are at the same speed.\\
B. the leftmost part of the picture\\
\red{C. the middle part of the picture}\\
D. the rightmost part of the picture
\end{answer}
\end{mycase}
\vspace{-2mm}
\captionof{figure}{Four examples illustrating dynamics fluid. The corresponding ability types are
perception, reasoning, perception and perception.}
\vspace{-3mm}
\label{fig:example_18}
\end{table*}
\begin{table*}[th!]
\fontsize{9.0pt}{\baselineskip}\selectfont
\linespread{0.9}\selectfont
\begin{mycase}{light_yellow}
\begin{question}{light_yellow}
Among the listed choices, which one outlines the proper sequence of events in candle burning?

image \#1: \includeimage{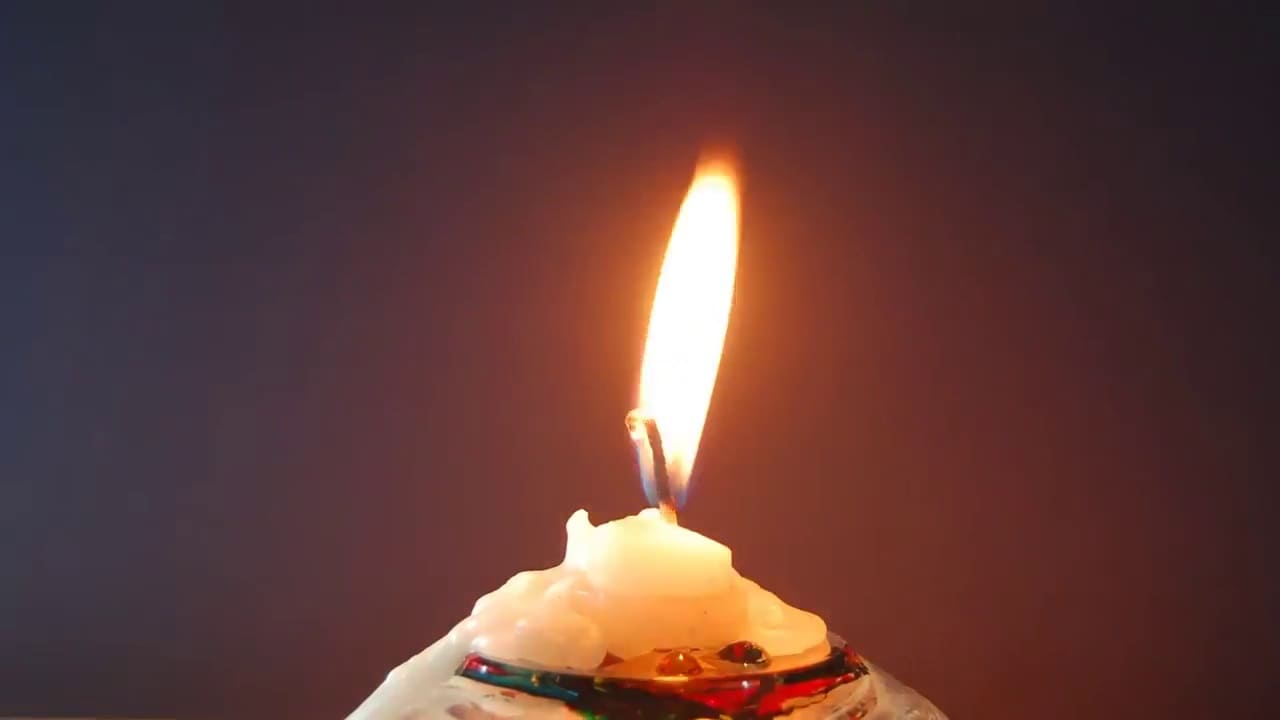}{5} 
image \#2: \includeimage{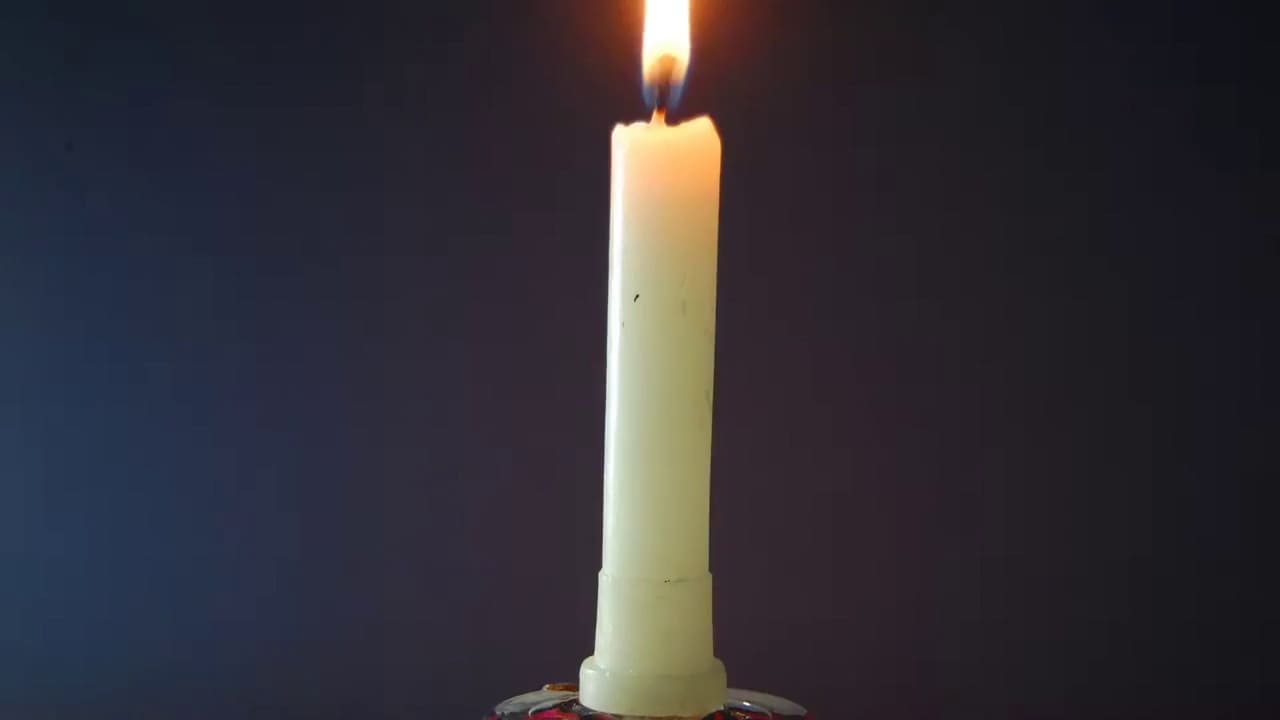}{5}
image \#3: \includeimage{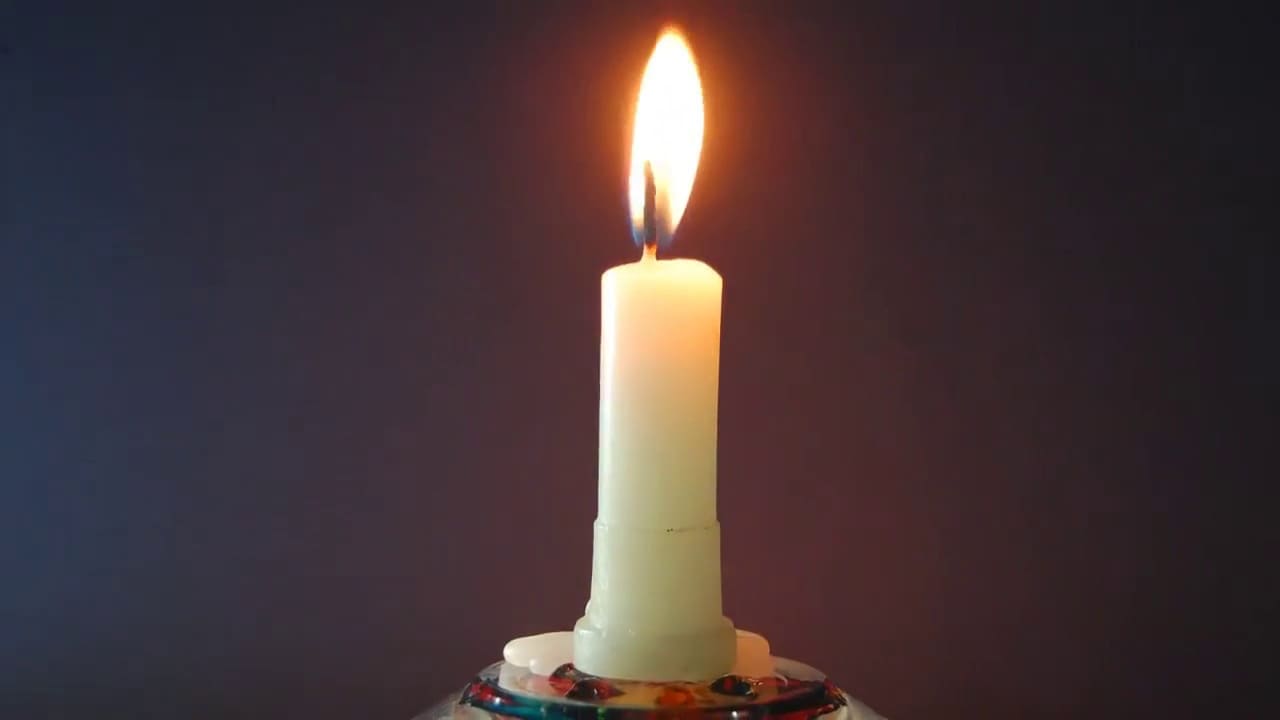}{5}
\end{question}
\begin{answer}{light_yellow}
A. 2 - 3 - 1\qquad B. 1 - 2 - 3\qquad C. 1 - 3 - 2\qquad  \red{D. 3 - 1 - 2}
\end{answer}
\begin{question}{light_yellow}
Which of the following options presents the correct order of occurrences in fruit rotting?

image \#1: \includeimage{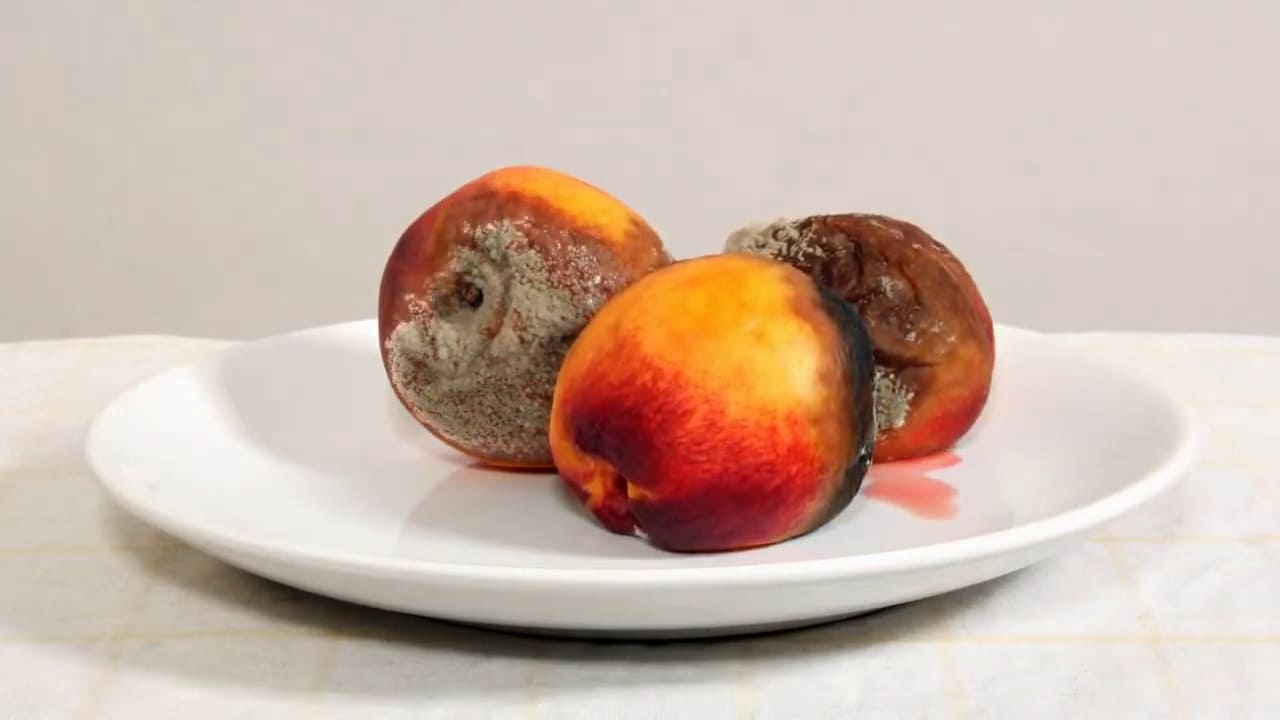}{5} 
image \#2: \includeimage{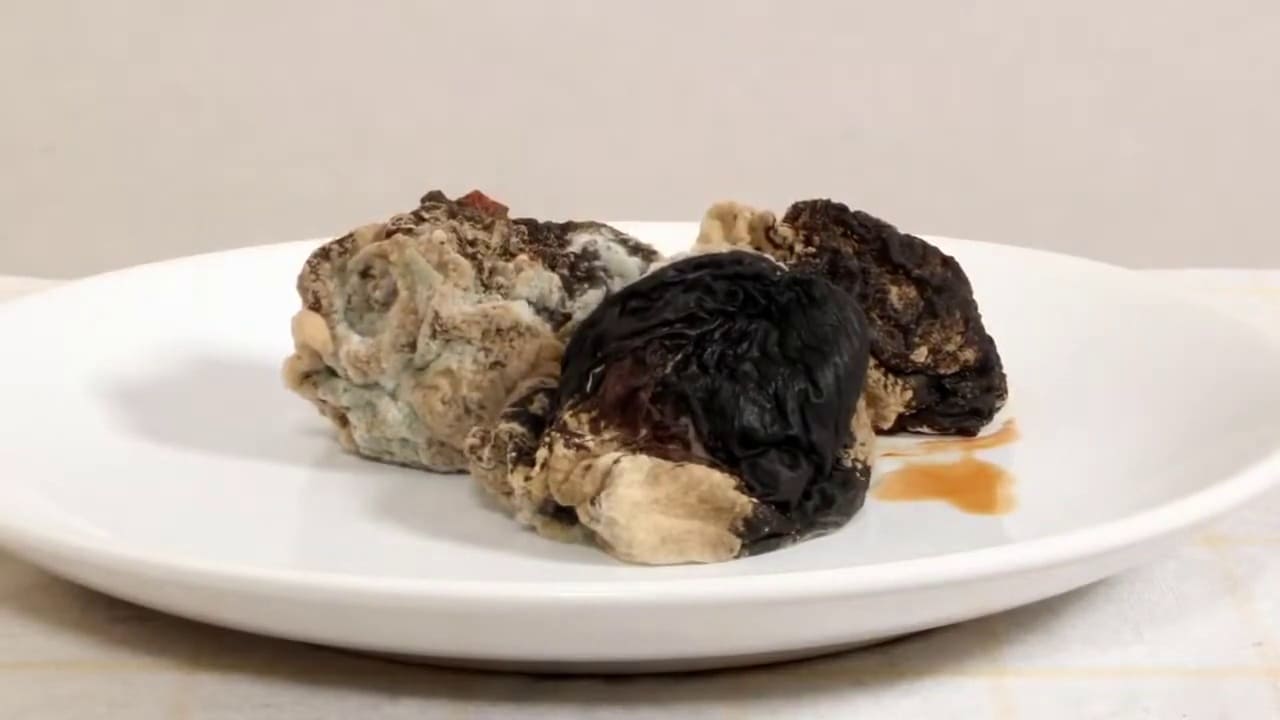}{5}
image \#3: \includeimage{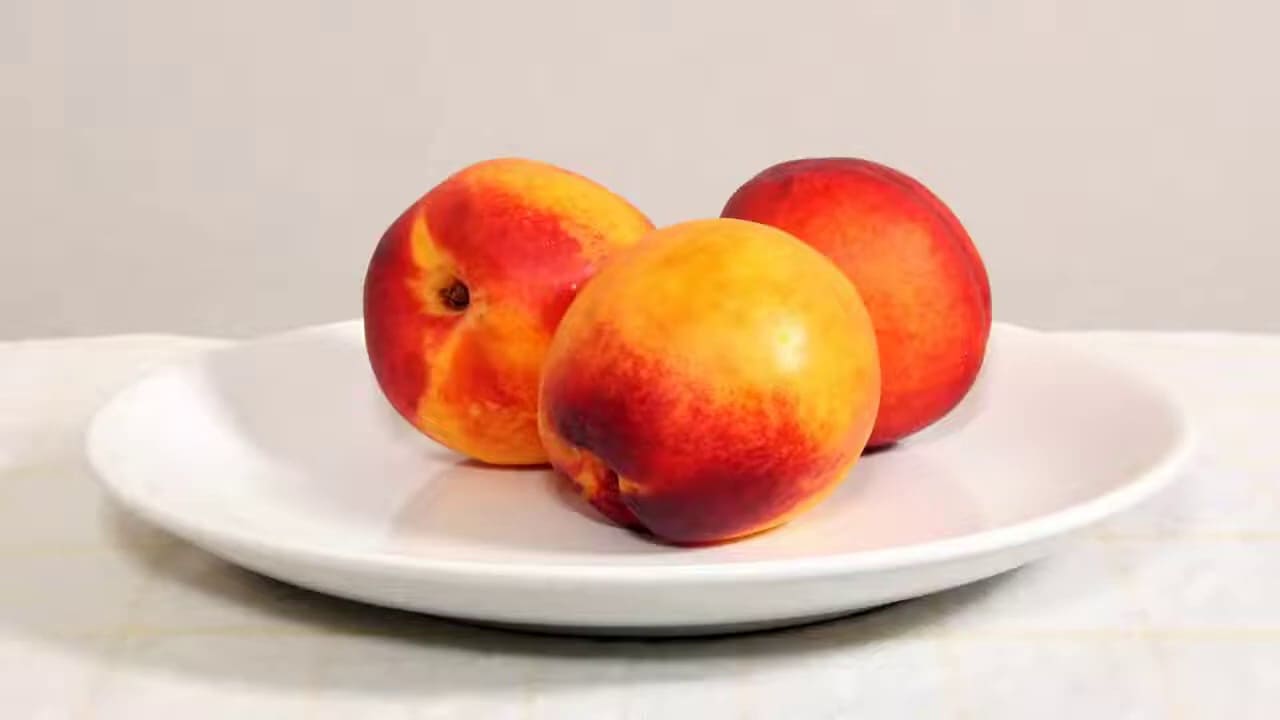}{5}
\end{question}
\begin{answer}{light_yellow}
A. 3 - 2 - 1\qquad \red{B. 2 - 3 - 1}\qquad C. 1 - 3 - 2\qquad D. 1 - 2 - 3
\end{answer}

\begin{question}{light_yellow}
\includeimage{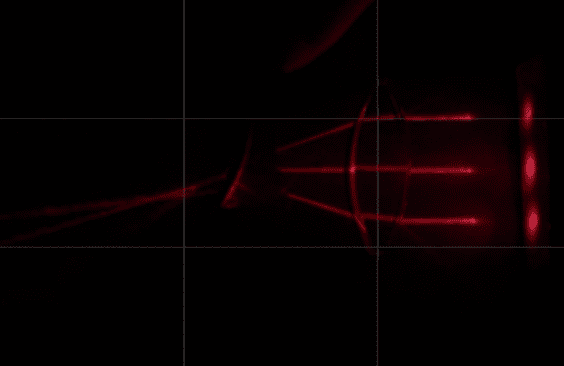}{7} The light first passes through the convex lens and then the concave lens. Slide the concave lens close to the convex lens. Which of the following options will correspond to the phenomenon in the picture?
\end{question}
\begin{answer}{light_yellow}
\red{A.} \includeimage{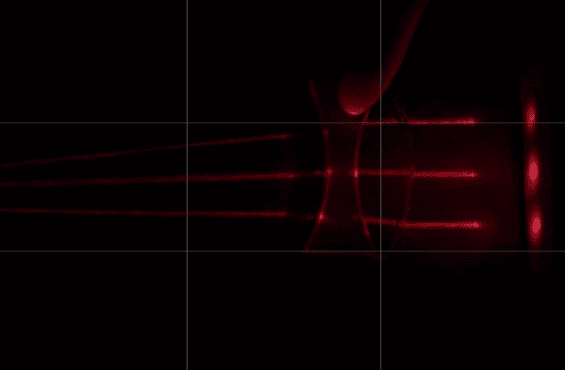}{7}\qquad 
B. \includeimage{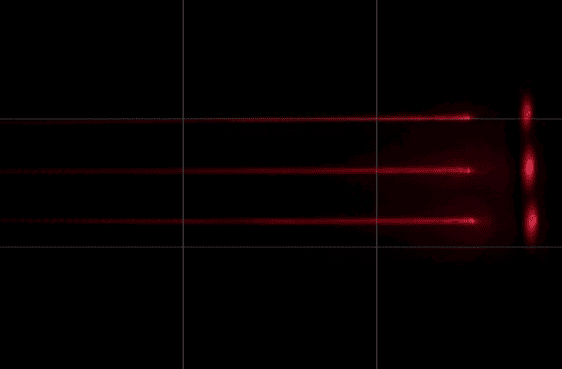}{7}\\
C. \includeimage{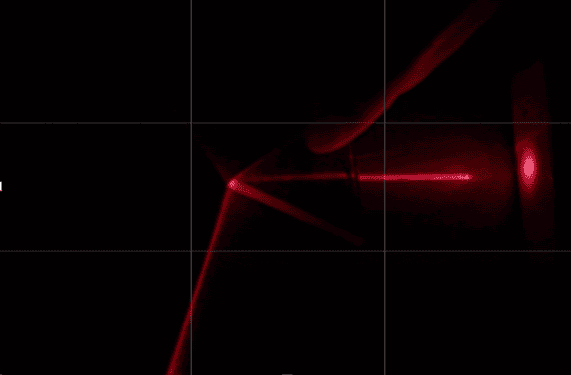}{7}\qquad 
D. \includeimage{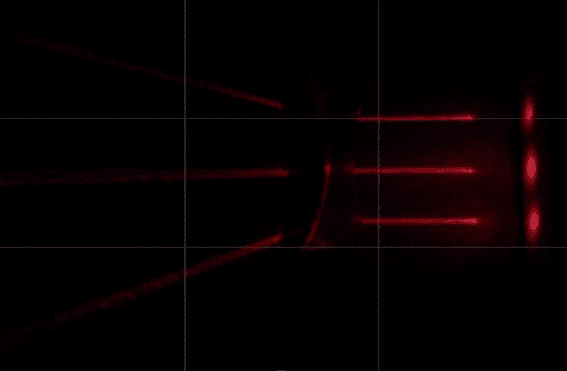}{7}
\end{answer}
\end{mycase}
\vspace{-2mm}
\captionof{figure}{Three examples illustrating dynamics others. The corresponding ability types are
reasoning, reasoning and prediction.}
\vspace{-3mm}
\label{fig:example_19}
\end{table*}

\newcommand{\roundedboxpink}[1]{
  \tikz[baseline=(char.base)]{
    \node[anchor=south west, rounded corners, text height=1.5ex, text depth=.25ex, fill=pink, draw=none, text=black, font=\bfseries] (char) {#1};
  }
}
\newcommand{\roundedboxgreen}[1]{
  \tikz[baseline=(char.base)]{
    \node[anchor=south west, rounded corners, text height=1.5ex, text depth=.25ex, fill=green!30, draw=none, text=black, font=\bfseries] (char) {#1};
  }
}
\newcommand{\roundedboxblue}[1]{
  \tikz[baseline=(char.base)]{
    \node[anchor=south west, rounded corners, text height=1.5ex, text depth=.25ex, fill=blue!30, draw=none, text=black, font=\bfseries] (char) {#1};
  }
}
\newcommand{\roundedboxyellow}[1]{
  \tikz[baseline=(char.base)]{
    \node[anchor=south west, rounded corners, text height=1.5ex, text depth=.25ex, fill=yellow!50, draw=none, text=black, font=\bfseries] (char) {#1};
  }
}
\newcommand{\roundedboxred}[1]{
  \tikz[baseline=(char.base)]{
    \node[anchor=south west, rounded corners, text height=1.5ex, text depth=.25ex, fill=red!30, draw=none, text=black, font=\bfseries] (char) {#1};
  }
}
\newcommand{\roundedboxpurple}[1]{
  \tikz[baseline=(char.base)]{
    \node[anchor=south west, rounded corners, text height=1.5ex, text depth=.25ex, fill=purple!50, draw=none, text=black, font=\bfseries] (char) {#1};
  }
}
\newcommand{\roundedboxbrown}[1]{
  \tikz[baseline=(char.base)]{
    \node[anchor=south west, rounded corners, text height=1.5ex, text depth=.25ex, fill=brown!30, draw=none, text=black, font=\bfseries] (char) {#1};
  }
}
\newcommand{\roundedboxorange}[1]{
  \tikz[baseline=(char.base)]{
    \node[anchor=south west, rounded corners, text height=1.5ex, text depth=.25ex, fill=orange!30, draw=none, text=black, font=\bfseries] (char) {#1};
  }
}
\newcommand{\roundedboxgray}[1]{
  \tikz[baseline=(char.base)]{
    \node[anchor=south west, rounded corners, text height=1.5ex, text depth=.25ex, fill=gray!50, draw=none, text=black, font=\bfseries] (char) {#1};
  }
}

\definecolor{customcolorred}{RGB}{225,159,156} 
\definecolor{customcolorgreen}{RGB}{5,204,151} 

\newcommand{\boxedred}[1]{
  \tikz[baseline=(char.base)]{
    \node[anchor=south west, rectangle, text height=1.5ex, text depth=.25ex, fill=customcolorred, draw=none, text=black, font=\bfseries] (char) {#1};
  }
}
\newcommand{\boxedgreen}[1]{
  \tikz[baseline=(char.base)]{
    \node[anchor=south west, rectangle, text height=1.5ex, text depth=.25ex, fill=customcolorgreen, draw=none, text=black, font=\bfseries] (char) {#1};
  }
}

\clearpage
\section{Error Study}~\label{append:error_case}
\subsection{Detailed Statics}\label{list:list_of_figures}

\begin{table*}[htp]
\centering
\caption{Table index of case study figures by meta-task with associated error categories.}
\label{tab:error_case_all}
\resizebox{1.\textwidth}{!}{%
\tiny
\begin{tabular}{lllllll}
\toprule
Case Figure & Meta-task & Subtask & GPT-4o & Gemini-1.5-flash & Phi-3V \\
     \midrule
      \textcolor{red}{Figure~\ref{fig:error_1}}    & Scene & Light & \roundedboxyellow{Perception Error} & \roundedboxyellow{Perception Error} & \roundedboxyellow{Perception Error} \\
      \textcolor{red}{Figure~\ref{fig:error_2}}& Dynamics & Collision & \roundedboxred{Reasoning Error} & \roundedboxyellow{Perception Error} & \roundedboxred{Reasoning Error}  \\
      \textcolor{red}{Figure~\ref{fig:error_3}}& Dynamics & Throwing & \roundedboxyellow{Perception Error} & \roundedboxyellow{Perception Error} & \roundedboxyellow{Perception Error} \\
      \textcolor{red}{Figure~\ref{fig:error_4}}& Scene & Light & \roundedboxyellow{Perception Error} & \roundedboxyellow{Perception Error} & \roundedboxyellow{Perception Error} \\
      \textcolor{red}{Figure~\ref{fig:error_5}}& Scene & Viewpoint & \roundedboxyellow{Perception Error} & \roundedboxred{Reasoning Error} & \roundedboxred{Reasoning Error} \\
      \textcolor{red}{Figure~\ref{fig:error_6}}& Scene & Viewpoint & \roundedboxyellow{Perception Error} & \roundedboxred{Reasoning Error} & \roundedboxgreen{Success} \\
      \textcolor{red}{Figure~\ref{fig:error_7}}& Dynamics & Chemistry & \roundedboxred{Reasoning Error} & \roundedboxgray{Refuse to Answer} & \roundedboxgreen{Success} \\
      \textcolor{red}{Figure~\ref{fig:error_8}}& Dynamics & Manipulation & \roundedboxyellow{Perception Error} & \roundedboxyellow{Perception Error} & \roundedboxyellow{Perception Error} \\
      \textcolor{red}{Figure~\ref{fig:error_9}}& Dynamics & Manipulation & \roundedboxyellow{Perception Error} & \roundedboxyellow{Perception Error} & \roundedboxyellow{Perception Error} \\
      \textcolor{red}{Figure~\ref{fig:error_10}}& Property & Attribute & \roundedboxyellow{Perception Error} & \roundedboxyellow{Perception Error} & \roundedboxred{Reasoning Error} \\
      \textcolor{red}{Figure~\ref{fig:error_11}}& Property & Color & \roundedboxyellow{Perception Error} & \roundedboxred{Reasoning Error} & \roundedboxyellow{Perception Error} \\
      \textcolor{red}{Figure~\ref{fig:error_12}}& Property & Mass & \roundedboxred{Reasoning Error} & \roundedboxgreen{Success} & \roundedboxyellow{Perception Error} \\
      \textcolor{red}{Figure~\ref{fig:error_13}}& Property & Mass & \roundedboxgreen{Success} & \roundedboxgreen{Success} & \roundedboxgreen{Success} \\
      \textcolor{red}{Figure~\ref{fig:error_14}}& Property & Number & \roundedboxyellow{Perception Error} & \roundedboxyellow{Perception Error} & \roundedboxgray{Refuse to Answer} \\
      \textcolor{red}{Figure~\ref{fig:error_15}}& Property & Number & \roundedboxyellow{Perception Error} & \roundedboxyellow{Perception Error} & \roundedboxgreen{Success} \\
      \textcolor{red}{Figure~\ref{fig:error_16}}& Property & Attribute & \roundedboxblue{Lack of Knowledge} & \roundedboxblue{Lack of Knowledge} & \roundedboxyellow{Perception Error} \\
      \textcolor{red}{Figure~\ref{fig:error_17}}& Relationships & Motion & \roundedboxyellow{Perception Error} & \roundedboxyellow{Perception Error} & \roundedboxred{Reasoning Error} \\
      \textcolor{red}{Figure~\ref{fig:error_18}}& Relationships & Motion & \roundedboxred{Reasoning Error} & \roundedboxred{Reasoning Error} & \roundedboxred{Reasoning Error} \\
      \textcolor{red}{Figure~\ref{fig:error_19}}& Relationships & Depth & \roundedboxred{Reasoning Error} & \roundedboxred{Reasoning Error} & \roundedboxgreen{Success} \\

      \textcolor{red}{Figure~\ref{fig:error_20}}& Relationships & Depth & \roundedboxgreen{Success} & \roundedboxgreen{Success} & \roundedboxgreen{Success} \\
      \textcolor{red}{Figure~\ref{fig:error_21}}& Relationships & Distance & \roundedboxyellow{Perception Error} & \roundedboxred{Reasoning Error} & \roundedboxgreen{Success} \\

      \textcolor{red}{Figure~\ref{fig:error_22}}& Relationships & Distance & \roundedboxred{Reasoning Error} & \roundedboxred{Reasoning Error} & \roundedboxblue{Lack of Knowledge} \\
      \textcolor{red}{Figure~\ref{fig:error_23}}& Relationships & Size & \roundedboxyellow{Perception Error} & \roundedboxred{Reasoning Error} & \roundedboxred{Reasoning Error} \\
      \textcolor{red}{Figure~\ref{fig:error_24}}& Dynamics & Others & \roundedboxblue{Lack of Knowledge} & \roundedboxblue{Lack of Knowledge} & \roundedboxblue{Lack of Knowledge} \\

      \textcolor{red}{Figure~\ref{fig:error_25}}& Relationships & Size & \roundedboxgreen{Success} & \roundedboxgreen{Success} & \roundedboxgreen{Success} \\
      \textcolor{red}{Figure~\ref{fig:error_26}}& Scene & Viewpoint & \roundedboxyellow{Perception Error} & \roundedboxyellow{Perception Error} & \roundedboxgreen{Success} \\

      \textcolor{red}{Figure~\ref{fig:error_27}}& Relationships & Location & \roundedboxyellow{Perception Error} & \roundedboxyellow{Perception Error} & \roundedboxorange{Fail to follow instruction} \\

      \textcolor{red}{Figure~\ref{fig:error_28}}& Relationships & Location & \roundedboxyellow{Perception Error} & \roundedboxyellow{Perception Error} & \roundedboxred{Reasoning Error} \\
      \textcolor{red}{Figure~\ref{fig:error_29}}& Scene & Viewpoint & \roundedboxyellow{Perception Error} & \roundedboxyellow{Perception Error} & \roundedboxgreen{Success} \\
      \textcolor{red}{Figure~\ref{fig:error_30}}& Scene & Temperature & \roundedboxyellow{Perception Error} & \roundedboxyellow{Perception Error} & \roundedboxyellow{Perception Error} \\
      \textcolor{red}{Figure~\ref{fig:error_31}}& Scene & Temperature & \roundedboxred{Reasoning Error} & \roundedboxred{Reasoning Error} & \roundedboxred{Reasoning Error} \\
      \textcolor{red}{Figure~\ref{fig:error_32}}& Dynamics & Air & \roundedboxyellow{Perception Error} & \roundedboxyellow{Perception Error} & \roundedboxyellow{Perception Error} \\
      \textcolor{red}{Figure~\ref{fig:error_33}}& Dynamics & Air & \roundedboxred{Reasoning Error} & \roundedboxred{Reasoning Error} & \roundedboxred{Reasoning Error} \\
      \textcolor{red}{Figure~\ref{fig:error_34}}& Dynamics & Manipulation & \roundedboxred{Reasoning Error} & \roundedboxred{Reasoning Error} & \roundedboxred{Reasoning Error} \\
      \textcolor{red}{Figure~\ref{fig:error_35}}& Relationships & Depth & \roundedboxred{Reasoning Error} & \roundedboxyellow{Perception Error} & \roundedboxyellow{Perception Error}\\
      \textcolor{red}{Figure~\ref{fig:error_36}}& Dynamics & Fluid & \roundedboxyellow{Perception Error} & \roundedboxyellow{Perception Error} & \roundedboxyellow{Perception Error}\\

      \textcolor{red}{Figure~\ref{fig:error_37}}& Dynamics & Others & \roundedboxyellow{Perception Error} & \roundedboxyellow{Perception Error} & \roundedboxyellow{Perception Error}\\

      \textcolor{red}{Figure~\ref{fig:error_38}}& Dynamics & Others & \roundedboxyellow{Perception Error} & \roundedboxyellow{Perception Error} & \roundedboxyellow{Perception Error}\\

      \textcolor{red}{Figure~\ref{fig:error_39}}& Dynamics & Collision & \roundedboxyellow{Perception Error} & \roundedboxyellow{Perception Error} & \roundedboxyellow{Perception Error}\\

      \textcolor{red}{Figure~\ref{fig:error_40}}& Scene & Light & \roundedboxred{Reasoning Error} & \roundedboxyellow{Perception Error} & \roundedboxgreen{Success}\\

      \textcolor{red}{Figure~\ref{fig:error_41}}& Scene & Viewpoint & \roundedboxyellow{Perception Error} & \roundedboxyellow{Perception Error} & \roundedboxyellow{Perception Error}\\

      \textcolor{red}{Figure~\ref{fig:error_42}}& Relationships & Location & \roundedboxgreen{Success} & \roundedboxyellow{Perception Error} & \roundedboxyellow{Perception Error}\\

      \textcolor{red}{Figure~\ref{fig:error_43}}& Property & Attribute & \roundedboxyellow{Perception Error} & \roundedboxred{Reasoning Error} & \roundedboxyellow{Perception Error}\\

      \textcolor{red}{Figure~\ref{fig:error_44}}& Dynamics & Manipulation & \roundedboxyellow{Perception Error} & \roundedboxgreen{Success} & \roundedboxyellow{Perception Error}\\

      \textcolor{red}{Figure~\ref{fig:error_45}}& Scene & Viewpoint & \roundedboxyellow{Perception Error} & \roundedboxyellow{Perception Error} & \roundedboxyellow{Perception Error}\\

      \textcolor{red}{Figure~\ref{fig:error_46}}& Scene & Light & \roundedboxyellow{Perception Error} & \roundedboxyellow{Perception Error} & \roundedboxyellow{Perception Error}\\

      \textcolor{red}{Figure~\ref{fig:error_47}}& Relationships & Location & \roundedboxyellow{Perception Error} & \roundedboxyellow{Perception Error} & \roundedboxyellow{Perception Error}\\

      \textcolor{red}{Figure~\ref{fig:error_48}}& Relationships & Location & \roundedboxyellow{Perception Error} & \roundedboxyellow{Perception Error} & \roundedboxyellow{Perception Error}\\

      \textcolor{red}{Figure~\ref{fig:error_49}}& Dynamics & Others & \roundedboxred{Reasoning Error} & \roundedboxred{Reasoning Error} & \roundedboxred{Reasoning Error}\\

      \textcolor{red}{Figure~\ref{fig:error_50}}& Scene & Viewpoint & \roundedboxyellow{Perception Error} & \roundedboxred{Reasoning Error} & \roundedboxgreen{Success}\\

      \textcolor{red}{Figure~\ref{fig:error_51}}& Scene & Light & \roundedboxred{Reasoning Error} & \roundedboxred{Reasoning Error} & \roundedboxred{Reasoning Error}\\

      \textcolor{red}{Figure~\ref{fig:error_52}}& Dynamics & Manipulation & \roundedboxyellow{Perception Error} & \roundedboxgreen{Success} & \roundedboxred{Reasoning Error}\\

      \textcolor{red}{Figure~\ref{fig:error_53}}& Dynamics & Collision & \roundedboxyellow{Perception Error} & \roundedboxyellow{Perception Error} & \roundedboxgreen{Success}\\

      \textcolor{red}{Figure~\ref{fig:error_54}}& Property & Attribute & \roundedboxyellow{Perception Error} & \roundedboxyellow{Perception Error} & \roundedboxred{Reasoning Error}\\

      \textcolor{red}{Figure~\ref{fig:error_55}}& Scene & Light & \roundedboxgreen{Success} & \roundedboxred{Reasoning Error} & \roundedboxgreen{Success}\\

      \textcolor{red}{Figure~\ref{fig:error_56}}& Scene & Light & \roundedboxyellow{Perception Error} & \roundedboxyellow{Perception Error} & \roundedboxyellow{Perception Error}\\

      \textcolor{red}{Figure~\ref{fig:error_57}}& Scene & Light & \roundedboxyellow{Perception Error} & \roundedboxyellow{Perception Error} & \roundedboxred{Reasoning Error}\\

      \textcolor{red}{Figure~\ref{fig:error_58}}& Scene & Light & \roundedboxgreen{Success} & \roundedboxyellow{Perception Error} & \roundedboxred{Reasoning Error}\\

      \textcolor{red}{Figure~\ref{fig:error_59}}& Scene & Light & \roundedboxred{Reasoning Error} & \roundedboxyellow{Perception Error} & \roundedboxred{Reasoning Error}\\

      \textcolor{red}{Figure~\ref{fig:error_60}}& Scene & Light & \roundedboxyellow{Perception Error} & \roundedboxyellow{Perception Error} & \roundedboxyellow{Perception Error}\\

\bottomrule
\end{tabular}%
}
\end{table*}

In this section, we present a case study analysis of the error types made by GPT-4V, Gemini-1.5-flash, and Phi-3V across various tasks. The errors are classified into the following five categories. Other less frequent error types are not included in this analysis. For the analysis, as mentioned in Section~\ref{exp:task_analysis}, we selected 500 samples for each model, but due to space limitations, we present only 60 of them here, as shown in Table~\ref{tab:error_case_all}.

\begin{figure}
    \centering
    \includegraphics[width=0.92\linewidth]{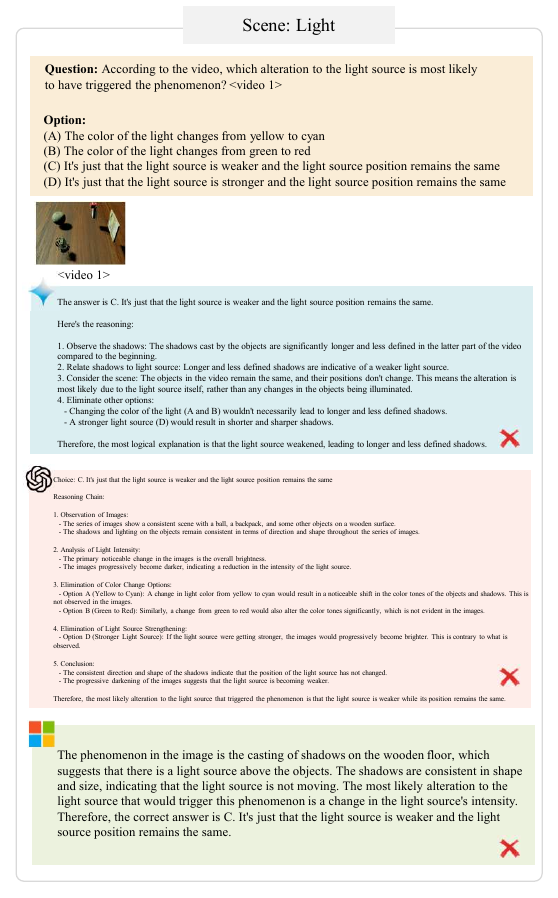}
    \caption{A sample case of physical scene understanding--light. \hyperref[list:list_of_figures]{Back to List of Figures}.}
    \label{fig:error_1}
\end{figure}

\begin{figure}
  \centering
  \includegraphics[width=0.92\linewidth]{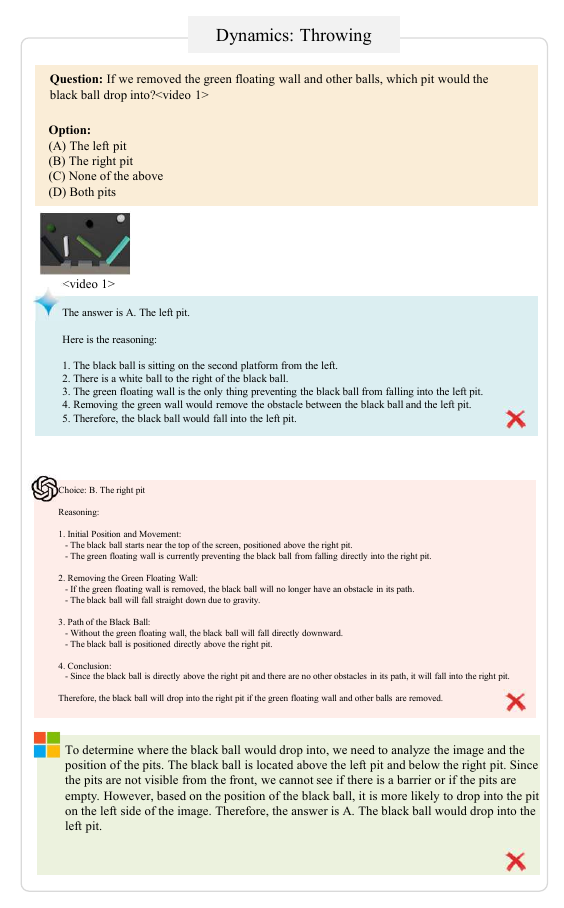}
  \caption{A sample case of physics-based dynamics——throwing. \hyperref[list:list_of_figures]{Back to List of Figures}.}
  \label{fig:error_3}
\end{figure}

\roundedboxyellow{Perception Error}: VLMs fail to recognize, classify, or detect the objects or content in images. VLMs are constrained by the representation power of visual encoders, and these errors account for the majority of them. See examples in Figure~\ref{fig:error_1}, Figure~\ref{fig:error_3}, \textit{etc.}

\begin{figure}
  \centering
  \includegraphics[width=0.92\linewidth]{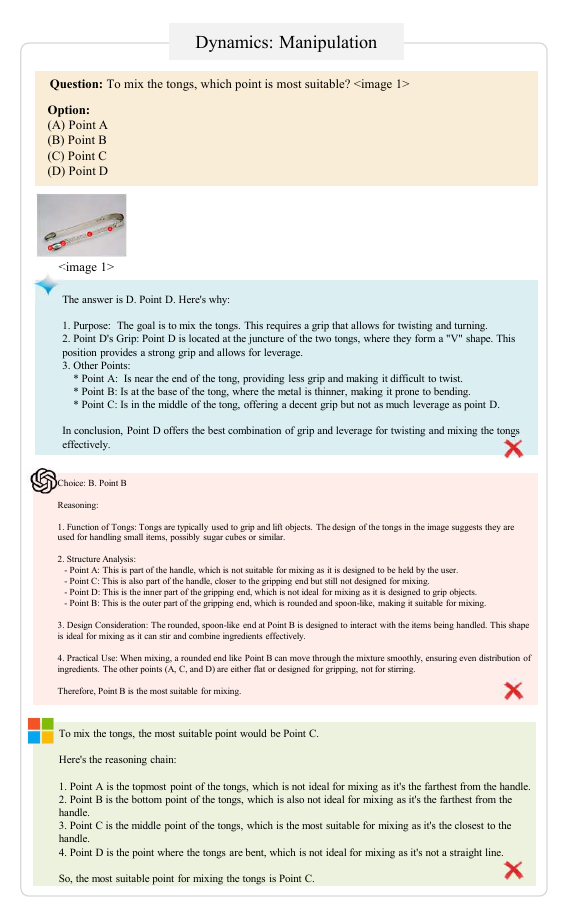}
  \caption{A sample case of physics-based dynamics——manipulation. \hyperref[list:list_of_figures]{Back to List of Figures}.}
  \label{fig:error_34}
\end{figure}

\roundedboxred{Reasoning Error}: VLMs can recognize the text and visual content exactly but make errors in reasoning, leading to incorrect results. See examples in Figure~\ref{fig:error_18}, Figure~\ref{fig:error_34}, \textit{etc.}

\begin{figure}
  \centering
  \includegraphics[width=0.92\linewidth]{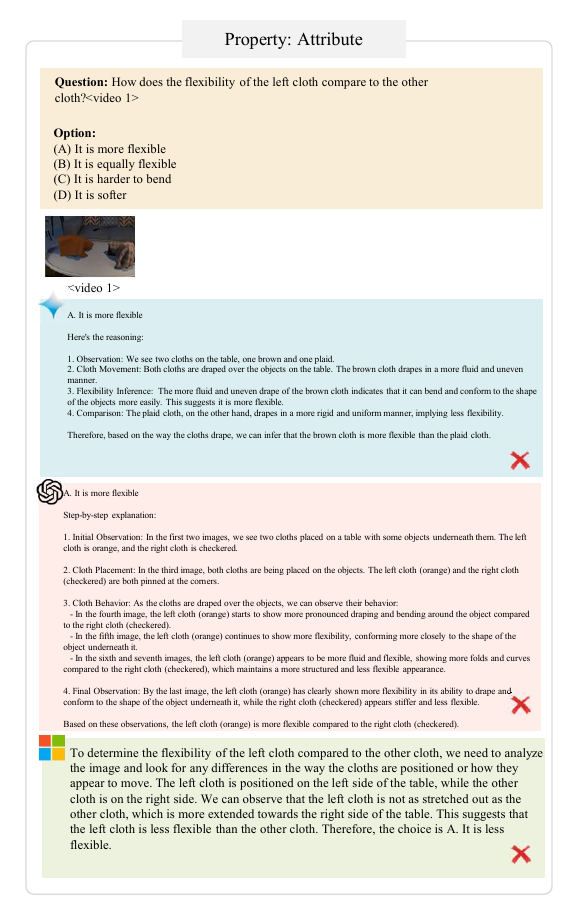}
  \caption{A sample case of physical object property--attribute. \hyperref[list:list_of_figures]{Back to List of Figures}.}
  \label{fig:error_10}
\end{figure}

\roundedboxblue{Lack of Knowledge}: VLMs do not have specific knowledge, so they finally get a wrong answer. See examples in Figure~\ref{fig:error_10}, Figure~\ref{fig:error_24}.

\begin{figure}
  \centering
  \includegraphics[width=0.92\linewidth]{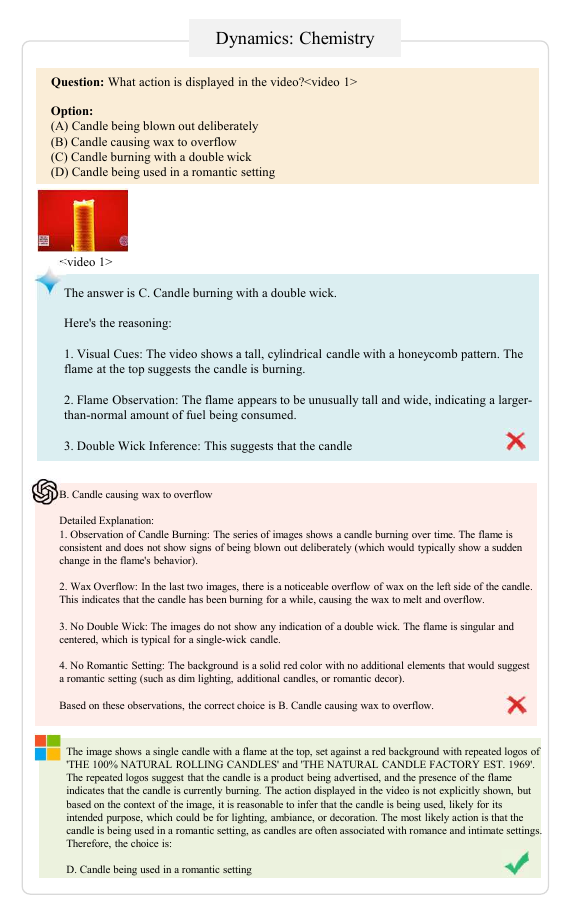}
  \caption{A sample case of physics-based dynamics——chemistry. \hyperref[list:list_of_figures]{Back to List of Figures}.}
  \label{fig:error_7}
\end{figure}

\roundedboxgray{Refuse to Answer}: VLMs refuse to answer questions and stop answering immediately. See examples in Figure~\ref{fig:error_7}.

\begin{figure}
  \centering
  \includegraphics[width=0.92\linewidth]{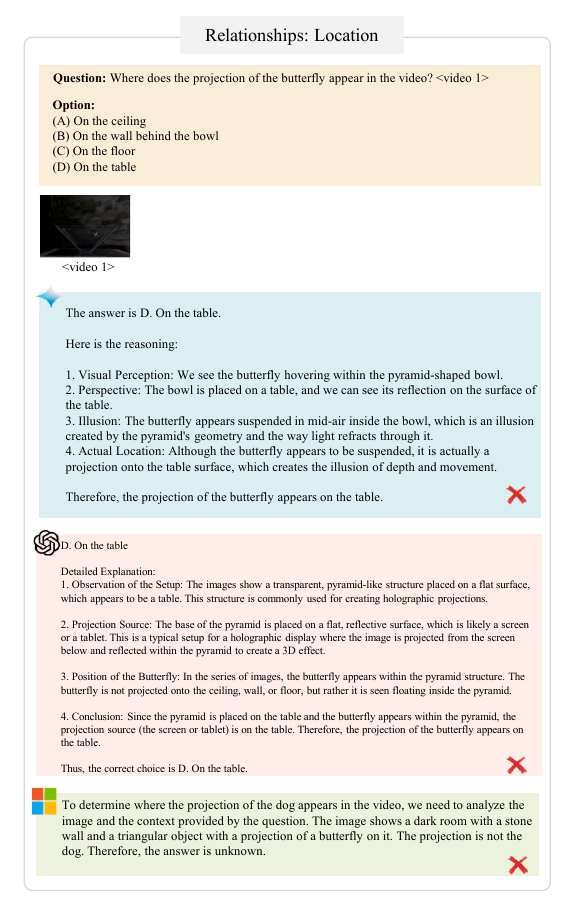}
  \caption{A sample case of physical object relationships--location. \hyperref[list:list_of_figures]{Back to List of Figures}.}
  \label{fig:error_27}
\end{figure}

\roundedboxorange{Fail to Follow Instruction}: VLMs fail to correctly understand instructions and provide erroneous answers. For example, VLMs may not understand the specified conditions in the instruction (see Figure~\ref{fig:error_27}).

\subsection{Main Reason Analysis}
\begin{figure}
  \centering
  \includegraphics[width=0.92\linewidth]{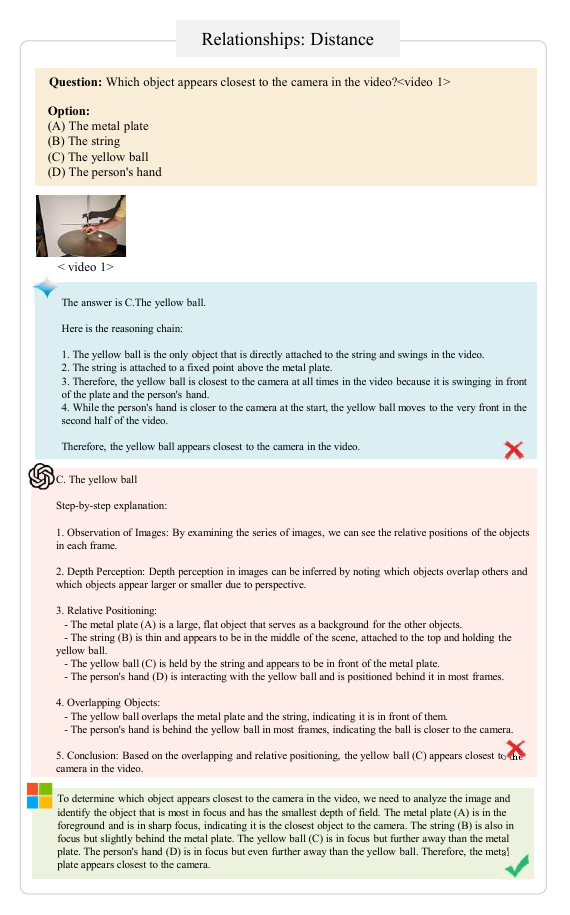}
  \caption{A sample case of physical object relationships--distance. \hyperref[list:list_of_figures]{Back to List of Figures}.}
  \label{fig:error_21}
\end{figure}

\textbf{Perceptual Errors}: Perceptual errors can be classified into two categories: basic perceptual errors and domain-specific perceptual errors. Basic perceptual errors occur when the model successfully understands the given information but fails to interpret fundamental visual objects correctly. In contrast, domain-specific perceptual errors arise when the model misinterprets visual inputs due to a lack of understanding of specialized conditions. Moreover, GPT-4V often exhibits a bias towards textual information, prioritizing text over visual inputs—a trend highlighted in recent studies~\cite{cui2023holisticanalysishallucinationgpt4vision}. A notable example is shown in Figure~\ref{fig:error_21}, where the model mistakenly identified the plate as part of the background, leading to incorrect depth estimation. This underscores the importance of seeking a balanced approach to enhance the model's interpretative capabilities.

\begin{figure}
  \centering
  \includegraphics[width=0.92\linewidth]{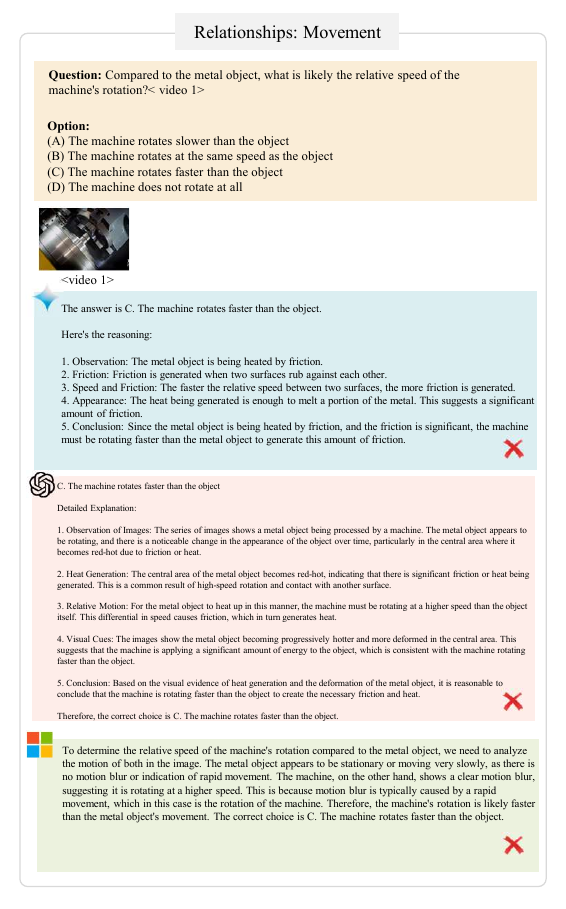}
  \caption{A sample case of physical object relationships--motion. \hyperref[list:list_of_figures]{Back to List of Figures}.}
  \label{fig:error_18}
\end{figure}

\begin{figure}
  \centering
  \includegraphics[width=0.92\linewidth]{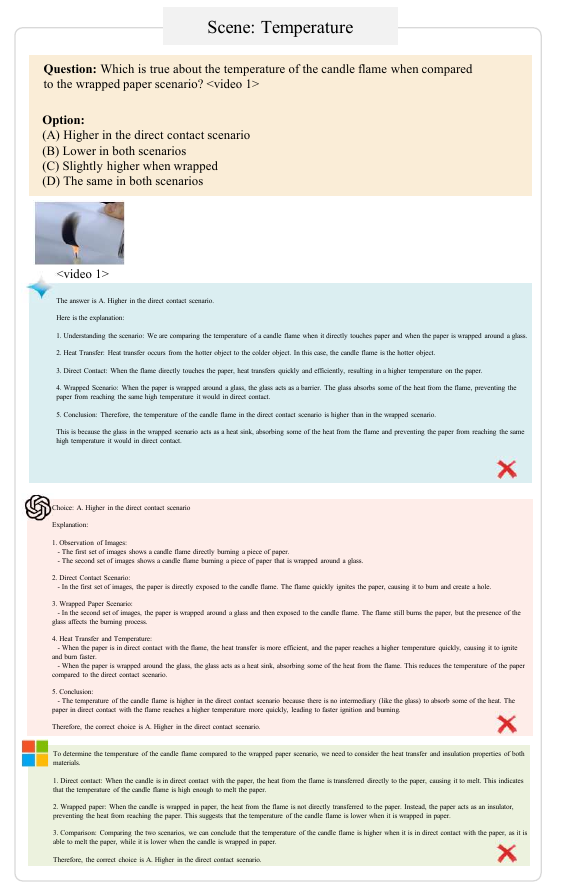}
  \caption{A sample case of physical scene understanding--temperature. \hyperref[list:list_of_figures]{Back to List of Figures}.}
  \label{fig:error_31}
\end{figure}

\textbf{Reasoning Errors}: Sometimes, even when the model correctly interprets the text, images, and question, it fails to establish a rigorous logical chain, leading to incorrect conclusions. These types of mistakes, known as reasoning errors, are illustrated in Figures~\ref{fig:error_18} and~\ref{fig:error_31}. In the first example, the model mistakenly assumes that friction can only exist if the machine rotates faster than the object, overlooking the presence of another object, which results in an incorrect answer. In the second example, while the model correctly identifies that the temperature of the wrapped paper is significantly lower than that of the other object, it incorrectly attributes this phenomenon to the flame's temperature, leading to an erroneous conclusion.

\begin{figure}
  \centering
  \includegraphics[width=0.92\linewidth]{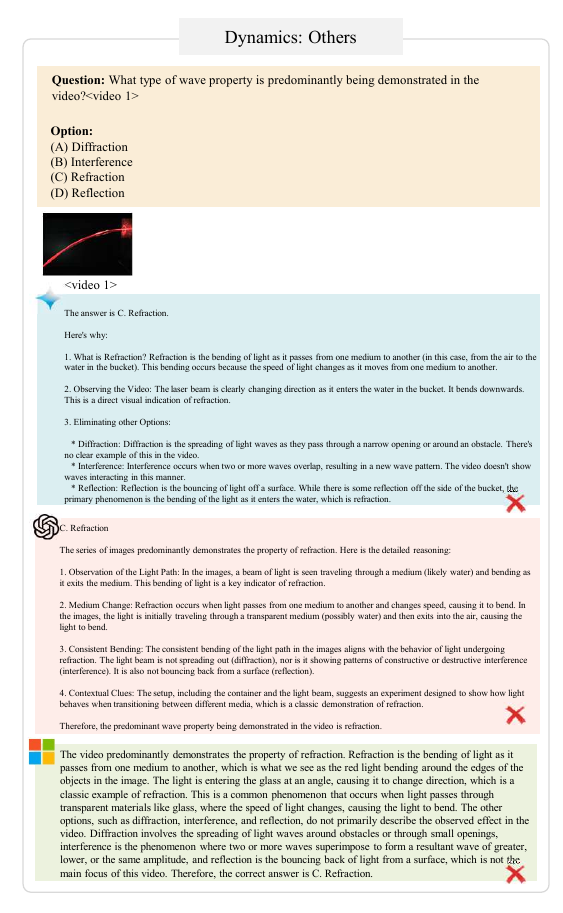}
  \caption{A sample case of physics-based dynamics——others. \hyperref[list:list_of_figures]{Back to List of Figures}.}
  \label{fig:error_24}
\end{figure}

\textbf{Lack of Knowledge}: Another major cause of errors is the model's lack of relevant knowledge. A notable example is shown in Figure~\ref{fig:error_24}, where the model incorrectly assumes that light bending in water is due to refraction. The model clearly does not understand the concept of total internal reflection in water, leading to an incorrect conclusion.

\textbf{Other Errors}: The remaining errors, such as those related to textual understanding, refusal to answer, annotation mistakes, and answer extraction issues, account for only a small proportion. However, they remain significant and should not be overlooked.

\subsection{Case Study}

\begin{figure}
  \centering
  \includegraphics[width=0.92\linewidth]{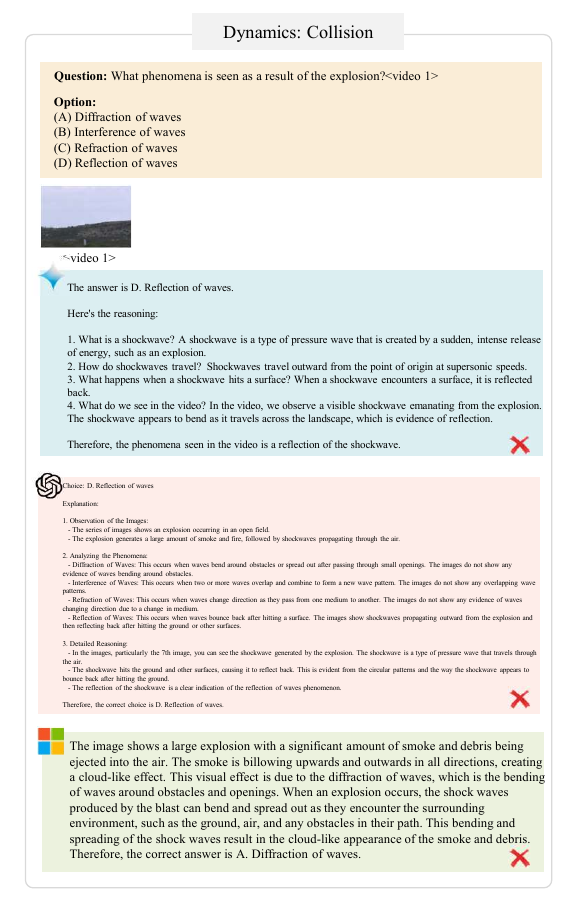}
  \caption{A sample case of physics-based dynamics——collision. \hyperref[list:list_of_figures]{Back to List of Figures}.}
  \label{fig:error_2}
\end{figure}

\begin{figure}
  \centering
  \includegraphics[width=0.92\linewidth]{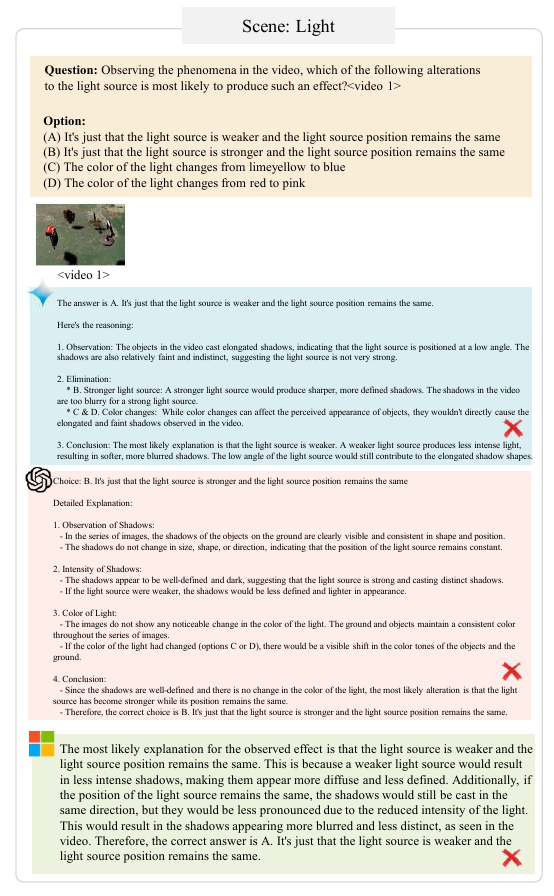}
  \caption{A sample case of physical scene understanding--light. \hyperref[list:list_of_figures]{Back to List of Figures}.}
  \label{fig:error_4}
\end{figure}

\begin{figure}
  \centering
  \includegraphics[width=0.92\linewidth]{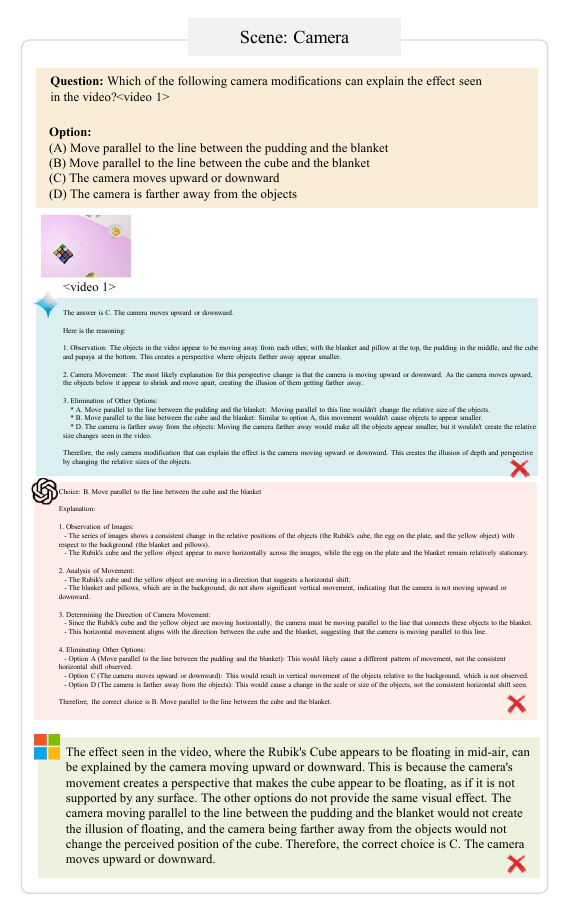}
  \caption{A sample case of physical scene understanding--viewpoint. \hyperref[list:list_of_figures]{Back to List of Figures}.}
  \label{fig:error_5}
\end{figure}

\begin{figure}
  \centering
  \includegraphics[width=0.92\linewidth]{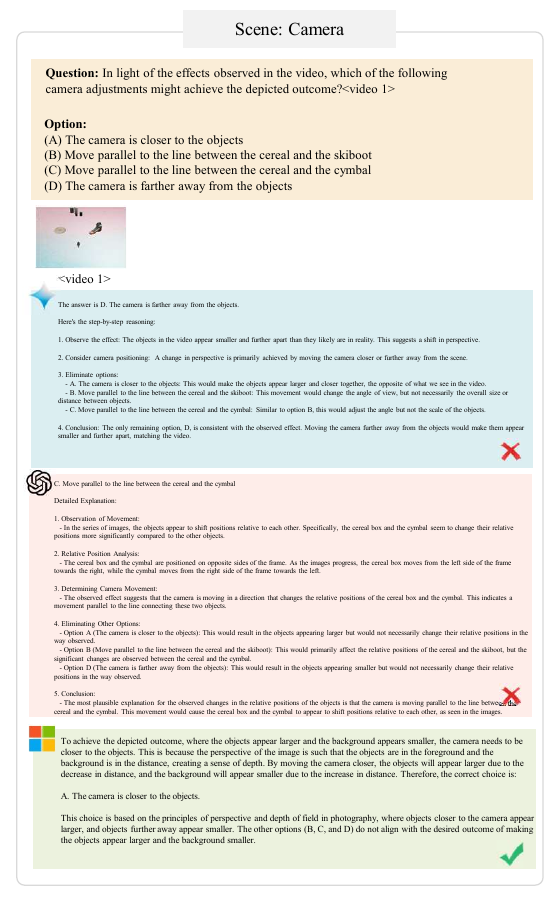}
  \caption{A sample case of physical scene understanding--viewpoint. \hyperref[list:list_of_figures]{Back to List of Figures}.}
  \label{fig:error_6}
\end{figure}

\begin{figure}
  \centering
  \includegraphics[width=0.92\linewidth]{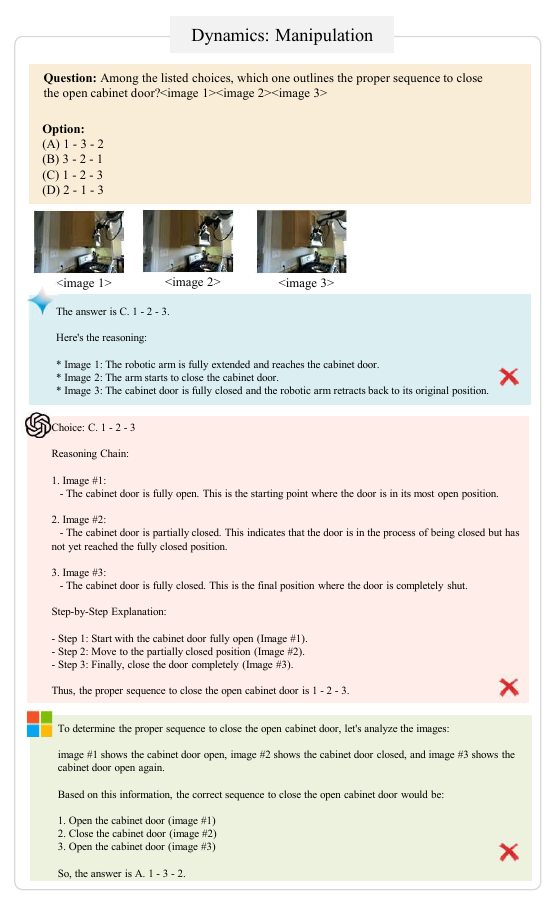}
  \caption{A sample case of physics-based dynamics——manipulation. \hyperref[list:list_of_figures]{Back to List of Figures}.}
  \label{fig:error_8}
\end{figure}

\begin{figure}
  \centering
  \includegraphics[width=0.92\linewidth]{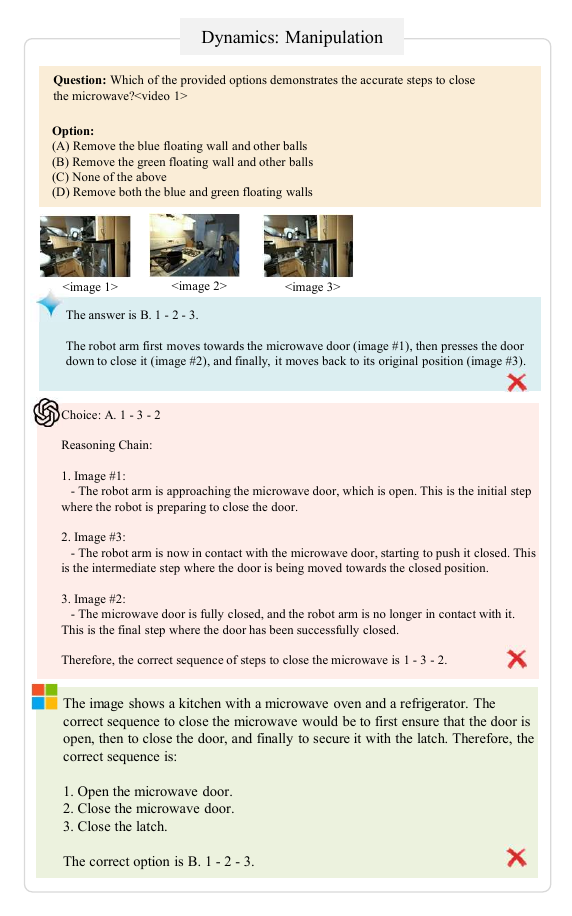}
  \caption{A sample case of physics-based dynamics——throwing. \hyperref[list:list_of_figures]{Back to List of Figures}.}
  \label{fig:error_9}
\end{figure}

\begin{figure}
  \centering
  \includegraphics[width=0.92\linewidth]{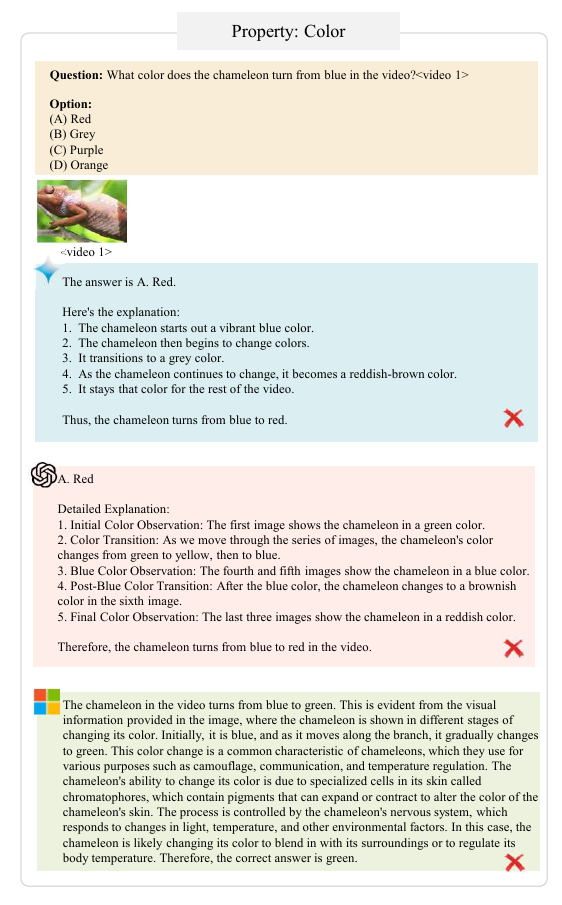}
  \caption{A sample case of physical object property--color. \hyperref[list:list_of_figures]{Back to List of Figures}.}
  \label{fig:error_11}
\end{figure}

\begin{figure}
  \centering
  \includegraphics[width=0.92\linewidth]{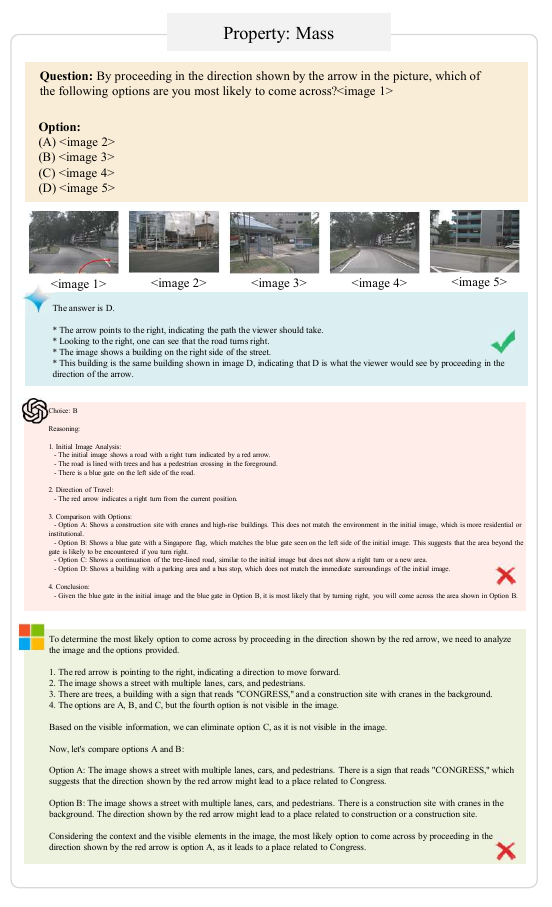}
  \caption{A sample case of physical object property--mass. \hyperref[list:list_of_figures]{Back to List of Figures}.}
  \label{fig:error_12}
\end{figure}

\begin{figure}
  \centering
  \includegraphics[width=0.92\linewidth]{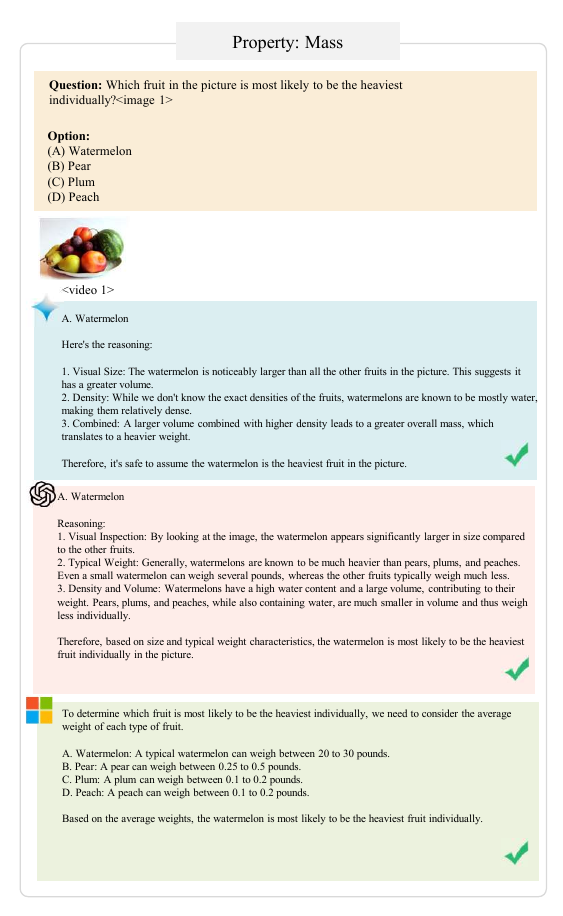}
  \caption{A sample case of physical object property--mass. \hyperref[list:list_of_figures]{Back to List of Figures}.}
  \label{fig:error_13}
\end{figure}

\begin{figure}
  \centering
  \includegraphics[width=0.92\linewidth]{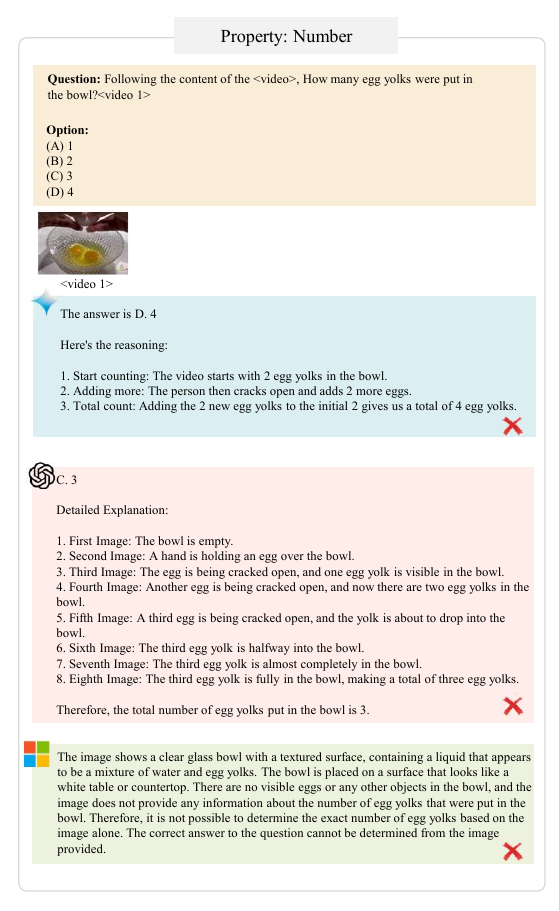}
  \caption{A sample case of physical object property--number. \hyperref[list:list_of_figures]{Back to List of Figures}.}
  \label{fig:error_14}
\end{figure}

\begin{figure}
  \centering
  \includegraphics[width=0.92\linewidth]{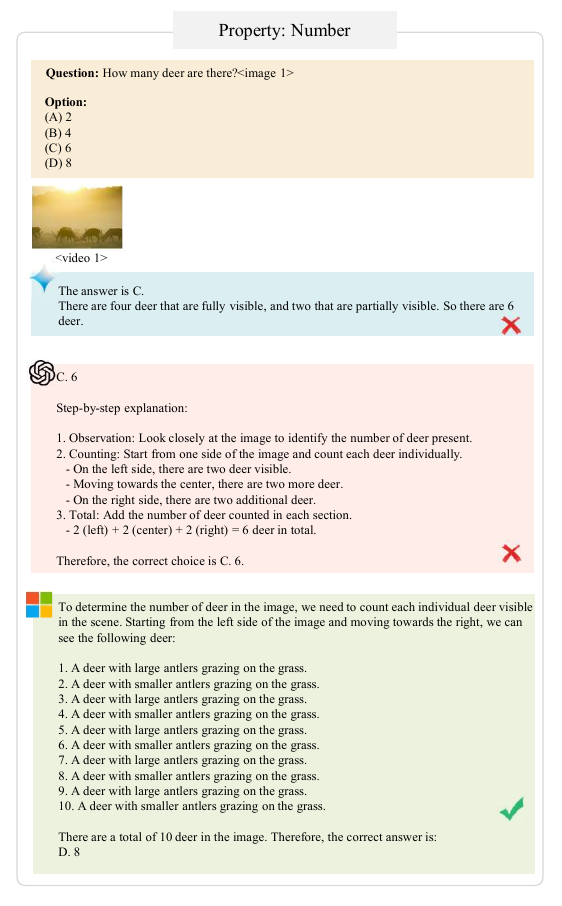}
  \caption{A sample case of physical object property--number. \hyperref[list:list_of_figures]{Back to List of Figures}.}
  \label{fig:error_15}
\end{figure}

\begin{figure}
  \centering
  \includegraphics[width=0.92\linewidth]{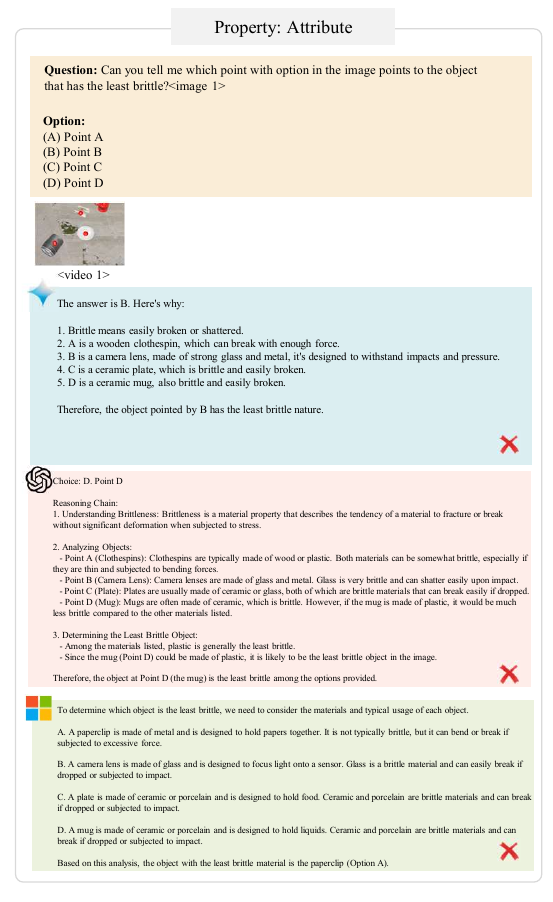}
  \caption{A sample case of physical object property--attribute. \hyperref[list:list_of_figures]{Back to List of Figures}.}
  \label{fig:error_16}
\end{figure}

\begin{figure}
  \centering
  \includegraphics[width=0.92\linewidth]{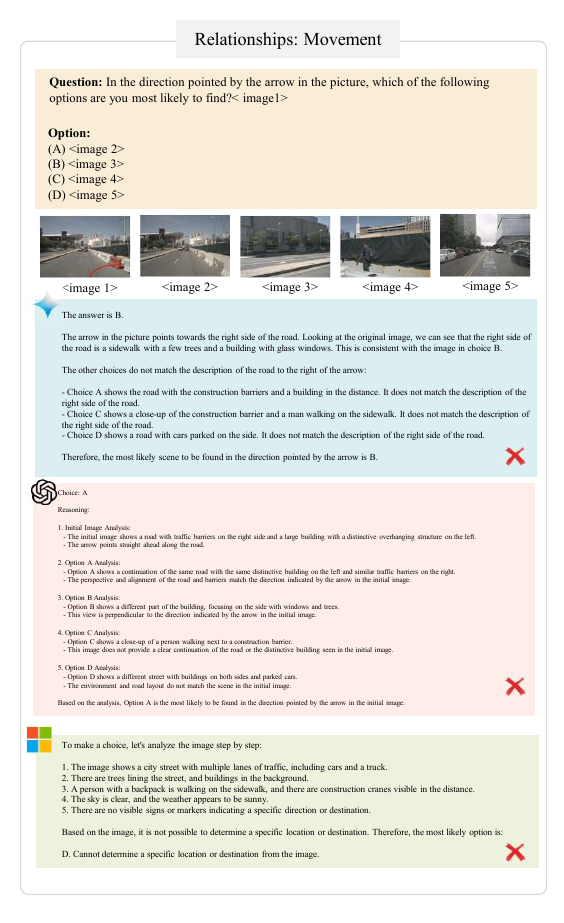}
  \caption{A sample case of physical object relationships--motion. \hyperref[list:list_of_figures]{Back to List of Figures}.}
  \label{fig:error_17}
\end{figure}

\begin{figure}
  \centering
  \includegraphics[width=0.92\linewidth]{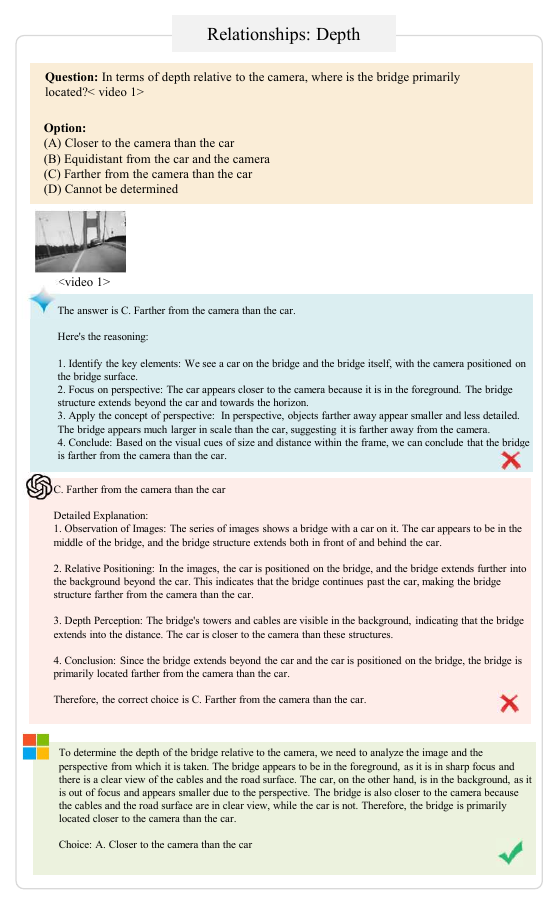}
  \caption{A sample case of physical object relationships--depth. \hyperref[list:list_of_figures]{Back to List of Figures}.}
  \label{fig:error_19}
\end{figure}

\begin{figure}
  \centering
  \includegraphics[width=0.92\linewidth]{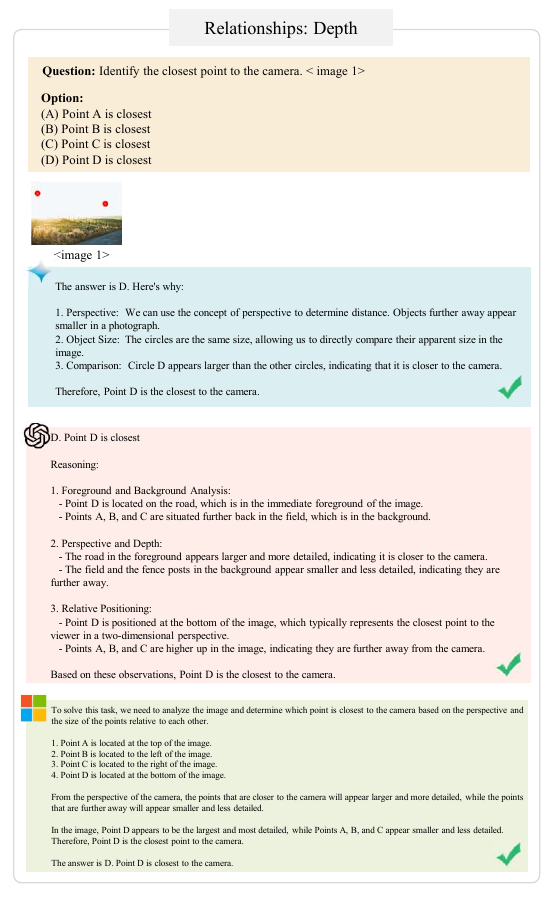}
  \caption{A sample case of physical object relationships--depth. \hyperref[list:list_of_figures]{Back to List of Figures}.}
  \label{fig:error_20}
\end{figure}

\begin{figure}
  \centering
  \includegraphics[width=0.92\linewidth]{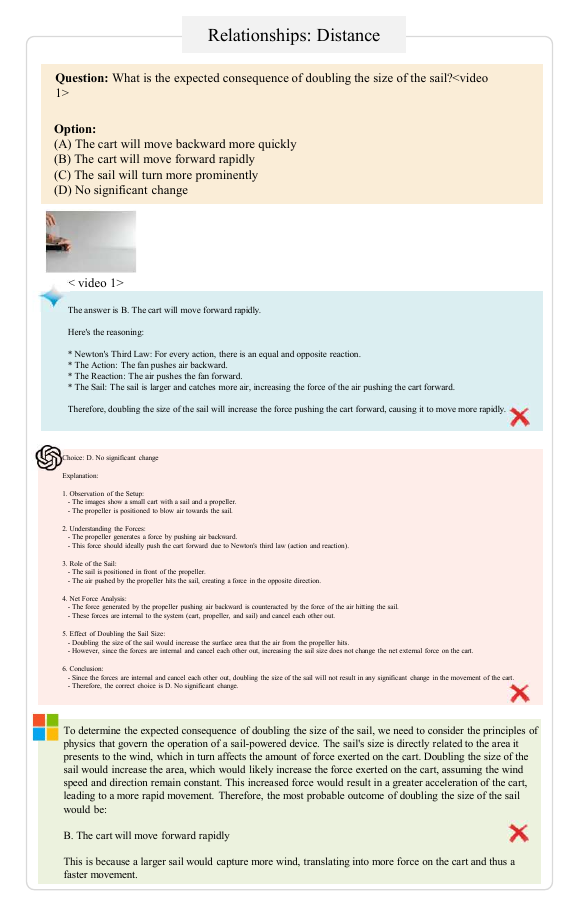}
  \caption{A sample case of physical object relationships--distance. \hyperref[list:list_of_figures]{Back to List of Figures}.}
  \label{fig:error_22}
\end{figure}

\begin{figure}
  \centering
  \includegraphics[width=0.92\linewidth]{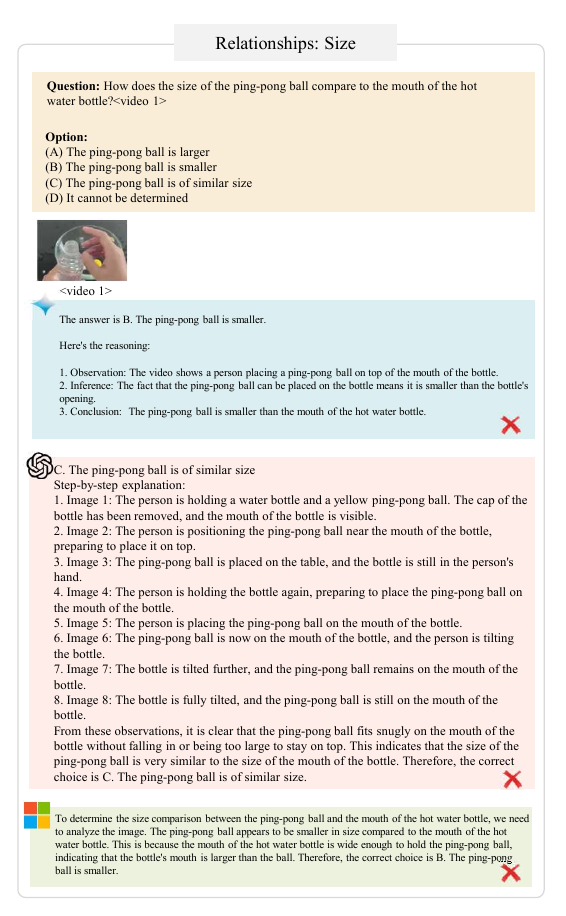}
  \caption{A sample case of physical object relationships--size. \hyperref[list:list_of_figures]{Back to List of Figures}.}
  \label{fig:error_23}
\end{figure}

\begin{figure}
  \centering
  \includegraphics[width=0.92\linewidth]{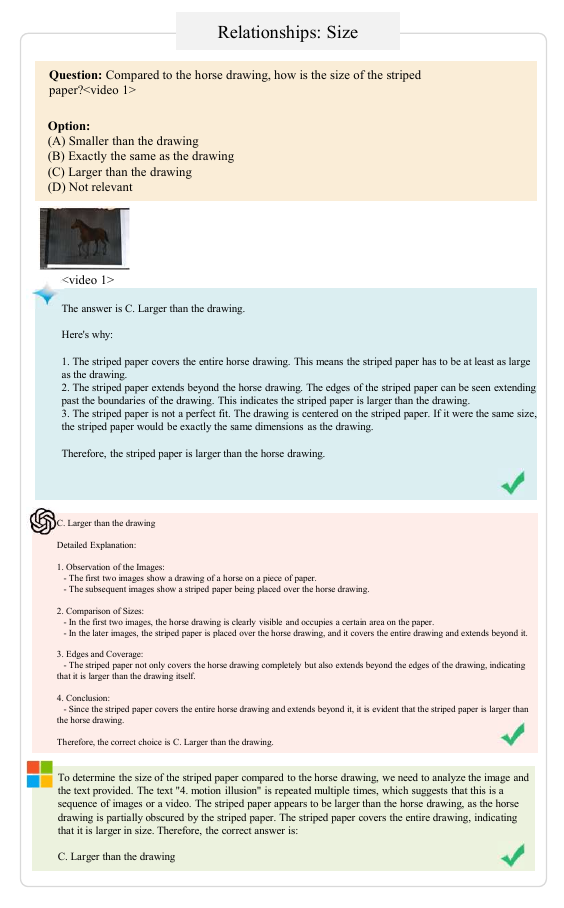}
  \caption{A sample case of physical object relationships--size. \hyperref[list:list_of_figures]{Back to List of Figures}.}
  \label{fig:error_25}
\end{figure}

\begin{figure}
  \centering
  \includegraphics[width=0.92\linewidth]{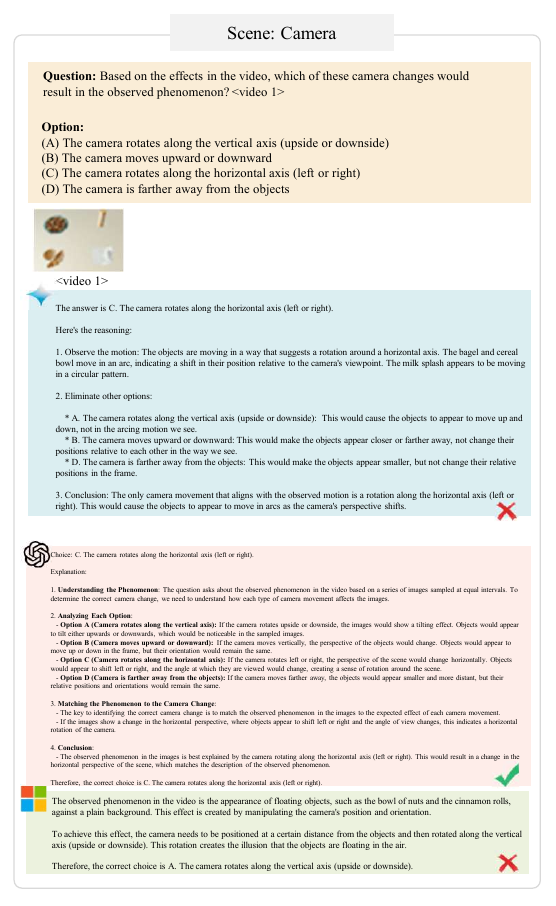}
  \caption{A sample case of physical scene understanding--viewpoint. \hyperref[list:list_of_figures]{Back to List of Figures}.}
  \label{fig:error_26}
\end{figure}

\begin{figure}
  \centering
  \includegraphics[width=0.92\linewidth]{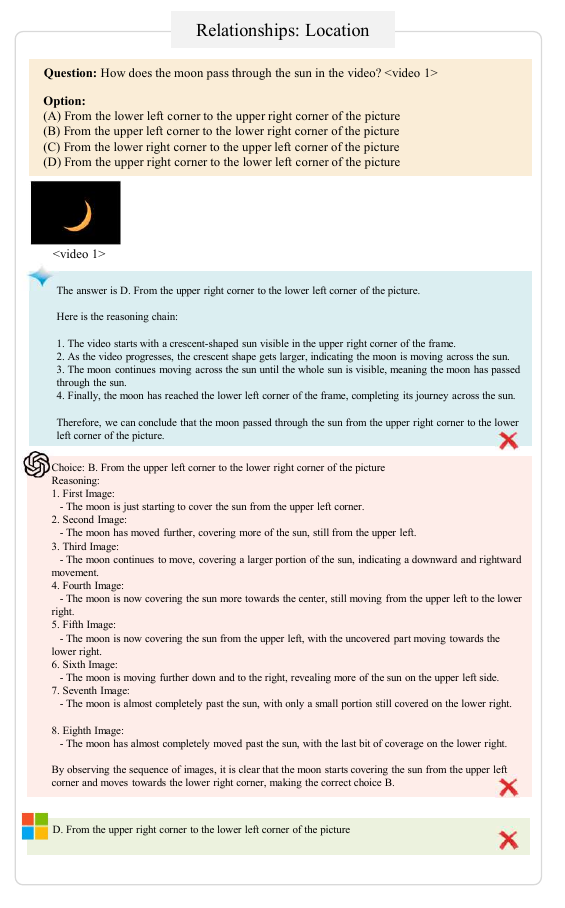}
  \caption{A sample case of physical object relationships--location. \hyperref[list:list_of_figures]{Back to List of Figures}.}
  \label{fig:error_28}
\end{figure}

\begin{figure}
  \centering
  \includegraphics[width=0.92\linewidth]{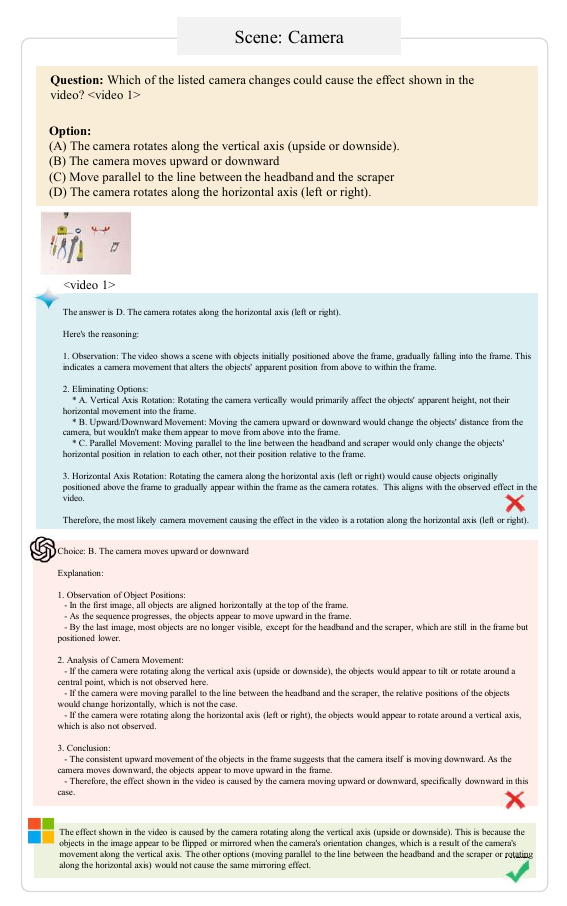}
  \caption{A sample case of physical scene understanding--light. \hyperref[list:list_of_figures]{Back to List of Figures}.}
  \label{fig:error_29}
\end{figure}

\begin{figure}
  \centering
  \includegraphics[width=0.92\linewidth]{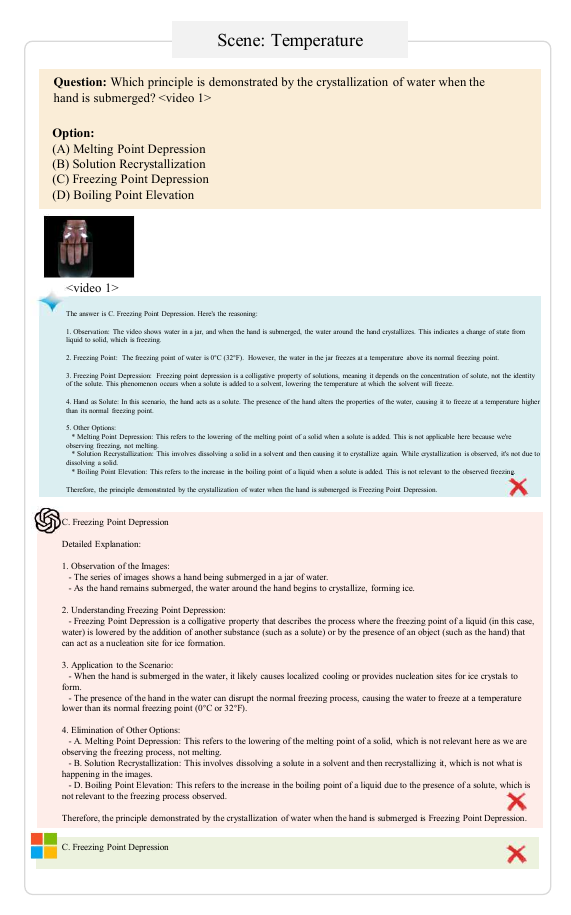}
  \caption{A sample case of physical scene understanding--temperature. \hyperref[list:list_of_figures]{Back to List of Figures}.}
  \label{fig:error_30}
\end{figure}

\begin{figure}
  \centering
  \includegraphics[width=0.92\linewidth]{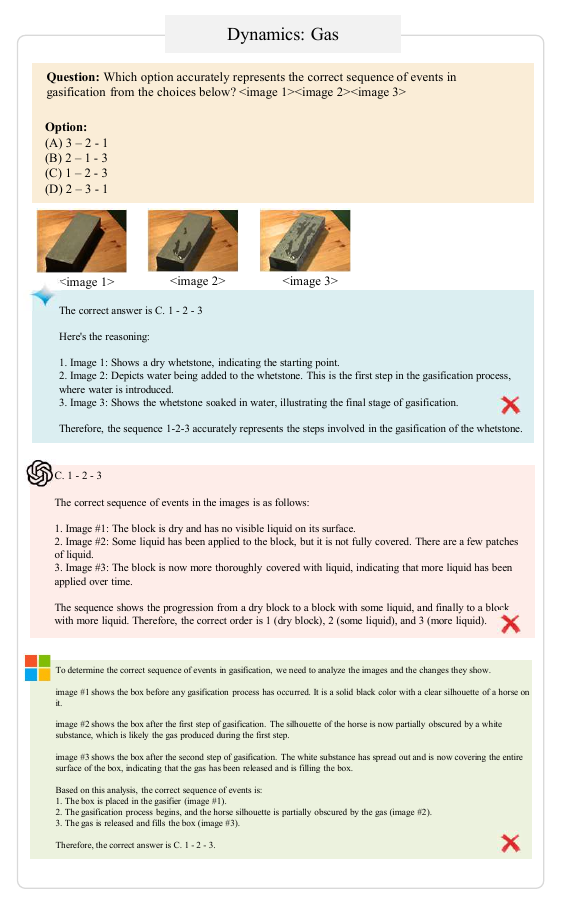}
  \caption{A sample case of physics-based dynamics——air. \hyperref[list:list_of_figures]{Back to List of Figures}.}
  \label{fig:error_32}
\end{figure}

\begin{figure}
  \centering
  \includegraphics[width=0.92\linewidth]{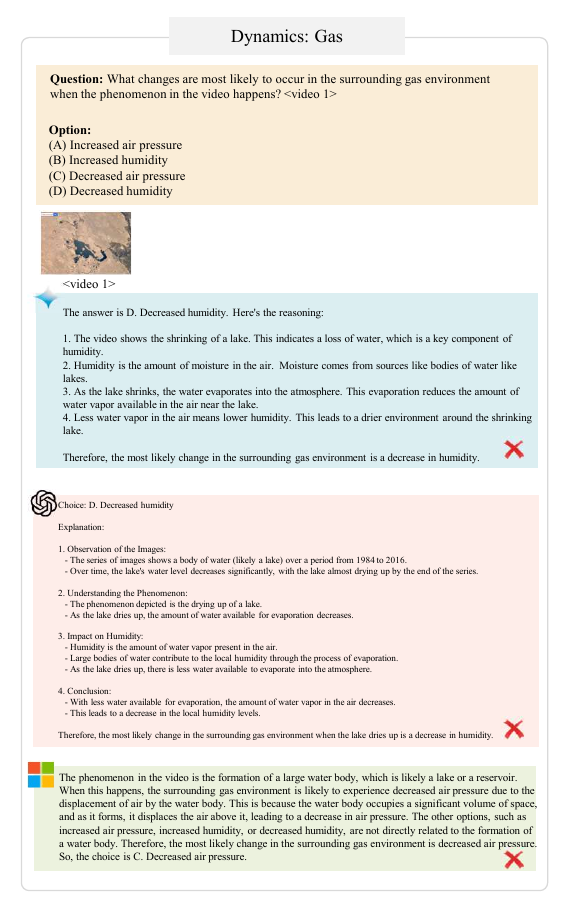}
  \caption{A sample case of physics-based dynamics——air. \hyperref[list:list_of_figures]{Back to List of Figures}.}
  \label{fig:error_33}
\end{figure}

\begin{figure}
  \centering
  \includegraphics[width=0.92\linewidth]{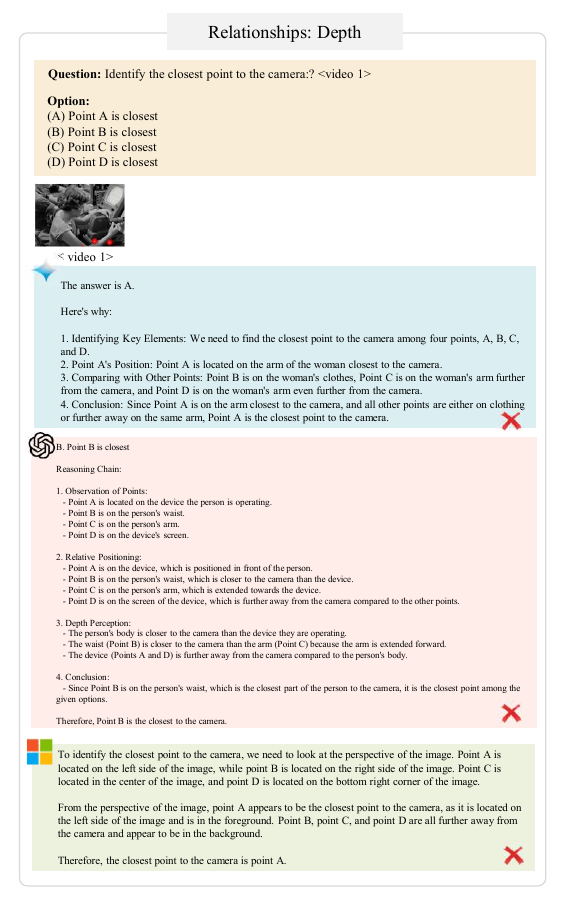}
  \caption{A sample case of physics-based dynamics——fluid. \hyperref[list:list_of_figures]{Back to List of Figures}.}
  \label{fig:error_35}
\end{figure}

\begin{figure}
  \centering
  \includegraphics[width=0.92\linewidth]{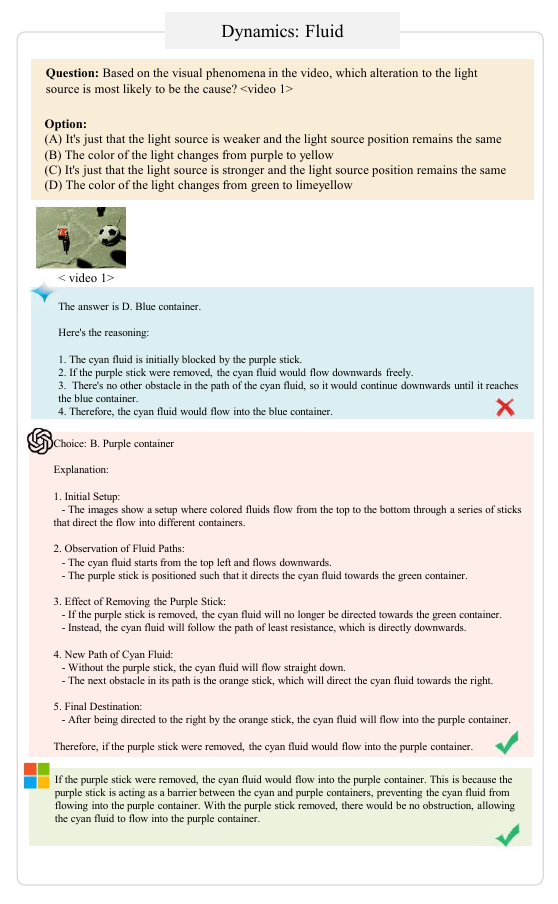}
  \caption{A sample case of physics-based dynamics——fluid. \hyperref[list:list_of_figures]{Back to List of Figures}.}
  \label{fig:error_36}
\end{figure}

\begin{figure}
  \centering
  \includegraphics[width=0.92\linewidth]{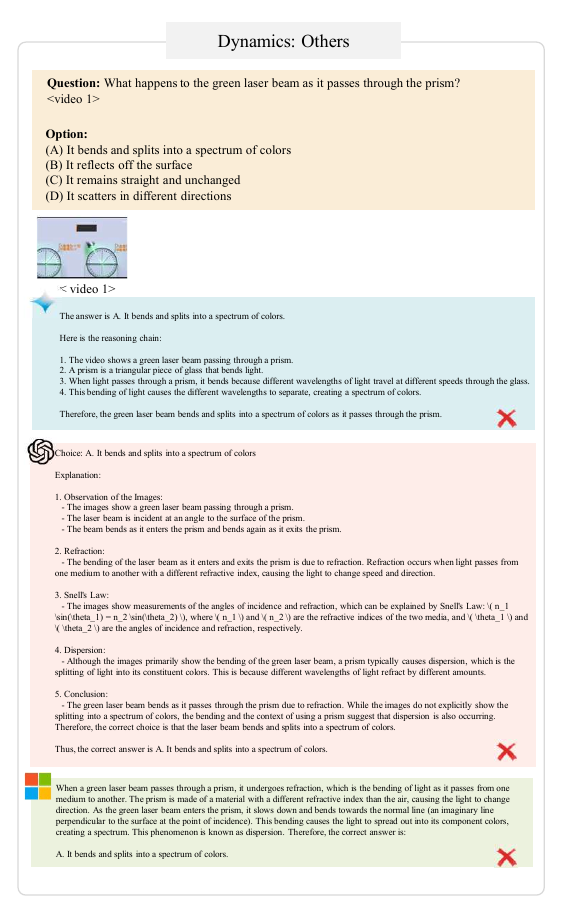}
  \caption{A sample case of physics-based dynamics——others. \hyperref[list:list_of_figures]{Back to List of Figures}.}
  \label{fig:error_37}
\end{figure}

\begin{figure}
  \centering
  \includegraphics[width=0.92\linewidth]{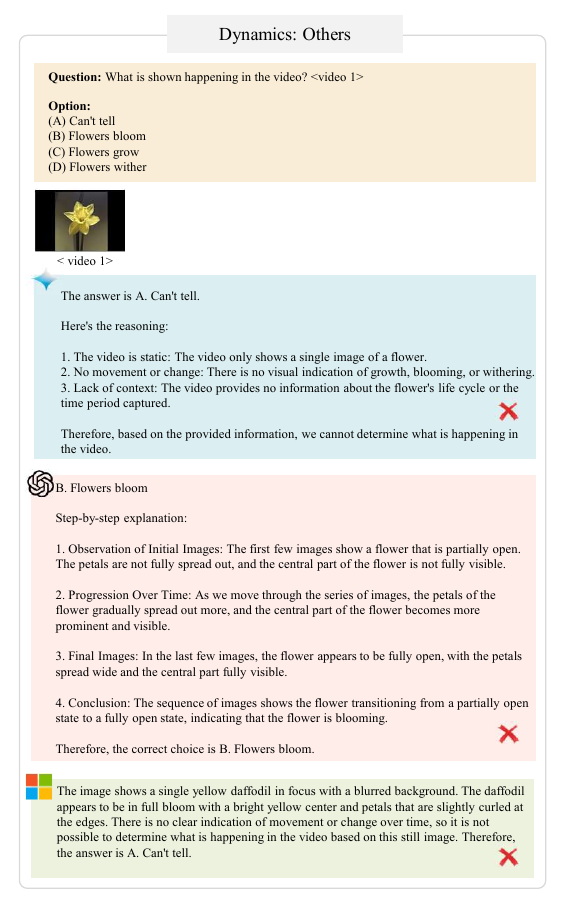}
  \caption{A sample case of physics-based dynamics——others. \hyperref[list:list_of_figures]{Back to List of Figures}.}
  \label{fig:error_38}
\end{figure}

\begin{figure}
  \centering
  \includegraphics[width=0.92\linewidth]{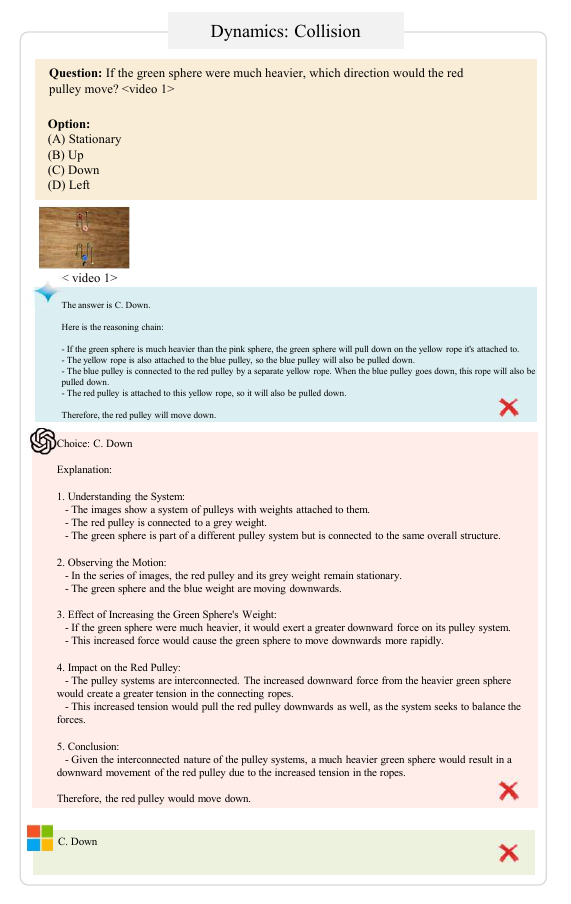}
  \caption{A sample case of physics-based dynamics——collision. \hyperref[list:list_of_figures]{Back to List of Figures}.}
  \label{fig:error_39}
\end{figure}

\begin{figure}
  \centering
  \includegraphics[width=0.92\linewidth]{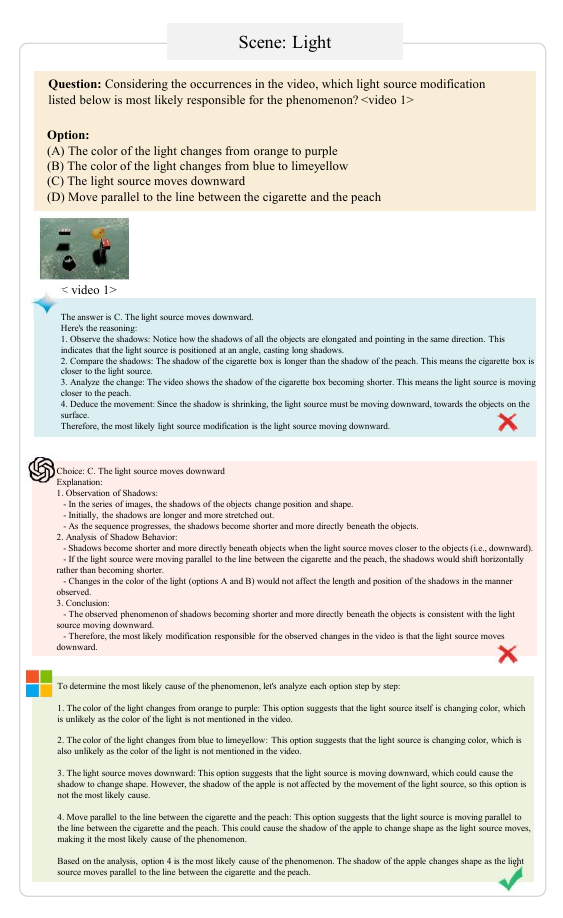}
  \caption{A sample case of physical scene understanding--light. \hyperref[list:list_of_figures]{Back to List of Figures}.}
  \label{fig:error_40}
\end{figure}

\begin{figure}
  \centering
  \includegraphics[width=0.92\linewidth]{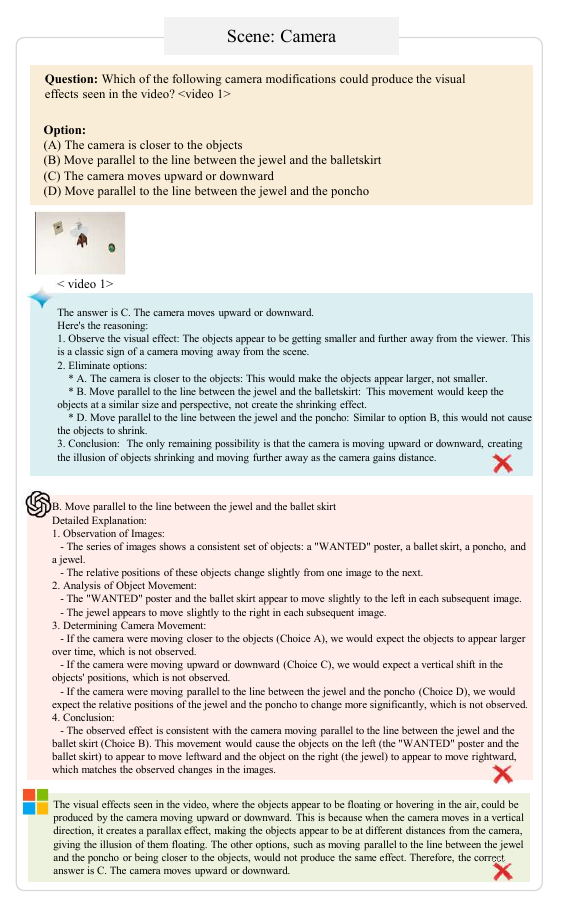}
  \caption{A sample case of physical scene understanding--viewpoint. \hyperref[list:list_of_figures]{Back to List of Figures}.}
  \label{fig:error_41}
\end{figure}

\begin{figure}
  \centering
  \includegraphics[width=0.92\linewidth]{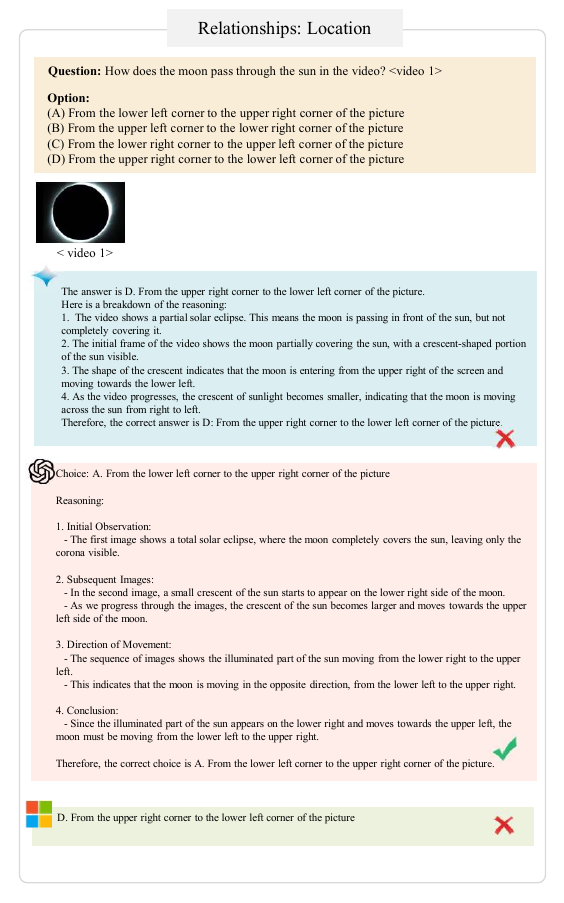}
  \caption{A sample case of physical object relationships--location. \hyperref[list:list_of_figures]{Back to List of Figures}.}
  \label{fig:error_42}
\end{figure}

\begin{figure}
  \centering
  \includegraphics[width=0.92\linewidth]{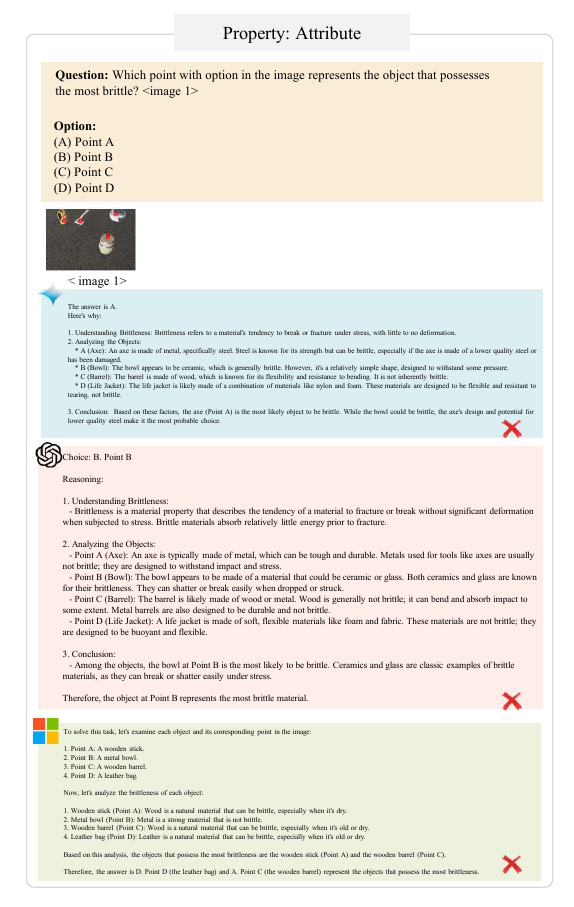}
  \caption{A sample case of physical object property--attribute. \hyperref[list:list_of_figures]{Back to List of Figures}.}
  \label{fig:error_43}
\end{figure}

\begin{figure}
  \centering
  \includegraphics[width=0.92\linewidth]{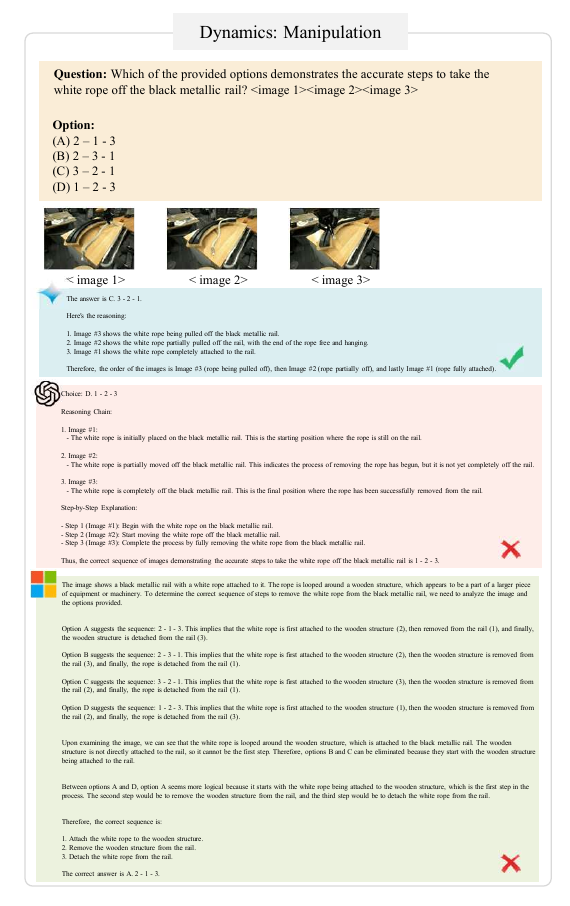}
  \caption{A sample case of physics-based dynamics——manipulation. \hyperref[list:list_of_figures]{Back to List of Figures}.}
  \label{fig:error_44}
\end{figure}

\begin{figure}
  \centering
  \includegraphics[width=0.92\linewidth]{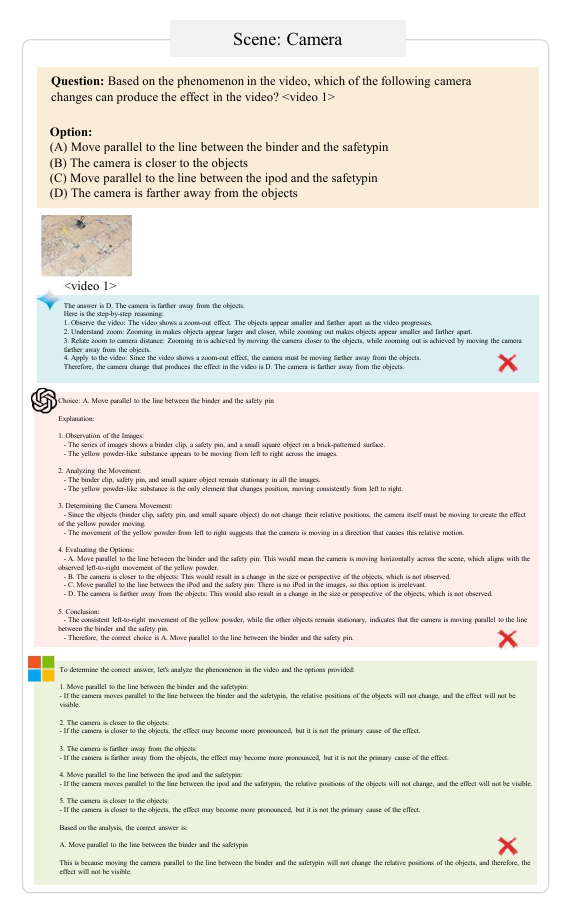}
  \caption{A sample case of physical scene understanding--viewpoint. \hyperref[list:list_of_figures]{Back to List of Figures}.}
  \label{fig:error_45}
\end{figure}

\begin{figure}
  \centering
  \includegraphics[width=0.92\linewidth]{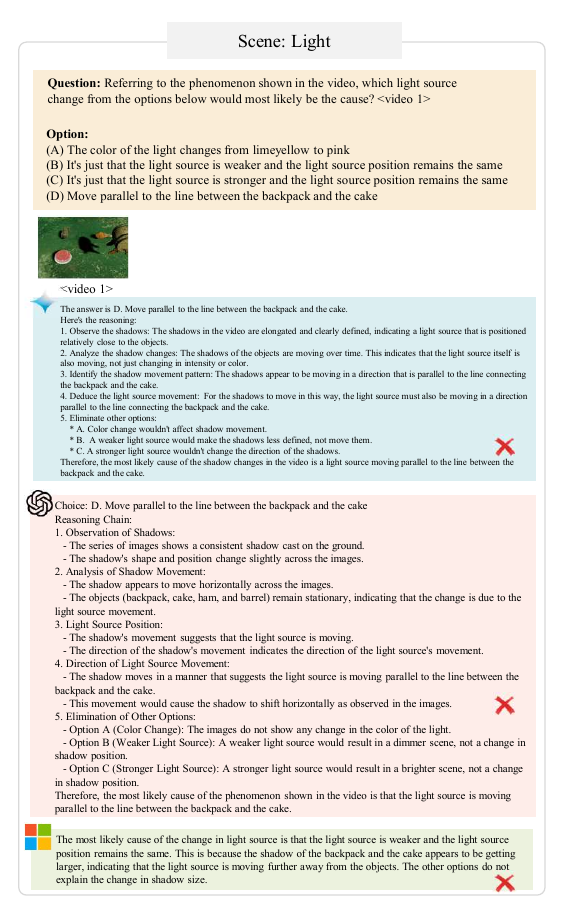}
  \caption{A sample case of physics-based dynamics——collision. \hyperref[list:list_of_figures]{Back to List of Figures}.}
  \label{fig:error_46}
\end{figure}

\begin{figure}
  \centering
  \includegraphics[width=0.92\linewidth]{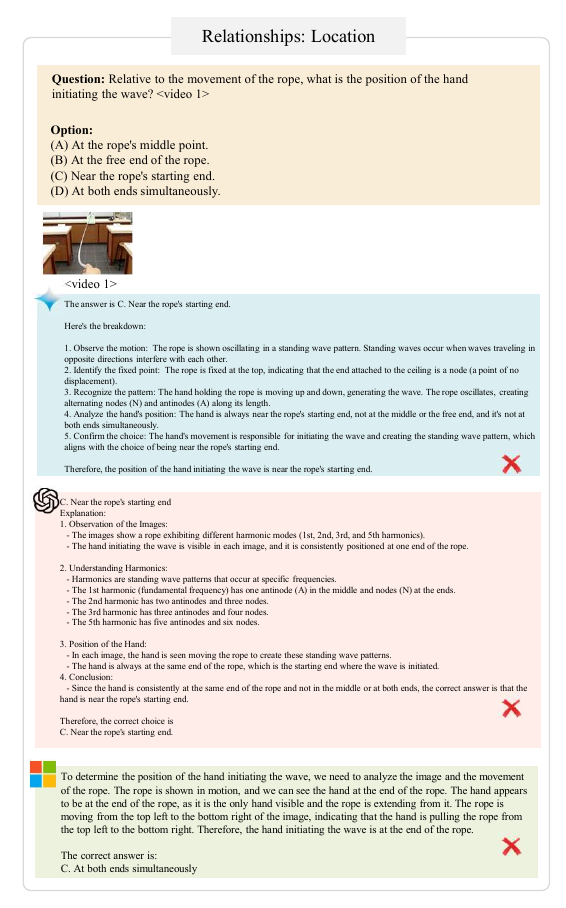}
  \caption{A sample case of physics-based dynamics——throwing. \hyperref[list:list_of_figures]{Back to List of Figures}.}
  \label{fig:error_47}
\end{figure}

\begin{figure}
  \centering
  \includegraphics[width=0.92\linewidth]{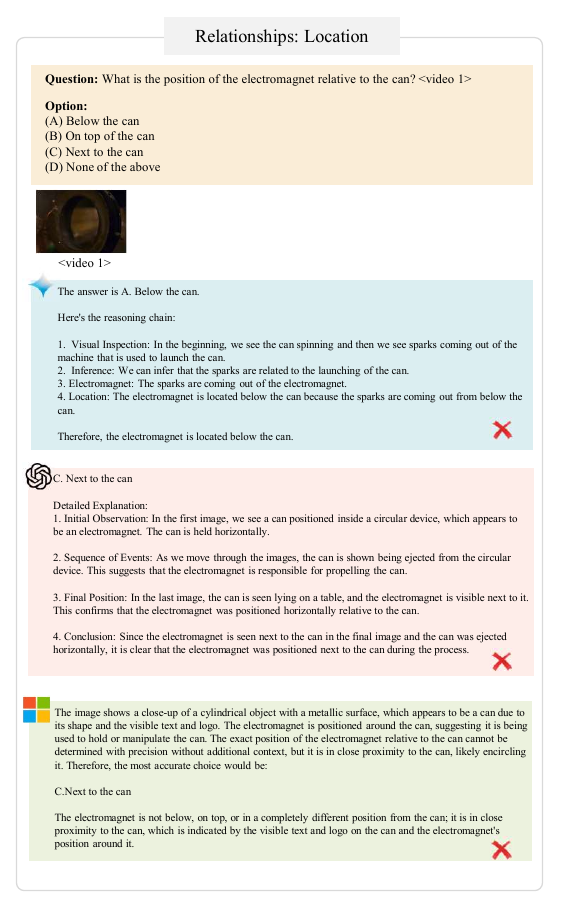}
  \caption{A sample case of physical object relationships--location. \hyperref[list:list_of_figures]{Back to List of Figures}.}
  \label{fig:error_48}
\end{figure}

\clearpage

\begin{figure}
  \centering
  \includegraphics[width=0.92\linewidth]{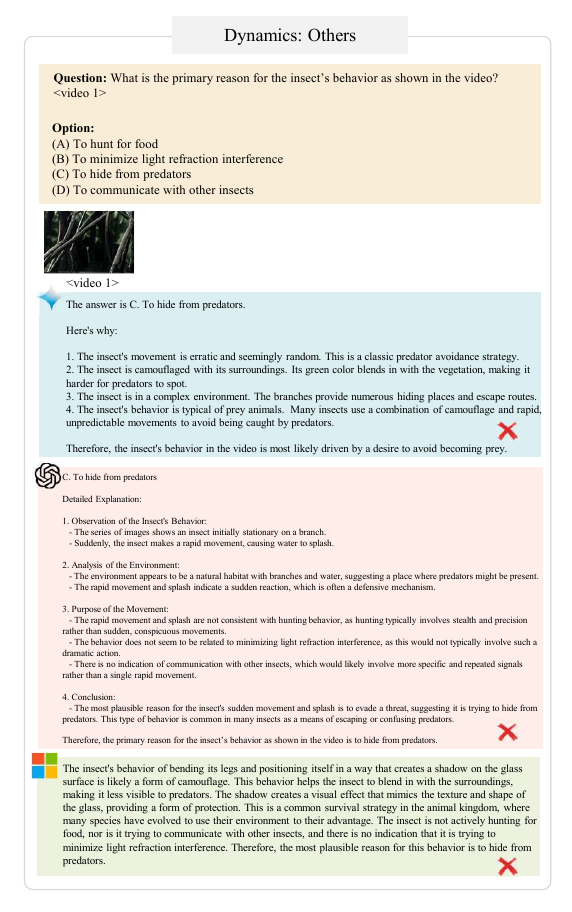}
  \caption{A sample case of physics-based dynamics——fluid. \hyperref[list:list_of_figures]{Back to List of Figures}.}
  \label{fig:error_49}
\end{figure}

\begin{figure}
  \centering
  \includegraphics[width=0.92\linewidth]{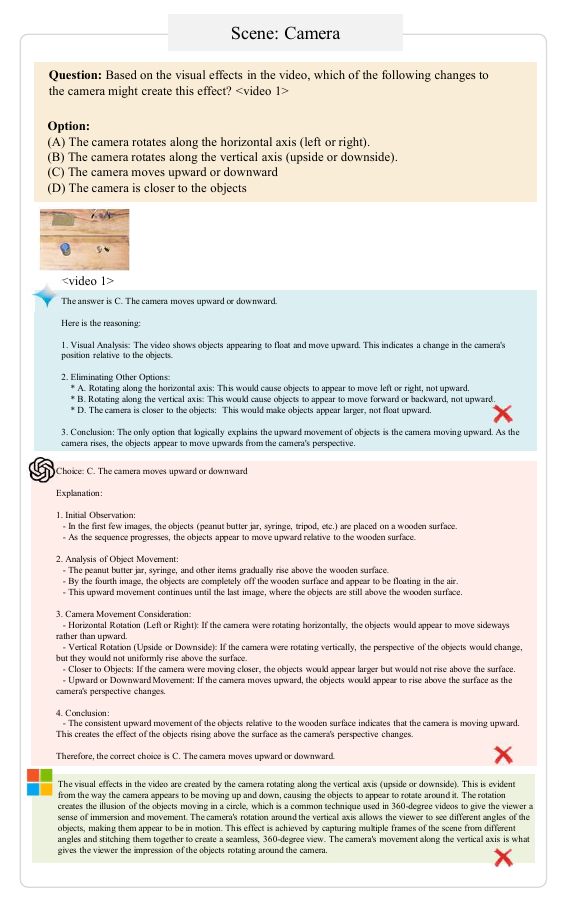}
  \caption{A sample case of physical scene understanding--viewpoint. \hyperref[list:list_of_figures]{Back to List of Figures}.}
  \label{fig:error_50}
\end{figure}

\begin{figure}
  \centering
  \includegraphics[width=0.92\linewidth]{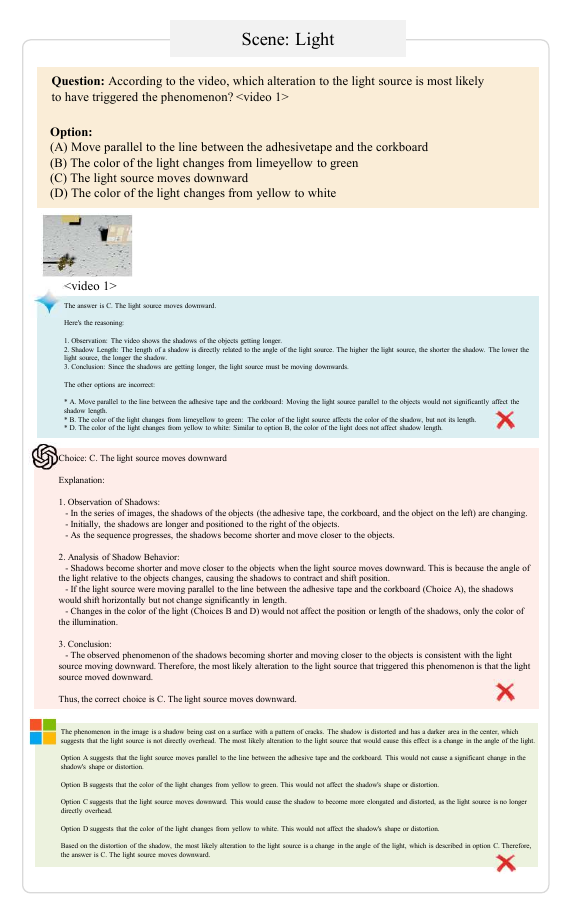}
  \caption{A sample case of physical scene understanding--light. \hyperref[list:list_of_figures]{Back to List of Figures}.}
  \label{fig:error_51}
\end{figure}

\begin{figure}
  \centering
  \includegraphics[width=0.92\linewidth]{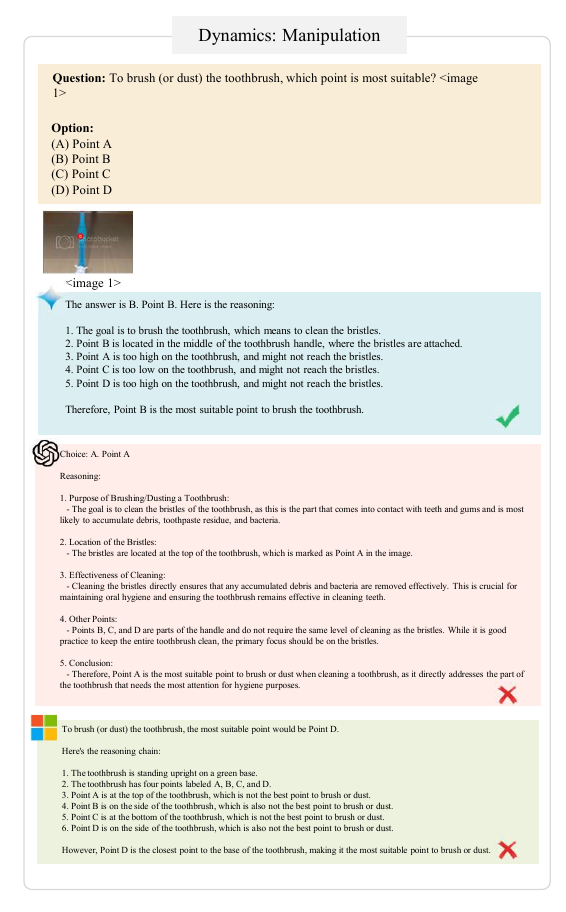}
  \caption{A sample case of physics-based dynamics——manipulation. \hyperref[list:list_of_figures]{Back to List of Figures}.}
  \label{fig:error_52}
\end{figure}

\begin{figure}
  \centering
  \includegraphics[width=0.92\linewidth]{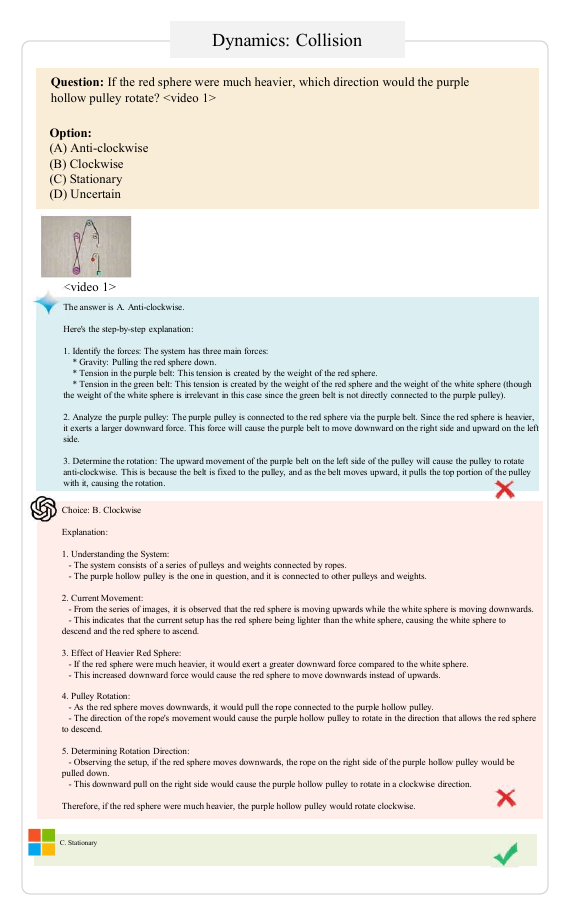}
  \caption{A sample case of physics-based dynamics——collision. \hyperref[list:list_of_figures]{Back to List of Figures}.}
  \label{fig:error_53}
\end{figure}

\begin{figure}
  \centering
  \includegraphics[width=0.92\linewidth]{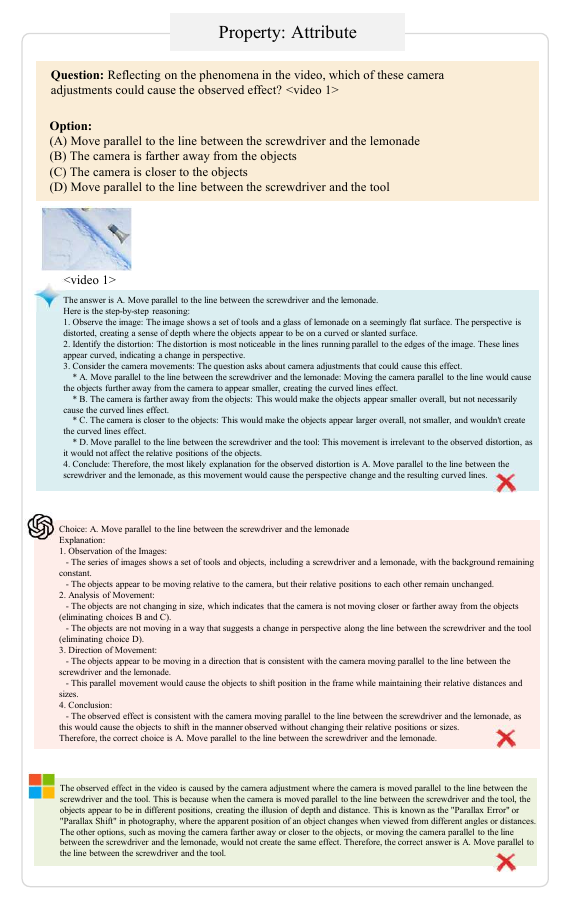}
  \caption{A sample case of physical object property--attribute. \hyperref[list:list_of_figures]{Back to List of Figures}.}
  \label{fig:error_54}
\end{figure}

\begin{figure}
  \centering
  \includegraphics[width=0.92\linewidth]{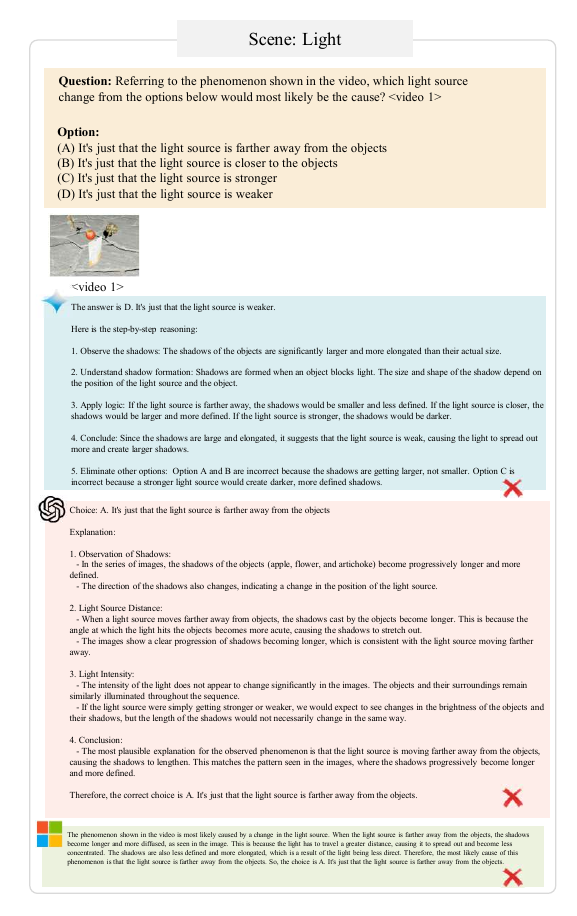}
  \caption{A sample case of physical scene understanding--light. \hyperref[list:list_of_figures]{Back to List of Figures}.}
  \label{fig:error_55}
\end{figure}

\begin{figure}
  \centering
  \includegraphics[width=0.92\linewidth]{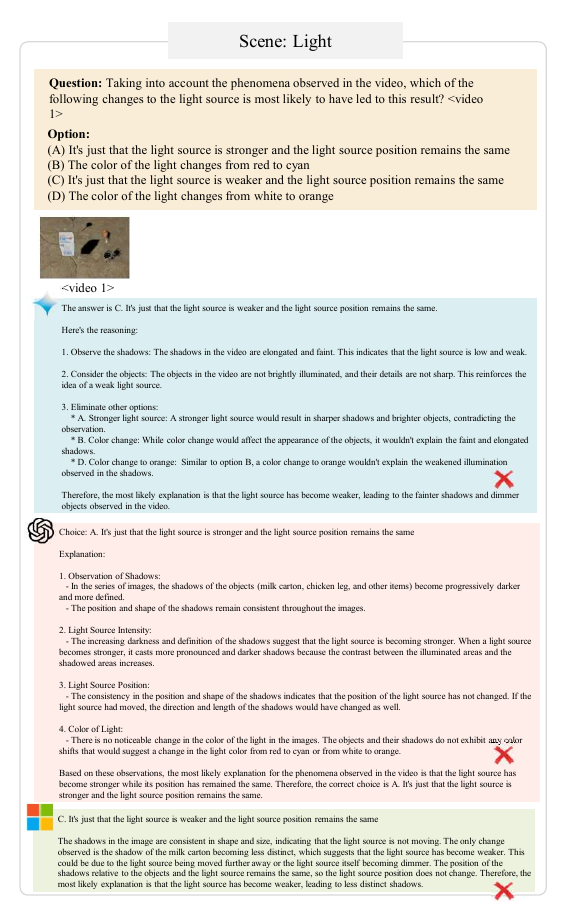}
  \caption{A sample case of physical object property--attribute. \hyperref[list:list_of_figures]{Back to List of Figures}.}
  \label{fig:error_56}
\end{figure}

\begin{figure}
  \centering
  \includegraphics[width=0.92\linewidth]{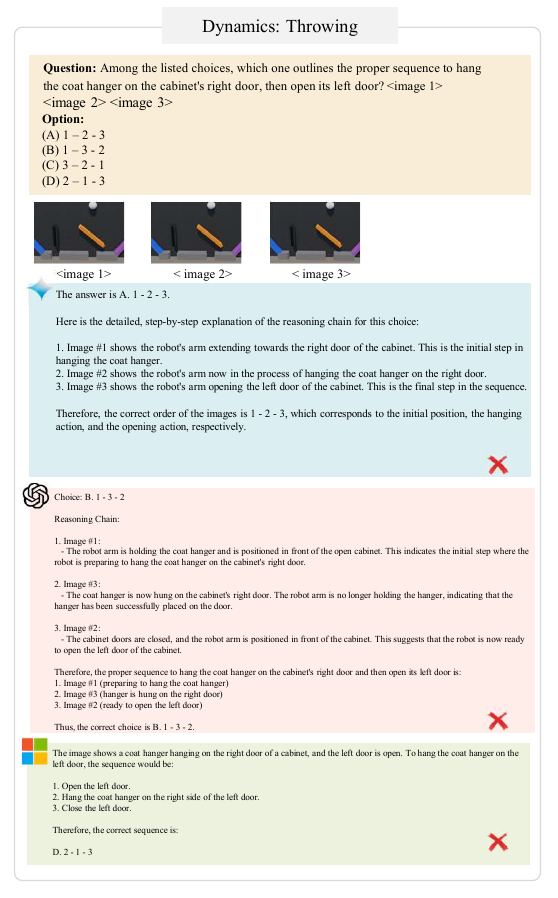}
  \caption{A sample case of physics-based dynamics——throwing. \hyperref[list:list_of_figures]{Back to List of Figures}.}
  \label{fig:error_57}
\end{figure}

\begin{figure}
  \centering
  \includegraphics[width=0.92\linewidth]{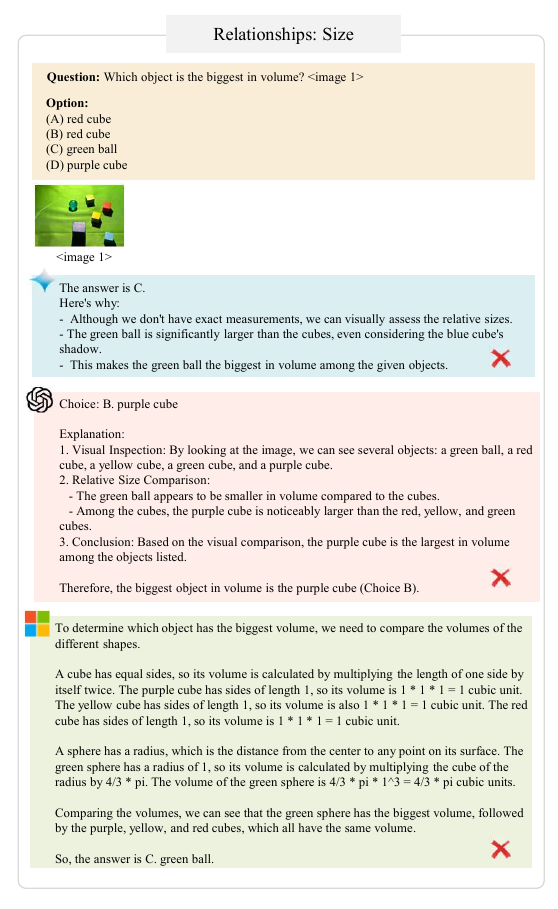}
  \caption{A sample case of physical object relationships--size. \hyperref[list:list_of_figures]{Back to List of Figures}.}
  \label{fig:error_58}
\end{figure}

\begin{figure}
  \centering
  \includegraphics[width=0.92\linewidth]{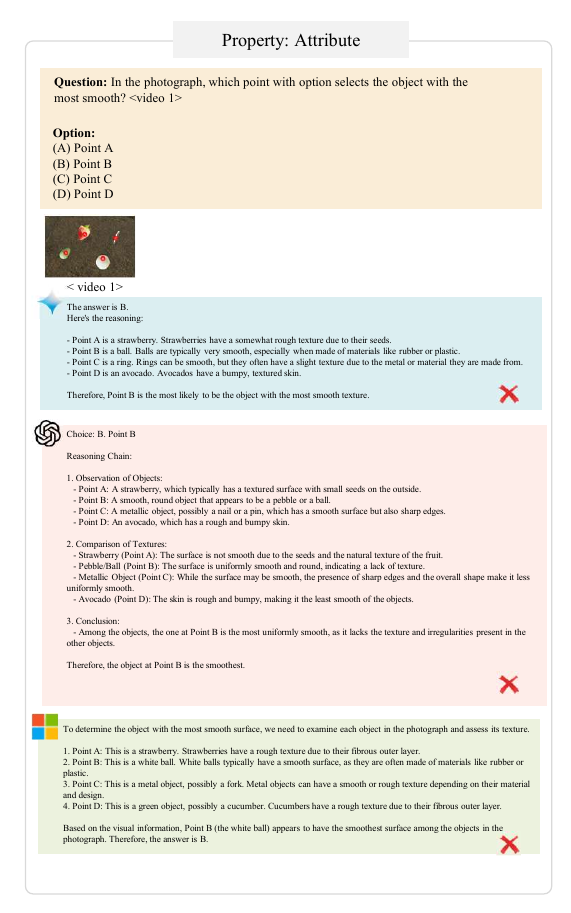}
  \caption{A sample case of physics-based dynamics——fluid. \hyperref[list:list_of_figures]{Back to List of Figures}.}
  \label{fig:error_59}
\end{figure}

\begin{figure}
  \centering
  \includegraphics[width=0.92\linewidth]{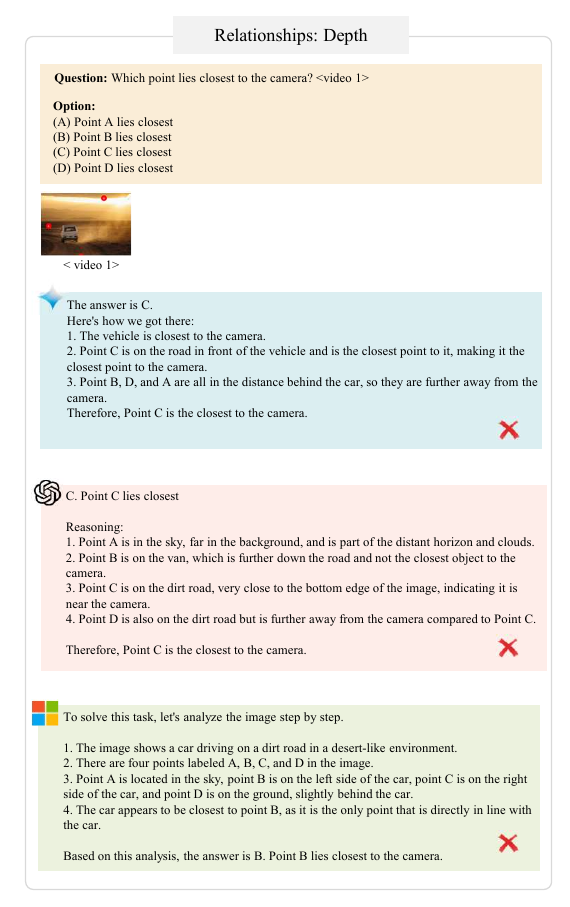}
  \caption{A sample case of physical object relationships--depth. \hyperref[list:list_of_figures]{Back to List of Figures}.}
  \label{fig:error_60}
\end{figure}

\clearpage
\vspace{-4mm}
\section{Discussion and Statement}~\label{app:statement}
\vspace{-4mm}
\subsection{Limitation}
Portions of our data are constructed based on pre-existing datasets, as detailed in Table~\ref{tab:data_ref}. We have made every effort to ensure that the images included in this paper comply with applicable copyright laws and are appropriately credited. Should you be the copyright holder of any image used in our work and believe that its usage conflicts with your licensing agreements, don't hesitate to get in touch with us directly. We are committed to promptly addressing any legitimate concerns.

Although PhysBench is categorized into 4 major categories and 19 subcategories, making it the first benchmark aimed at evaluating a vision-language model's understanding of the physical world, it still does not encompass all aspects of the real physical environment. We invested thousands of hours in data collection, organization, and annotation, following multiple rigorous processes and repeated reviews to ensure data quality. Nevertheless, minor issues may persist in a small portion of the dataset. 

In the experiments described in Section~\ref{exp:task_analysis} and~\ref{sec:phsagent}, we utilized GPT-4o-mini to extract answers, which, while currently being one of the most reliable and reproducible methods, may still unavoidably introduce hallucinations. Considering the challenges of using LLMs for evaluation, such as hallucinations and knowledge limitations, assessing open-ended formats is particularly difficult~\citep{yu2023mmvet, ge2024demon24, li2023fine}. Moreover, automatically evaluating the quality of reasoning processes presents additional challenges~\citep{lu2024mathvista}. To address these difficulties and simplify testing, we adopted a multiple-choice format. Nevertheless, PhysBench remains highly challenging, with even the most advanced GPT-4o model achieving less than 50\% accuracy. We believe that exploring more complex evaluation formats is a crucial direction for future research.

These challenges are not unique to our dataset but are common across various datasets in VLMs. Nevertheless, we believe that the potential benefits outweigh the associated risks, promoting continued advancement and societal progress. To the best of our knowledge, PhysBench and PhysAgent represent the first effort aimed at benchmarking and enhancing VLMs for physical world understanding, marking a significant step forward in the field. We are committed to the ongoing refinement of our dataset to further improve machine intelligence’s comprehension of the physical world, advancing embodied AI toward human-level capabilities.

\subsection{Boarder Impact}
The broader impact of PhysBench and PhysAgent carries both potential benefits and risks upon deployment and release. 
While some considerations pertain specifically to the nature of the dataset, others reflect broader challenges inherent to instruction-following vision-language models (VLMs). Below, we outline key risks and corresponding mitigation strategies.

\textbf{Biases.} PhysAgent may inherit biases from its foundational models, both in vision and language foundation models. These biases can manifest in skewed outcomes or unfair representations, necessitating careful evaluation and mitigation efforts.

\textbf{Anticipated Societal Implications.} A major societal concern is the potential misuse of the dataset and system, including the generation of fabricated content, which may contribute to misinformation, privacy infringements, and other detrimental outcomes. To mitigate these risks, strict adherence to ethical guidelines and ongoing oversight are essential.

\textbf{Environmental Considerations.} In alignment with environmental sustainability goals, we commit to publicly releasing the dataset and scripts to reduce unnecessary carbon emissions by regenerating similar datasets. Throughout our experiments, we ensure compliance with model and data licensing requirements.

\subsection{Ethics Statement}
This study does not raise any ethical concerns, as it exclusively utilizes publicly available pre-existing models, with no involvement of subjective evaluations. All research presented in this paper strictly adheres to the ethical guidelines set forth by the ICLR Code of Ethics.

\subsection{Reproducibility Statement}
We have adhered to the standard baseline settings employed by existing evaluation benchmarks or the original testing benchmarks of specific models. 
All necessary implementation details of our method are provided in Appendix~\ref{app:setup} and~\ref{app:more_data}. 
Furthermore, we are committed to releasing the data and code under an open-access license, accompanied by comprehensive instructions to ensure the accurate reproduction of the primary experimental results presented in this paper. All research conducted complies fully with the ICLR Reproducibility Requirements.

\section{Latest Results}\label{app:latex}
The models listed in Table~\ref{main_experiment} are current as of August 2024. Given the rapid evolution of VLMs, we evaluated an additional 36 models in December 2024, as shown in Table~\ref{main_experiment_new}. A description of these newly added models and their hyperparameters, along with those of the original 39 models, can be found in Appendix~\ref{app:hyper}. Due to space limitations, we have not combined the two tables, but a consolidated version is available on our project page at \href{https://physbench.github.io/}{\color{blue}\textbf{Our Project Page}\xspace}. We will continue to update the results to reflect the latest advancements in VLMs.

In Figure~\ref{fig:scaletable}(a), the largest model size tested was 13B. In this experiment, we have updated to larger models to evaluate whether increasing the model size significantly impacts performance on PhysBench. Based on the new models added in Table~\ref{main_experiment}, we present Figure~\ref{fig:scale_new} below.

\begin{figure}[th!]
	\centering  
	\vspace{-3mm}
	\includegraphics[width=\linewidth]{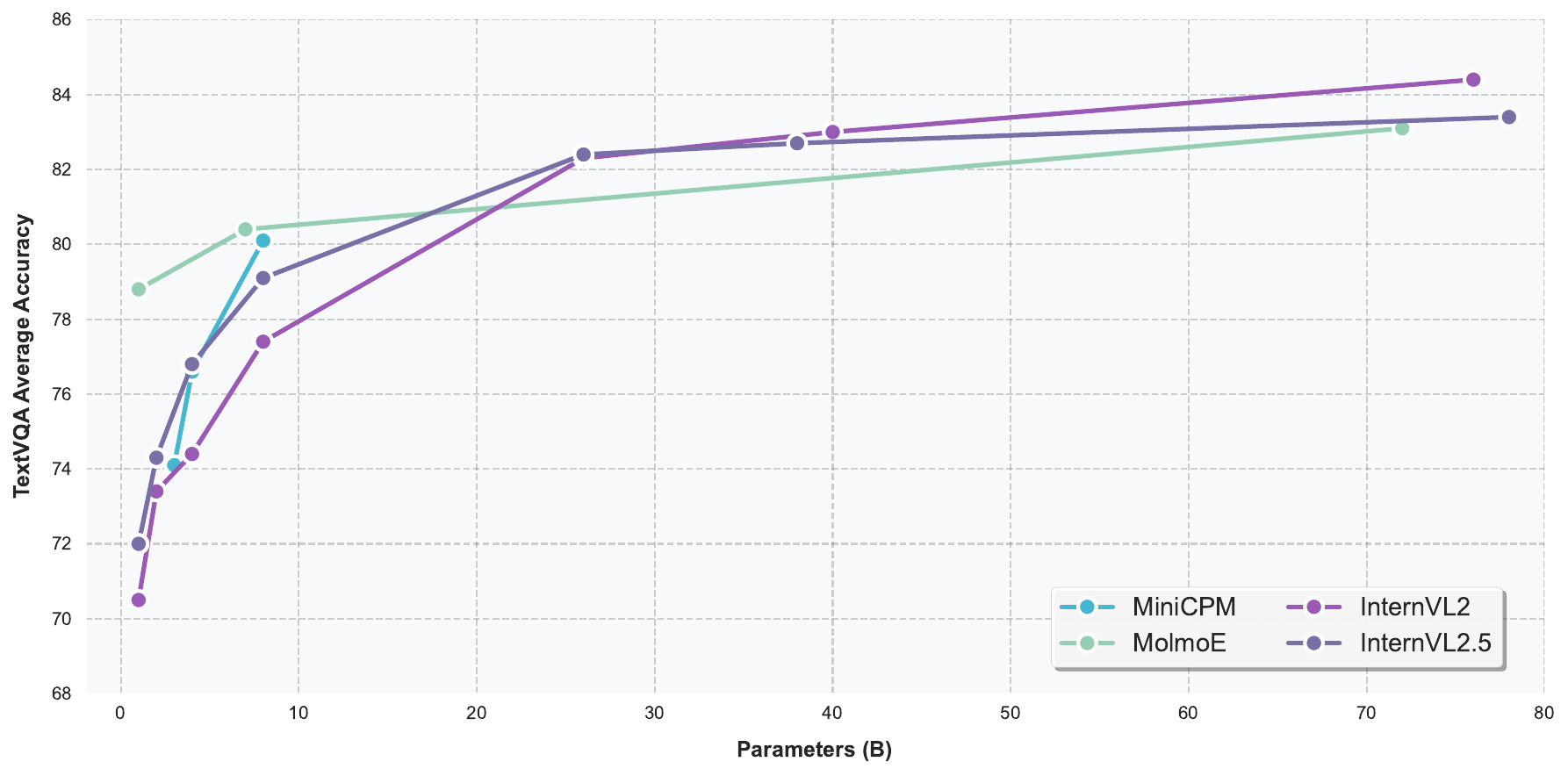}
	\vspace{-5mm}
        \caption{The performance of models of different sizes on TextVQA~\citep{singh2019towards}.}\label{fig:scale_new_tqa}  
	\vspace{-4mm}
\end{figure}

\begin{figure}[th!]
	\centering  
	\vspace{-3mm}
	\includegraphics[width=\linewidth]{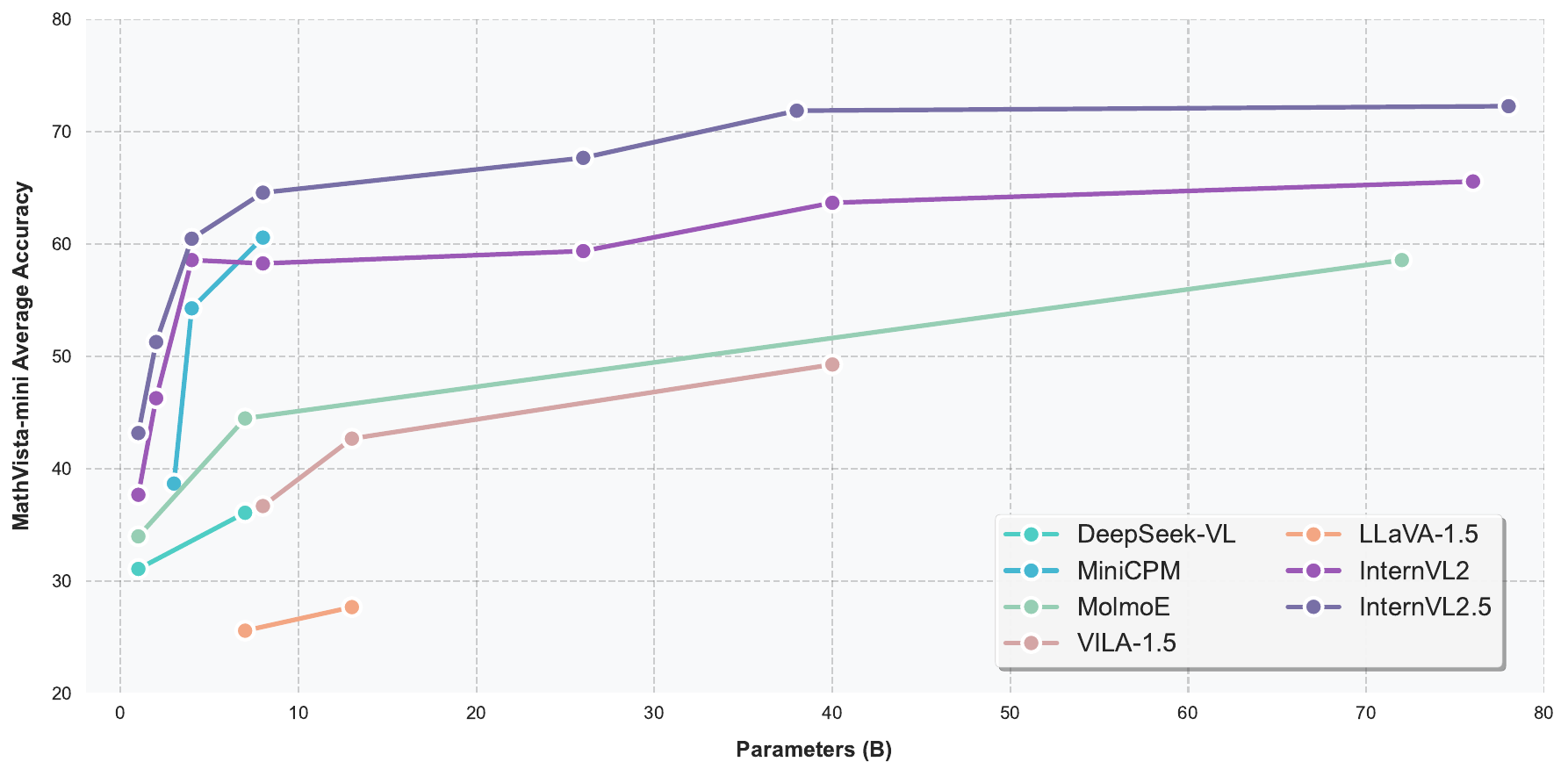}
	\vspace{-5mm}
        \caption{The performance of models of different sizes on MathVista~\citep{lu2024mathvista}.}\label{fig:scale_new_math}  
	\vspace{-4mm}
\end{figure}

\begin{figure}[th!]
	\centering  
	\vspace{-3mm}
	\includegraphics[width=\linewidth]{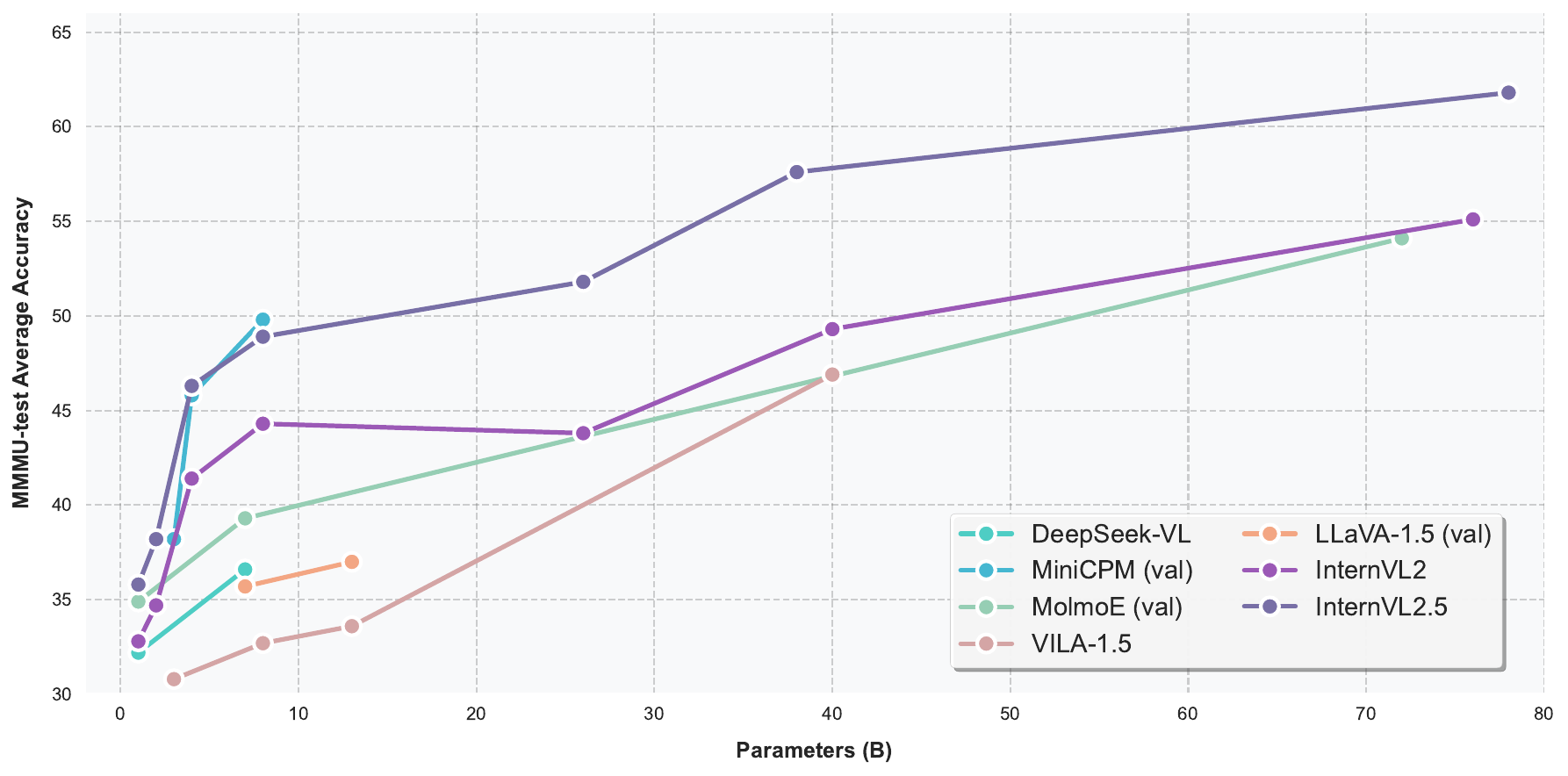}
	\vspace{-5mm}
        \caption{The performance of models of different sizes on MMMU~\citep{yue2023mmmu}.}\label{fig:scale_new_mmmu}  
	\vspace{-4mm}
\end{figure}

The results in Figure~\ref{fig:scale_new} reveal that models ranging from 1B to larger sizes generally show improvements, while the performance between 3B and 13B does not exhibit significant gains, and in some cases, such as with VILA and PLLaVA, a decrease in performance is observed. This observation constitutes the primary area of interest in Figure~\ref{fig:scaletable}(a). Notably, at larger scales, particularly at 26B and 40B, we observe substantial improvements compared to previous sizes. However, the scalability patterns in PhysBench differ from traditional VQA benchmarks, which typically demonstrate a strong positive correlation between model size and performance. We selected widely-used benchmarks including TextVQA~\citep{singh2019towards}, MathVista~\citep{lu2024mathvista}, and MMMU~\citep{yue2023mmmu} for comparison. The relationship between model size and performance on these benchmarks is illustrated in Figures~\ref{fig:scale_new_tqa}, ~\ref{fig:scale_new_math}, and ~\ref{fig:scale_new_mmmu}, which demonstrate this consistent scaling pattern.

\begin{figure}[th!]
	\centering  
	\vspace{-3mm}
	\includegraphics[width=\linewidth]{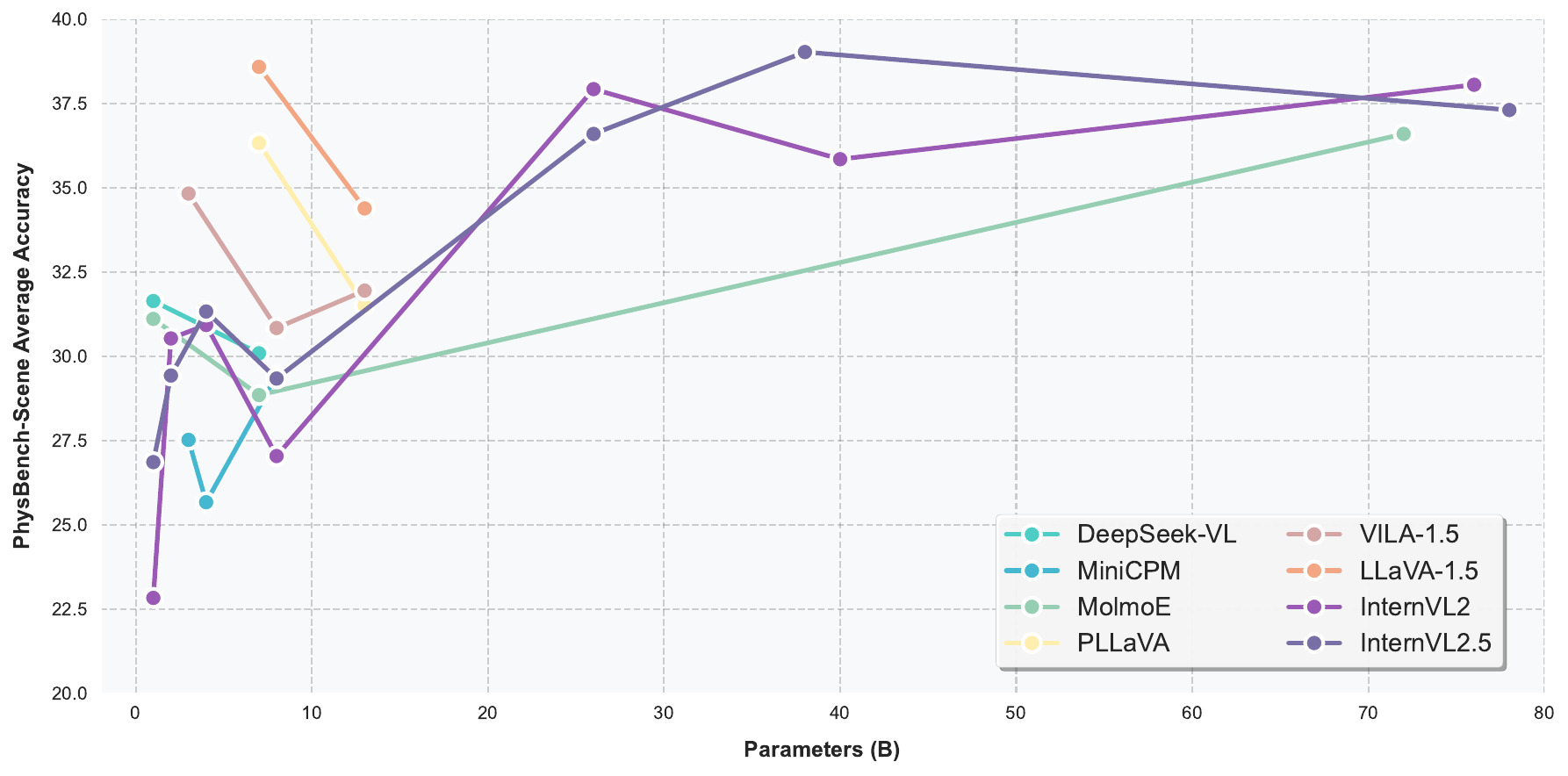}
	\vspace{-5mm}
        \caption{The performance of models of different sizes on PhysBench-Scene.}\label{fig:scale_new_sce}  
	\vspace{-4mm}
\end{figure}

\begin{figure}[th!]
	\centering  
	\vspace{-3mm}
	\includegraphics[width=\linewidth]{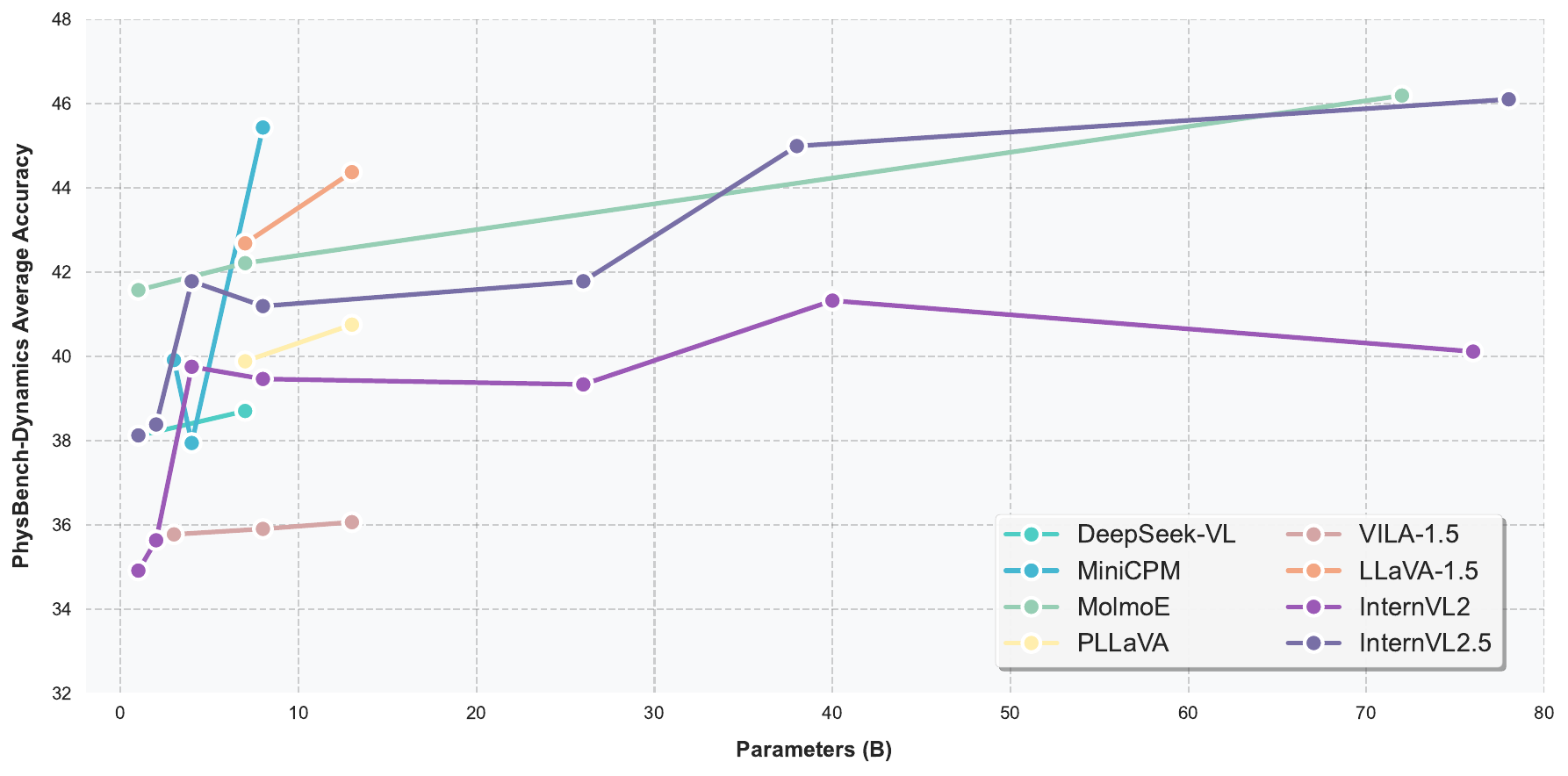}
	\vspace{-5mm}
        \caption{The performance of models of different sizes on PhysBench-Dynamics.}\label{fig:scale_new_dyn}  
	\vspace{-4mm}
\end{figure}

In contrast, PhysBench exhibits less pronounced scalability compared to these benchmarks, with performance not always correlating positively with model size, as shown in Figures~\ref{fig:scale_new}, ~\ref{fig:scale_new_sce}, and ~\ref{fig:scale_new_dyn}. This phenomenon is particularly evident in the Scene subcategory of PhysBench, where performance remains relatively stagnant in the 5-20B parameter range, only showing notable improvements with models exceeding 25B parameters, as demonstrated in Figure~\ref{fig:scale_new_sce}.

\begin{figure}[th!]
	\centering  
	\vspace{-3mm}
	\includegraphics[width=\linewidth]{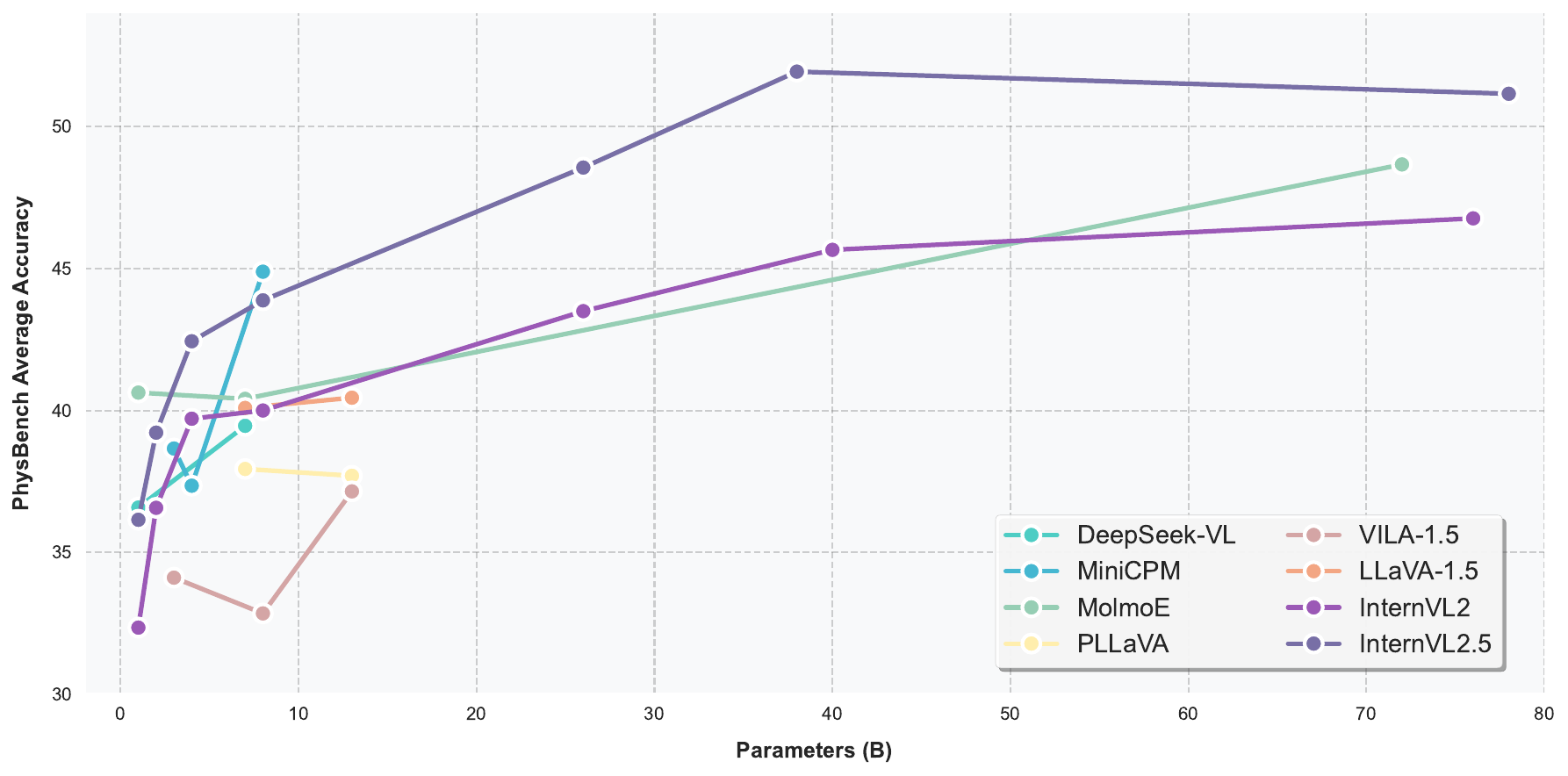}
	\vspace{-5mm}
        \caption{The performance of models of different sizes on PhysBench.}\label{fig:scale_new}  
	\vspace{-4mm}
\end{figure}

\begin{table}[th!]
    \setlength{\tabcolsep}{3pt}
    \centering
    \scalebox{0.84}{
\begin{tabular}{lccccccc} \hline
    & Size & Format & \faGear Property &  \faLocationArrow Relationships & \faTree Scene & \faFlask Dynamics & \textbf{Avg}\\\hline
\multicolumn{7}{c}{Image VLM} \\\hline
MolmoE-1B~\cite{deitke2024molmo} &   1B  &merge  &43.17  & 58.06& 31.12& 41.58&40.63\\
MolmoE-7B~\cite{deitke2024molmo} &   7B  &merge  &42.35  &63.33 &28.86 &42.22 &40.41\\
MolmoE-7B-D~\cite{deitke2024molmo} &   7B & merge  &49.48  &67.78 &30.81 &47.18 &45.18\\
MolmoE-72B~\cite{deitke2024molmo} &   72B & merge  &55.13  &72.08 &36.61 &46.20 &48.67 \\
MiniCPM2~\citep{yao2024minicpmvgpt4vlevelmllm} &   3B &  merge &45.62  &44.17&27.53 &39.92 &38.66\\
MiniCPM2.5~\citep{yao2024minicpmvgpt4vlevelmllm} &   4B&merge    &44.26  &47.08 &25.68 &37.95 &37.35\\
\hline
  \end{tabular}
    }
    \vspace{-1mm}
    \caption{\textbf{Evaluation results for latest 36 VLMs.} The evaluation of General VLMs is based on the data from Video and Image VLM evaluations, with the addition of interleaved data. 
    ``Seq`` refers to sequential input of frames of videos, while ``merge`` refers to merging video frames into a single image. Results of original 39 VLMs can be seen in Table~\ref{main_experiment}.
    }\label{main_experiment_new}
    \vspace{-4mm}
\end{table}

\begin{table}[th!]
    \setlength{\tabcolsep}{3pt}
    \centering
    \scalebox{0.68}{
\begin{tabular}{lcccccccccc} \hline
\textbf{Model}& Size & \textbf{Identify} & \textbf{Comparison} & \textbf{Static} & \textbf{Dynamic} & \textbf{Perception} & \textbf{Prediction} & \textbf{Judgment} & \textbf{Reasoning} & \textbf{Avg}\\\hline
\multicolumn{11}{c}{Image VLM} \\\hline
MolmoE-1B~\cite{deitke2024molmo} &   1B   & 56.76 & 32.42 & 64.74 & 39.07 & 48.22 & 37.84 & 41.88 & 29.22 & 40.63 \\
MolmoE-7B~\cite{deitke2024molmo} &   7B   & 54.84 & 32.49 & 72.01 & 37.75 & 49.85 & 39.22 & 45.01 & 25.20 & 40.41 \\
MolmoE-7B-D~\cite{deitke2024molmo} &   7B   & 67.04 & 35.63 & 74.44 & 49.67 & 54.74 & 43.82 & 49.00 & 27.19 & 45.18 \\
MolmoE-72B~\cite{deitke2024molmo} &   72B  & 68.70 & 44.51 & 76.49 & 58.94 & 54.33 & 43.13 & 50.71 & 33.00 & 48.67 \\
MiniCPM2~\citep{yao2024minicpmvgpt4vlevelmllm} &   3B   & 62.41 & 32.49 & 45.71 & 38.41 & 47.40 & 36.33 & 44.86 & 24.12 & 38.66 \\
MiniCPM2.5~\citep{yao2024minicpmvgpt4vlevelmllm} &   4B   & 49.48 & 40.07 & 45.90 & 51.66 & 40.88 & 36.06 & 41.43 & 23.62 & 37.35 \\
\hline
\end{tabular}
    }
    \caption{\textbf{Evaluation Results for 36 new VLMs Categorized by Ability Dimensions}. PhysBench includes two evaluation dimensions: ability and task. Table~\ref{main_experiment_new} presents results categorized by the task-type dimension.}
    \label{tab:app_exp_main_ability_new}
\end{table}

\begin{table}[th!]
    \setlength{\tabcolsep}{3pt}
    \centering
    \scalebox{0.93}{
\begin{tabular}{lccccc}\hline
\textbf{Model}      &\textbf{Size}          & \textbf{Number} & \textbf{Mass} & \textbf{Color} & \textbf{Attribute} \\\hline
\multicolumn{6}{c}{Image VLM} \\\hline
MolmoE-1B~\cite{deitke2024molmo} &   1B   &  56.85 & 23.89 & 52.67 & 39.39 \\
MolmoE-7B~\cite{deitke2024molmo} &   7B   & 48.01 & 30.03 & 56.67 & 39.46 \\
MolmoE-7B-D~\cite{deitke2024molmo} &   7B   & 65.16 & 36.52 & 63.33 & 42.73 \\
MolmoE-72B~\cite{deitke2024molmo} &   72B  & 63.95 & 51.19 & 64.00 & 50.25 \\
MiniCPM2~\citep{yao2024minicpmvgpt4vlevelmllm} &   3B   & 56.52 & 42.66 & 66.56 & 37.26 \\
MiniCPM2.5~\citep{yao2024minicpmvgpt4vlevelmllm} &   4B   & 46.26 & 50.17 & 45.82 & 41.80 \\
\hline
\end{tabular}
    }
    \caption{
    Evaluation Results for 36 new VLMs in PhysBench Physical Object \faGear Property Sub-task (the fifth last column of Table~\ref{main_experiment_new}).}
    \label{tab:app_exp_main_pp_new}
\end{table}

\begin{table}[th!]
    \setlength{\tabcolsep}{3pt}
    \centering
    \scalebox{0.93}{
   \begin{tabular}{lcccccc} 
\hline
\textbf{Model}                & \textbf{Model Size} & \textbf{Size} & \textbf{Location} & \textbf{Depth} & \textbf{Distance} & \textbf{Motion} \\\hline
\multicolumn{7}{c}{Image VLM} \\\hline
MolmoE-1B~\cite{deitke2024molmo} &   1B   &  39.29 & 51.13 & 72.28 & 48.33 & 50.56 \\
MolmoE-7B~\cite{deitke2024molmo} &   7B   & 42.86 & 61.65 & 75.44 & 53.33 & 50.56 \\
MolmoE-7B-D~\cite{deitke2024molmo} &   7B   &  58.93 & 60.15 & 79.30 & 66.67 & 59.55 \\
MolmoE-72B~\cite{deitke2024molmo} &   72B  & 76.79 & 67.29 & 76.49 & 68.33 & 74.16 \\
MiniCPM2~\citep{yao2024minicpmvgpt4vlevelmllm} &   3B   & 50.00 & 45.49 & 39.65 & 46.67 & 59.55 \\
MiniCPM2.5~\citep{yao2024minicpmvgpt4vlevelmllm} &   4B   & 66.07 & 36.84 & 51.23 & 43.33 & 67.42 \\
\hline
\end{tabular}
    }
    \caption{Evaluation Results for 36 new VLMs in PhysBench Object  \faLocationArrow Relationships Sub-task (the forth last column of Table~\ref{main_experiment_new}).
    } 
    \label{tab:app_exp_main_spa_new}
\end{table}

\begin{table}[th!]
    \setlength{\tabcolsep}{3pt}
    \centering
    \scalebox{1.0}{
        \begin{tabular}{lccccc}
        \hline
        \textbf{Model} &\textbf{Size}               & \textbf{Temperature} & \textbf{Viewpoint} & \textbf{Air} & \textbf{Light} \\\hline
        \multicolumn{6}{c}{Image VLM} \\\hline
        MolmoE-1B~\cite{deitke2024molmo} &   1B   & 50.00 & 30.73 & 26.00 & 30.21 \\
        MolmoE-7B~\cite{deitke2024molmo} &   7B   & 64.71 & 21.11 & 36.00 & 33.76 \\
        MolmoE-7B-D~\cite{deitke2024molmo} &   7B   & 79.41 & 42.70 & 86.00 & 26.84 \\
        MolmoE-72B~\cite{deitke2024molmo} &   72B  & 76.79 & 67.29 & 76.49 & 68.33 \\
        MiniCPM2~\citep{yao2024minicpmvgpt4vlevelmllm} &   3B   & 66.18 & 20.83 & 40.00 & 31.12 \\
        MiniCPM2.5~\citep{yao2024minicpmvgpt4vlevelmllm} &   4B   & 48.53 & 13.85 & 42.00 & 35.21 \\
\hline
        \end{tabular}
    }
    \caption{Evaluation Results for 36 new VLMs in PhysBench Physical \faTree Scene Understanding Sub-task (the third last column of Table~\ref{main_experiment_new}.}
    \label{tab:app_exp_main_env_new}
\end{table}

\begin{table}[th!]
    \setlength{\tabcolsep}{3pt}
    \centering
    \scalebox{0.8}{
        \begin{tabular}{lccccccc}
        \hline
        \textbf{Model} & \textbf{Size}     & \textbf{Collision} & \textbf{Throwing} & \textbf{Manipulation} & \textbf{Fluid} & \textbf{Chemistry} & \textbf{Others} \\\hline
            \multicolumn{8}{c}{Image VLM} \\\hline
            MolmoE-1B~\cite{deitke2024molmo} &   1B   & 32.72 & 33.02 & 43.47 & 44.75 & 71.62 & 57.48 \\
            MolmoE-7B~\cite{deitke2024molmo} &   7B   & 28.26 & 28.74 & 36.68 & 53.55 & 67.57 & 67.01 \\
            MolmoE-7B-D~\cite{deitke2024molmo} &   7B   & 38.56 & 39.43 & 39.45 & 50.15 & 68.92 & 72.11 \\
            MolmoE-72B~\cite{deitke2024molmo} &   72B  & 37.02 & 40.38 & 30.90 & 48.30 & 63.51 & 80.61 \\
            MiniCPM2~\citep{yao2024minicpmvgpt4vlevelmllm} &   3B   & 34.82 & 34.20 & 34.92 & 34.88 & 67.57 & 67.69 \\
            MiniCPM2.5~\citep{yao2024minicpmvgpt4vlevelmllm} &   4B   & 433.44 & 36.82 & 27.89 & 35.03 & 54.05 & 60.88 \\
\hline
        \end{tabular}
    }
    \caption{Evaluation Results for 36 new VLMs in PhysBench Physics-based \faFlask Dynamics Sub-task (the second last column of Table~\ref{main_experiment_new}).}
    \label{tab:app_exp_main_phe_new}
\end{table}
\end{document}